\documentclass[twoside,11pt]{article}

\usepackage{melba}
\usepackage{subcaption, multicol, multirow}
\usepackage{algorithm, algorithmic}
\usepackage{booktabs}
\usepackage{microtype}

\graphicspath{ {figures/} }
\hypersetup{hidelinks}

\melbaheading{011}{https://www.melba-journal.org/article/27647}{2021}{1--25}{01/2021}{10/2021}{Haque, Imran, Wang and Terzopoulos}{}{}

\firstpageno{1}

\title{Generalized Multi-Task Learning from Substantially Unlabeled Multi-Source Medical Image Data}

\ShortHeadings{Multi-Task Learning from Substantially Unlabeled Multi-Source Image Data}{Haque, Imran, Wang and Terzopoulos}

\author{\name Ayaan Haque\thanks{Co-first author} \email ayaanzhaque@gmail.com\\
	\addr Saratoga High School, Saratoga, CA, USA
	\AND
	\name Abdullah-Al-Zubaer Imran\footnotemark[\value{footnote}] \email aimran@stanford.edu\\
	\addr Stanford University, Stanford, CA, USA
	\AND
	\name Adam Wang \email adamwang@stanford.edu\\
	\addr Stanford University, Stanford, CA, USA
	\AND
	\name Demetri Terzopoulos \email dt@cs.ucla.edu \\
	\addr University of California, Los Angeles, CA, USA\\
	\addr VoxelCloud, Inc., Los Angeles, CA, USA
}

\begin{document}

\maketitle

\begin{abstract}
Deep learning-based models, when trained in a fully-supervised manner, can be effective in performing complex image analysis tasks, although contingent upon the availability of large labeled datasets. Especially in the medical imaging domain, however, expert image annotation is expensive, time-consuming, and prone to variability. Semi-supervised learning from limited quantities of labeled data has shown promise as an alternative. Maximizing knowledge gains from copious unlabeled data benefits semi-supervised learning models. Moreover, learning multiple tasks within the same model further improves its generalizability. We propose MultiMix, a new multi-task learning model that jointly learns disease classification and anatomical segmentation in a semi-supervised manner, while preserving explainability through a novel saliency bridge between the two tasks. Our experiments with varying quantities of multi-source labeled data in the training sets confirm the effectiveness of MultiMix in the simultaneous classification of pneumonia and segmentation of the lungs in chest X-ray images. Moreover, both in-domain and cross-domain evaluations across these tasks further showcase the potential of our model to adapt to challenging generalization scenarios.
\end{abstract}


\section{Introduction}

Learning-based medical image analysis has become widespread with the advent of deep learning. However, most deep learning models rely on large pools of labeled data. Especially in the medical imaging domain, obtaining copious labeled imagery is often infeasible, as annotation requires substantial domain expertise and manual labor. Therefore, developing large-scale deep learning methodologies for medical image analysis tasks is challenging. In confronting the limited labeled data problem, Semi-Supervised Learning (SSL) has been gaining attention. In semi-supervised learning, unlabeled training examples are leveraged in combination with labeled examples to maximize information gains \citep{chapelle2009semi}. Specifically within the medical domain, where collecting data is generally easier than annotating those data, the use of deep learning for medical image analysis tasks can be fostered by leveraging semi-supervised learning.

Recent research has yielded a variety of semi-supervised learning techniques \citep{imran2020fully}. Pseudo-labeling \citep{lee2013pseudo} trains a model with labeled data and unlabeled data simultaneously, generating labels for the unlabeled data by assuming the model-predicted labels to be reliable. Similarly, entropy minimization \citep{grandvalet2005semi} trains so as to match the predicted data distribution of unlabeled data with that of the labeled data, under the assumption that unlabeled examples should yield prediction distributions that are similar to those from labeled examples \citep{ouali2020overview}. Domain adaptation \citep{beijbom2012domain} is a form of inductive transfer learning, where a model is trained on labeled data from the source domain as well as labeled plus unlabeled data from the target domain, which improves model generalization for the target domain, but lacks clinical value if the target domain data is inaccessible during training.

Thus, progress has been made in learning from limited labeled data, although mainly within the confines of single-task learning. In particular, individual medical imaging tasks, such as diagnostic classification and anatomical segmentation, have been addressed using state-of-the-art Convolutional Neural Network (CNN) models \citep{anwar2018medical}; e.g., for medical image segmentation, encoder-decoder networks \citep{ronneberger2015unet}, variational auto-encoder networks, \citep{myronenko20183d}, context encoder networks \citep{gu2019conext}, multiscale adversarial learning \citep{imran2020progressive}, etc.

By contrast, Multi-Task Learning (MTL) is defined as optimizing more than one loss in a single model such that multiple related tasks are performed by sharing the learned representation \citep{ruder2017overview}. Jointly training multiple tasks within a model improves the generalizability of the model as each of the tasks regularizes the others \citep{Caruana1993MultitaskLA}. Assuming that training data with limited annotations come from different distributions for different tasks, multi-task learning may be useful in such scenarios for learning in a scarcely-supervised manner \citep{imran2020partly, imran2021multi}.  

Combining the objectives of substantially unlabeled data training and multi-task learning, Semi-Supervised Multi-Task Learning (SSMTL) is a promising research area in the context of medical image analysis. While there have been prior efforts on multi-tasking \citep{mehta2018ynet, Girard_2019}, rarely do they focus on incorporating semi-supervised learning particularly within the medical realm. \citet{liu2008semi} proposed a general semi-supervised multi-tasking method that uses soft-parameter sharing to allow multiple classification tasks in a single model. \citet{gao2019feature} performed multi-tasking on tasks within the same medical domain by exploiting feature transfer. Adversarial learning \citep{salimans2016improved} combines a classifier with a discriminator to perform semi-supervised, adversarial multi-tasking. \citet{imran2019semi} introduced semi-supervised multi-task learning using adversarial learning and attention masking. \citet{Zhou_2019_CVPR} proposed a semi-supervised multi-tasking model that uses an attention mechanism to grade segmented retinal images. None of the aforecited works, however, take into consideration the disparity in the training data distributions for multiple tasks.

To learn diagnostic classification and anatomical segmentation jointly from substantially unlabeled multi-source data, we propose \emph{MultiMix}, a novel, better-generalized multi-tasking model that incorporates confidence-based augmentation and a module that bridges the classification and segmentation tasks. This \emph{saliency bridge module} produces a saliency map by computing the gradient of the class score with respect to the input image, thus not only enabling the analysis of the model's predictions, but also improving the model's performance of both tasks. While the explainability of any deep learning model can be based on visualizing saliency maps \citep{simonyan2014deep,Zhang2016BridgingSD, hu2019sunet}, to our knowledge a saliency bridge between two shared tasks within a single model has not previously been explored. We demonstrate that the saliency bridge module in conjunction with a simple yet effective semi-supervised learning method in a multi-tasking setting can yield improved and consistent performance across multiple domains.

This article is a revised and extended version of our ISBI publication \citep{haque2021multimix}.\footnote{With an augmented literature review, a more detailed explanation of the methods, model architecture, and training algorithm, further details about the datasets, saliency map visualizations from multiple datasets, and additional results and discussion supported by quantitative (performance metrics tables) and qualitative (mask predictions, Bland Altman plots, ROC curves, consistency plots) characteristics.} Our main contributions may be summarized as follows:
\begin{itemize}
\item 
A new semi-supervised learning model, MultiMix, that exploits confidence-based data augmentation and consistency regularization to jointly learn diagnostic classification and anatomical segmentation from multi-source, multi-domain medical image datasets.

\item 
Incorporation of an innovative saliency bridge module connecting the segmentation and classification branches of the model, resulting in the improved performance of both tasks.

\item 
Substantiation of the improved generalizability (both in-domain and cross-domain) of the proposed model via experimentation with varied quantities of labeled data and mixed data sources related to multiple tasks, specifically in the classification of pneumonia and the simultaneous segmentation of the lungs in chest X-ray images.

\item
MultiMix software made available at~\url{https://github.com/ayaanzhaque/MultiMix}.
\end{itemize}

\section{The MultiMix Model}

To formulate our approach, we assume unknown data distributions $p(X^c, C)$ over images $X^c$ and class labels $C$ as well as $p(X^s,S)$ over images $X^s$ and segmentation labels $S$. Hence, segmentation labels for the $X^c$ images and class labels for the $X^s$ images are unavailable. We also assume access to labeled training sets $\mathcal{D}^c_l$ sampled i.i.d.~from $p(X^c, C)$ and $\mathcal{D}^s_l$ sampled i.i.d.~from $p(X^s,S)$, along with unlabeled training sets $\mathcal{D}^c_u$ sampled i.i.d.~from $p(X^c)$ and $\mathcal{D}^s_u$ sampled i.i.d.~from $p(X^s)$, after marginalizing out $C$ and $S$, respectively. 

\begin{figure}
    \centering
    \includegraphics[width=\linewidth]{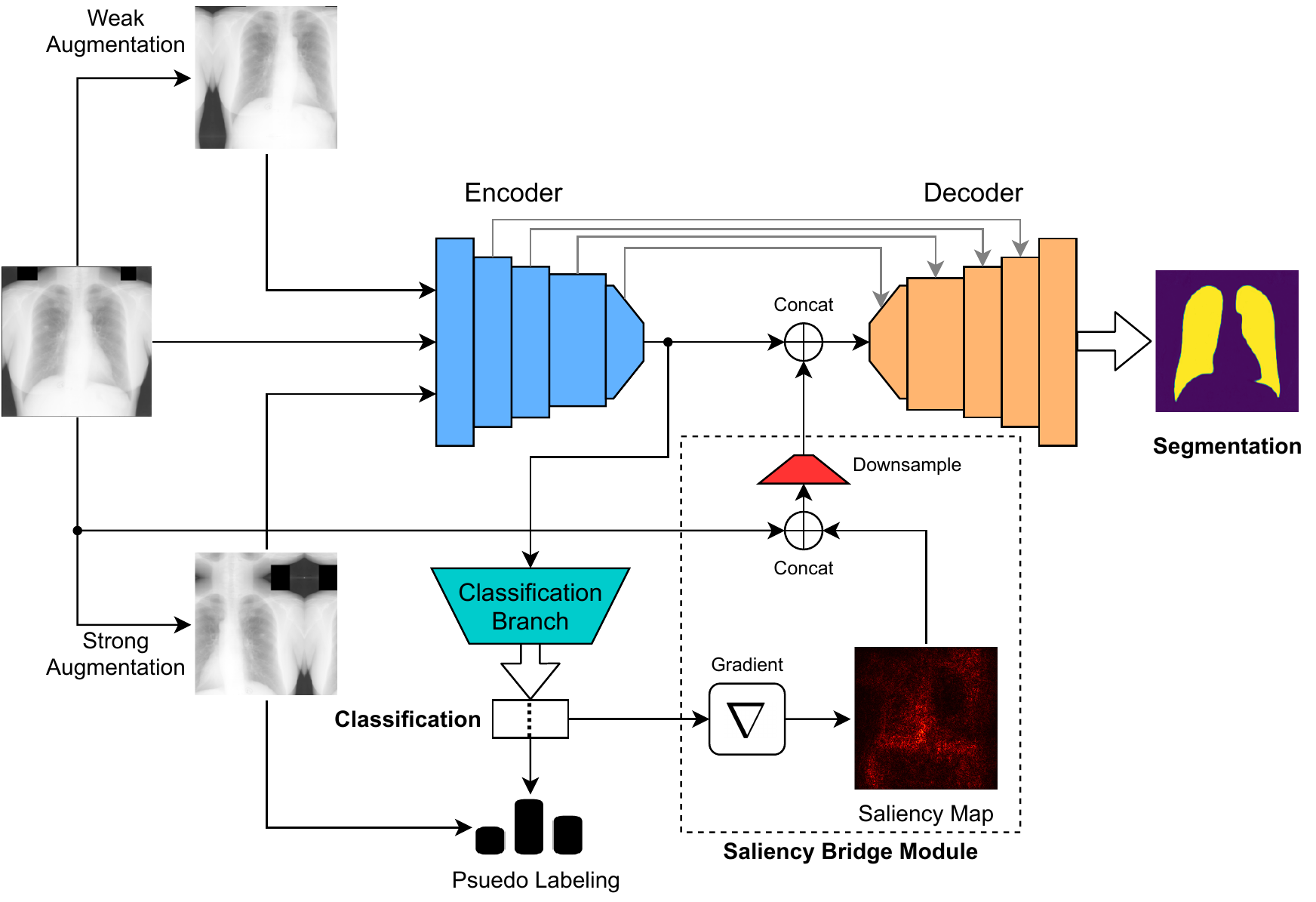}
    \caption{Schematic of the MultiMix model. \emph{Classification:} Using predictions on unlabeled weakly augmented images, pseudo-labels are generated with confidence, and loss is computed with these labels and the strongly augmented versions of those images. \emph{Segmentation:} Saliency maps generated from the class predictions are concatenated via the saliency bridge module to guide the decoder in generating the segmentation masks.}
    \label{fig:model}
\end{figure}

In our MultiMix model (Figure~\ref{fig:model}), we utilize a U-Net-like \citep{ronneberger2015unet} encoder-decoder architecture for image deconstruction and reconstruction. The encoder functions similarly to a standard CNN. To perform multi-tasking, we use pooling layers followed by fully-connected layers, allowing the encoder to output class predictions through the classification branch of the model. Furthermore, in the segmentaton branch of the model, the segmentation predictions are obtained as the output of the decoder.

MultiMix performs multi-tasking in a semi-supervised learning manner, assuming the training data for the two tasks come from disparate distributions. It is well established that a multi-tasking model usually outperforms its single-task counterparts \citep{imran2019semi, imran2020partly, imran2020fully}. The shared encoder in the MultiMix model learns features useful for addressing both the classification and segmentation tasks. This joint representation learning enables the model to avoid overfitting and generalize better. Most importantly, it exploits the relatedness of the tasks, which is crucial for effective multi-tasking.

In the following sections, we describe the classification and segmentation branches of the MultiMix model, explain the saliency bridge module that bridges the two branches, and specify the MultiMix training procedure.

\subsection{Classification Branch}

For semi-supervised classification, we leverage data augmentation and pseudo-labeling. Inspired by the work of \citet{sohn2020fixmatch}, for each unlabeled image we perform two degrees of augmentation: weak and strong. The former consists of standard augmentations---both random horizontal flipping and random cropping---and is applied to the labeled data as well, whereas the latter is performed by randomly applying any number of augmentations from a pool of ``heavy'' augmentations.\footnote{This pool includes augmentations such as horizontal flip, crop, autocontrast, brightness, contrast, equalize, identity, posterize, rotate, sharpness, shearX, shearY, solarize, translateX, and TranslateY. Autocontrast, brightness, contrast, and equalize are all severe image intensity modifications.} An unlabeled image $x^c_u$ is first weakly augmented, $x^c_w = \text{WAug}(x^c_u)$, and a pseudo-label $c_p = \text{arg\,max}(\hat{c}_w) \ge t$ is synthesized from $x^c_w$ using the model prediction $\hat{c}_w$. The image-label pair is retained only if the confidence with which the model generates the pseudo-label exceeds the experimentally tuned threshold $t$, thus deterring learning from poor and incorrect labels. Second, $x^c_g = \text{GAug}(x^c_u)$ are strongly augmented versions of $x^c_u$.

Our training strategy promotes effective learning from large amounts of unlabeled data, which is challenging. At first, the predictions are less reliable as the model begins to learn mainly from the labeled data, but the model gains confidence with the generation of labels for the unlabeled images and, as a result, it becomes more proficient. Since the unlabeled examples are incrementally added to the training set, subject to the threshold, the model learns to predict more accurately in a progressive manner and, with increasing confidence, the performance of the model improves at an increasingly higher rate. Furthermore, employing two degrees of augmentation enables the model to maximize its knowledge gain from the unlabeled data due to the enhanced image diversity through what is known as consistency learning as, in theory, two augmented versions of the same image should yield the same prediction, which is encouraged using an unsupervised loss. In other words, imposing on the model, through an unsupervised loss, to produce the same predictions on images subjected to two different degrees of augmentation results in better classification performance.

The classification objective
\begin{equation}
\label{eqn:class-loss}
       L^c(c_l, \hat{c}_l, c_p, \hat{c}_g) = L_l(c_l, \hat{c}_l) + \lambda L_u(c_p, \hat{c}_g)
\end{equation}
includes a supervised loss component $L_l$ for the labeled data, which uses cross-entropy between the reference class label $c_l$ and the model prediction $\hat{c}_l$, as well as an unsupervised loss component $L_u$ for the unlabeled data, which uses cross-entropy between the pseudo-label $c_p$ and the model prediction $\hat{c}_g$. 

Note that the model is trained to ignore GAug as it is provided the pseudo-label $c_p$. Since the underlying data distributions are the same for both augmentations, it is compelled to learn that for the sake of consistency. Weak augmentations are used to produce reliable and usable pseudo-labels whereas strong augmentations are used to provide a difficult challenge for the model. This difficulty forces the model to learn more effective representations in order to be accurate, and it also prevents overfitting from minimizing the loss too early. With the assumption that the weakly augmented image has the correct label to be associated with the strongly augmented image, the model is empowered to discern the augmentations in the image, and its performance improves as a result, by learning the underlying features crucial to the diagnosis. This helps achieve better generalization despite the differences in data distributions across different domains. By teaching the model to learn only the more salient representations that will exist to some extent in all domains, it can generalize and be effective across domains.

\subsection{Segmentation Branch}

For segmentation, the predictions are made through the encoder-decoder architecture with skip connections. For the labeled samples $x^s_l$, we calculate the direct segmentation loss in the form of Dice loss $L_{l}(s_l,\hat{s}_l)$ between the reference lung mask $s_l$ and predicted segmentation $\hat{s}_l$. Since we do not have the segmentation masks for the unlabeled examples $x^s_u$, we cannot directly calculate the segmentation loss for them. To ensure consistency, we compute the KL divergence $L_{u}(\hat{s}_l, \hat{s}_u)$ between segmentation predictions for the labeled examples and unlabeled examples $\hat{s}_u$. This penalizes the model for making predictions that increasingly differ from those of the labeled data, which helps the model fit the unlabeled data. The total segmentation objective is therefore
\begin{equation}
\label{eqn:seg-loss}
       L^s(s_l, \hat{s}_l, \hat{s}_u) =  \alpha L_{l}(s_l, \hat{s}_l) + \beta L_{u}(\hat{s}_l, \hat{s}_u),
\end{equation}
where $\alpha$ and $\beta$ are weights.

\subsection{Saliency Bridge Module}

We incorporate a saliency bridge module to bridge between the classification and segmentation branches of the MultiMix model, as indicated in Figure~\ref{fig:model}. To learn which image regions are most relevant to classification, saliency maps 
\begin{equation}
\label{eqn:saliency}
y_l = \text{Saliency}(\hat{c}^s_l) \quad\text{and}\quad
y_u = \text{Saliency}(\hat{c}^s_u),
\end{equation}
where $\hat{c}^s_l$ and $\hat{c}^s_u$ denote the class predictions for the input images $x^s_l$ and $x^s_u$, respectively,
are generated from the classification branch by computing the gradient of the predicted class with respect to the input image.\footnote{These saliency maps should not be confused with simultaneous segmentation and saliency detection or prediction, where a semantic segmentation model is trained to produce saliency maps to accompany the output segmentation masks; e.g., \citep{zeng2019joint}. Our saliency bridge module is novel in that it performs a saliency analysis of MultiMix's classification branch and leverages it to improve the performance of its semantic segmentation branch.} It cannot be directly known if the image samples in $\mathcal{D}^s$ represent normal or diseased cases, thus $x^s_l$ and $x^s_u$ are considered to be unlabeled for the classification task. Therefore, the saliency maps generated via the class prediction are not true segmentation maps, but they will nonetheless highlight the lungs or lung regions relevant to the particular disease class (see Appendix~\ref{app:saliency-vis}).

The outputs of the saliency bridge module,
\begin{equation}
    \label{eqn:sal_out}
    b_l = y_l \oplus x^s_l \quad\text{and}\quad
    b_u = y_u \oplus x^s_u,
\end{equation}
obtained by concatenating the saliency maps with the associated input images, are further downsampled before they are concatenated with the encoder-decoder bottleneck in the segmentation branch. This results in a tighter connection between the classification and segmentation tasks and improves the effectiveness of the bridge module, which retains important information from the encoder that may otherwise be lost because of the repeated convolutions. The saliency maps serve to guide the segmentation during the decoding phase, yielding improved segmentation while learning from limited labeled data. With improving classification performance, the saliency maps become more accurate, thus yielding improved segmentations, since the shared parameters responsible for improved classification produce a feedback loop that allows both tasks to improve jointly. 

Conventionally, saliency maps are used to analyze which features and areas of the image are most relevant for classification, thereby enhancing understanding of the model's learning process. Similarly, our saliency module is explainable, as it is a relevant connection between the classification and segmentation tasks (although model explainability in and of itself is not the main focus of our work). Since the saliency maps are comparable to segmentation masks, it is sensible to employ them to guide the decoder in the task of segmentation. Multi-tasking requires the tasks to be somewhat related, so our task-relevant bridge fosters a tighter bond between classification and segmentation.

\subsection{MultiMix Training Procedure}

\def\batch#1{{}^{#1}\!}

\begin{algorithm}[t]
\caption{MultiMix Mini-Batch Training}
\label{alg:train}
\begin{algorithmic}
\REQUIRE 
\STATE Training set of labeled classification data $\mathcal{D}^c_l$
\STATE Training set of labeled segmentation data $\mathcal{D}^s_l$
\STATE Training set of unlabeled classification data $\mathcal{D}^c_u$ 
\STATE Training set of unlabeled segmentation data $\mathcal{D}^s_u$
\STATE Network architecture $\mathcal{F}_\theta$ with learnable parameters $\theta$
\STATE Minibatch size $m$
\REPEAT
\STATE Create labeled classification minibatch:
$\{\batch{1}x^c_l,\dots,\batch{m}x^c_l\} \sim\mathcal{D}^c_l$ 
\STATE Create labeled segmentation minibatch:
$\{\batch{1}x^s_l,\dots,\batch{m}x^s_l\} \sim\mathcal{D}^s_l$ 
\STATE Create unlabeled classification minibatch:
$\{\batch{1}x^c_u,\dots,\batch{m}x^c_u\} \sim\mathcal{D}^c_u$ 
\STATE Create unlabeled segmentation minibatch:
$\{\batch{1}x^s_u,\dots,\batch{m}x^s_u\} \sim\mathcal{D}^s_u$
\vspace{3pt}
\STATE Compute predictions for the labeled data: $\batch{i}\hat{c}_l \leftarrow \mathcal F_{\theta}(\batch{i}x^c_l); ~ \batch{i}\hat{s}_l \leftarrow \mathcal F_{\theta}(\batch{i}x^s_l)$
\STATE Generate weakly-augmented samples:
$\batch{i}x^c_w \leftarrow \text{WAug}(\batch{i}x^c_u)$ 
\STATE Generate strongly-augmented samples: 
$\batch{i}x^c_g \leftarrow \text{GAug}(\batch{i}x^c_u)$
\STATE Compute predictions for the unlabeled data: $\batch{i}\hat{c}_{w} \leftarrow \mathcal F_{\theta}(\batch{i}x^c_w); ~ \batch{i}\hat{c}_{g} \leftarrow \mathcal F_{\theta}(\batch{i}x^c_g); ~ \batch{i}\hat{s}_u \leftarrow \mathcal F_{\theta}(\batch{i}x^s_u)$
\vspace{-14pt}
\STATE Compute pseudo label: $\batch{i}c_p \leftarrow \text{arg\,max}(\batch{i}\hat{c}_w) \ge t$
\smallskip
\STATE Update $\mathcal{F}_\theta$ by backpropagating the loss gradient $\nabla\!_\theta L$
\UNTIL{convergence}
\end{algorithmic}
\end{algorithm}

Algorithm~\ref{alg:train} presents the main steps of the MultiMix training procedure applied to labeled and unlabeled classification and segmentation training data. The model is trained simultaneously on the classification objective (\ref{eqn:class-loss}) and segmentation objective (\ref{eqn:seg-loss}) using the following total loss for a minibatch size of $m$:
\begin{equation}
    L = \frac{1}{m} \sum_{i=1}^m \biggl(L^c\left(\batch{i}c_l, \batch{i}\hat{c}_l, \batch{i}c_p, \batch{i}\hat{c}_g\right) + L^s\left(\batch{i}s_l, \batch{i}\hat{s}_l, \batch{i}\hat{s}_u\right)\biggr)
\end{equation}

\section{Experimental Evaluation}
\label{sec:exp}

\subsection{Data}

\begin{figure} \centering
\subcaptionbox{\label{fig:class-images}}{
  \begin{tabular}{cc}
  Normal & Abnormal \\
  \multicolumn{2}{c}{CheX} \\
  \includegraphics[width=0.2\linewidth, trim={4cm 1cm 4cm 2cm},clip]{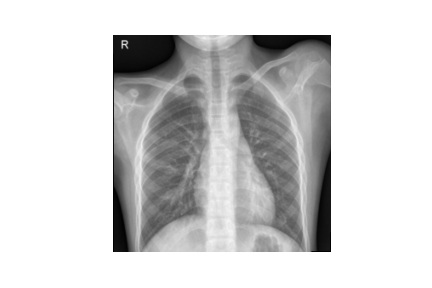} &
  \includegraphics[width=0.2\linewidth, trim={4cm 1cm 4cm 2cm},clip]{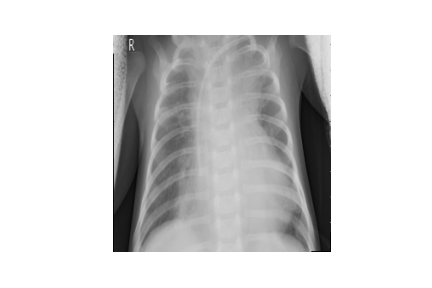}\\
  \multicolumn{2}{c}{NIHX} \\
  \includegraphics[width=0.2\linewidth, trim={4cm 1cm 4cm 2cm},clip]{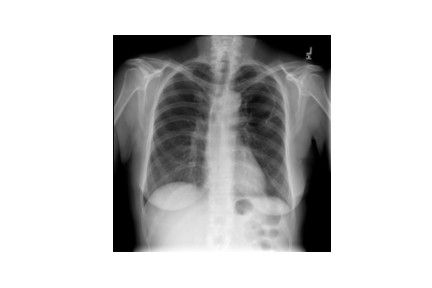} &
  \includegraphics[width=0.2\linewidth, trim={4cm 1cm 4cm 2cm},clip]{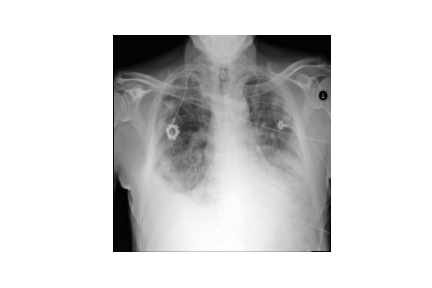}
\end{tabular}} 
\hfill
\subcaptionbox{\label{fig:distributions}}{\includegraphics[width=0.48\linewidth, trim={70 35 60 0}, clip]{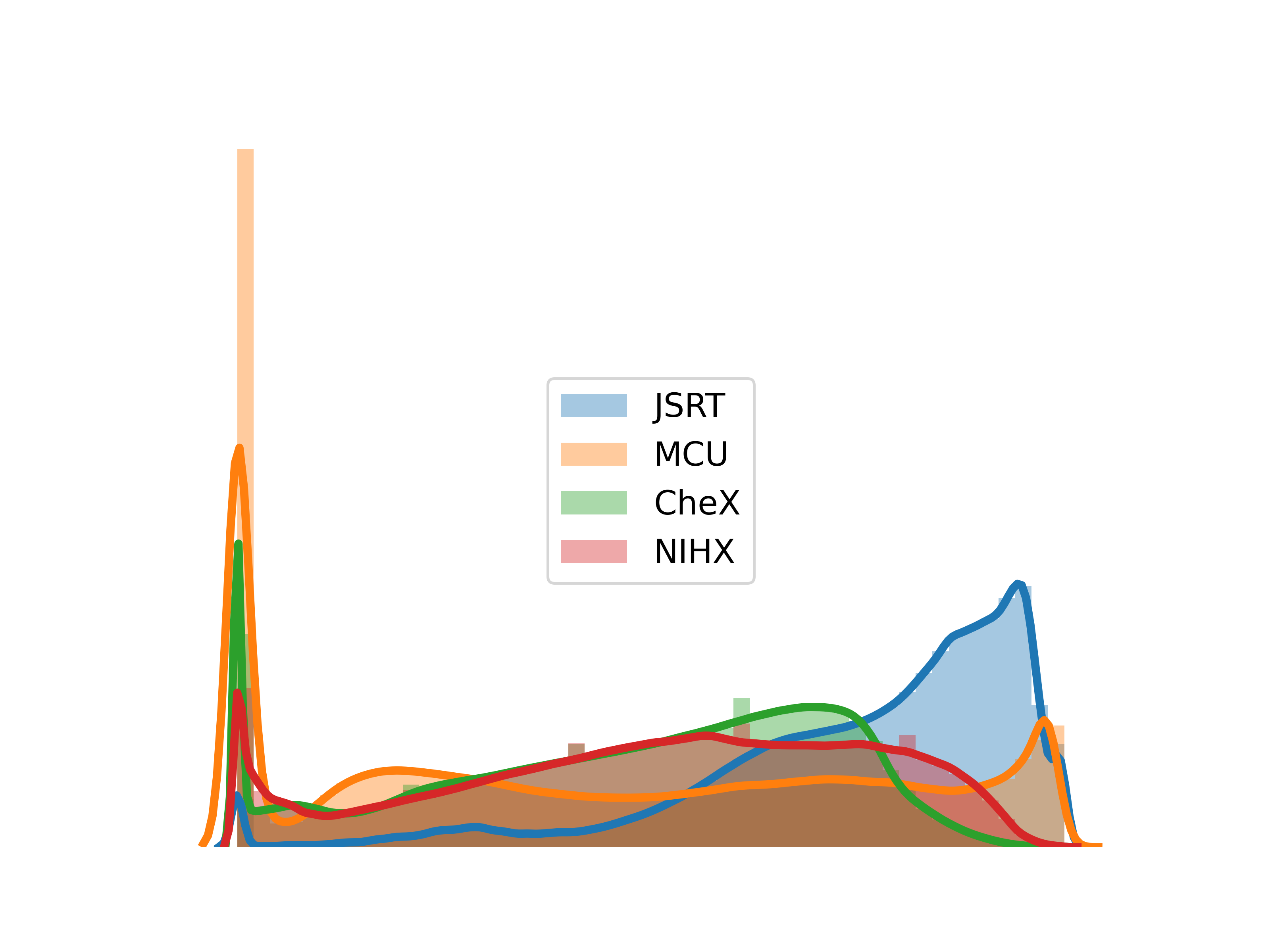}}
\caption{(a) Sample (normal, abnormal) images from the CheX and NIHX datasets. (b) Intensity distributions of the four chest X-ray image datasets.}
\end{figure}

Models were trained and tested in the combined classification and segmentation tasks using chest X-ray images from two different sources: pneumonia detection (CheX) \citep{kermany2018identifying} and the Japanese Society of Radiological Technology (JSRT) \citep{shiraishi2000development}. We further validated the models using the Montgomery County chest X-rays (MCU) \citep{jaeger2014two} and a subset of the NIH chest X-ray dataset (NIHX) \citep{wang2017chestx} (Figure~\ref{fig:class-images}). Table~\ref{table:data} presents some details about the datasets used in our experiments. In addition to the diversity in the source, image quality, size, and proportion of normal and abnormal images, the disparity in the intensity distributions of the four datasets is also evident (Figure~\ref{fig:distributions}). All the images were normalized and resized to $256\times256\times1$ before passing them to the models.

\begin{table}
\setlength{\tabcolsep}{4pt}
\centering
\caption{Details of the datasets used for training and testing.}
\label{table:data}
\begin{tabular}{@{} lc lc rrr c rrr @{}}
\toprule
Mode & \phantom{a} & Dataset & \phantom{a} & Total & Normal & Abnormal & \phantom{a} & Train & Val & Test\\
\midrule
\multirow{2}{*}{in-domain} && JSRT && 247 & -- & -- && 111 & 13 & 123\\
&& CheX && 5,856 & 1583 & 4273 && 5216 & 16 & 624\\
\midrule
\multirow{2}{*}{cross-domain} && MCU && 138 & -- & -- && 93 & 10 & 35\\
&& NIHX && 4185 & 2754 & 1431 && -- & -- & 4185\\
\bottomrule
\end{tabular}
\end{table}

\subsection{Implementation Details}
\label{sec:implementation-details}

\paragraph{Baselines:} We used the U-Net and encoder-only (Enc) networks separately for the single-task baseline models in both the fully supervised and semi-supervised schemes. Using the same backbone network, we also trained a multi-tasking U-Net with the classification branch (UMTL). All these models incorporate an INorm, LReLU, and dropout at every convolutional block (see Appendix~\ref{app:architecture}). Moreover, we performed ablation experiments to assess the impact of each key piece of our MultiMix model: single-task Enc-SSL (encoder with confidence-based augmentation SSL), single-task Enc-MM (an implementation of MixMatch \citep{berthelot2019mixmatch}), UMTL-S (UMTL with saliency bridge), UMTL-SSL (UMTL with SSL classification), and UMTL-SSL-S (UMTL with saliency bridge and confidence-based augmentation). 

\paragraph{Augmentations:} We performed random horizontal flip and $32\times32$ crop in WAug for the examples in $\mathcal{D}^c_u$. On the other hand, GAug was applied through a random combination from the pool of augmentations: random horizontal flip, crop ($32\times32$), autocontrast, brightness, contrast, equalize, identity, posterize, rotate ($30^{\circ}$), sharpness, shearX, shearY, solarize, translateX (30\%), and translateY (30\%). 

\paragraph{Training:} All the models (single-task or multi-task) were trained on varying $|\mathcal{D}^s_l|$ (10, 50, full), and $|\mathcal{D}^c_l|$ (100, 1000, full). Each experiment was repeated 5 times and the average performance is reported. We implemented the models using Python and the PyTorch framework and trained using an Nvidia K80 GPU. 

\paragraph{Hyper-parameters:} We used the Adam optimizer with adaptive learning rates of 0.1 every 8 epochs and an initial learning rate of 0.0001. A negative slope of 0.2 was applied to Leaky ReLU, and the dropout was set to 0.25. We set $t=0.7$, $\lambda=0.25$, $\alpha=5.0$ (for smaller $|\mathcal{D}^s_l|$) and $\beta=0.01$. Each model was trained with a mini-batch size of $m=10$. All model-specific hyperparameters were experimentally tuned. We found that the performance of the model varied only minimally subject to the different choices. 

\paragraph{Evaluation:} For classification, along with the overall accuracy (Acc), we recorded the class-wise F1 scores (F1-N for normal and F1-P for pneumonia). To evaluate segmentation performance, we used the Dice similarity (DS), Jaccard similarity (JS), structural similarity measure (SSIM), average Hausdorff distance (HD), precision (P), and Recall (R) scores.

\begin{figure} \centering
\subcaptionbox{in-domain}{\includegraphics[width=0.49\linewidth]{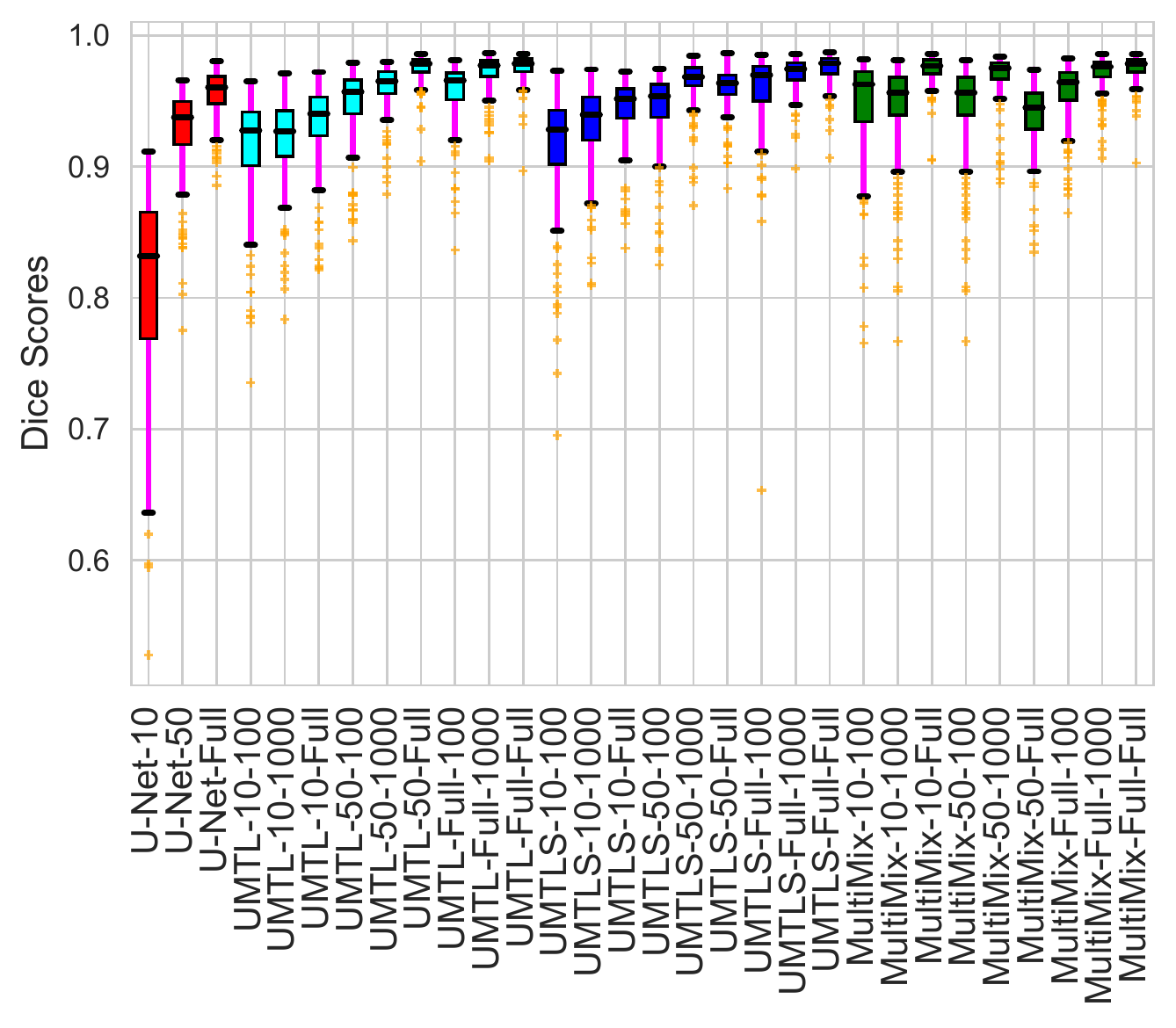}} \hfill
\subcaptionbox{cross-domain}{\includegraphics[width=0.49\linewidth]{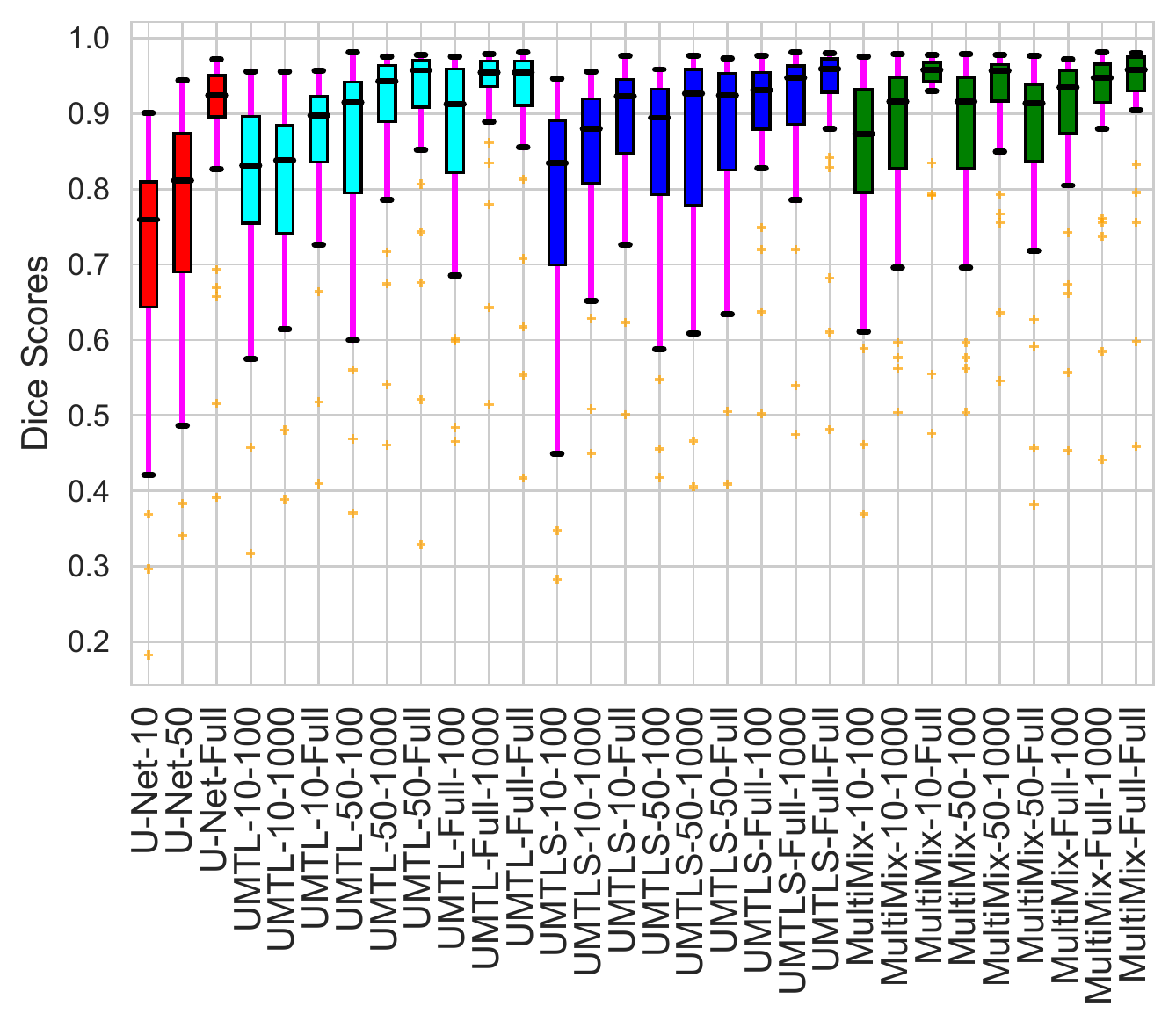}}
\caption{Distributions of the Dice scores demonstrate the superiority of the MultiMix model over the baseline models in segmenting lungs from the chest X-ray images in both domains.}
\label{fig:consistency}
\end{figure}

\begin{table}
\centering
\setlength{\tabcolsep}{1pt}
\caption{Classification and segmentation performance figures with varying label proportions in in-domain evaluations: CheX (classification) and JSRT (segmentation) datasets. The best scores from fully-supervised models are underlined and the best scores from semi-supervised models are bolded. Scores are given as $\text{mean}\pm\text{std}$.}
\label{table:results-in}
\resizebox{\linewidth}{!}{
\begin{tabular}{@{} lc cc ccc c cc cccccc @{}}
            \toprule
          \multirow{2}{*}{Model}
           &
           \phantom{a}
           &
           \multirow{2}{*}{$|\mathcal{D}^c_l|$}
           &
           \phantom{a}
           &
           \multicolumn{3}{c}{Classification}
           &
           \phantom{a}
           &
           \multirow{2}{*}{$|\mathcal{D}^s_l|$}
           &
           \phantom{a}
           &
          \multicolumn{6}{c}{Segmentation}
          \\
          \cmidrule{5-7} \cmidrule{11-16}
          &&&&
          Acc & F1-Nor & F1-Abn
          &&&&
          DS & JS & SSIM & HD & P & R\\
          \midrule
          \multirow{3}{*}{\rotatebox{45}{U-Net}} 
           &&
           ---
           &&
           --- & --- & --- 
           &&
           10
           &&
           $0.634\pm0.017$ & $0.695\pm0.024$ & $0.810\pm0.019$ & $2.899\pm0.272$ & $0.779\pm0.021$ & $0.865\pm0.023$ 
           \\
           &&
           ---
           &&
           --- & --- & --- 
           &&
           50
           &&
           $0.855\pm0.004$ & $0.854\pm0.008$ & $0.904\pm0.003$ & $0.341\pm0.071$ & $0.918\pm0.009$ & $0.925\pm0.009$ 
           \\
           &&
           ---
           &&
           --- & --- & --- 
           &&
           Full
           &&
           $0.915\pm0.001$ & $0.906\pm0.002$ & $0.929\pm0.002$ & $0.104\pm0.025$ & $0.949\pm0.007$ & $0.953\pm0.005$ 
           \\
          \midrule
          \multirow{3}{*}{\rotatebox{45}{Enc}} 
           &&
           100
           &&
           $0.732\pm0.044$ & $0.424\pm0.122$ & $0.806\pm0.026$ 
           &&
           ---
           &&
           --- & --- & --- & --- & --- & ---
           \\
           &&
           1000
           &&
           $0.773\pm0.037$ & $0.546\pm0.020$ & $0.842\pm0.018$ 
           &&
           ---
           &&
           --- & --- & --- & --- & --- & ---
           \\
           &&
           Full
           &&
           $0.737\pm0.021$ & $0.534\pm0.058$ & $0.838\pm0.012$ 
           &&
           ---
           &&
           --- & --- & --- & --- & --- & ---
           \\
        \midrule
          \multirow{3}{*}{\rotatebox{45}{Enc-MM}} 
           &&
           100
           &&
           $0.738\pm0.043$ & $0.452\pm0.103$ & $0.800\pm0.024$ 
           &&
           ---
           &&
           --- & --- & --- & --- & --- & ---
           \\
           &&
           1000
           &&
           $0.745\pm0.036$ & $0.560\pm0.078$ & $0.584\pm0.101$ 
           &&
           ---
           &&
           --- & --- & --- & --- & --- & ---
           \\
           &&
           Full
           &&
           $0.751\pm0.036$ & $0.605\pm0.065$ & $0.846\pm0.025$ 
           &&
           ---
           &&
           --- & --- & --- & --- & --- & ---
           \\
           \midrule
          \multirow{3}{*}{\rotatebox{45}{Enc-SSL}} 
           &&
           100
           &&
           $0.780\pm0.035$ & $0.570\pm0.083$ & $0.844\pm0.021$ 
           &&
           ---
           &&
           --- & --- & --- & --- & --- & ---
           \\
           &&
           1000
           &&
           $0.822\pm0.027$ & $0.692\pm0.058$ & $0.876\pm0.015$ 
           &&
           ---
           &&
           --- & --- & --- & --- & --- & ---
           \\
           &&
           Full
           &&
           $0.817\pm0.016$ & $0.680\pm0.042$ & $0.872\pm0.013$ 
           &&
           ---
           &&
           --- & --- & --- & --- & --- & ---
           \\
           \midrule
          \multirow{9}{*}{\rotatebox{45}{UMTL}} 
           &&
           100
           &&
           $0.707\pm0.024$ & $0.443\pm0.071$ & $0.797\pm0.016$ 
           &&
           10
           &&
           $0.626\pm0.008$ & $0.871\pm0.014$ & $0.908\pm0.010$ & $4.323\pm1.268$ & $0.900\pm0.014$ & $0.964\pm0.007$ 
           \\
           &&
           100
           &&
           $0.655\pm0.052$ & $0.683\pm0.129$ & $0.853\pm0.030$ 
           &&
           50
           &&
           $0.647\pm0.022$ & $0.854\pm0.040$ & $0.881\pm0.025$ & $4.733\pm1.238$ & $0.864\pm0.043$ & $0.989\pm0.004$ 
           \\
           &&
           100
           &&
           $0.706\pm0.046$ & $0.416\pm0.138$ & $0.804\pm0.028$ 
           &&
           Full
           &&
           $0.696\pm0.014$ & $0.872\pm0.0252$ & $0.911\pm0.016$ & $3.908\pm0.795$ & $0.892\pm0.025$ & $0.986\pm0.004$ 
           \\
           &&
           1000
           &&
           $0.750\pm0.010$ & $0.490\pm0.020$ & $0.825\pm0.005$ 
           &&
           10
           &&
           $0.761\pm0.004$ & $0.904\pm0.009$ & $0.926\pm0.002$ & $3.050\pm0.531$ & $0.924\pm0.001$ & $0.977\pm0.009$ 
           \\
           &&
           1000
           &&
           $0.749\pm0.024$ & $0.510\pm0.064$ & $0.833\pm0.009$ 
           &&
           50
           &&
           $0.768\pm0.001$ & $0.927\pm0.003$ & $0.938\pm0.002$ & $2.606\pm0.205$ & $0.940\pm0.004$ & $0.985\pm0.001$ 
           \\
           &&
           1000
           &&
           $0.747\pm0.045$ & $0.530\pm0.140$ & $0.840\pm0.024$ 
           &&
           Full
           &&
           $0.759\pm0.005$ & $0.928\pm0.010$ & $0.930\pm0.008$ & $2.955\pm0.483$ & $0.924\pm0.015$ & $0.981\pm0.006$ 
           \\
           &&
           Full
           &&
           $0.744\pm0.011$ & $0.515\pm0.022$ & $0.828\pm0.016$ 
           &&
           10
           &&
           $0.909\pm0.021$ & $0.919\pm0.037$ & $0.521\pm0.028$ & $0.903\pm0.296$ & $0.912\pm0.050$ & $0.962\pm0.013$ 
           \\
           &&
           Full
           &&
           $0.738\pm0.004$ & $0.438\pm0.013$ & $0.820\pm0.000$ 
           &&
           50
           &&
           $0.930\pm0.000$ & $0.948\pm0.000$ & $0.954\pm0.000$ & $0.444\pm0.142$ & $0.969\pm0.001$ & $0.977\pm0.001$ 
           \\
           &&
           Full
           &&
           $0.731\pm0.018$ & $0.447\pm0.067$ & $0.822\pm0.009$ 
           &&
           Full
           &&
           $0.932\pm0.000$ & $0.951\pm0.000$ & $0.957\pm0.000$ & $\underline{0.372}\pm0.052$ & $0.965\pm0.001$ & $0.977\pm0.001$ 
           \\
           \midrule
           \multirow{9}{*}{\rotatebox{45}{UMTL-S}} 
           &&
           100
           &&
           $0.704\pm0.052$ & $0.358\pm0.223$ & $0.806\pm0.024$ 
           &&
           10
           &&
           $0.922\pm0.007$ & $0.848\pm0.013$ & $0.891\pm0.009$ & $4.005\pm0.413$ & $0.871\pm0.017$ & $0.966\pm0.005$ 
           \\
           &&
           100
           &&
           $0.701\pm0.033$ & $0.336\pm0.130$ & $0.796\pm0.019$ 
           &&
           50
           &&
           $0.926\pm0.002$ & $0.867\pm0.003$ & $0.894\pm0.003$ & $4.393\pm0.217$ & $0.873\pm0.005$ & $0.891\pm0.002$ 
           \\
           &&
           100
           &&
           $0.713\pm0.041$ & $0.442\pm0.164$ & $0.794\pm0.025$ 
           &&
           Full
           &&
           $0.931\pm0.003$ & $0.890\pm0.006$ & $0.920\pm0.003$ & $3.983\pm0.375$ & $0.906\pm0.008$ & $0.980\pm0.004$ 
           \\
           &&
           1000
           &&
           $0.740\pm0.020$ & $0.482\pm0.052$ & $0.828\pm0.012$ 
           &&
           10
           &&
           $0.948\pm0.001$ & $0.908\pm0.003$ & $0.924\pm0.003$ & $2.546\pm0.217$ & $0.931\pm0.005$ & $0.972\pm0.002$ 
           \\
           &&
           1000
           &&
           $0.771\pm0.041$ & $0.566\pm0.010$ & $0.844\pm0.024$ 
           &&
           50
           &&
           $0.965\pm0.001$ & $0.931\pm0.003$ & $0.941\pm0.001$ & $2.083\pm0.217$ & $0.949\pm0.005$ & $0.981\pm0.002$ 
           \\
           &&
           1000
           &&
           $0.742\pm0.019$ & $0.497\pm0.059$ & $0.830\pm0.014$ 
           &&
           Full
           &&
           $0.962\pm0.005$ & $0.925\pm0.010$ & $0.935\pm0.008$ & $1.758\pm0.132$ & $0.958\pm0.015$ & $0.985\pm0.005$ 
           \\
           &&
           Full
           &&
           $0.747\pm0.006$ & $0.500\pm0.021$ & $0.830\pm0.006$ 
           &&
           10
           &&
           $0.955\pm0.020$ & $0.914\pm0.035$ & $0.936\pm0.027$ & $0.568\pm0.136$ & $0.954\pm0.037$ & $0.956\pm0.005$ 
           \\
           &&
           Full
           &&
           $0.737\pm0.016$ & $0.433\pm0.054$ & $0.820\pm0.008$ 
           &&
           50
           &&
           $0.972\pm0.006$ & $0.944\pm0.011$ & $0.953\pm0.009$ & $0.560\pm0.427$ & $0.966\pm0.014$ & $0.977\pm0.004$ 
           \\
           &&
           Full
           &&
           $0.723\pm0.005$ & $0.413\pm0.019$ & $0.817\pm0.005$ 
           &&
           Full
           &&
           $0.974\pm0.000$ & $0.953\pm0.000$ & $0.957\pm0.000$ & $0.539\pm0.437$ & $0.967\pm0.002$ & $0.981\pm0.001$ 
           \\
            \midrule
            \multirow{9}{*}{\rotatebox{45}{UMTL-SSL}} 
           &&
           100
           &&
           $0.790\pm0.043$ & $0.618\pm0.105$ & $0.856\pm0.024$ 
           &&
           10
           &&
           $0.906\pm0.002$ & $0.925\pm0.004$ & $0.940\pm0.002$ & $0.626\pm0.280$ & $0.954\pm0.006$ & $0.953\pm0.003$ 
           \\
           &&
           100
           &&
           $0.818\pm0.039$ & $0.688\pm0.087$ & $0.872\pm0.024$ 
           &&
           50
           &&
           $0.919\pm0.001$ & $0.946\pm0.001$ & $0.952\pm0.001$ & $0.561\pm0.115$ & $0.962\pm0.003$ & $0.963\pm0.002$ 
           \\
           &&
           100
           &&
           $0.852\pm0.039$ & $0.670\pm0.095$ & $0.868\pm0.022$ 
           &&
           Full
           &&
           $0.937\pm0.001$ & $0.954\pm0.001$ & $0.958\pm0.001$ & $0.613\pm0.386$ & $0.969\pm0.004$ & $0.981\pm0.002$
           \\
           &&
           1000
           &&
           $0.794\pm0.020$ & $0.630\pm0.046$ & $0.860\pm0.012$ 
           &&
           10
           &&
           $0.893\pm0.000$ & $0.926\pm0.001$ & $0.941\pm0.001$ & $0.524\pm0.107$ & $0.961\pm0.002$ & $0.962\pm0.001$ 
           \\
           &&
           1000
           &&
           $0.822\pm0.038$ & $0.693\pm0.096$ & $0.877\pm0.026$ 
           &&
           50
           &&
           $0.903\pm0.000$ & $0.945\pm0.000$ & $0.952\pm0.001$ & $0.712\pm0.167$ & $0.963\pm0.002$ & $0.980\pm0.002$ 
           \\
           &&
           1000
           &&
           $0.818\pm0.005$ & $0.707\pm0.019$ & $0.867\pm0.005$ 
           &&
           Full
           &&
           $0.899\pm0.001$ & $0.953\pm0.001$ & $0.958\pm0.001$ & $0.724\pm0.400$ & $0.968\pm0.005$ & $0.982\pm0.003$ 
           \\
           &&
           Full
           &&
           $0.812\pm0.022$ & $0.688\pm0.050$ & $0.870\pm0.012$ 
           &&
           10
           &&
           $0.905\pm0.005$ & $0.921\pm0.008$ & $0.935\pm0.004$ & $0.627\pm0.150$ & $0.946\pm0.014$ & $0.973\pm0.009$ 
           \\
           &&
           Full
           &&
           $0.813\pm0.012$ & $0.683\pm0.020$ & $0.873\pm0.008$ 
           &&
           50
           &&
           $0.927\pm0.001$ & $0.947\pm0.001$ & $0.954\pm0.001$ & $\textbf{0.397}\pm0.172$ & $0.968\pm0.001$ & $0.977\pm0.001$ 
           \\
           &&
           Full
           &&
           $0.816\pm0.008$ & $0.678\pm0.019$ & $0.873\pm0.004$ 
           &&
           Full
           &&
           $0.935\pm0.001$ & $\underline{0.954}\pm0.001$ & $0.958\pm0.001$ & $0.625\pm0.208$ & $\underline{0.970}\pm0.002$ & $0.981\pm0.001$ 
           \\
            \midrule
            \multirow{9}{*}{\rotatebox{45}{UMTL-SSL-S}} 
           &&
           100
           &&
           $0.798\pm0.030$ & $0.628\pm0.081$ & $0.860\pm0.018$ 
           &&
           10
           &&
           $0.951\pm0.004$ & $0.911\pm0.008$ & $0.935\pm0.004$ & $0.792\pm0.313$ & $0.940\pm0.006$ & $0.963\pm0.006$ 
           \\
           &&
           100
           &&
           $0.834\pm0.033$ & $0.696\pm0.074$ & $0.874\pm0.019$ 
           &&
           50
           &&
           $0.972\pm0.001$ & $0.946\pm0.001$ & $0.952\pm0.001$ & $0.727\pm0.340$ & $0.965\pm0.002$ & $0.977\pm0.002$ 
           \\
           &&
           100
           &&
           $0.817\pm0.036$ & $0.688\pm0.100$ & $0.860\pm0.021$ 
           &&
           Full
           &&
           $0.975\pm0.001$ & $0.951\pm0.001$ & $0.954\pm0.001$ & $0.812\pm0.315$ & $0.968\pm0.002$ & $0.981\pm0.002$ 
           \\
           &&
           1000
           &&
           $0.806\pm0.020$ & $0.652\pm0.055$ & $0.872\pm0.014$ 
           &&
           10
           &&
           $0.956\pm0.002$ & $0.916\pm0.004$ & $0.937\pm0.003$ & $0.852\pm0.275$ & $0.943\pm0.005$ & $0.966\pm0.003$ 
           \\
           &&
           1000
           &&
           $0.808\pm.013$ & $0.662\pm0.038$ & $0.862\pm0.010$ 
           &&
           50
           &&
           $0.971\pm0.000$ & $0.944\pm0.001$ & $0.952\pm0.000$ & $0.917\pm0.239$ & $0.965\pm0.002$ & $0.978\pm0.003$ 
           \\
           &&
           1000
           &&
           $0.801\pm0.020$ & $0.646\pm0.049$ & $0.862\pm0.010$ 
           &&
           Full
           &&
           $0.975\pm0.001$ & $0.952\pm0.001$ & $0.954\pm0.001$ & $0.753\pm0.228$ & $0.969\pm0.001$ & $0.981\pm0.001$ 
           \\
           &&
           Full
           &&
           $0.796\pm0.033$ & $0.632\pm0.086$ & $0.864\pm0.018$ 
           &&
           10
           &&
           $0.960\pm0.002$ & $0.923\pm0.004$ & $0.940\pm0.002$ & $0.782\pm0.229$ & $0.954\pm0.005$ & $0.967\pm0.003$ 
           \\
           &&
           Full
           &&
           $0.808\pm0.014$ & $0.662\pm0.030$ & $0.868\pm0.007$ 
           &&
           50
           &&
           $0.972\pm0.001$ & $0.945\pm0.001$ & $0.953\pm0.001$ & $0.645\pm0.196$ & $0.966\pm0.003$ & $0.978\pm0.003$ 
           \\
           &&
           Full
           &&
           $0.800\pm0.016$ & $0.632\pm0.038$ & $0.628\pm0.009$ 
           &&
           Full
           &&
           $0.961\pm0.008$ & $0.924\pm0.016$ & $0.940\pm0.009$ & $0.392\pm0.337$ & $0.948\pm0.014$ & $0.969\pm0.007$ 
           \\
            \midrule
            \multirow{9}{*}{\rotatebox{45}{MultiMix}} 
           &&
           100
           &&
           $0.800\pm0.025$ & $0.594\pm0.064$ & $0.856\pm0.015$ 
           &&
           10
           &&
           $0.954\pm0.004$ & $0.920\pm0.008$ & $0.938\pm0.004$ & $0.695\pm0.198$ & $0.949\pm0.010$  & $0.969\pm0.007$ 
           \\
           &&
           100
           &&
           $0.824\pm0.022$ & $0.613\pm0.056$ & $0.854\pm0.014$ 
           &&
           50
           &&
           $0.971\pm0.001$ & $0.943\pm0.002$ & $0.951\pm0.001$ & $0.681\pm0.086$ & $0.964\pm0.003$ & $0.976\pm0.002$ 
           \\
           &&
           100
           &&
           $0.792\pm0.035$ & $0.593\pm0.101$ & $0.854\pm0.021$ 
           &&
           Full
           &&
           $0.973\pm0.012$ & $0.948\pm0.022$ & $0.954\pm0.015$ & $0.636\pm0.070$ & $0.966\pm0.025$ & $0.981\pm0.004$ 
           \\
           &&
           1000
           &&
           $0.817\pm0.016$ & $0.647\pm0.038$ & $0.865\pm0.006$ 
           &&
           10
           &&
           $0.954\pm0.004$ & $0.910\pm0.008$ & $0.932\pm0.004$ & $0.902\pm0.186$ & $0.942\pm0.005$ & $0.968\pm0.007$ 
           \\
           &&
           1000
           &&
           $0.825\pm0.016$ & $0.650\pm0.033$ & $0.860\pm0.011$ 
           &&
           50
           &&
           $0.970\pm0.001$ & $0.941\pm0.002$ & $0.950\pm0.001$ & $0.811\pm0.112$ & $0.964\pm0.004$ & $0.977\pm0.002$ 
           \\
           &&
           1000
           &&
           $0.830\pm0.048$ & $0.586\pm0.138$ & $0.856\pm0.029$ 
           &&
           Full
           &&
           $\textbf{0.974}\pm0.011$ & $0.919\pm0.020$ & $0.953\pm0.014$ & $0.643\pm0.126$ & $0.933\pm0.024$ & $0.984\pm0.004$ 
           \\
           &&
           Full
           &&
           $0.840\pm0.025$ & $0.730\pm0.060$ & $0.880\pm0.016$ 
           &&
           10
           &&
           $0.954\pm0.002$ & $0.913\pm0.004$ & $0.935\pm0.001$ & $0.621\pm0.123$ & $0.949\pm0.006$ & $0.968\pm0.006$ 
           \\
           &&
           Full
           &&
           $\textbf{0.854}\pm0.022$ & $\textbf{0.760}\pm0.055$ & $\textbf{0.890}\pm0.014$ 
           &&
           50
           &&
           $0.972\pm0.001$ & $\textbf{0.950}\pm0.001$ & $\textbf{0.956}\pm0.001$ & $0.692\pm0.036$ & $\textbf{0.970}\pm0.003$ & $\textbf{0.980}\pm0.003$
           \\
           &&
           Full
           &&
           $\underline{0.843}\pm0.024$ & $\underline{0.740}\pm0.065$ & $\underline{0.890}\pm0.017$ 
           &&
           Full
           &&
           $\underline{0.975}\pm0.000$ & $0.952\pm0.001$ & $\underline{0.960}\pm0.001$ & $0.528\pm0.037$ & $\underline{0.970}\pm0.001$ & $\underline{0.982}\pm0.001$ 
           \\
\bottomrule
\end{tabular}
}
\end{table}

\begin{table}
\centering
\setlength{\tabcolsep}{1pt}
\caption{Classification and segmentation performance figures with varying label proportions in cross-domain evaluations: NIHX (classification) and MCU (segmentation) datasets. The best scores from fully-supervised models are underlined and the best scores from semi-supervised models are bolded. Scores are given as $\text{mean}\pm\text{std}$.}
\label{table:results-cross}
\resizebox{\linewidth}{!}{
\begin{tabular}{@{} lc cc ccc c cc cccccc @{}}
            \toprule
          \multirow{2}{*}{Model}
           &
           \phantom{a}
           &
           \multirow{2}{*}{$|\mathcal{D}^c_l|$}
           &
           \phantom{a}
           &
           \multicolumn{3}{c}{Classification}
           &
           \phantom{a}
           &
           \multirow{2}{*}{$|\mathcal{D}^s_l|$}
           &
           \phantom{a}
           &
          \multicolumn{6}{c}{Segmentation}
          \\
          \cmidrule{5-7} \cmidrule{11-16}
          &&&&
          Acc & F1-Nor & F1-Abn
          &&&&
          DS & JS & SSIM & HD & P & R\\
          \midrule
          \multirow{3}{*}{\rotatebox{45}{U-Net}} 
           &&
           ---
           &&
           --- & --- & --- 
           &&
           10
           &&
           $0.555\pm0.047$ & $0.480\pm0.053$ & $0.680\pm0.059$ & $8.691\pm1.100$ & $0.553\pm0.070$ & $0.866\pm0.032$ 
           \\
           &&
           ---
           &&
           --- & --- & --- 
           &&
           50
           &&
           $0.763\pm0.026$ & $0.736\pm0.037$ & $0.870\pm0.019$ & $2.895\pm0.832$ & $0.752\pm0.035$ & $0.887\pm0.019$ 
           \\
           &&
           ---
           &&
           --- & --- & --- 
           &&
           Full
           &&
           $0.838\pm0.023$ & $0.906\pm0.035$ & $0.929\pm0.017$ & $1.414\pm0.529$ & $0.793\pm0.041$ & $0.910\pm0.013$ 
           \\
          \midrule
          \multirow{3}{*}{\rotatebox{45}{Enc}} 
           &&
           100
           &&
           $0.352\pm0.035$ & $0.070\pm0.131$ & $0.506\pm0.008$ 
           &&
           ---
           &&
           --- & --- & --- & --- & --- & ---
           \\
           &&
           1000
           &&
           $0.390\pm0.037$ & $0.192\pm0.124$ & $0.508\pm0.007$ 
           &&
           ---
           &&
           --- & --- & --- & --- & --- & ---
           \\
           &&
           Full
           &&
           $0.434\pm0.026$ & $0.296\pm0.068$ & $0.524\pm0.005$ 
           &&
           ---
           &&
           --- & --- & --- & --- & --- & ---
           \\
           \midrule
          \multirow{3}{*}{\rotatebox{45}{Enc-MM}} 
           &&
           100
           &&
           $0.360\pm0.030$ & $0.110\pm0.065$ & $0.500\pm0.024$ 
           &&
           ---
           &&
           --- & --- & --- & --- & --- & ---
           \\
           &&
           1000
           &&
           $0.406\pm0.078$ & $0.242\pm0.114$ & $0.460\pm0.010$ 
           &&
           ---
           &&
           --- & --- & --- & --- & --- & ---
           \\
           &&
           Full
           &&
           $0.452\pm0.040$ & $0.316\pm0.068$ & $0.502\pm0.030$ 
           &&
           ---
           &&
           --- & --- & --- & --- & --- & ---
           \\
           \midrule
          \multirow{3}{*}{\rotatebox{45}{Enc-SSL}} 
           &&
           100
           &&
           $0.402\pm0.052$ & $0.222\pm0.136$ & $0.510\pm0.012$ 
           &&
           ---
           &&
           --- & --- & --- & --- & --- & ---
           \\
           &&
           1000
           &&
           $0.486\pm0.050$ & $0.380\pm0.109$ & $0.530\pm0.010$ 
           &&
           ---
           &&
           --- & --- & --- & --- & --- & ---
           \\
           &&
           Full
           &&
           $0.510\pm0.024$ & $0.472\pm0.056$ & $0.538\pm0.004$ 
           &&
           ---
           &&
           --- & --- & --- & --- & --- & ---
           \\
           \midrule
          \multirow{9}{*}{\rotatebox{45}{UMTL}} 
           &&
           100
           &&
           $0.350\pm0.034$ & $0.045\pm0.123$ & $0.510\pm0.003$ 
           &&
           10
           &&
           $0.586\pm0.023$ & $0.708\pm0.035$ & $0.836\pm0.020$ & $7.156\pm2.316$ & $0.731\pm0.032$ & $0.950\pm0.014$ 
           \\
           &&
           100
           &&
           $0.363\pm0.029$ & $0.085\pm0.096$ & $0.515\pm0.005$ 
           &&
           50
           &&
           $0.580\pm0.040$ & $0.684\pm0.061$ & $0.825\pm0.038$ & $7.013\pm1.576$ & $0.697\pm0.065$ & $0.975\pm0.012$ 
           \\
           &&
           100
           &&
           $0.342\pm0.049$ & $0.015\pm0.153$ & $0.508\pm0.006$ 
           &&
           Full
           &&
           $0.607\pm0.023$ & $0.742\pm0.037$ & $0.863\pm0.021$ & $6.398\pm1.331$ & $0.759\pm0.041$ & $0.968\pm0.007$ 
           \\
           &&
           1000
           &&
           $0.413\pm0.004$ & $0.263\pm0.033$ & $0.507\pm0.005$ 
           &&
           10
           &&
           $0.676\pm0.020$ & $0.674\pm0.030$ & $0.833\pm0.016$ & $3.268\pm0.768$ & $0.712\pm0.025$ & $0.927\pm0.013$ 
           \\
           &&
           1000
           &&
           $0.400\pm0.022$ & $0.203\pm0.069$ & $0.513\pm0.005$ 
           &&
           50
           &&
           $0.704\pm0.025$ & $0.811\pm0.040$ & $0.896\pm0.015$ & $3.232\pm0.229$ & $0.828\pm0.034$ & $0.964\pm0.014$ 
           \\
           &&
           1000
           &&
           $0.430\pm0.041$ & $0.293\pm0.110$ & $0.517\pm0.005$ 
           &&
           Full
           &&
           $0.638\pm0.015$ & $0.795\pm0.024$ & $0.890\pm0.013$ & $3.893\pm0.465$ & $0.810\pm0.031$ & $0.966\pm0.010$ 
           \\
           &&
           Full
           &&
           $0.455\pm0.015$ & $0.365\pm0.038$ & $0.525\pm0.004$ 
           &&
           10
           &&
           $0.737\pm0.037$ & $0.765\pm0.056$ & $0.879\pm0.028$ & $0.917\pm0.478$ & $0.801\pm0.059$ & $0.930\pm0.013$ 
           \\
           &&
           Full
           &&
           $0.444\pm0.026$ & $0.332\pm0.075$ & $0.522\pm0.007$ 
           &&
           50
           &&
           $0.868\pm0.022$ & $0.793\pm0.038$ & $0.894\pm0.014$ & $0.742\pm0.212$ & $0.898\pm0.0378$ & $0.946\pm0.004$ 
           \\
           &&
           Full
           &&
           $0.443\pm0.038$ & $0.328\pm0.099$ & $0.520\pm0.010$ 
           &&
           Full
           &&
           $0.854\pm0.018$ & $0.828\pm0.030$ & $0.913\pm0.012$ & $0.792\pm0.490$ & $0.866\pm0.026$ & $0.942\pm0.008$ 
           \\
           \midrule
           \multirow{9}{*}{\rotatebox{45}{UMTL-S}} 
           &&
           100
           &&
           $0.344\pm0.021$ & $0.006\pm0.074$ & $0.510\pm0.005$ 
           &&
           10
           &&
           $0.797\pm0.029$ & $0.670\pm0.042$ & $0.807\pm0.027$ & $5.754\pm1.047$ & $0.698\pm0.043$ & $0.938\pm0.009$ 
           \\
           &&
           100
           &&
           $0.364\pm0.035$ & $0.098\pm0.019$ & $0.506\pm0.006$ 
           &&
           50
           &&
           $0.828\pm0.032$ & $0.715\pm0.049$ & $0.826\pm0.033$ & $6.412\pm1.753$ & $0.731\pm0.053$ & $0.971\pm0.009$ 
           \\
           &&
           100
           &&
           $0.342\pm0.016$ & $0.008\pm0.070$ & $0.510\pm0.000$ 
           &&
           Full
           &&
           $0.838\pm0.041$ & $0.715\pm0.063$ & $0.834\pm0.043$ & $6.321\pm1.573$ & $0.740\pm0.068$ & $0.966\pm0.009$ 
           \\
           &&
           1000
           &&
           $0.378\pm0.017$ & $0.138\pm0.057$ & $0.512\pm0.007$ 
           &&
           10
           &&
           $0.844\pm0.018$ & $0.718\pm0.027$ & $0.854\pm0.014$ & $3.921\pm0.480$ & $0.754\pm0.023$ & $0.939\pm0.011$ 
           \\
           &&
           1000
           &&
           $0.392\pm0.024$ & $0.186\pm0.078$ & $0.514\pm0.005$ 
           &&
           50
           &&
           $0.883\pm0.016$ & $0.793\pm0.025$ & $0.888\pm0.011$ & $3.017\pm0.191$ & $0.821\pm0.027$ & $0.959\pm0.002$ 
           \\
           &&
           1000
           &&
           $0.370\pm0.014$ & $0.130\pm0.057$ & $0.510\pm0.000$ 
           &&
           Full
           &&
           $0.898\pm0.016$ & $0.831\pm0.027$ & $0.905\pm0.014$ & $4.150\pm1.269$ & $0.845\pm0.031$ & $0.970\pm0.010$ 
           \\
           &&
           Full
           &&
           $0.470\pm0.026$ & $0.398\pm0.015$ & $0.524\pm0.010$ 
           &&
           10
           &&
           $0.881\pm0.019$ & $0.785\pm0.031$ & $0.888\pm0.014$ & $0.862\pm0.199$ & $0.830\pm0.034$ & $0.939\pm0.009$ 
           \\
           &&
           Full
           &&
           $0.413\pm0.009$ & $0.270\pm0.008$ & $0.510\pm0.014$ 
           &&
           50
           &&
           $0.917\pm0.009$ & $0.848\pm0.014$ & $0.919\pm0.006$ & $0.658\pm0.227$ & $0.966\pm0.012$ & $0.888\pm0.007$ 
           \\
           &&
           Full
           &&
           $0.433\pm0.026$ & $0.315\pm0.007$ & $0.513\pm0.015$ 
           &&
           Full
           &&
           $0.916\pm0.008$ & $0.850\pm0.013$ & $0.921\pm0.005$ & $0.882\pm0.151$ & $0.886\pm0.014$ & $0.952\pm0.002$ 
           \\
            \midrule
            \multirow{9}{*}{\rotatebox{45}{UMTL-SSL}} 
           &&
           100
           &&
           $0.442\pm0.045$ & $0.316\pm0.004$ & $0.524\pm0.007$ 
           &&
           10
           &&
           $0.833\pm0.025$ & $0.778\pm0.042$ & $0.884\pm0.019$ & $0.895\pm0.415$ & $0.810\pm0.042$ & $0.948\pm0.007$ 
           \\
           &&
           100
           &&
           $0.398\pm0.010$ & $0.166\pm0.004$ & $0.520\pm0.019$ 
           &&
           50
           &&
           $0.853\pm0.023$ & $0.839\pm0.038$ & $0.907\pm0.017$ & $0.851\pm0.157$ & $0.864\pm0.039$ & $0.952\pm0.007$ 
           \\
           &&
           100
           &&
           $0.385\pm0.062$ & $0.165\pm0.006$ & $0.515\pm0.013$ 
           &&
           Full
           &&
           $0.841\pm0.023$ & $0.818\pm0.039$ & $0.911\pm0.015$ & $0.853\pm0.410$ & $0.854\pm0.039$ & $0.949\pm0.007$ 
           \\
           &&
           1000
           &&
           $0.445\pm0.038$ & $0.333\pm0.091$ & $0.525\pm0.005$ 
           &&
           10
           &&
           $0.818\pm0.032$ & $0.781\pm0.051$ & $0.892\pm0.023$ & $1.085\pm0.464$ & $0.825\pm0.051$ & $0.938\pm0.008$ 
           \\
           &&
           1000
           &&
           $0.526\pm0.080$ & $0.486\pm0.017$ & $0.544\pm0.016$ 
           &&
           50
           &&
           $0.826\pm0.018$ & $0.804\pm0.029$ & $0.904\pm0.014$ & $0.811\pm0.137$ & $0.792\pm0.031$ & $0.949\pm0.006$ 
           \\
           &&
           1000
           &&
           $0.485\pm0.043$ & $0.413\pm0.097$ & $0.538\pm0.008$ 
           &&
           Full
           &&
           $0.843\pm0.010$ & $0.837\pm0.018$ & $0.924\pm0.007$ & $0.983\pm0.429$ & $0.882\pm0.016$ & $0.953\pm0.006$ 
           \\
           &&
           Full
           &&
           $0.526\pm0.023$ & $0.504\pm0.036$ & $0.546\pm0.012$ 
           &&
           10
           &&
           $0.824\pm0.020$ & $0.765\pm0.0312$ & $0.873\pm0.017$ & $0.994\pm0.379$ & $0.790\pm0.031$ & $0.943\pm0.010$ 
           \\
           &&
           Full
           &&
           $0.530\pm0.030$ & $0.514\pm0.058$ & $0.542\pm0.012$ 
           &&
           50
           &&
           $0.867\pm0.020$ & $0.839\pm0.034$ & $0.917\pm0.013$ & $0.566\pm0.282$ & $0.881\pm0.033$ & $0.945\pm0.005$ 
           \\
           &&
           Full
           &&
           $\underline{0.520}\pm0.031$ & $\underline{0.490}\pm0.061$ & $0.542\pm0.007$ 
           &&
           Full
           &&
           $0.884\pm0.011$ & $0.884\pm0.021$ & $0.934\pm0.008$ & $0.599\pm0.201$ & $0.918\pm0.021$ & $0.955\pm0.003$ 
           \\
            \midrule
            \multirow{9}{*}{\rotatebox{45}{UMTL-SSL-S}} 
           &&
           100
           &&
           $0.370\pm0.046$ & $0.114\pm0.129$ & $0.510\pm0.011$ 
           &&
           10
           &&
           $0.853\pm0.026$ & $0.747\pm0.041$ & $0.866\pm0.019$ & $1.048\pm0.186$ & $0.782\pm0.041$ & $0.944\pm0.010$ 
           \\
           &&
           100
           &&
           $0.400\pm0.067$ & $0.192\pm0.171$ & $0.518\pm0.012$ 
           &&
           50
           &&
           $0.889\pm0.019$ & $0.799\pm0.031$ & $0.899\pm0.014$ & $0.854\pm0.298$ & $0.834\pm0.031$ & $0.950\pm0.006$ 
           \\
           &&
           100
           &&
           $0.370\pm0.061$ & $0.114\pm0.159$ & $0.514\pm0.016$ 
           &&
           Full
           &&
           $0.915\pm0.019$ & $0.848\pm0.032$ & $0.920\pm0.013$ & $0.987\pm0.328$ & $0.880\pm0.033$ & $0.956\pm0.003$ 
           \\
           &&
           1000
           &&
           $0.432\pm0.043$ & $0.286\pm0.019$ & $0.524\pm0.010$ 
           &&
           10
           &&
           $0.871\pm0.034$ & $0.785\pm0.054$ & $0.884\pm0.025$ & $1.327\pm0.135$ & $0.818\pm0.053$ & $0.944\pm0.008$ 
           \\
           &&
           1000
           &&
           $0.458\pm0.077$ & $0.342\pm0.186$ & $0.530\pm0.013$ 
           &&
           50
           &&
           $0.893\pm0.008$ & $0.803\pm0.014$ & $0.895\pm0.006$ & $1.123\pm0.215$ & $0.835\pm0.015$ & $0.946\pm0.005$ 
           \\
           &&
           1000
           &&
           $0.462\pm0.050$ & $0.350\pm0.115$ & $0.536\pm0.014$ 
           &&
           Full
           &&
           $0.930\pm0.008$ & $0.860\pm0.013$ & $0.925\pm0.006$ & $1.042\pm0.206$ & $0.912\pm0.013$ & $0.955\pm0.003$ 
           \\
           &&
           Full
           &&
           $0.482\pm0.042$ & $0.412\pm0.087$ & $0.536\pm0.008$ 
           &&
           10
           &&
           $0.880\pm0.031$ & $0.765\pm0.050$ & $0.885\pm0.022$ & $0.745\pm0.219$ & $0.818\pm0.045$ & $0.941\pm0.014$ 
           \\
           &&
           Full
           &&
           $0.490\pm0.021$ & $0.426\pm0.041$ & $0.540\pm0.006$ 
           &&
           50
           &&
           $0.912\pm0.014$ & $0.845\pm0.024$ & $0.909\pm0.009$ & $0.956\pm0.215$ & $0.881\pm0.025$ & $0.952\pm0.002$ 
           \\
           &&
           Full
           &&
           $0.510\pm0.038$ & $0.474\pm0.071$ & $0.540\pm0.011$ 
           &&
           Full
           &&
           $0.875\pm0.032$ & $0.809\pm0.053$ & $0.875\pm0.026$ & $0.722\pm0.335$ & $0.851\pm0.057$ & $0.944\pm0.008$ 
           \\
            \midrule
            \multirow{9}{*}{\rotatebox{45}{MultiMix}} 
           &&
           100
           &&
           $0.440\pm0.058$ & $0.164\pm0.019$ & $0.510\pm0.014$ 
           &&
           10
           &&
           $0.857\pm0.028$ & $0.732\pm0.044$ & $0.863\pm0.018$ & $1.227\pm0.534$ & $0.767\pm0.044$ & $0.943\pm0.016$ 
           \\
           &&
           100
           &&
           $0.370\pm0.086$ & $0.036\pm0.003$ & $0.510\pm0.013$ 
           &&
           50
           &&
           $0.889\pm0.021$ & $0.790\pm0.036$ & $0.890\pm0.015$ & $1.061\pm0.434$ & $0.866\pm0.035$ & $0.947\pm0.008$  
           \\
           &&
           100
           &&
           $0.500\pm0.080$ & $0.300\pm0.002$ & $0.510\pm0.006$ 
           &&
           Full
           &&
           $0.899\pm0.022$ & $0.825\pm0.036$ & $0.906\pm0.017$ & $0.647\pm0.074$ & $0.852\pm0.040$ & $0.952\pm0.012$ 
           \\
           &&
           1000
           &&
           $0.520\pm0.041$ & $0.386\pm0.009$ & $0.530\pm0.014$ 
           &&
           10
           &&
           $0.862\pm0.017$ & $0.775\pm0.026$ & $0.878\pm0.011$ & $1.307\pm0.325$ & $0.816\pm0.029$ & $0.939\pm0.006$ 
           \\
           &&
           1000
           &&
           $0.540\pm0.018$ & $0.500\pm0.036$ & $0.536\pm0.005$ 
           &&
           50
           &&
           $0.912\pm0.018$ & $0.831\pm0.031$ & $0.907\pm0.012$ & $1.293\pm0.375$ & $0.865\pm0.030$ & $0.955\pm0.007$ 
           \\
           &&
           1000
           &&
           $\textbf{0.570}\pm0.088$ & $\textbf{0.620}\pm0.003$ & $0.510\pm0.008$ 
           &&
           Full
           &&
           $\textbf{0.936}\pm0.026$ & $\textbf{0.880}\pm0.043$ & $\textbf{0.932}\pm0.022$ & $0.803\pm0.178$ & $0.917\pm0.050$ & $\textbf{0.979}\pm0.008$ 
           \\
           &&
           Full
           &&
           $0.550\pm0.038$ & $0.430\pm0.006$ & $0.534\pm0.010$ 
           &&
           10
           &&
           $0.886\pm0.013$ & $0.802\pm0.022$ & $0.894\pm0.011$ & $0.746\pm0.284$ & $0.839\pm0.028$ & $0.948\pm0.007$ 
           \\
           &&
           Full
           &&
           $0.560\pm0.040$ & $0.570\pm0.008$ & $\textbf{0.550}\pm0.007$ 
           &&
           50
           &&
           $0.935\pm0.017$ & $0.878\pm0.030$ & $0.930\pm0.012$ & $\textbf{0.515}\pm0.232$ & $\textbf{0.928}\pm0.033$ & $0.957\pm0.005$ 
           \\
           &&
           Full
           &&
           $\underline{0.520}\pm0.022$ & $\underline{0.490}\pm0.064$ & $\underline{0.550}\pm0.008$ 
           &&
           Full
           &&
           $\underline{0.943}\pm0.009$ & $\underline{0.892}\pm0.015$ & $\underline{0.937}\pm0.006$ & $\underline{0.417}\pm0.181$ & $\underline{0.928}\pm0.016$ & $\underline{0.958}\pm0.002$ 
           \\
\bottomrule
\end{tabular}
}
\end{table}

\begin{figure} \centering
\subcaptionbox{JSRT (in-domain)}{
  \begin{tabular}{ccc}
    & \scriptsize Ground Truth & \\
    & \includegraphics[width=0.134\linewidth, trim={110 40 105 30},clip]{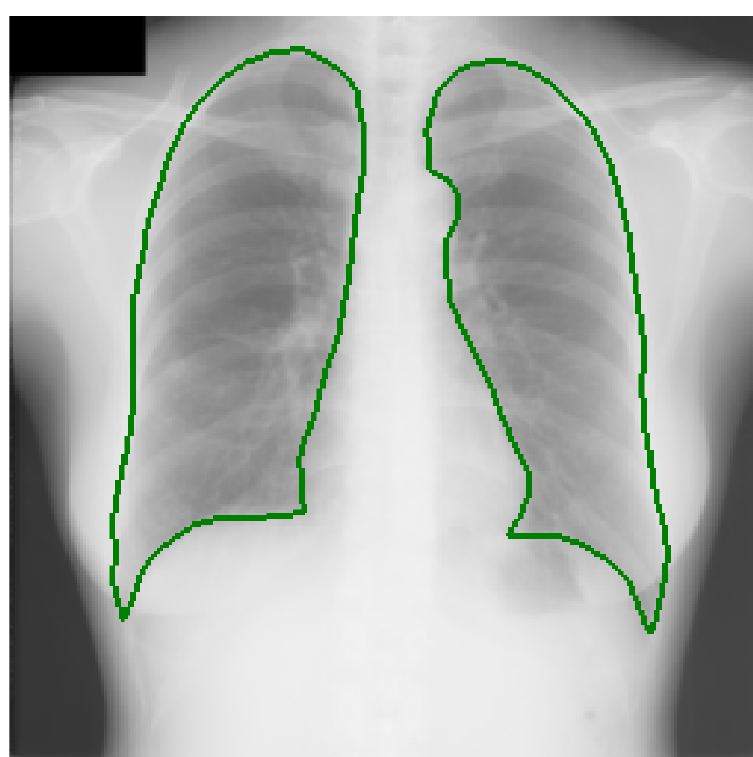} & \\[6pt]
    $|\mathcal{D}^s| = 10$ & $|\mathcal{D}^s| = 50$ & $|\mathcal{D}^s| = \text{\small Full}$ \\
    \multicolumn{3}{c}{\small U-Net} \\
    \includegraphics[width=0.134\linewidth, trim={110 40 105 30},clip]{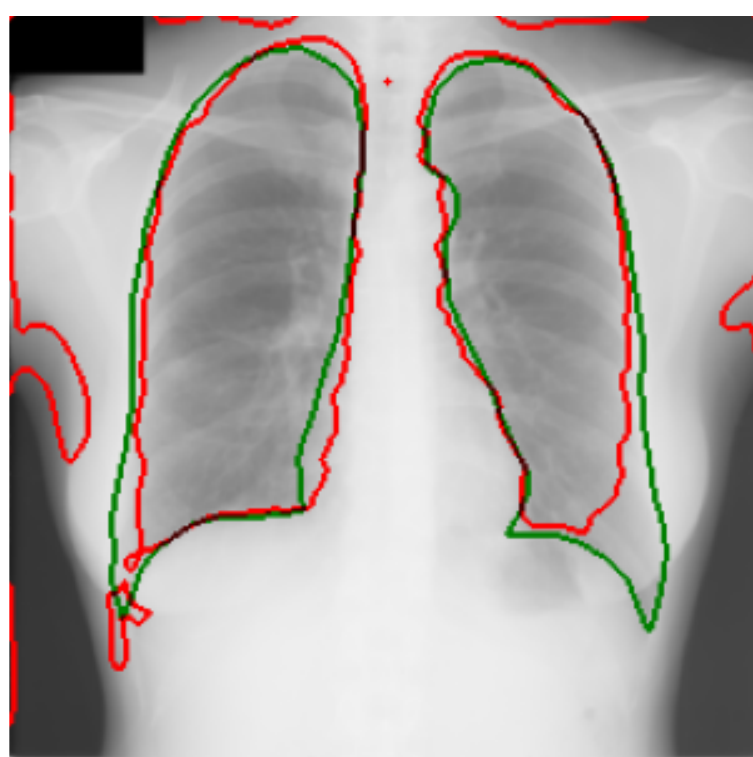} &
    \includegraphics[width=0.134\linewidth, trim={110 40 105 30},clip]{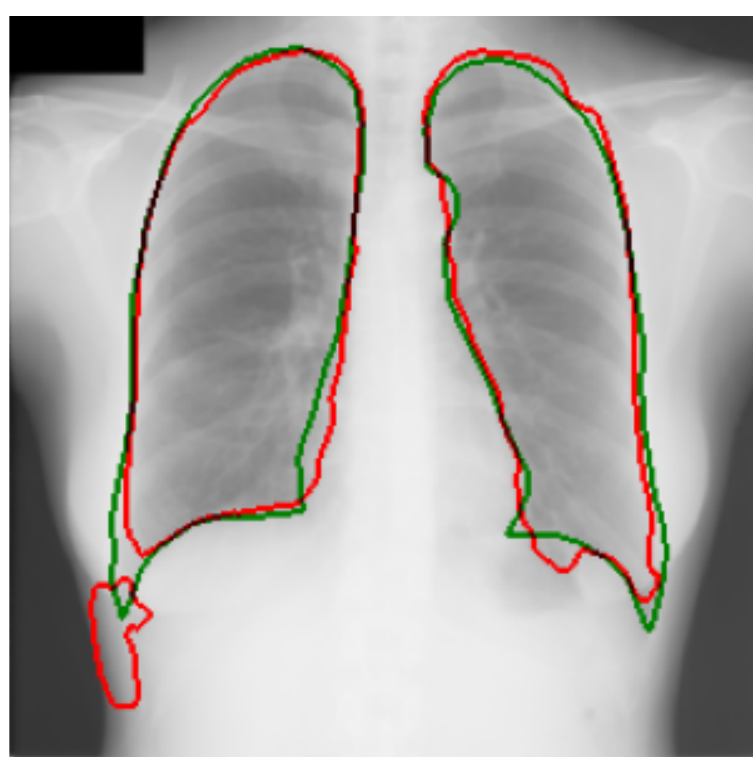} &
    \includegraphics[width=0.134\linewidth, trim={110 40 105 30},clip]{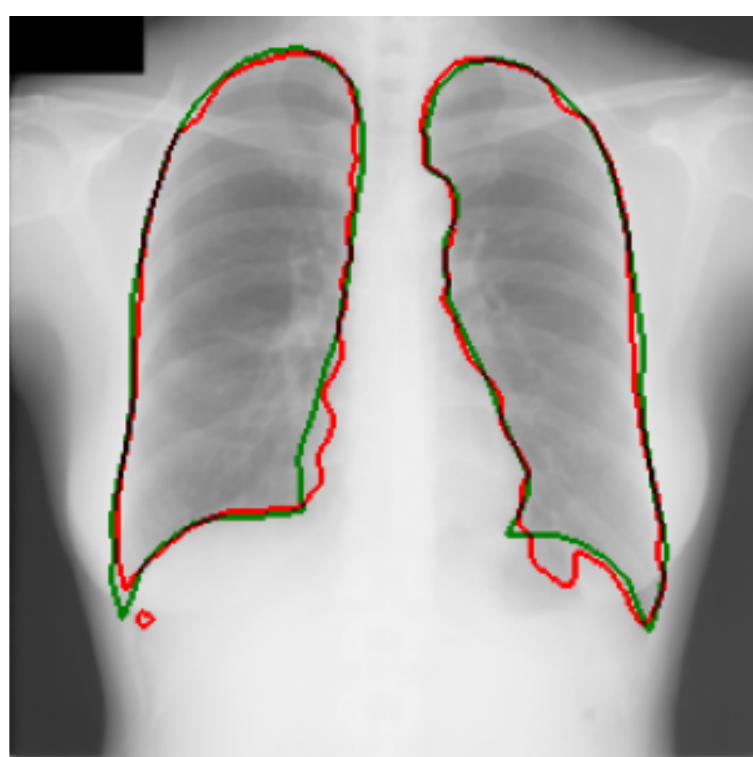}
    \\
    \multicolumn{3}{c}{\small UMTL} \\
    \includegraphics[width=0.134\linewidth, trim={110 40 105 30},clip]{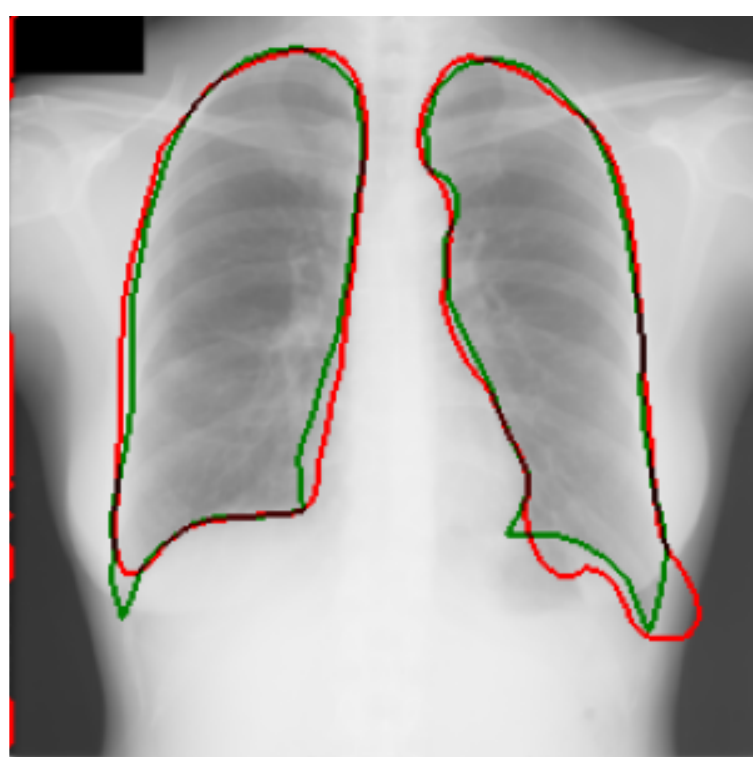} &
    \includegraphics[width=0.134\linewidth, trim={110 40 105 30},clip]{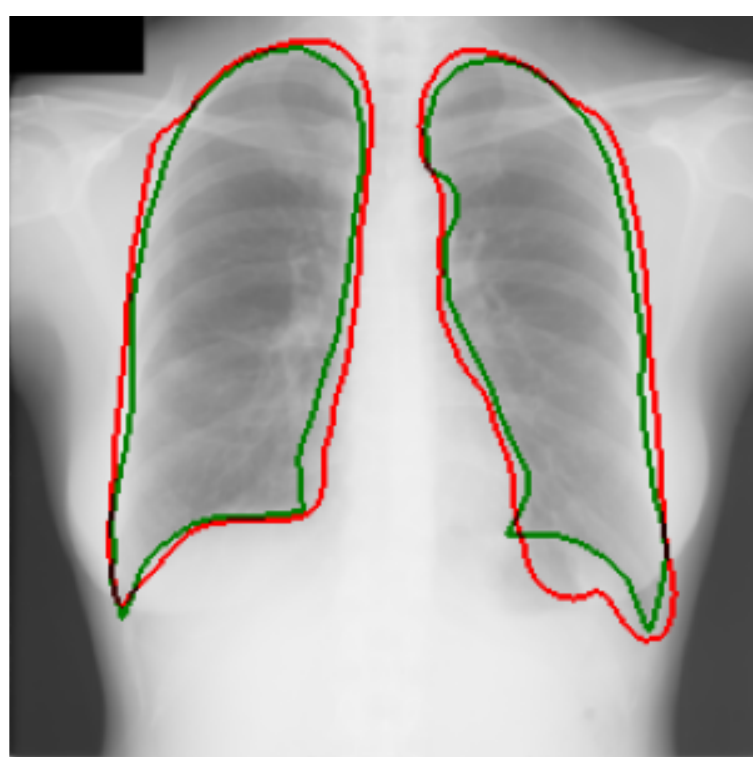} &
    \includegraphics[width=0.134\linewidth, trim={110 40 105 30},clip]{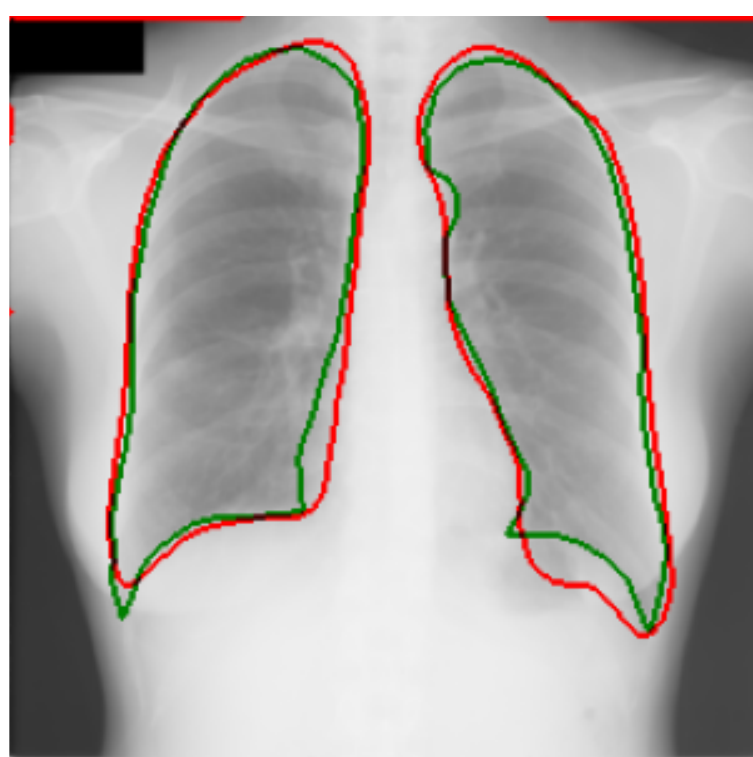}
    \\
    \multicolumn{3}{c}{\small UMTLS} \\
    \includegraphics[width=0.134\linewidth, trim={110 40 105 30},clip]{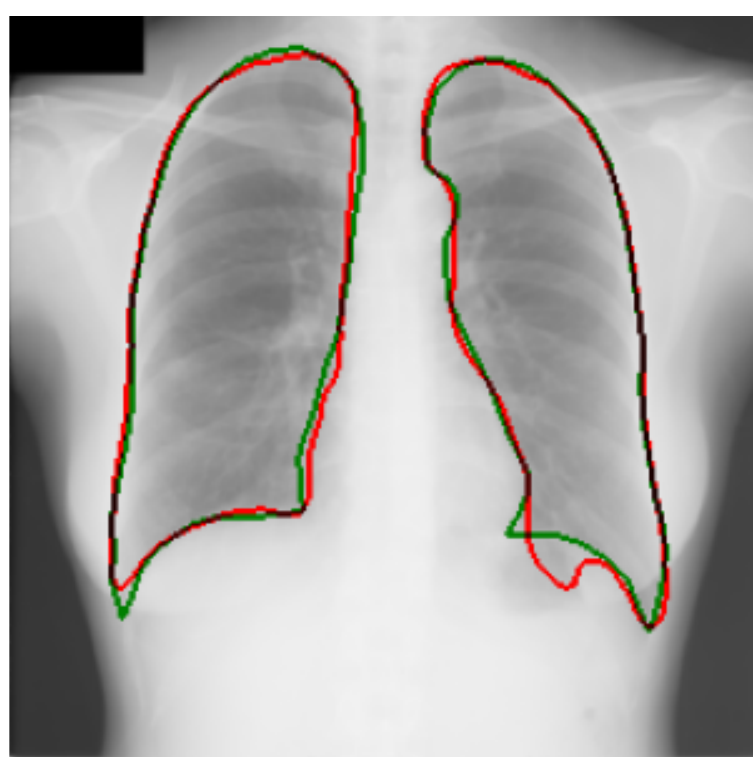} &
    \includegraphics[width=0.134\linewidth, trim={110 40 105 30},clip]{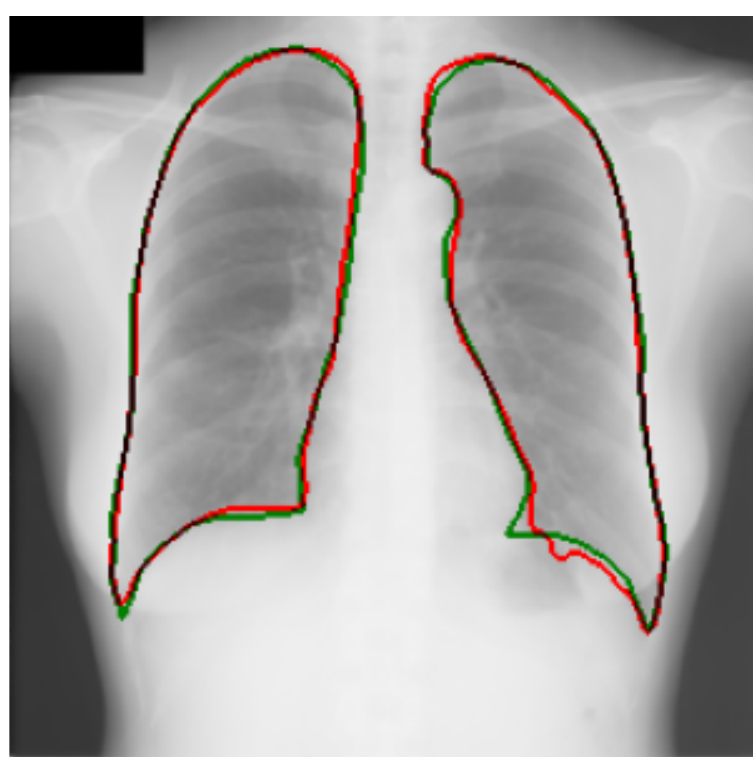} &
    \includegraphics[width=0.134\linewidth, trim={110 40 105 30},clip]{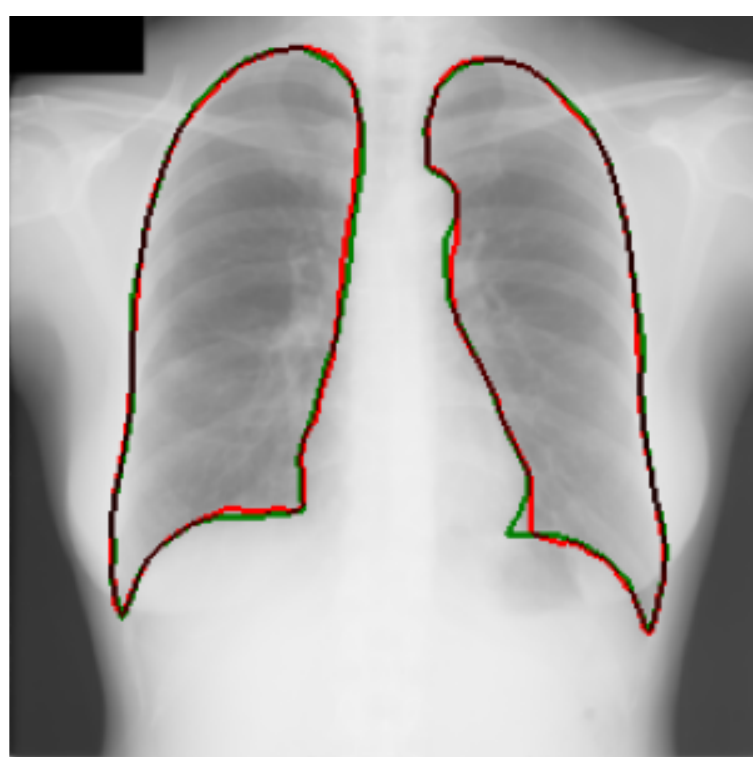}
    \\
    \multicolumn{3}{c}{\small UMTL-SSL} \\
    \includegraphics[width=0.134\linewidth, trim={110 40 105 30},clip]{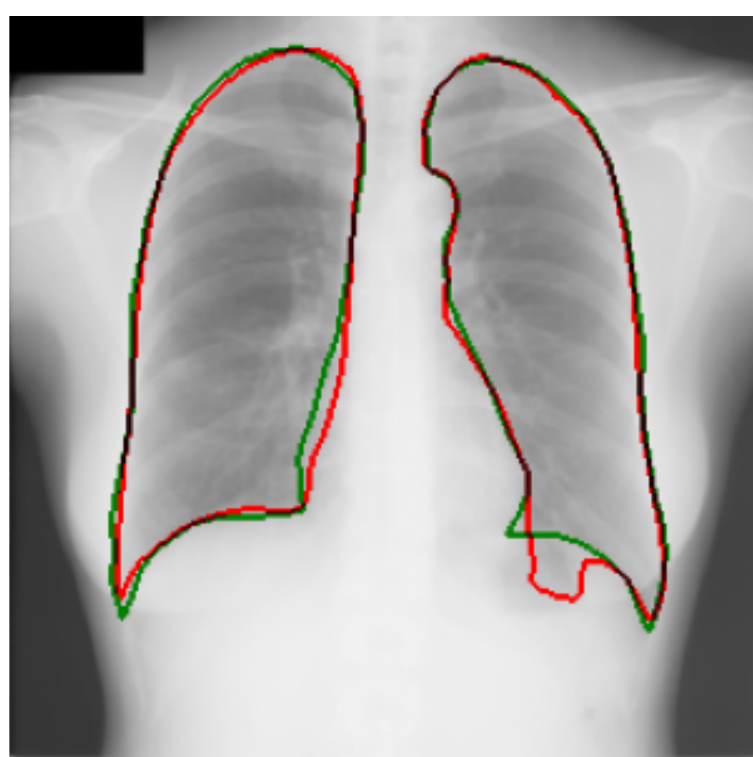} &
    \includegraphics[width=0.134\linewidth, trim={110 40 105 30},clip]{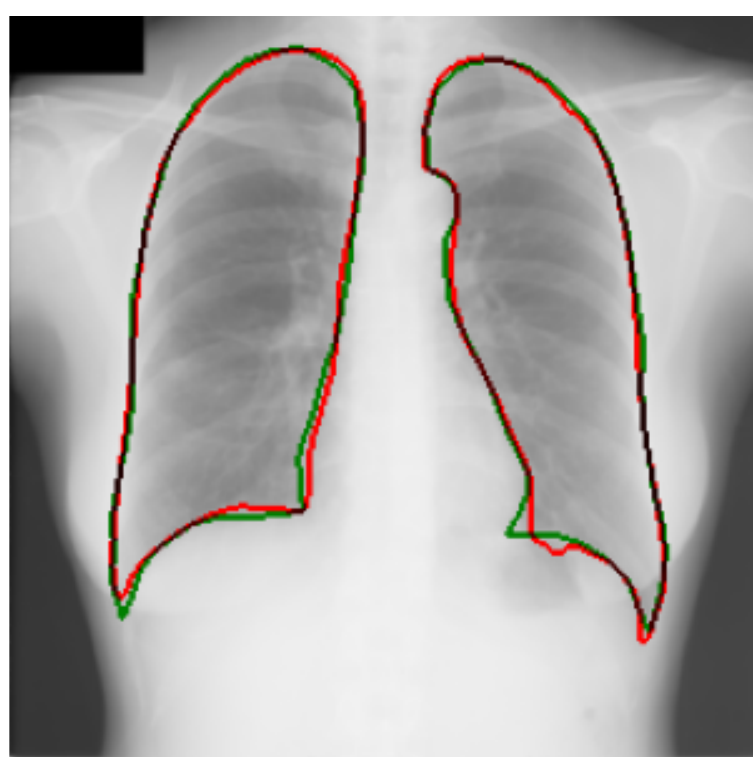} &
    \includegraphics[width=0.134\linewidth, trim={110 40 105 30},clip]{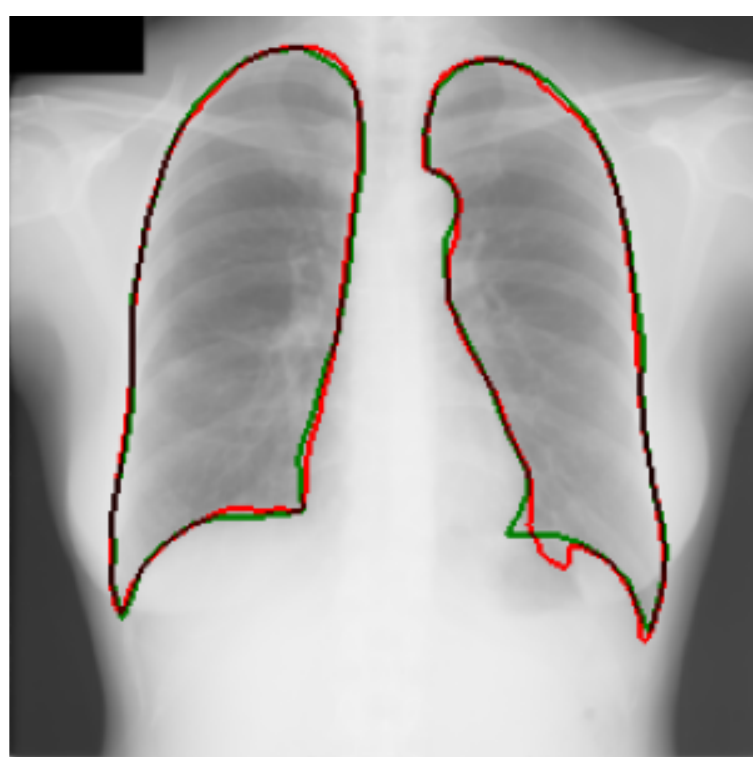}
    \\
    \multicolumn{3}{c}{\small UMTL-SSL-S} \\
    \includegraphics[width=0.134\linewidth, trim={110 40 105 30},clip]{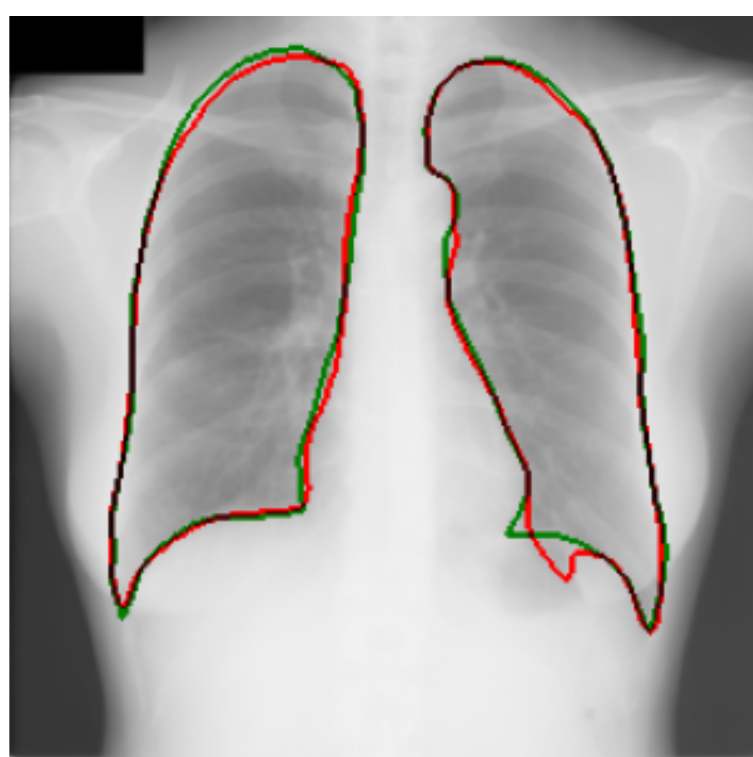} &
    \includegraphics[width=0.134\linewidth, trim={110 40 105 30},clip]{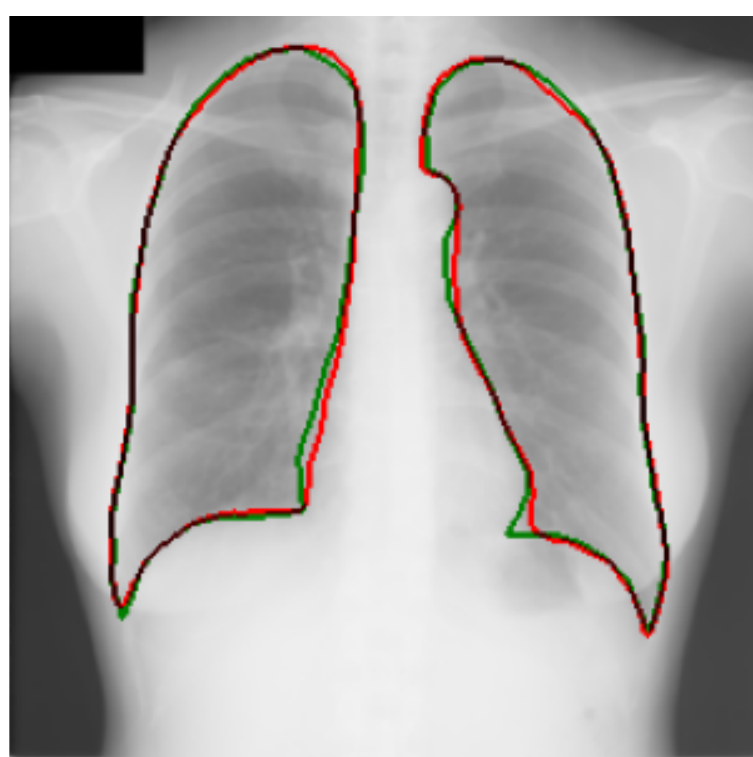} &
    \includegraphics[width=0.134\linewidth, trim={110 40 105 30},clip]{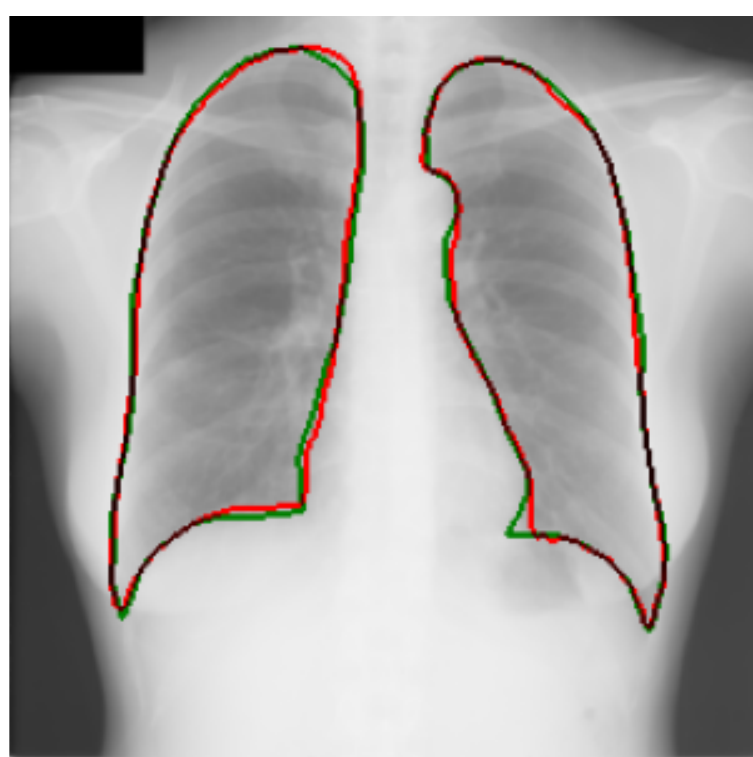}
    \\
    \multicolumn{3}{c}{\small MultiMix} \\
    \includegraphics[width=0.134\linewidth, trim={110 40 105 30},clip]{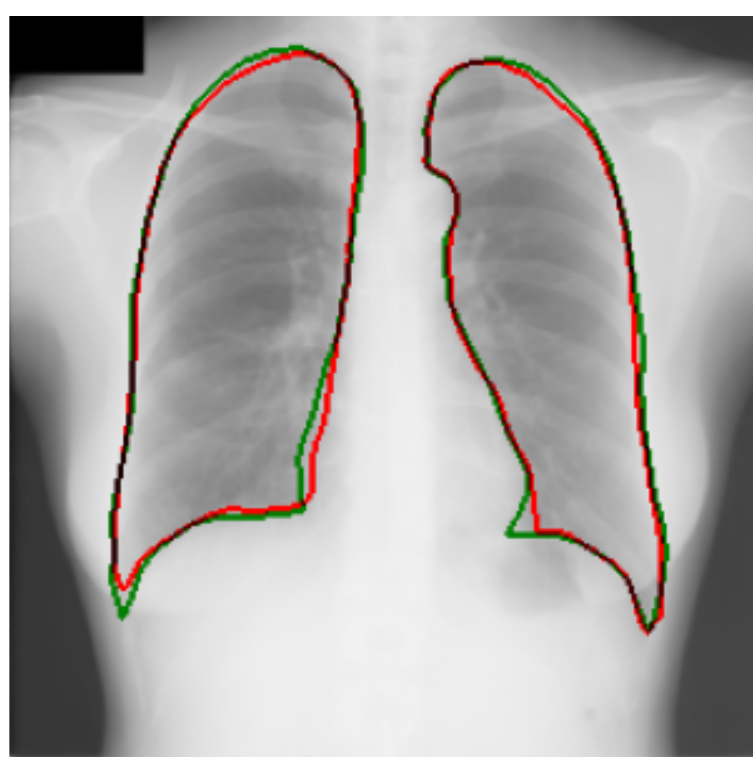} &
    \includegraphics[width=0.134\linewidth, trim={110 40 105 30},clip]{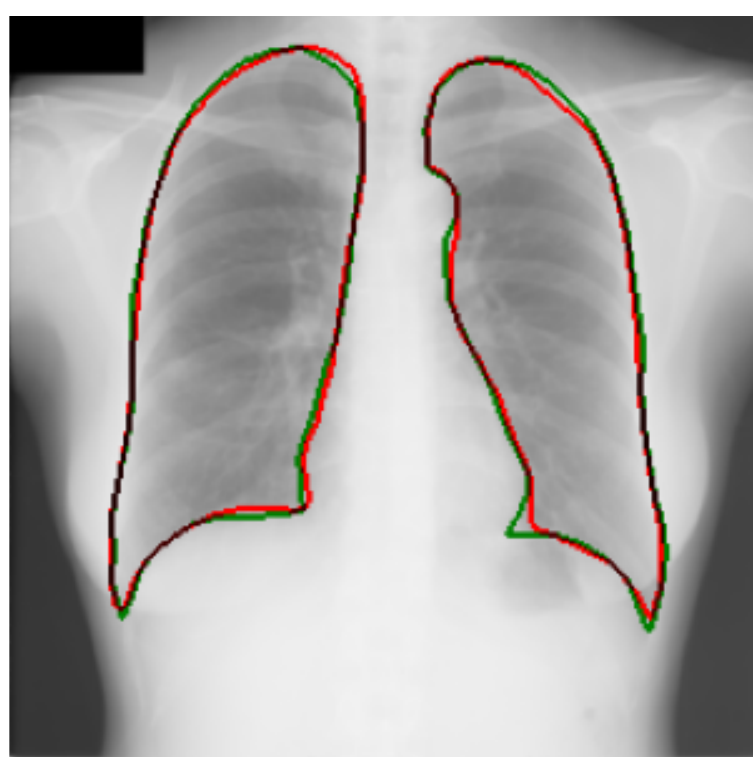} &
    \includegraphics[width=0.134\linewidth, trim={110 40 105 30},clip]{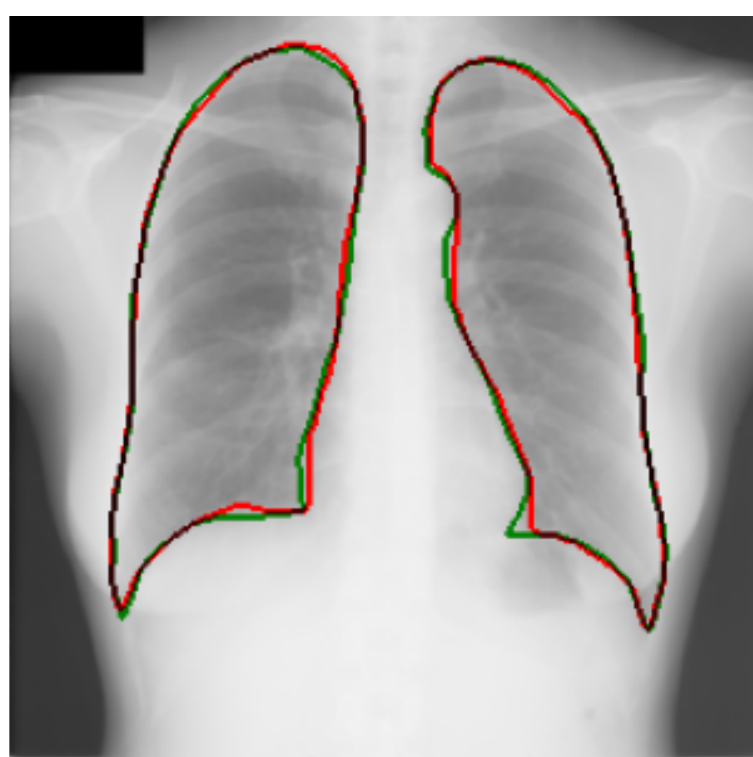}
\end{tabular}}
\hfill
\subcaptionbox{MCU (cross-domain)}{
  \begin{tabular}{ccc}
    & \scriptsize Ground Truth & \\
    & \includegraphics[width=0.134\linewidth, trim={110 40 105 30},clip]{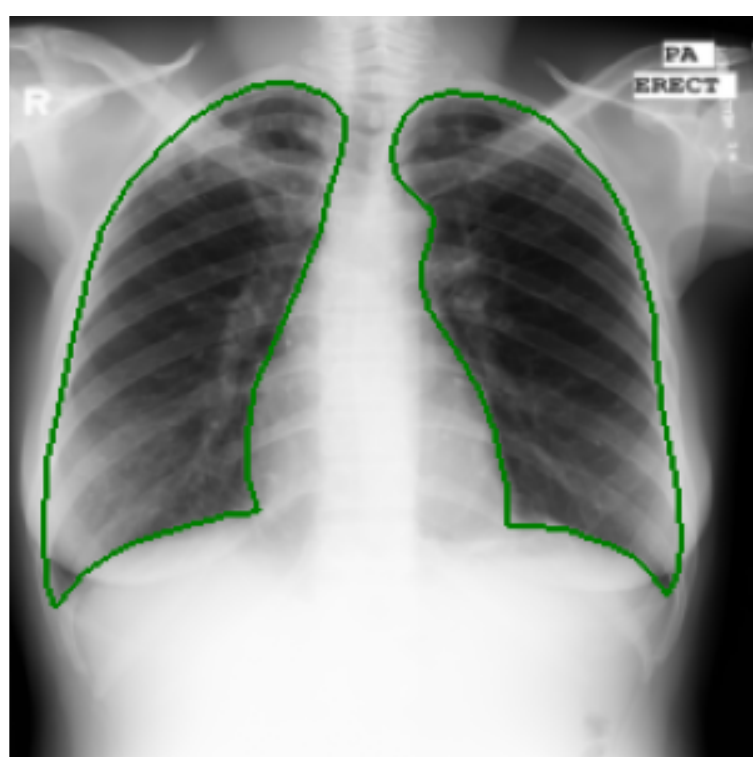} & \\[6pt]
    $|\mathcal{D}^s| = 10$ & $|\mathcal{D}^s| = 50$ & $|\mathcal{D}^s| = \text{\small Full}$ \\
    \multicolumn{3}{c}{\small U-Net} \\
    \includegraphics[width=0.134\linewidth, trim={110 40 105 30},clip]{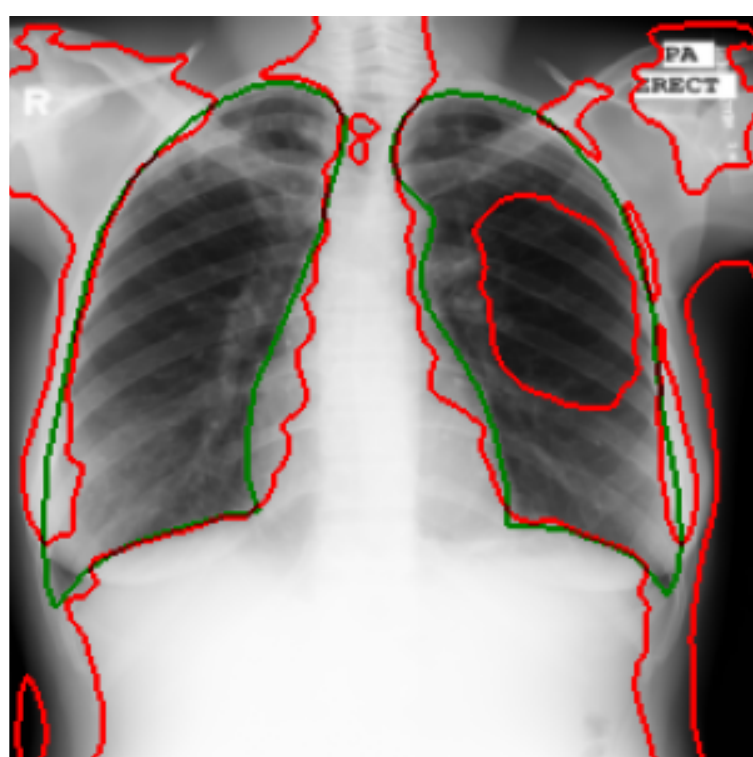} &
    \includegraphics[width=0.134\linewidth, trim={4cm 1cm 3cm  1cm},clip]{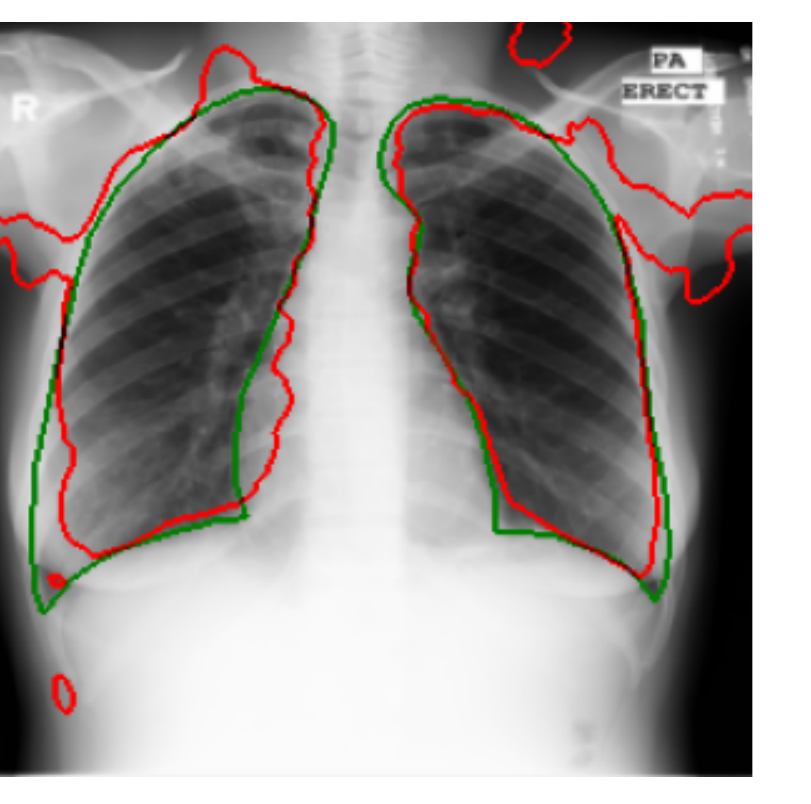} &
    \includegraphics[width=0.134\linewidth, trim={110 40 105 30},clip]{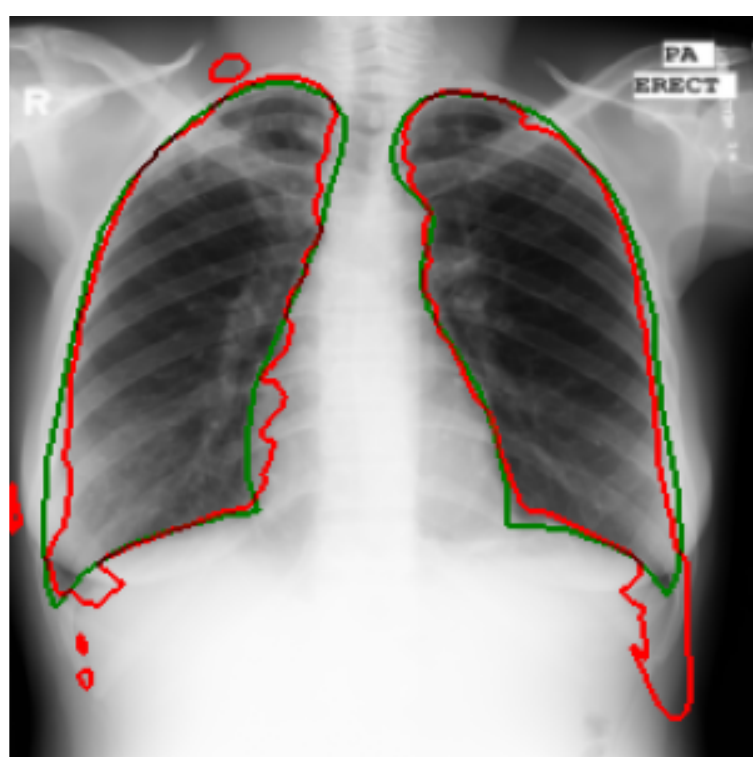}
    \\
    \multicolumn{3}{c}{\small UMTL} \\
    \includegraphics[width=0.134\linewidth, trim={110 40 105 30},clip]{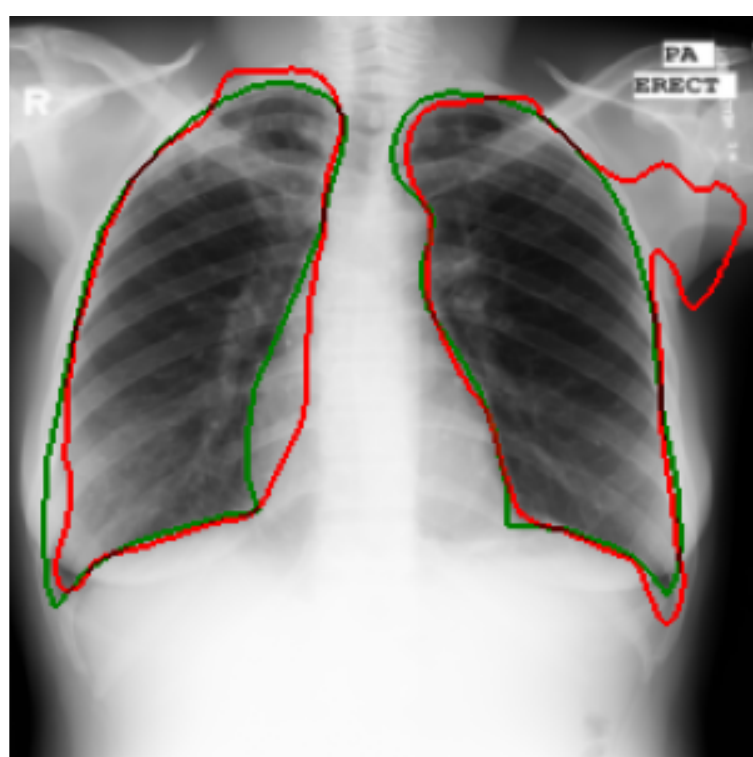} &
    \includegraphics[width=0.134\linewidth, trim={110 40 105 30},clip]{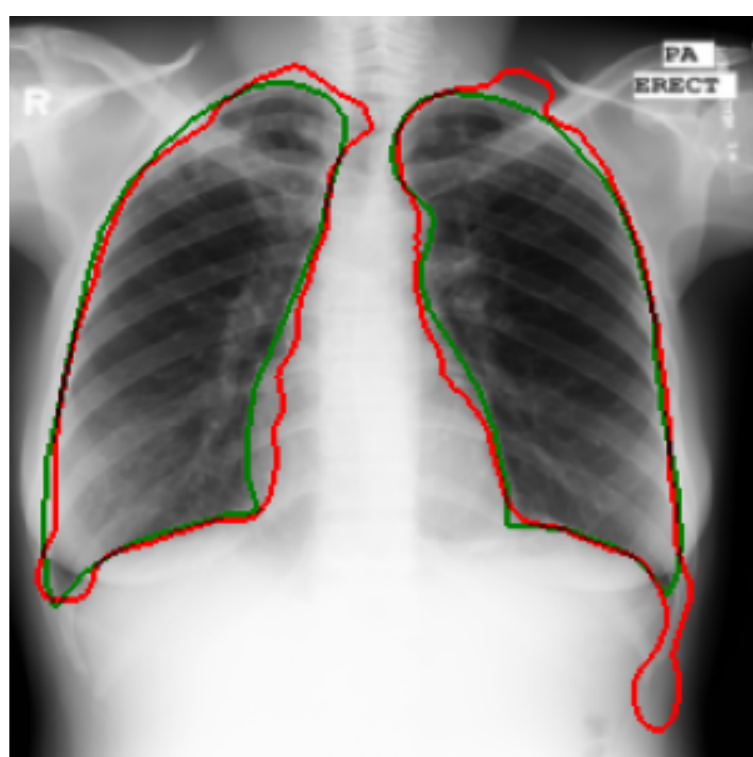} &
    \includegraphics[width=0.134\linewidth, trim={110 40 105 30},clip]{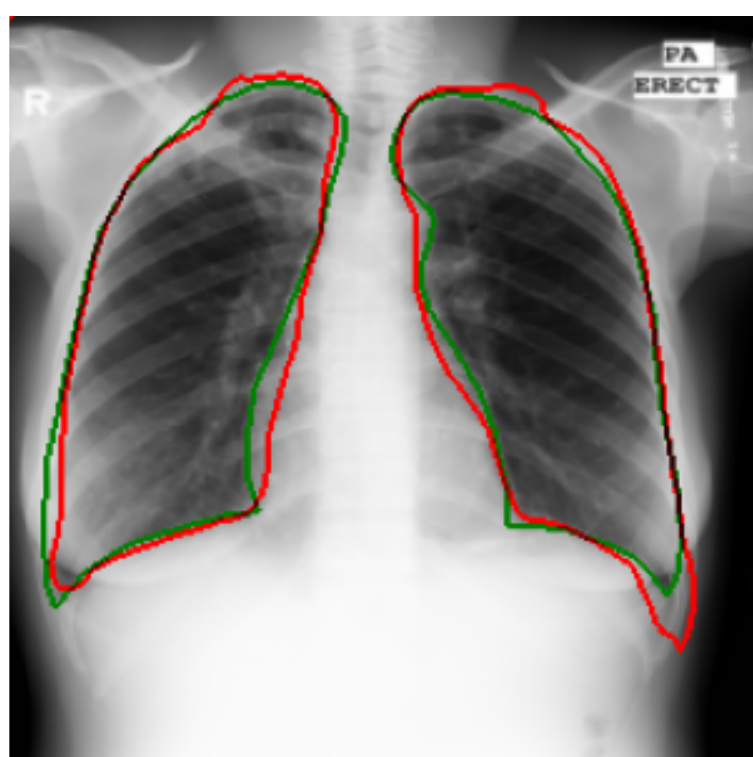}
    \\
    \multicolumn{3}{c}{\small UMTLS} \\
    \includegraphics[width=0.134\linewidth, trim={110 40 105 30},clip]{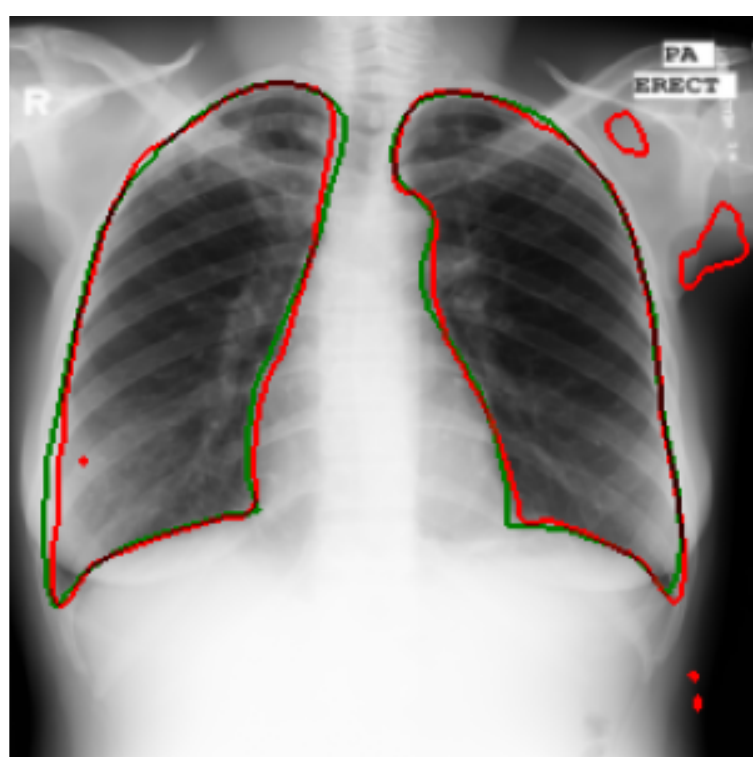} &
    \includegraphics[width=0.134\linewidth, trim={110 40 105 30},clip]{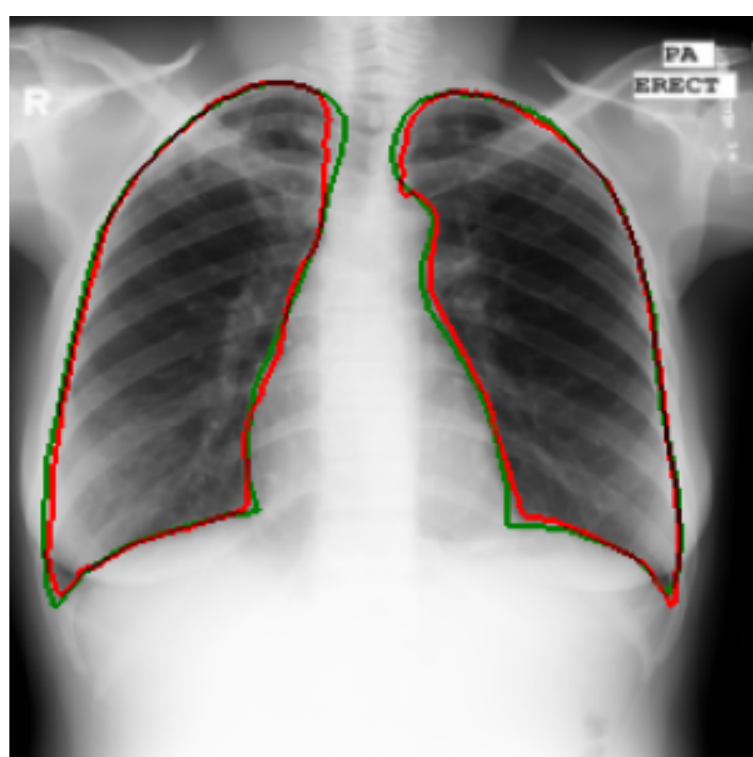} &
    \includegraphics[width=0.134\linewidth, trim={110 40 105 30},clip]{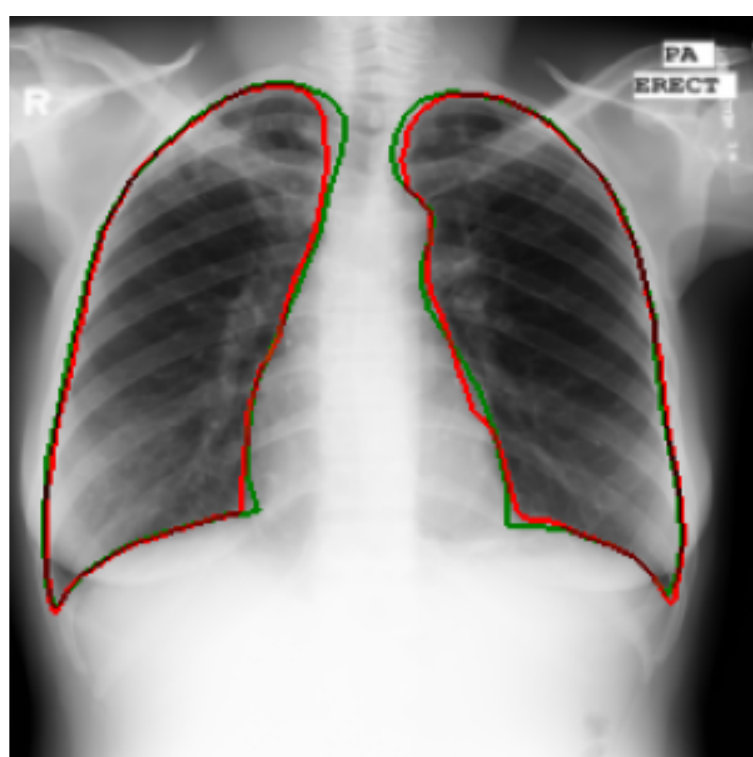}
    \\
    \multicolumn{3}{c}{\small UMTL-SSL} \\
    \includegraphics[width=0.134\linewidth, trim={110 40 105 30},clip]{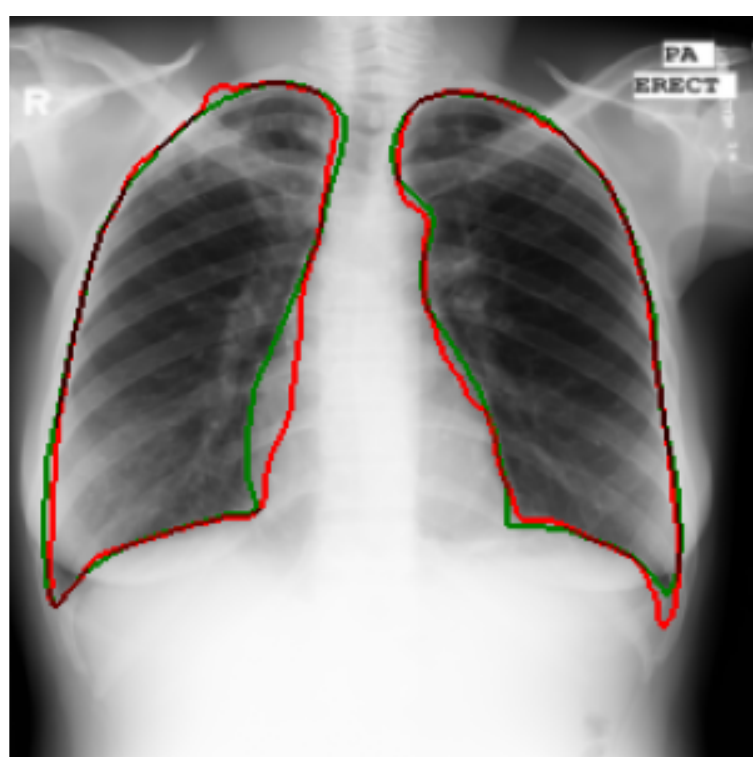} &
    \includegraphics[width=0.134\linewidth, trim={110 40 105 30},clip]{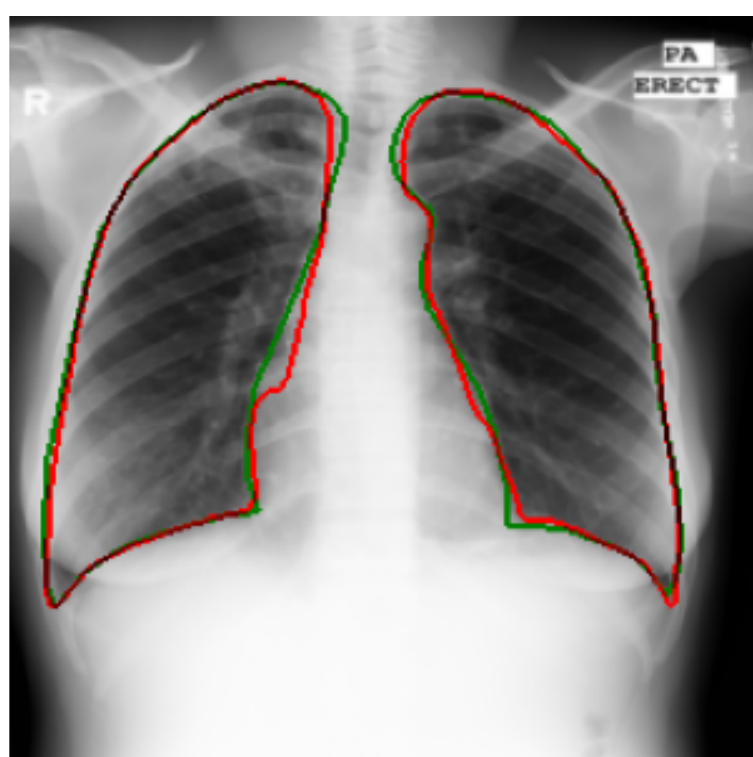} &
    \includegraphics[width=0.134\linewidth, trim={110 40 105 30},clip]{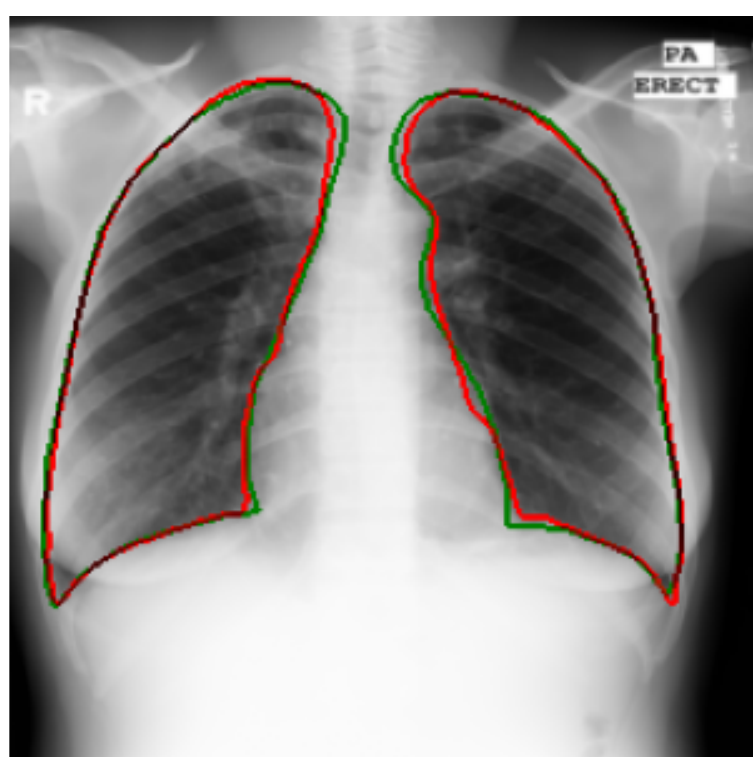}
    \\
    \multicolumn{3}{c}{\small UMTL-SSL-S} \\
    \includegraphics[width=0.134\linewidth, trim={110 40 105 30},clip]{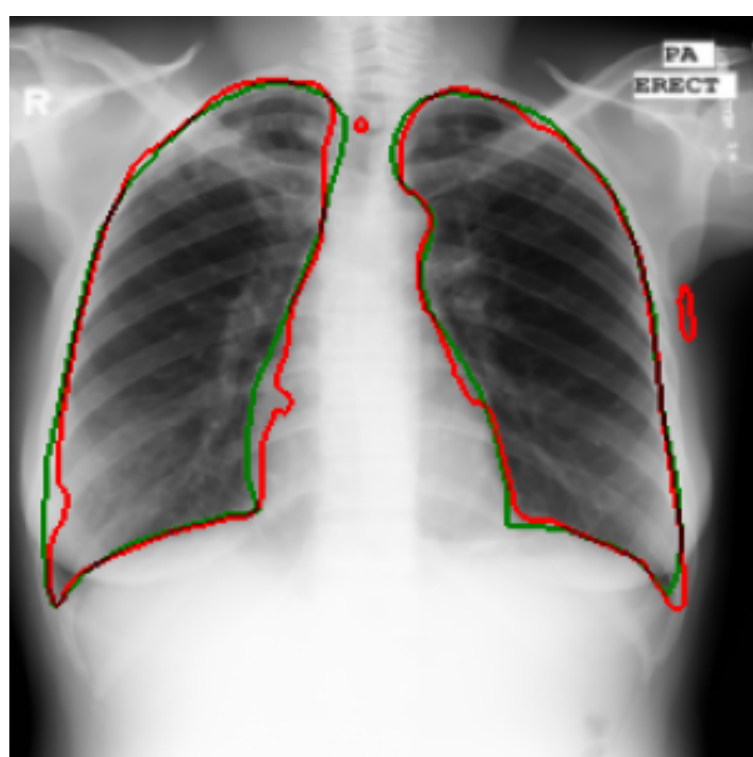} &
    \includegraphics[width=0.134\linewidth, trim={110 40 105 30},clip]{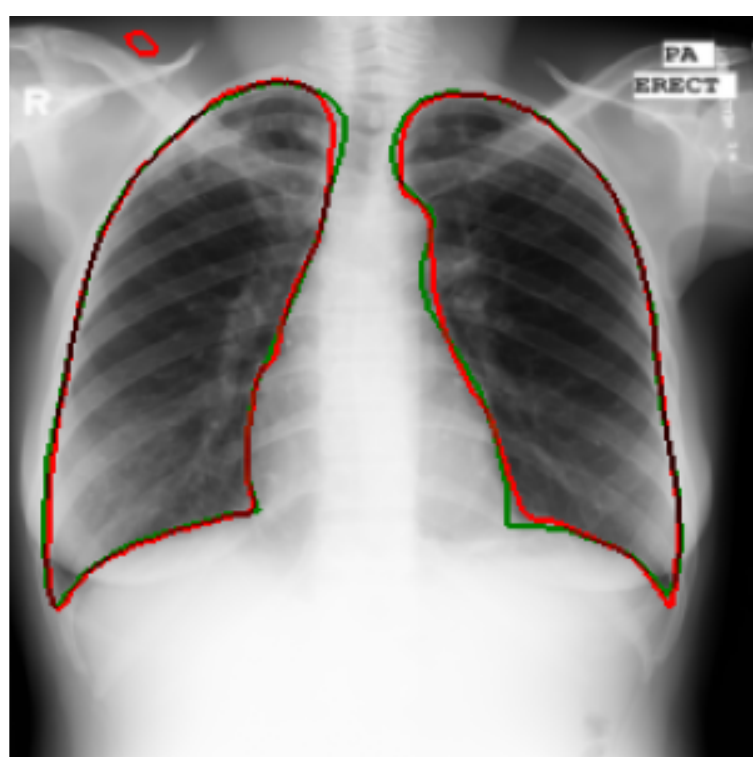} &
    \includegraphics[width=0.134\linewidth, trim={110 40 105 30},clip]{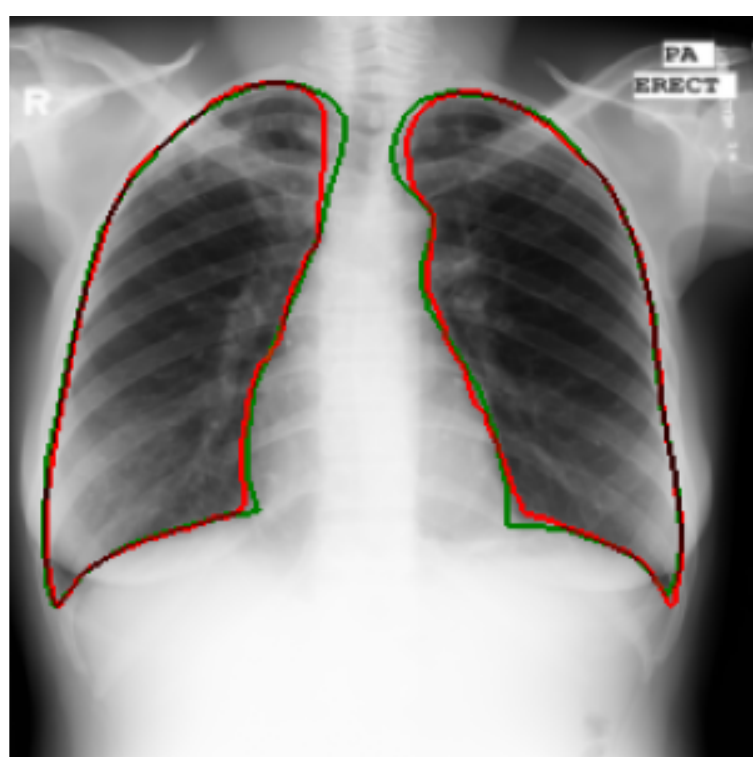}
    \\
    \multicolumn{3}{c}{\small MultiMix} \\
    \includegraphics[width=0.134\linewidth, trim={110 40 105 30},clip]{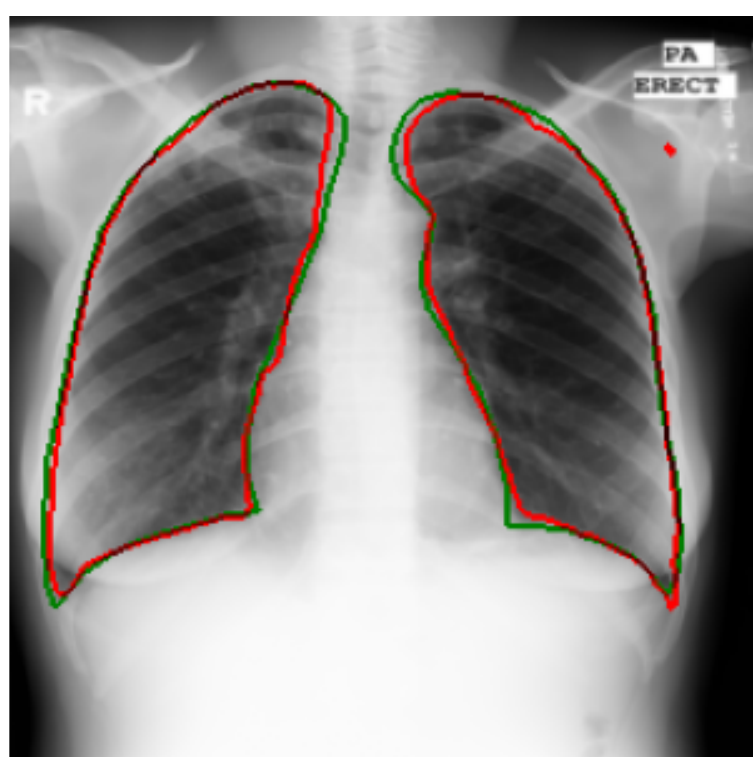} &
    \includegraphics[width=0.134\linewidth, trim={110 40 105 30},clip]{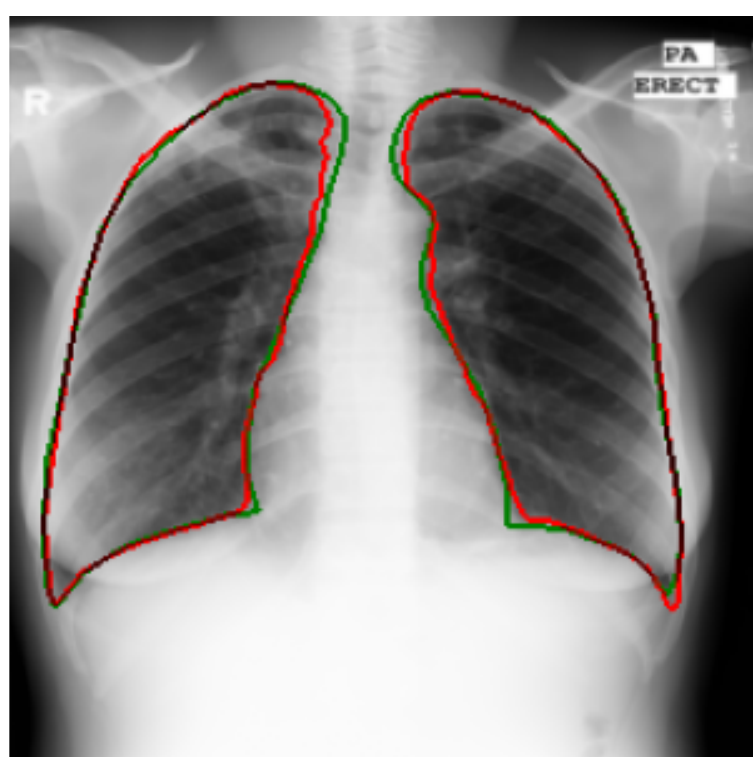} &
    \includegraphics[width=0.134\linewidth, trim={110 40 105 30},clip]{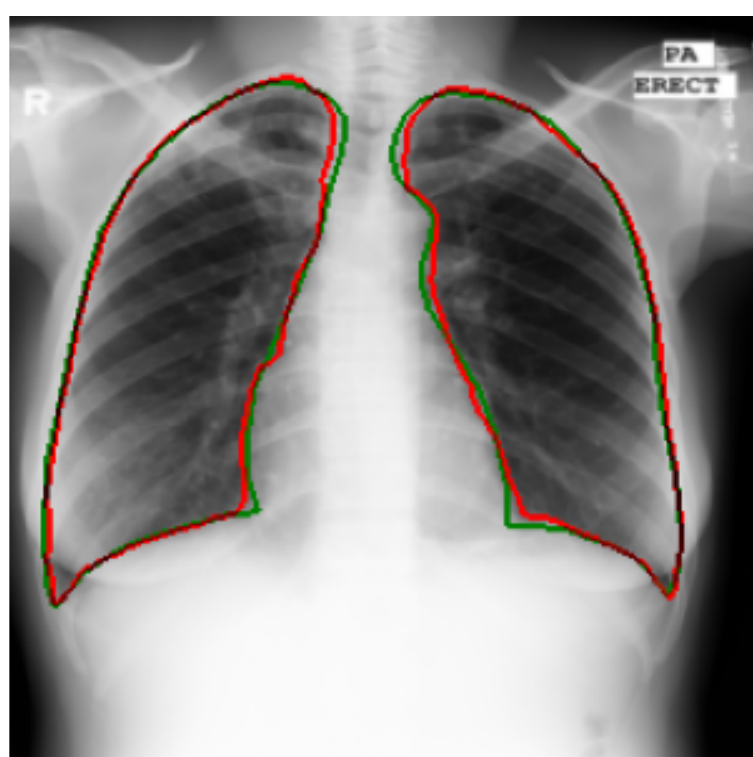}
\end{tabular}}
\caption{Visualization of the ground truth reference (green) and predicted (red) segmentation boundaries in a chest X-ray reveals the superiority of MultiMix.}
\label{fig:lung_vis}
\end{figure}

\begin{figure} \centering 
\subcaptionbox{in-domain \label{fig:altman-in}}{
 \begin{tabular}{ccc} 
  \includegraphics[width=0.28\linewidth]{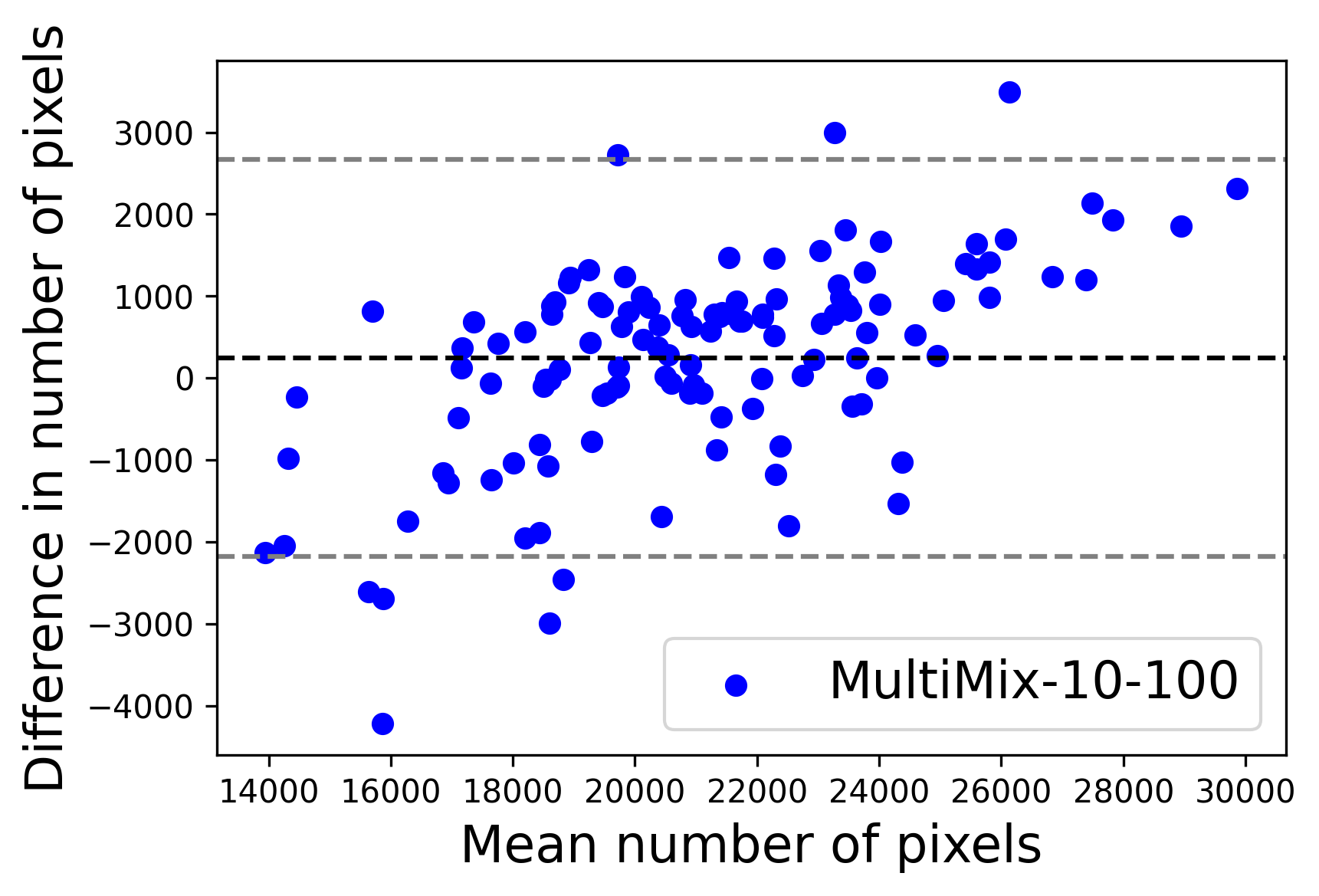}\quad
  \includegraphics[width=0.28\linewidth]{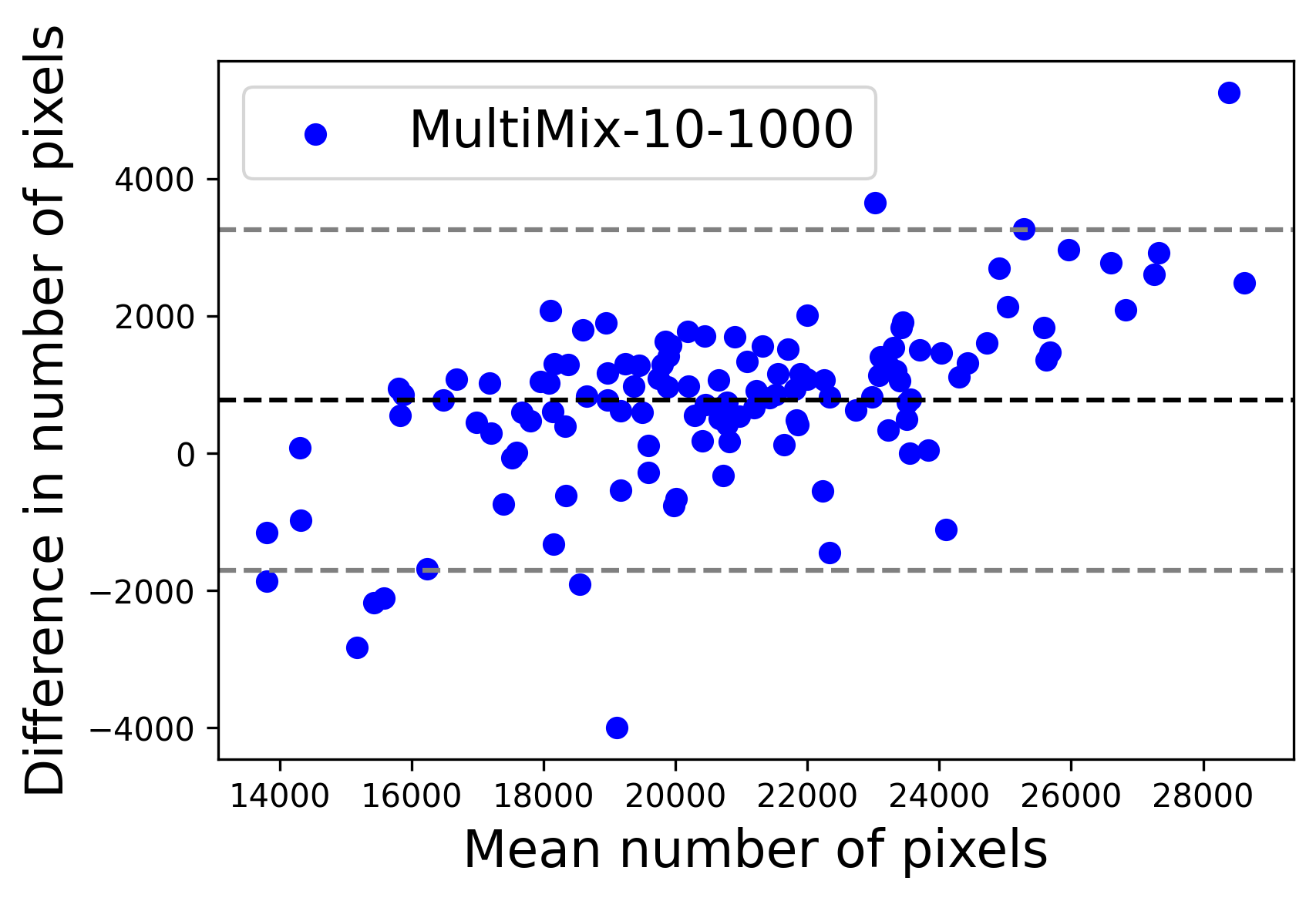}\quad
  \includegraphics[width=0.28\linewidth]{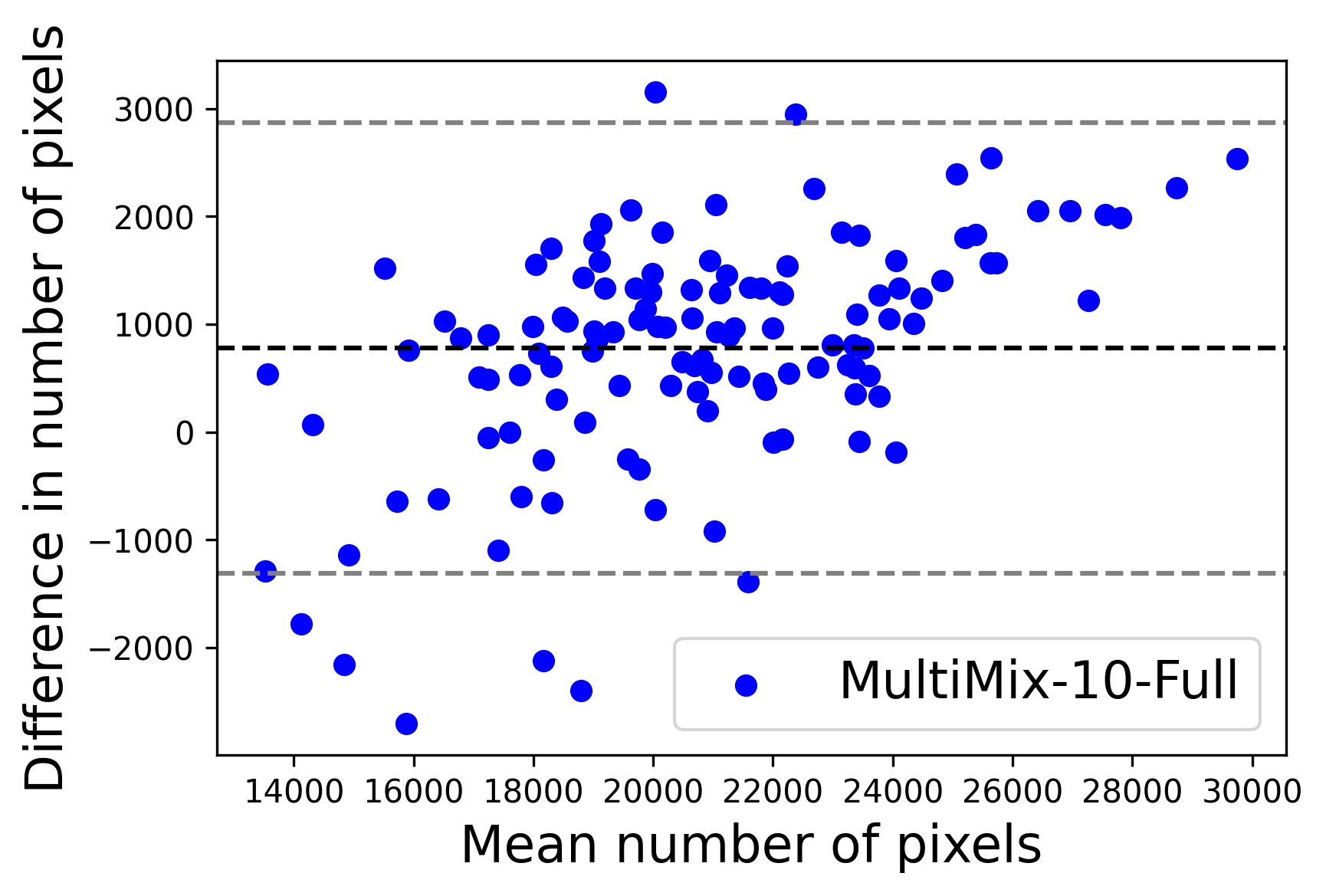} \\
  \includegraphics[width=0.28\linewidth]{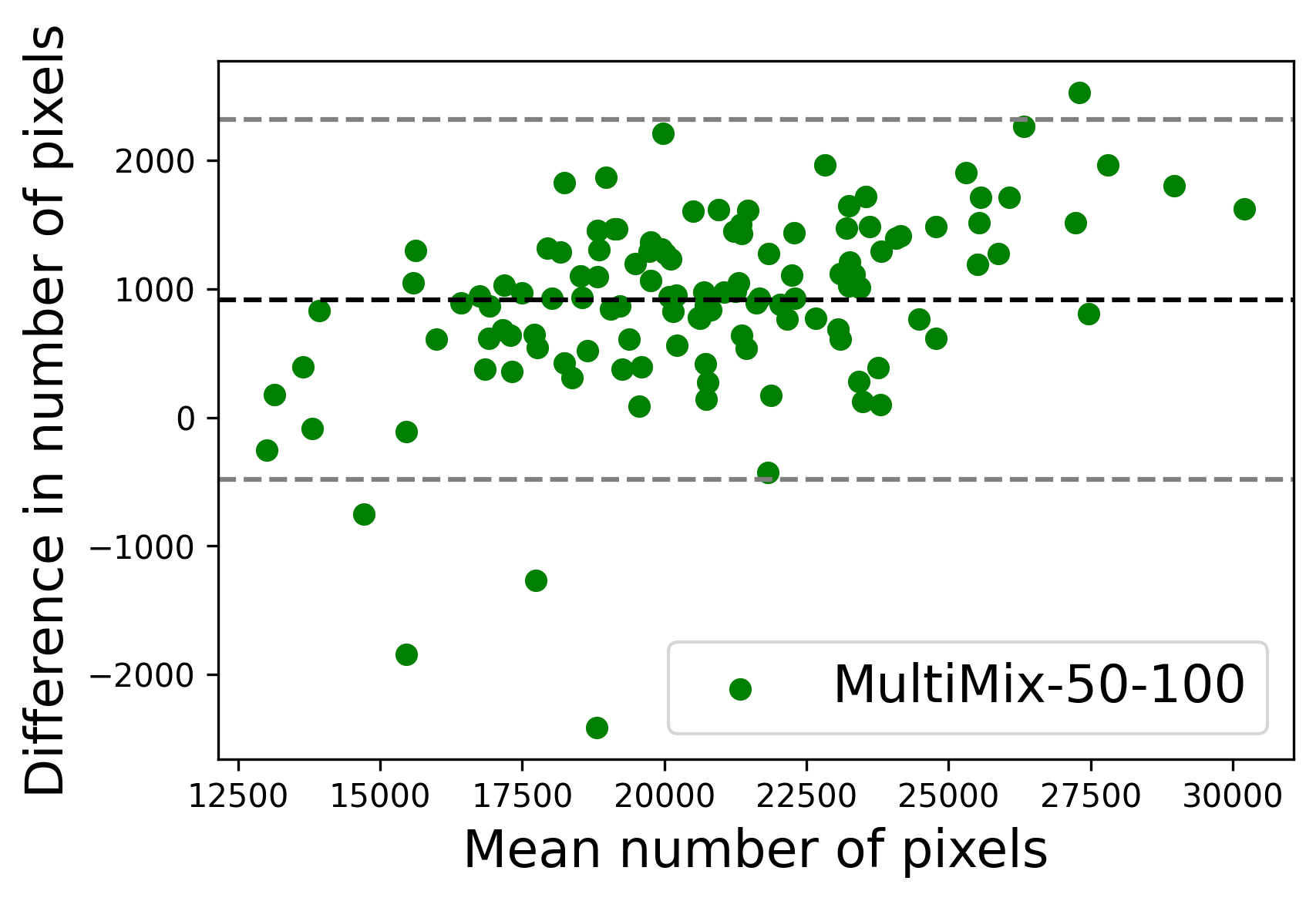}\quad
  \includegraphics[width=0.28\linewidth]{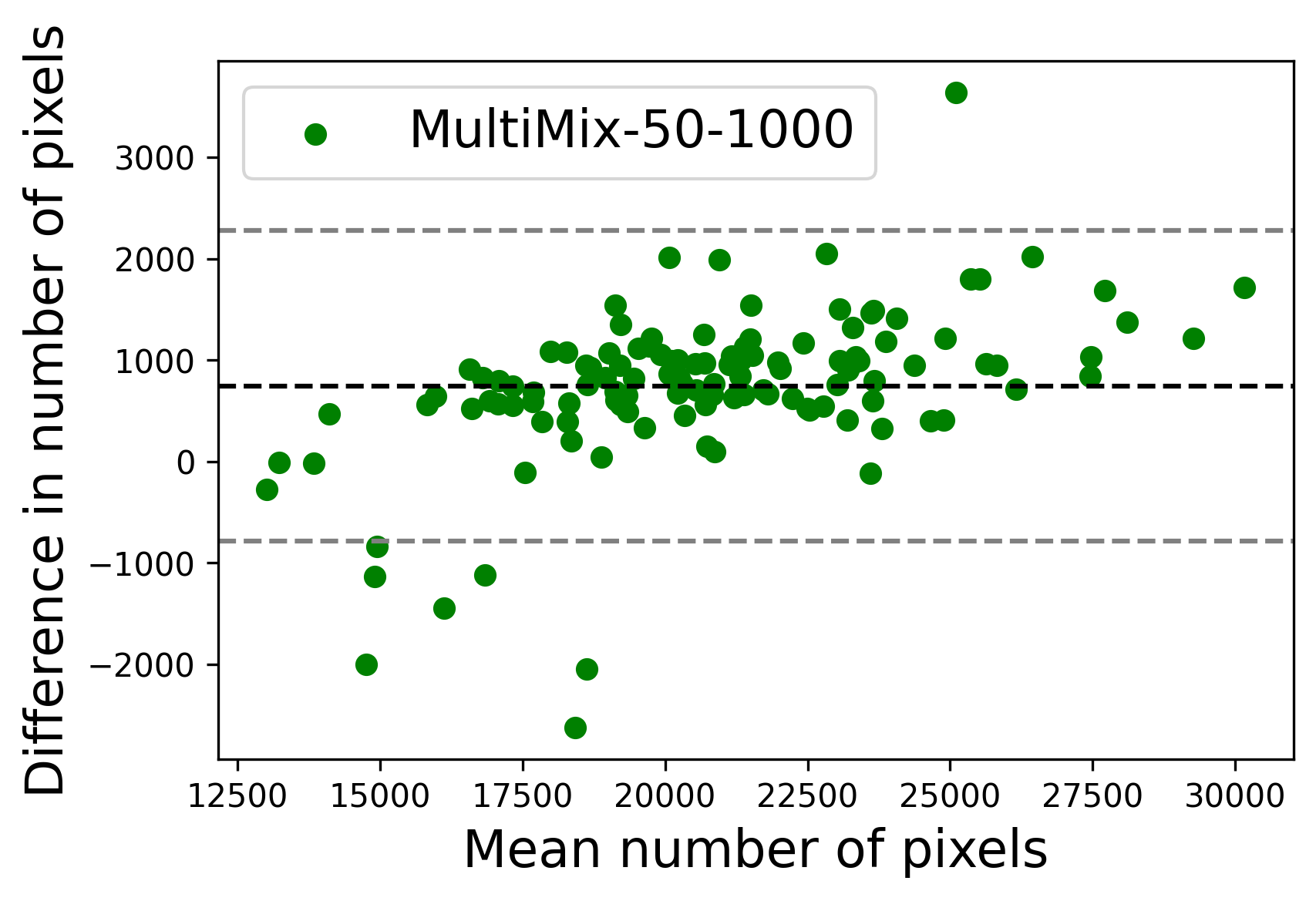}\quad
  \includegraphics[width=0.28\linewidth]{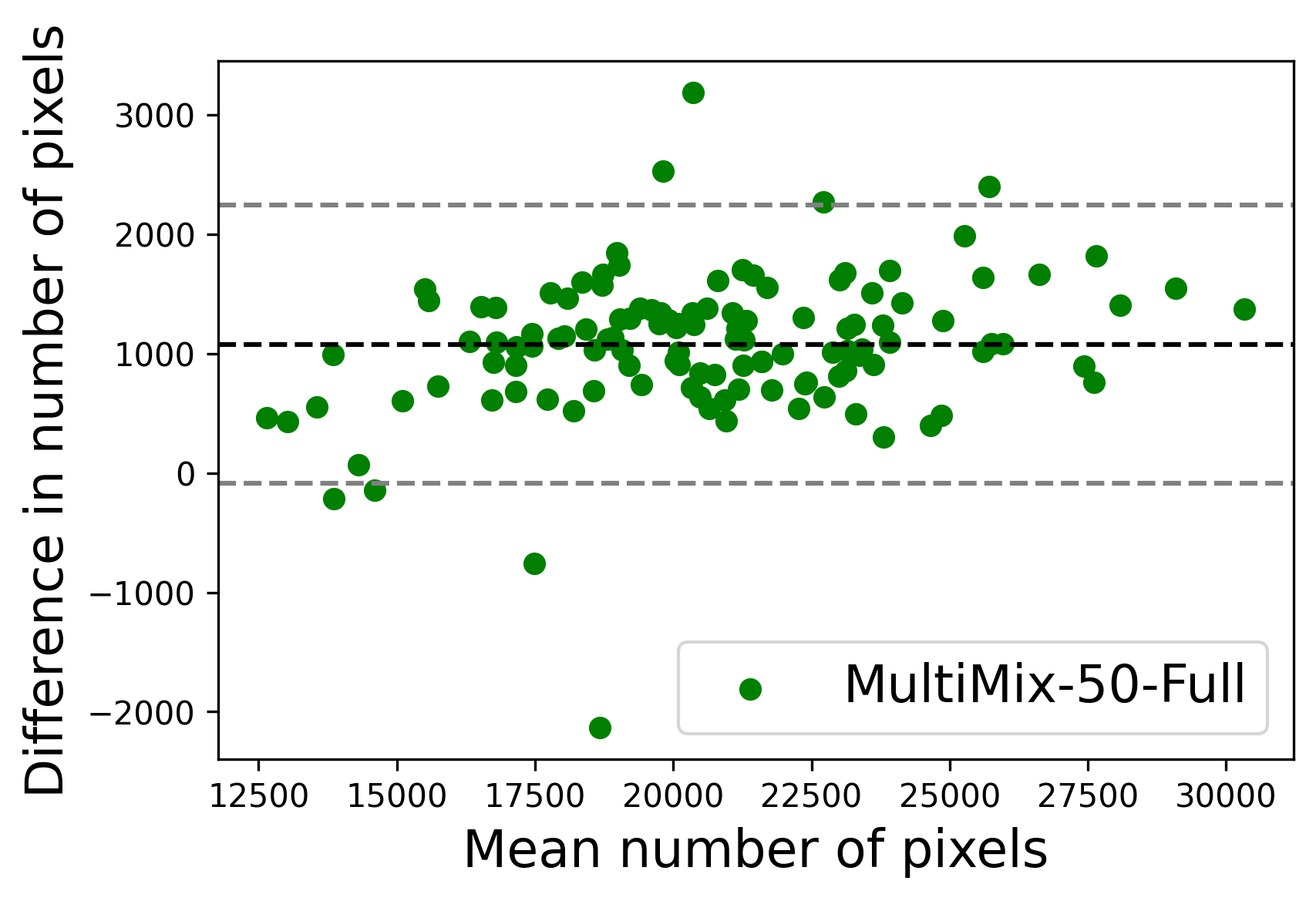} \\
  \includegraphics[width=0.28\linewidth]{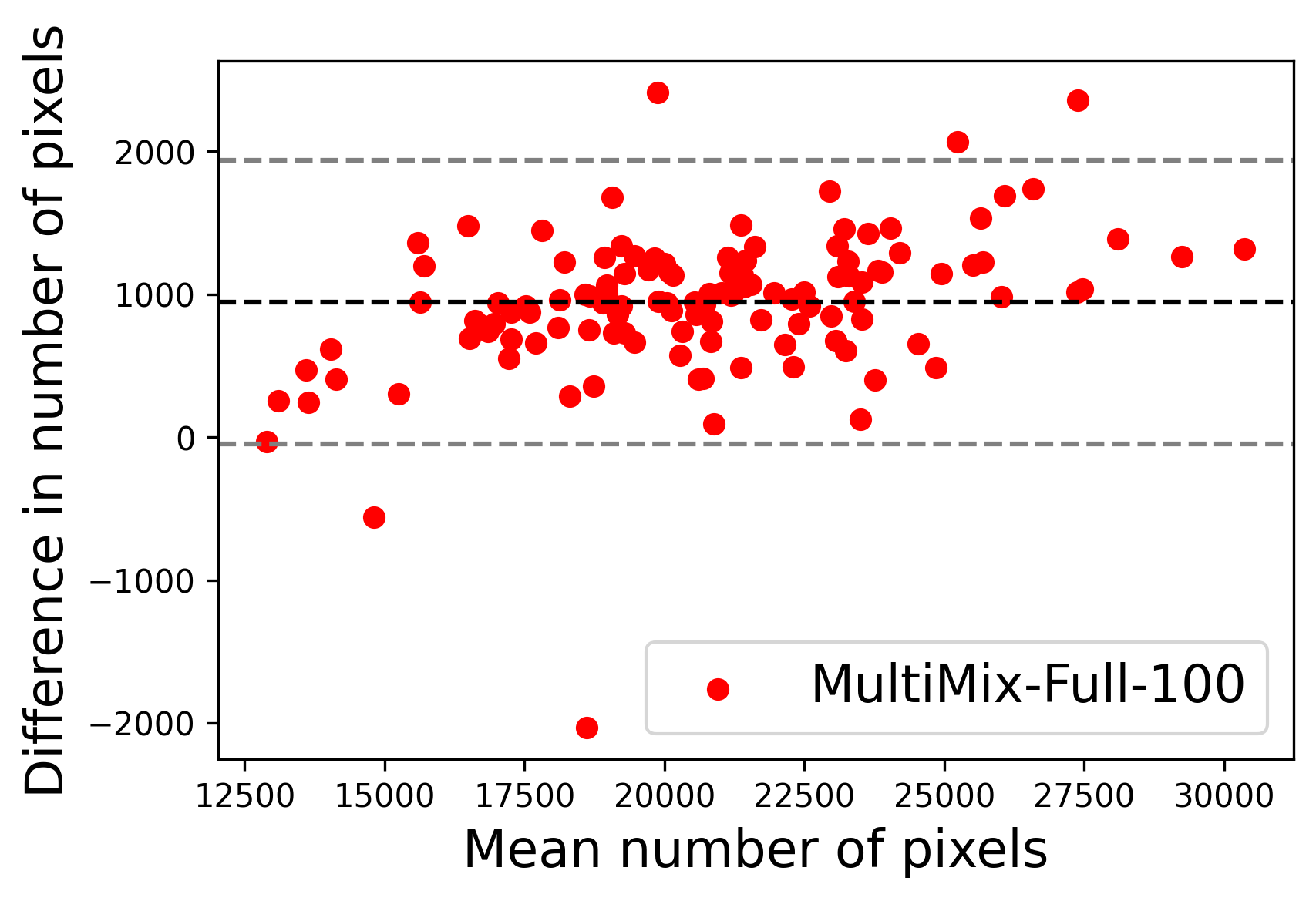}\quad
  \includegraphics[width=0.28\linewidth]{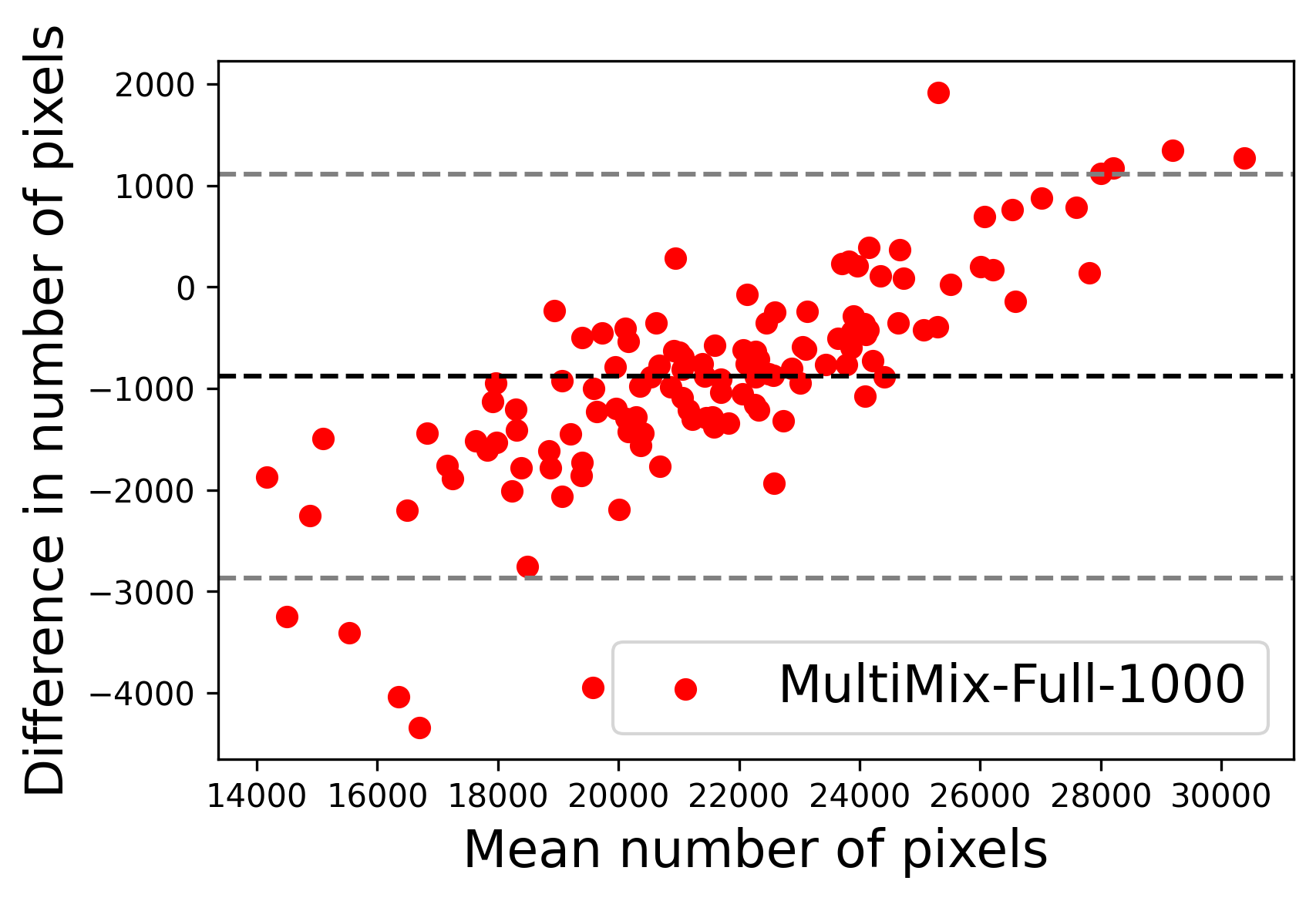}\quad
  \includegraphics[width=0.28\linewidth]{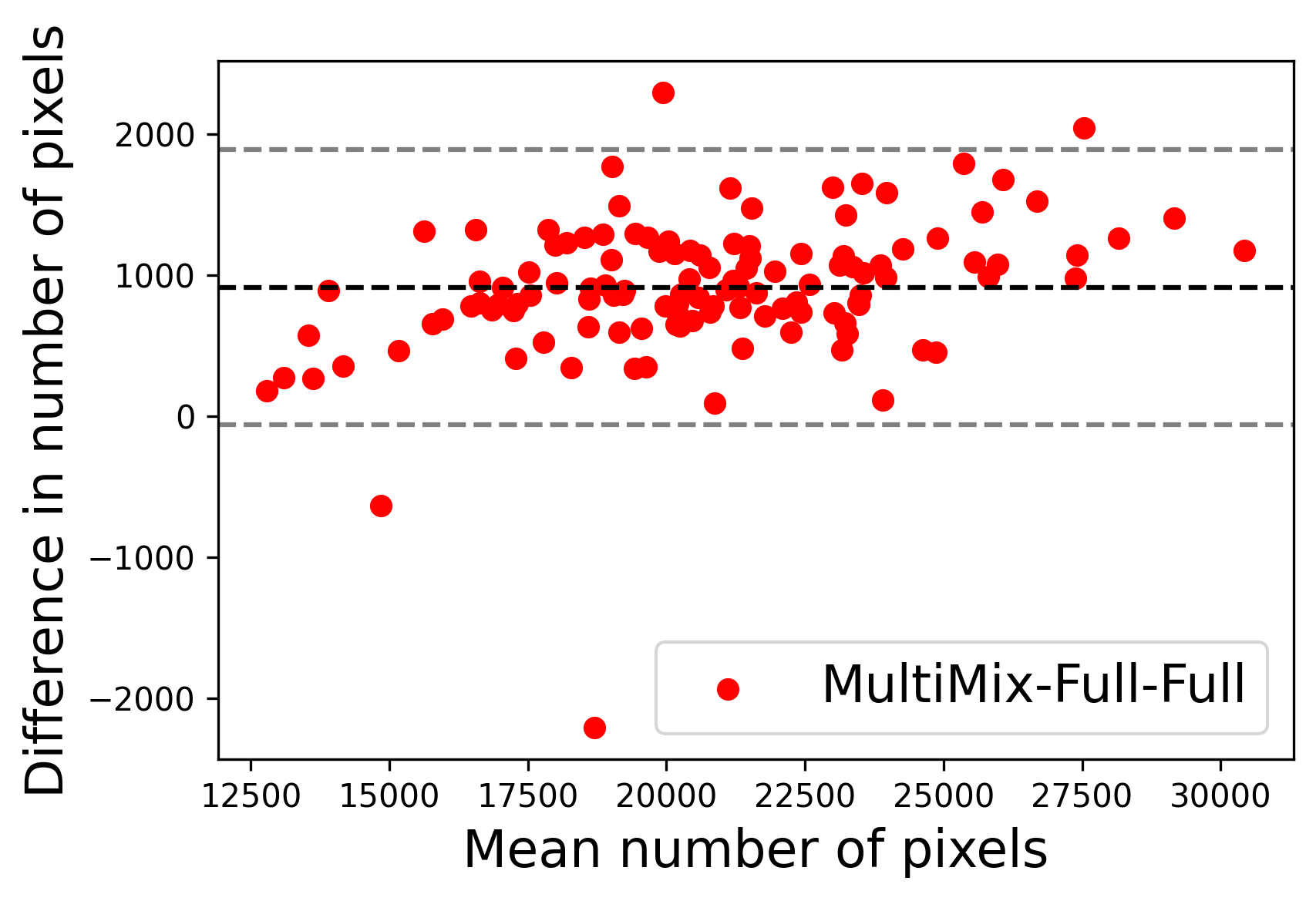}
 \end{tabular}
 } \\[10pt]
\subcaptionbox{cross-domain \label{fig:altman-cross}}{
 \begin{tabular}{ccc} \setlength{\tabcolsep}{1pt}
  \includegraphics[width=0.28\linewidth]{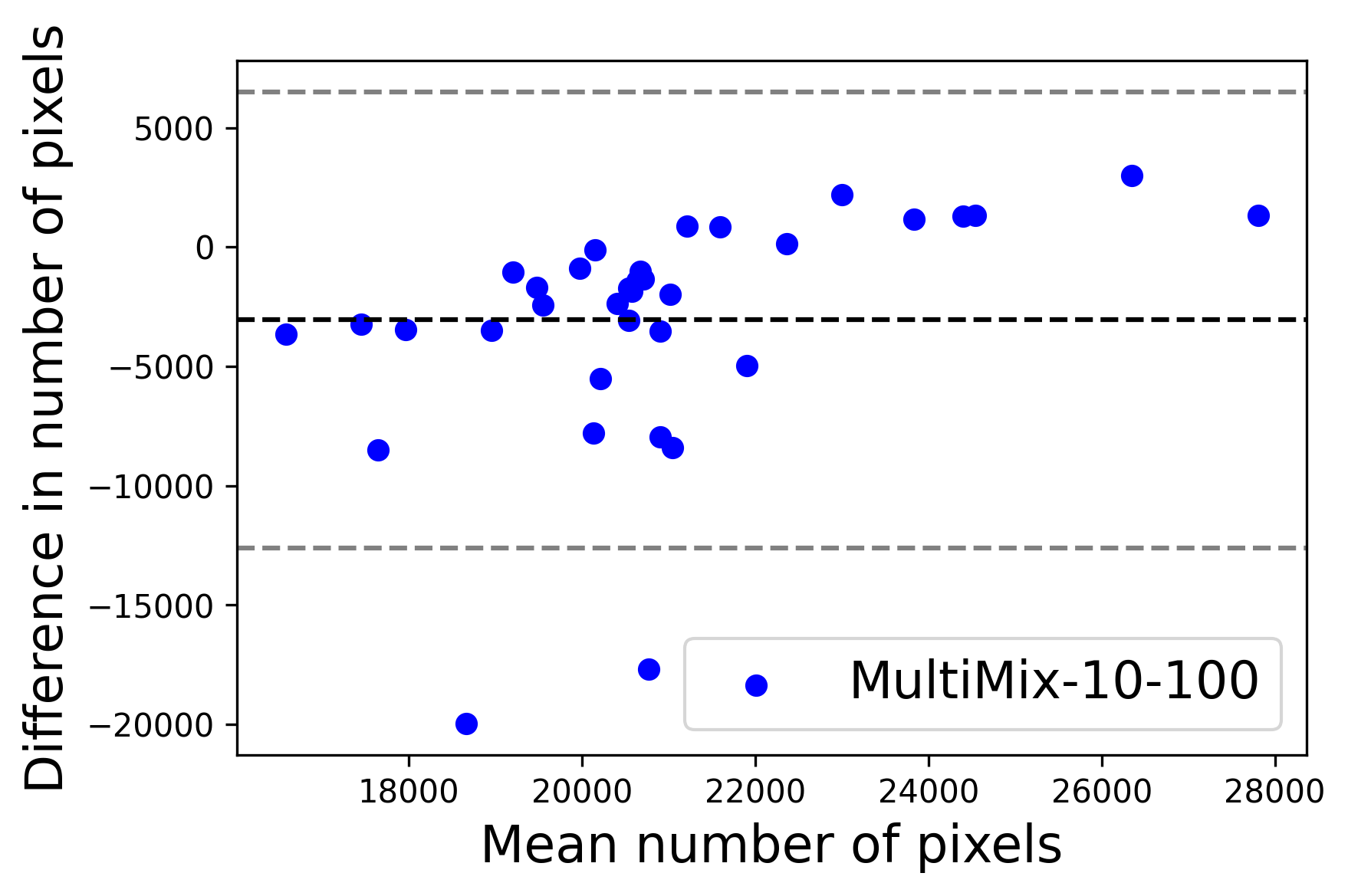}\quad
  \includegraphics[width=0.28\linewidth]{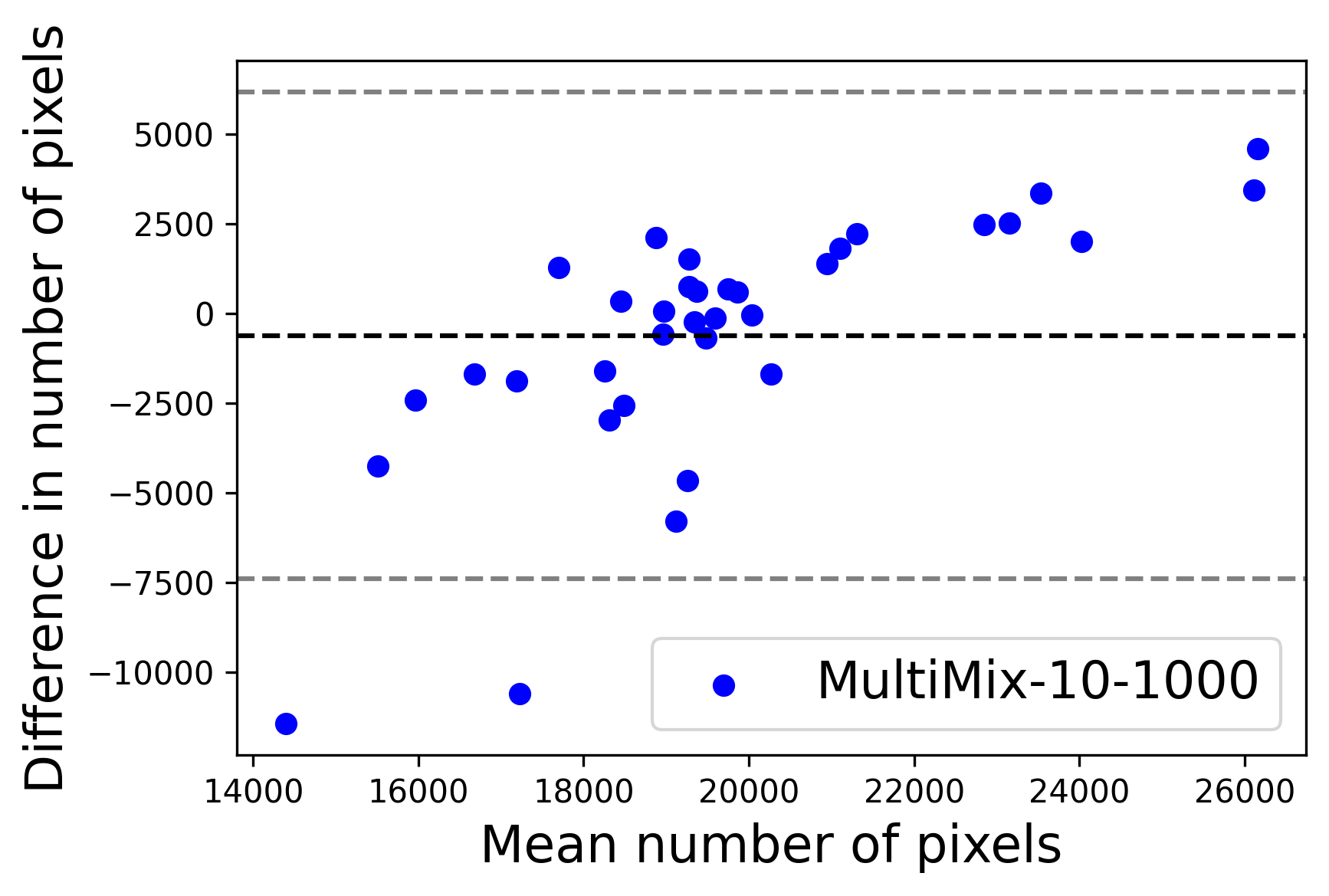}\quad
  \includegraphics[width=0.28\linewidth]{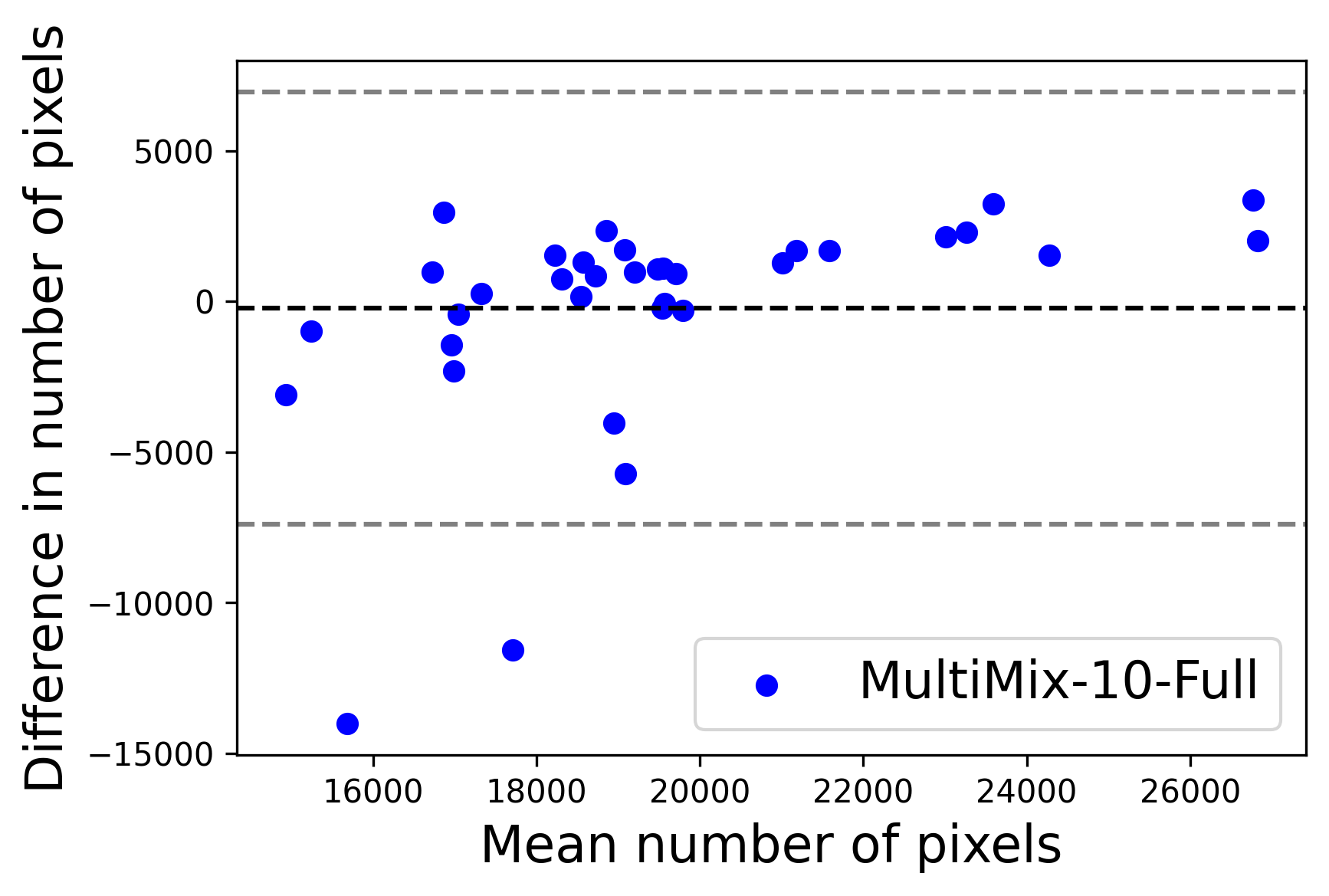} \\
  \includegraphics[width=0.28\linewidth]{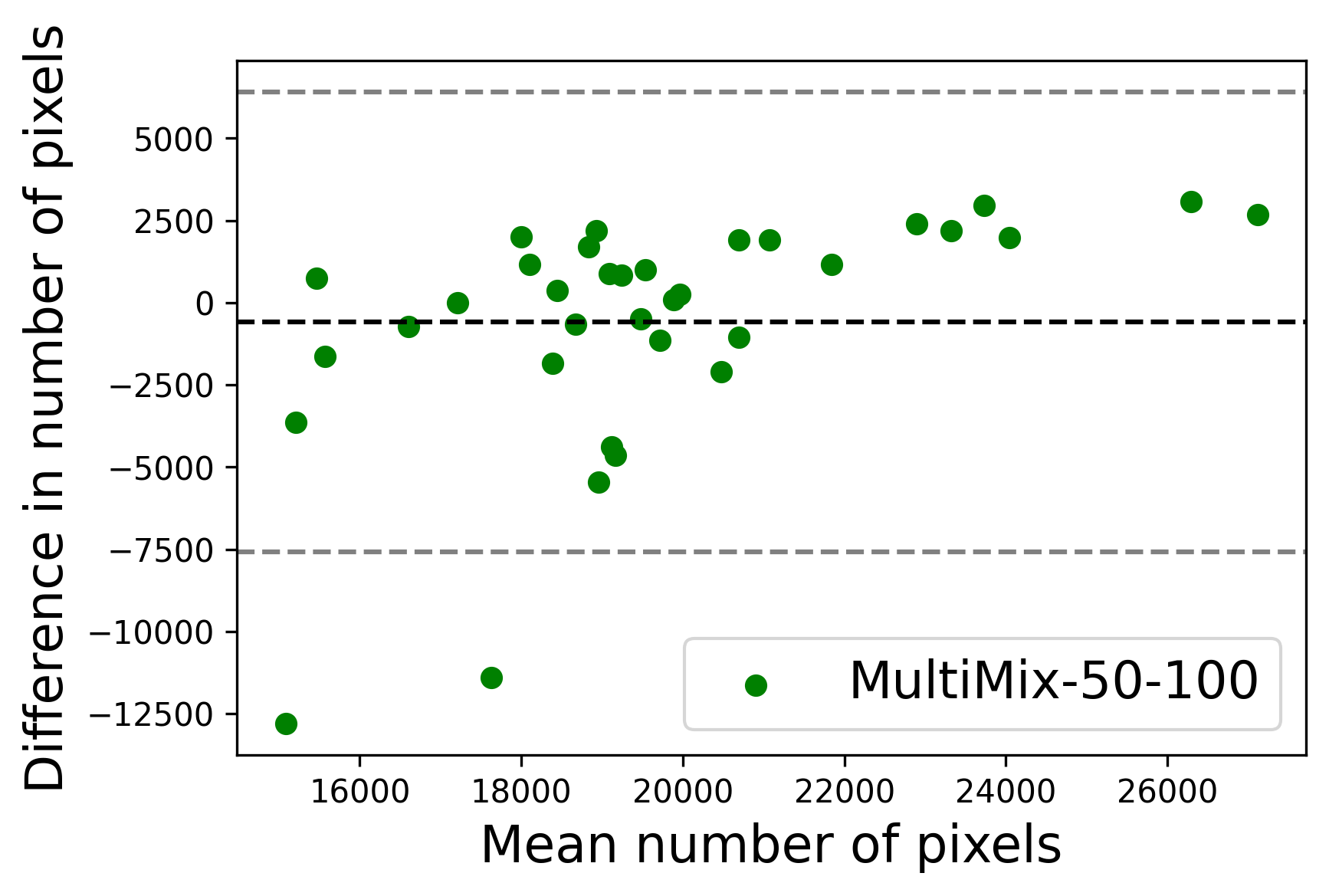}\quad
  \includegraphics[width=0.28\linewidth]{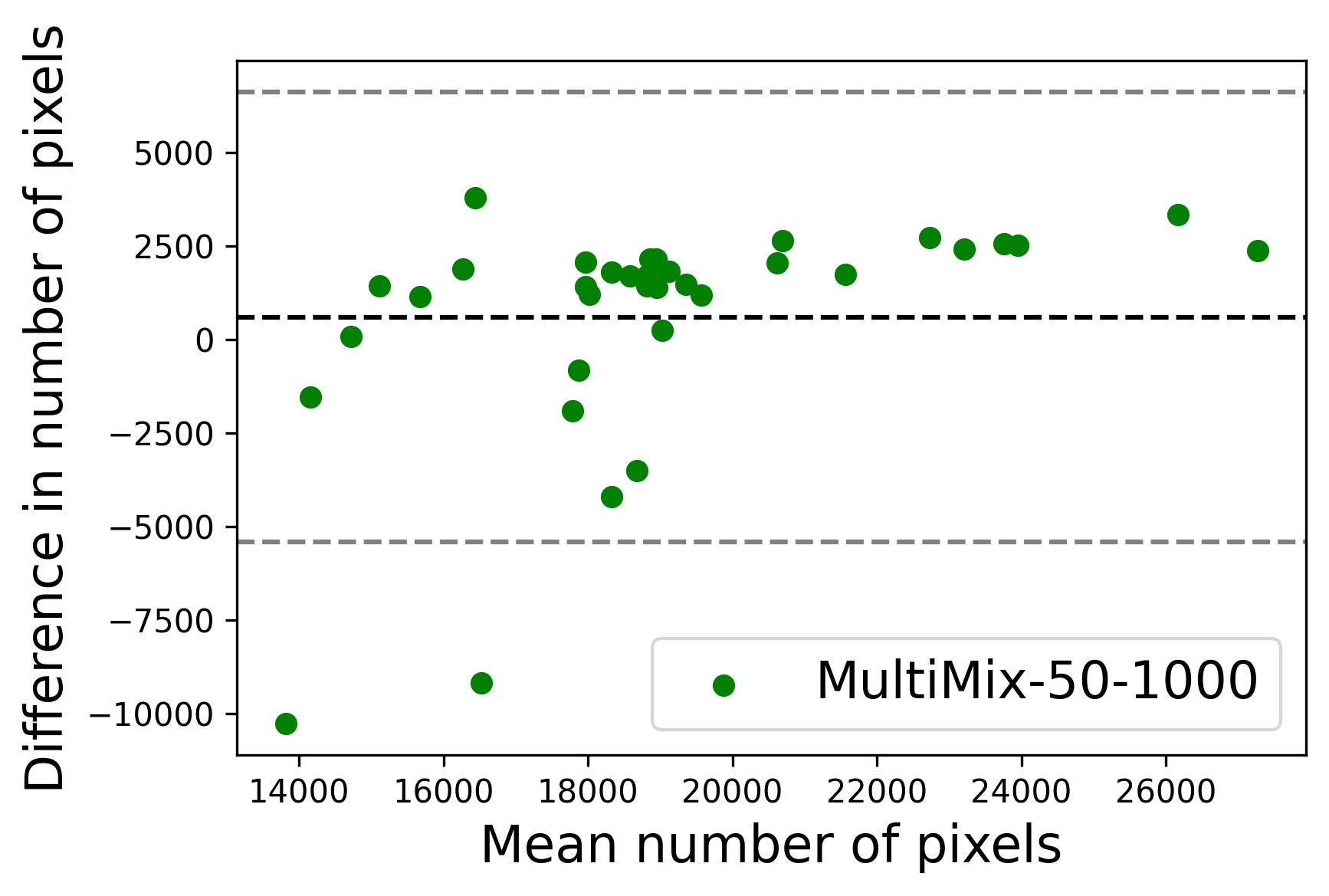}\quad
  \includegraphics[width=0.28\linewidth]{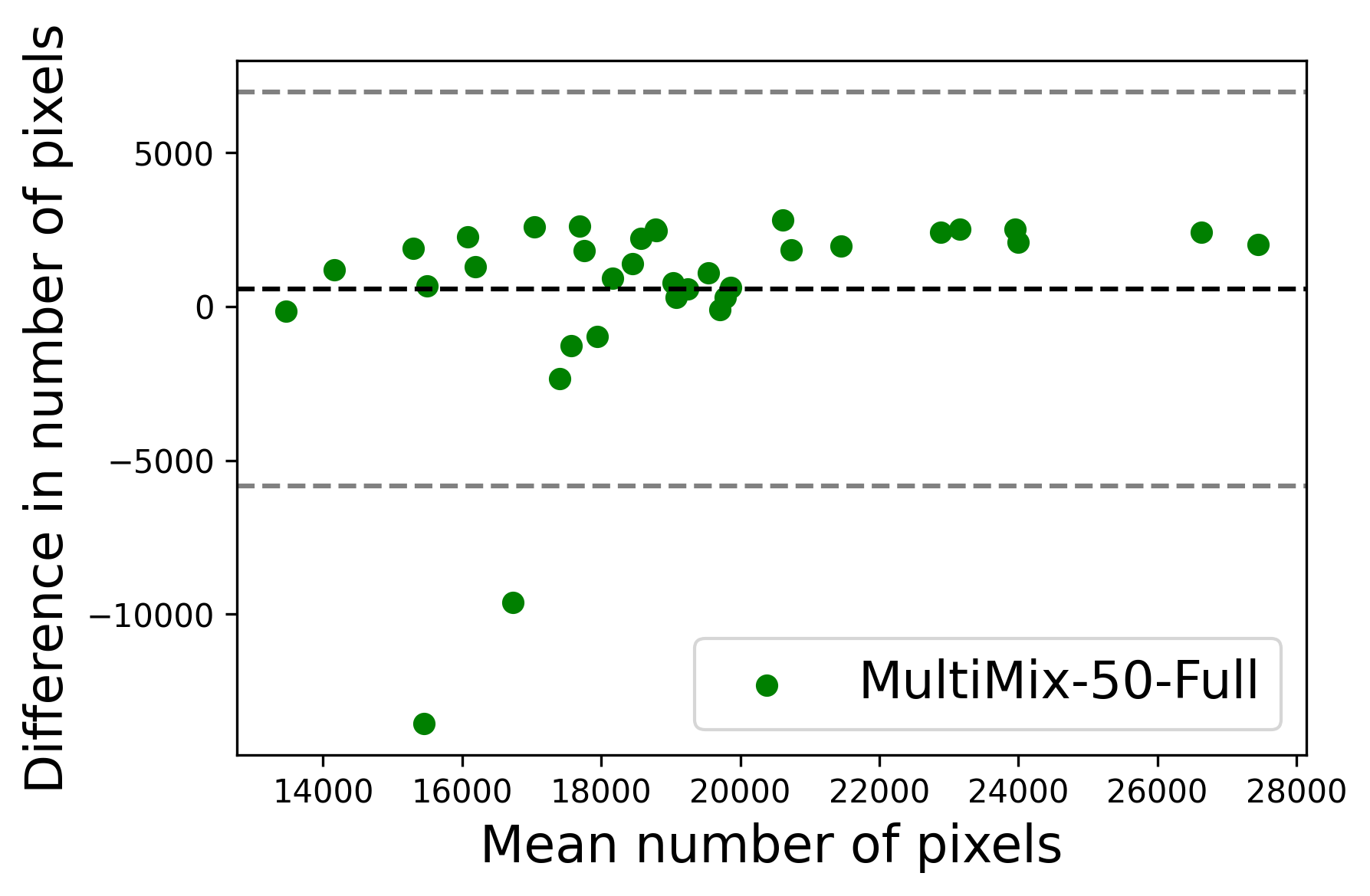} \\
  \includegraphics[width=0.28\linewidth]{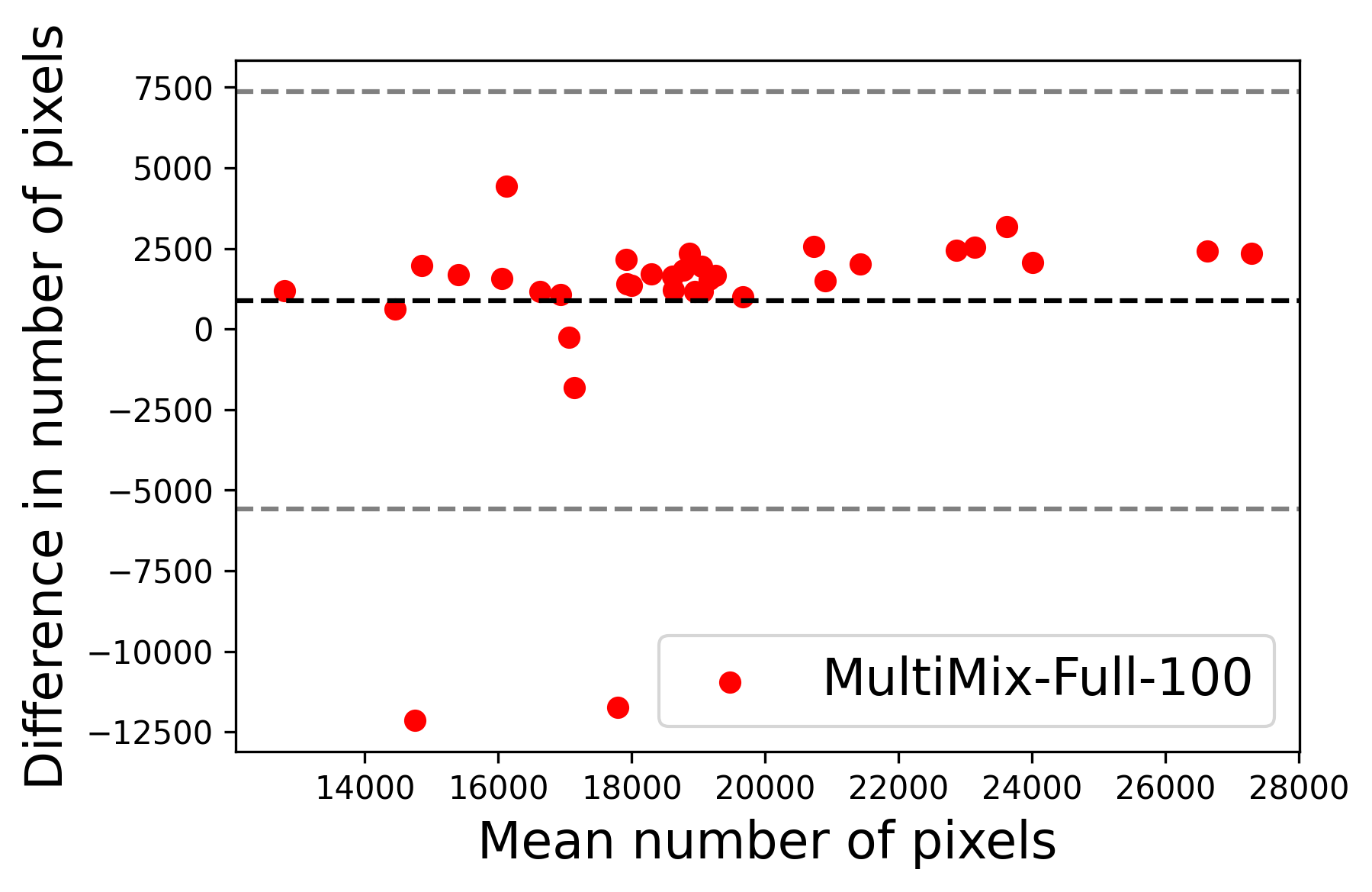}\quad
  \includegraphics[width=0.28\linewidth]{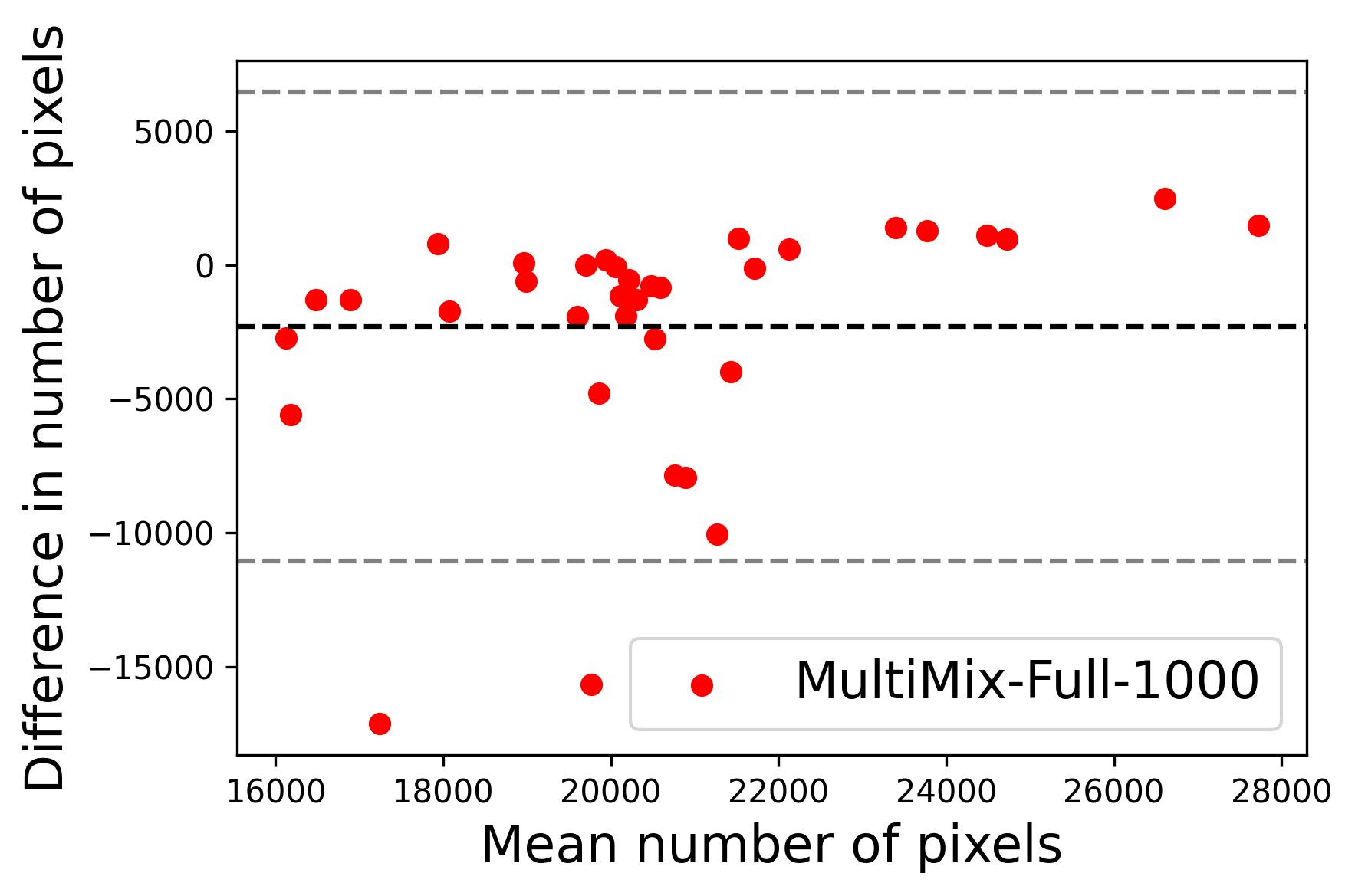}\quad
  \includegraphics[width=0.28\linewidth]{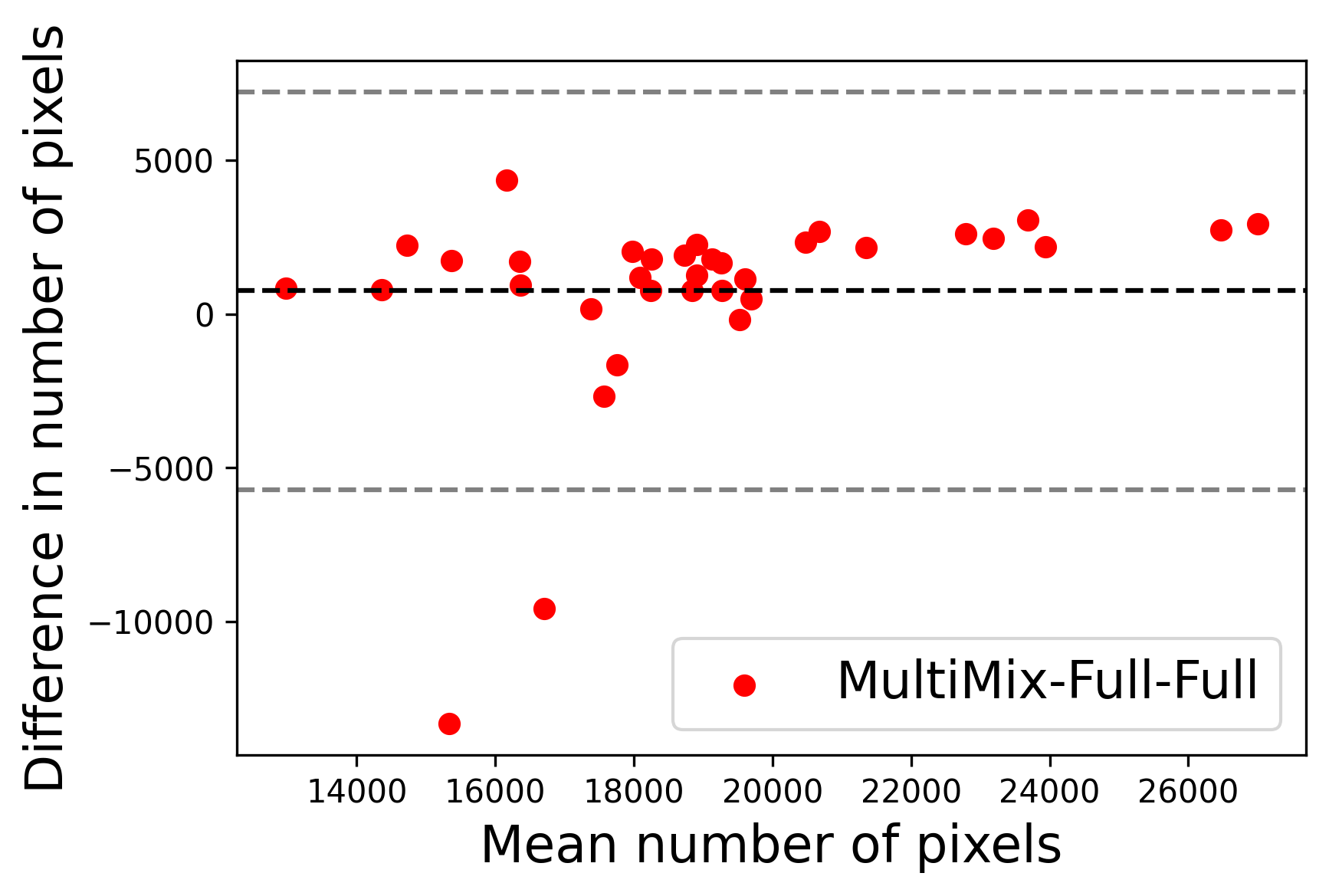}
 \end{tabular}
}
\caption{Bland-Altman plots at varying training labels show good agreement between the number of ground truth pixels and MultiMix-predicted pixels for the (a) in-domain and (b) cross-domain evaluations, as well as consistent improvement with increasing quantities of labeled data.}
\label{fig:altman}
\end{figure}

\subsection{Results and Discussion}

As is revealed by the results in Table~\ref{table:results-in}, the performance of our model improves with the inclusion of each of the novel components in the backbone network. For the classification task, our confidence-based augmentation approach for semi-supervised learning yields significantly improved performance compared to the baseline models. Even with the $\min|\mathcal{D}^c_l|$ and $\min|\mathcal{D}^s_l|$, our MultiMix-100-10 model outperforms the fully-supervised baseline (Enc) in classifying the normal and abnormal chest X-rays. As is confirmed by the Student's t-test, MultiMix exhibits significant improvements over the classification baselines Enc, Enc-SSL, and UMTL ($p < 0.05$).

For the segmentation task, the inclusion of the saliency bridge module yields large improvements over the baseline U-Net and UMTL models. Again, with $\min|\mathcal{D}^s_l|$, we observed a 30\% performance gain over its counterparts, which proves the effectiveness of our MultiMix model. The improvement in Dice scores of MultiMix with minimal supervision over the segmentation baselines of U-Net, UMTL, and UMTL-S is statistically significant ($p < 0.05$), confirming the quantitative efficacy of MultiMix. Figure~\ref{fig:consistency} shows improved and consistent segmentation performance by the MultiMix model over the baselines. For a fair comparison, we used the same backbone U-Net and the same classification branch for all the models. 

In Figure~\ref{fig:lung_vis}, the segmented lung boundary visualizations also show good agreement with the reference masks by MultiMix over the other models (also see Appendix~\ref{app:seg-vis}). For both the in-domain and cross-domain segmentations, we observe that the predicted boundaries are almost identical with the reference boundaries, as they substantially overlap. Moreover, the noise in the predictions is mitigated with the introduction of each additional component into the intermediate models, which justifies the value of those components in the MultiMix model. The good agreement between the ground truth lung masks and the MultiMix predicted segmentation masks is confirmed by the Bland-Altman plots for varying quantities of labeled data, shown in Figure~\ref{fig:altman-in}.

The generalization test through the cross-domain datasets (MCU and NIHX) demonstrates the effectiveness of the MultiMix model. It consistently performs well against both domains with improved generalizability in either task. As reported in Table~\ref{table:results-cross}, the performance of MultiMix is as promising as in the in-domain evaluations. MultiMix achieved better scores in the classification task over all the baseline models. Due to the significant differences in the NIHX and CheX datasets, the scores are not as good as the in-domain results, yet our model performs significantly better than the other classification models Enc, Enc-SSL, and UMTL ($p < 0.05$). For the segmentation task, our MultiMix model again achieved better scores in all the various metrics, with improved consistency over the baselines (Figure~\ref{fig:consistency}). Just like for the in-domain results, MultiMix shows significant improvements in Dice scores over the segmentation baselines U-Net, UMTL, and UMTL-S ($p < 0.05$), thus proving the generalizability of our method.
The Bland-Altman plots in Figure~\ref{fig:altman-cross} for the MultiMix model in the cross-domain segmentation evaluations with varying quantities of labeled data confirms the observed good agreement between the ground truth lung segmentation masks and the MultiMix-predicted segmentation masks.

\begin{figure} \centering
\subcaptionbox{in-domain}{\includegraphics[width=0.49\linewidth, trim={0 0 40 0},clip]{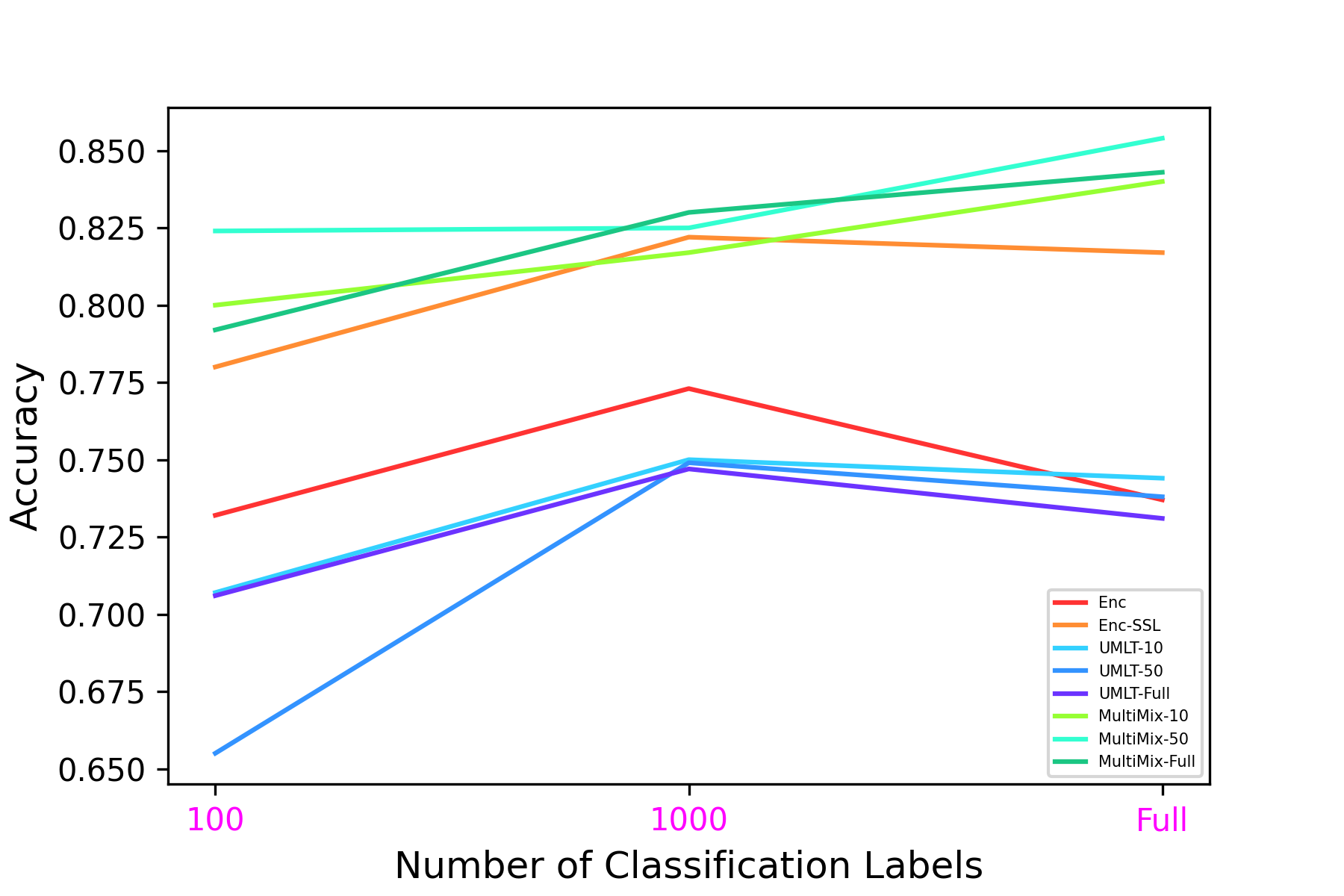}} \hfill
\subcaptionbox{cross-domain}{\includegraphics[width=0.49\linewidth, trim={0 0 40 0},clip]{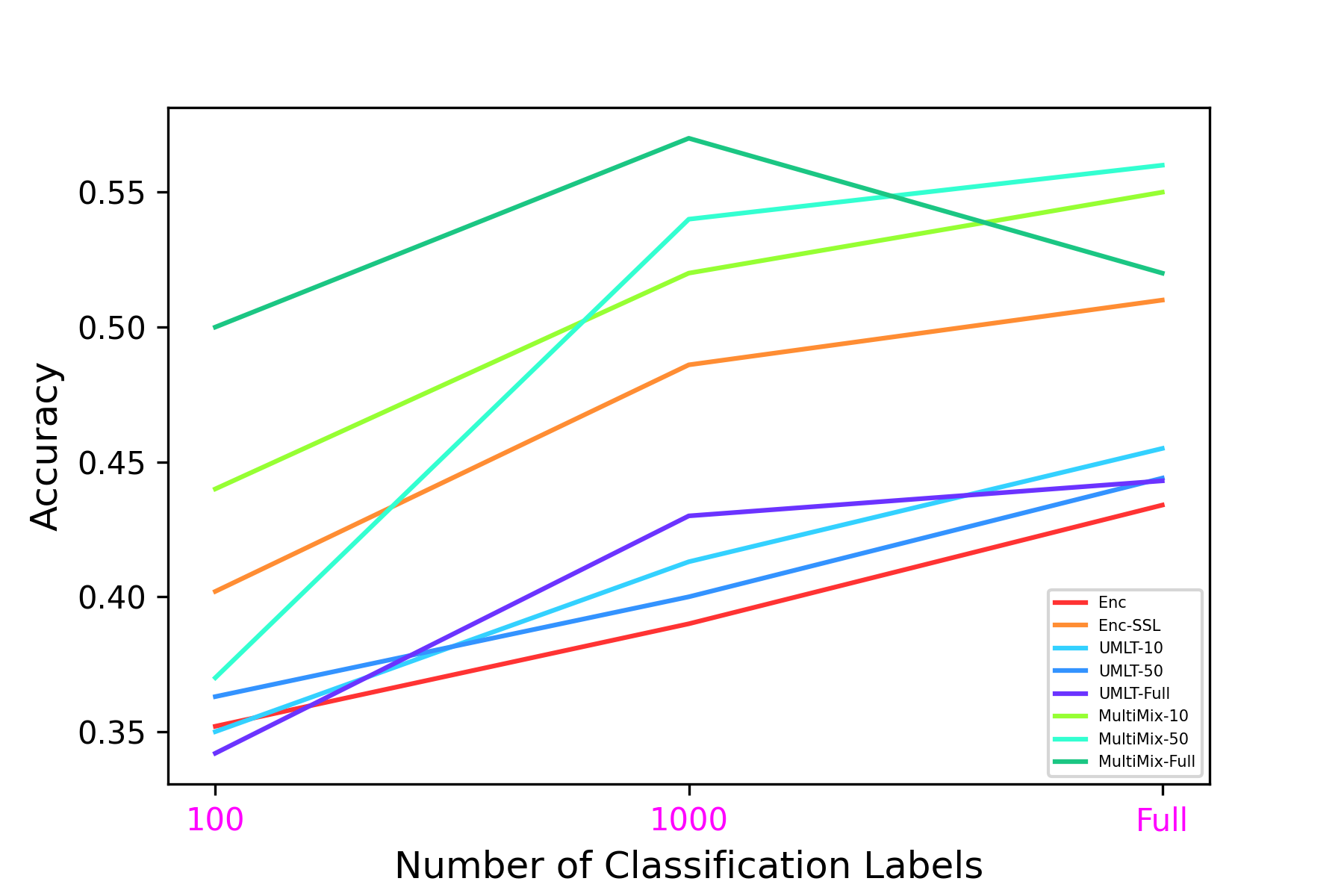}}
\caption{Classification accuracies of different supervised and semi-supervised baselines at different training datasizes. The in-domain and cross-domain plots show that MultiMix has higher accuracy and consistency over the baselines.}
\label{fig:acc_increasing}
\end{figure}

\begin{figure} \centering
\subcaptionbox{in-domain}{\includegraphics[width=0.49\linewidth]{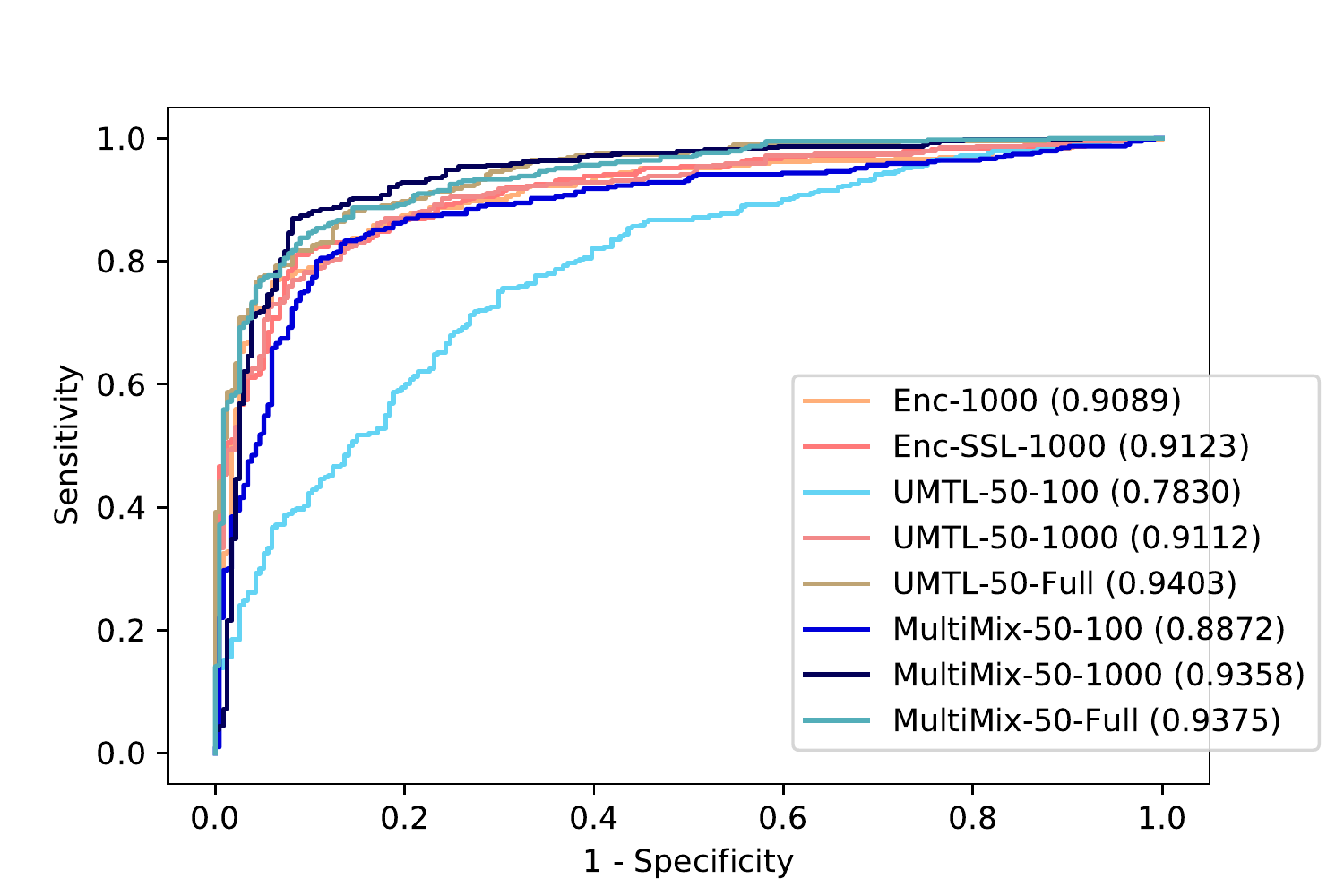}} \hfill
\subcaptionbox{cross-domain}{\includegraphics[width=0.49\linewidth]{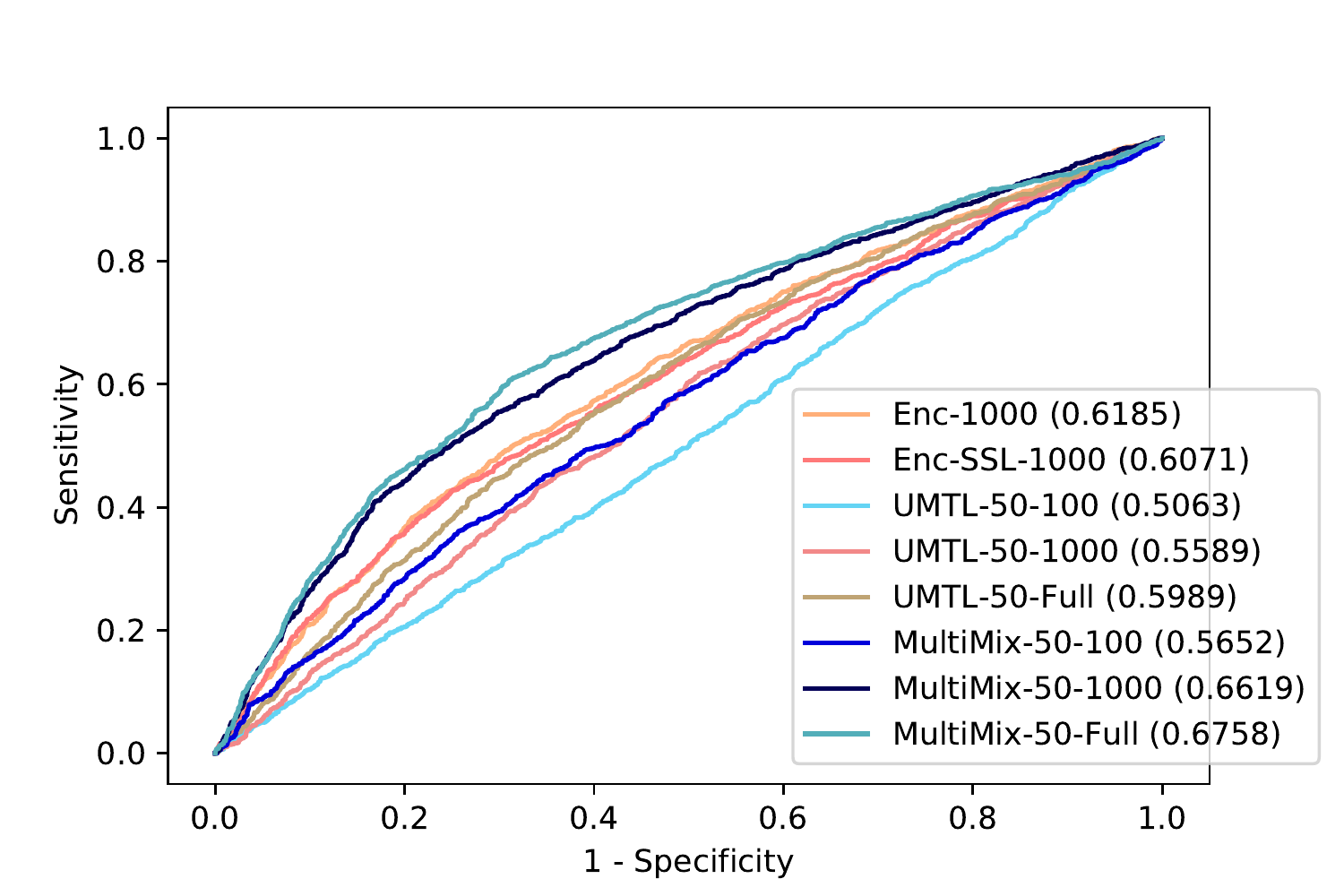}}
\caption{ROC curves for supervised and semi-supervised baselines with 50 segmentation labels show higher AUC values from our MultiMix for in-domain and cross-domain evaluations.}
\label{fig:ROC}
\end{figure}

Figure~\ref{fig:acc_increasing} demonstrates the superiority and better consistency of our MultiMix models over the baselines in classifying normal and abnormal (pneumonia) X-rays. Figure~\ref{fig:ROC} further showcases the superior classification performance of MultiMix over the baseline single-task and multi-task models. Figure~\ref{fig:acc_increasing} and Figure~\ref{fig:ROC} together show that MultiMix outperforms all baselines in many different metrics, as the Accuracy and Area Under Curve (AUC) values confirm the superiority of the MultiMix model. With regard to the cross-domain ROC curves, however, although MultiMix has the best relative performance when compared to the baselines, the absolute performance of the algorithm indicates room for improvement.
     
\section{Conclusions}

We have presented MultiMix, a novel semi-supervised, multi-task learning model that jointly learns classification and segmentation tasks. Through the incorporation of confidence-guided data augmentation and a novel saliency bridge module, MultiMix performs improved and consistent pneumonia detection and lung segmentation when trained on multi-source chest X-ray datasets with varying quantities of ground truth labels. Our thorough experimentation using four different chest X-ray datasets demonstrated the effectiveness of MultiMix both in in-domain and cross-domain evaluations, for both tasks; in fact, outperforming a number of baseline models. 

Beyond chest X-rays---which is the most frequently performed radiologic procedure worldwide, comprising 40\% of all imaging tests, or 1.4 billion annually \citep{who2016communicating}---our future work will focus on generalizing the MultiMix concept, particularly the saliency bridge module, to other applications and imaging modalities, including volumetric images.

\ethics{Appropriate ethical standards were maintained in writing this manuscript and conducting the reported research, following all applicable laws and regulations regarding the treatment of animals or human subjects.}

\coi{DT is a founder of VoxelCloud, Inc., Los Angeles, CA, USA.}

\newpage

\appendix

\section{Model Architecture}
\label{app:architecture}

Architectural details of the MultiMix model are presented in Table~\ref{tab:arch-encoder} for the encoder network and in Table~\ref{tab:arch-decoder} for the decoder network. The encoder and decoder incorporate double-convolution blocks; the encoder has 5 blocks and the decoder has 4. Each block includes a 2D convolutional layer, an instance normalization layer, and a Leaky ReLU activation, and this sequence repeats in each block. 

In the encoder, each double-convolution block is followed by a dropout layer and a maxpooling layer. The encoder finally branches to a classification branch, which includes a maxpooling layer (5), an average pooling layer, followed by a fully-connected layer for classification prediction.

The decoder begins with an upsampling layer. Next, in the first double-convolution layer, the downsampled saliency maps and original inputs are concatenated. The increase in dimensions at the beginning of each decoder block are due to the skip connections. These convolutional layers are also followed by a dropout layer. This sequence is repeated for 3 more layers. To output the final segmentation prediction, the decoder finishes with a single convolutional layer that downsamples to a single channel. 

\section{Segmentation Visualization}
\label{app:seg-vis}

\begin{figure} \centering
\subcaptionbox{JSRT (in-domain)}{
  \begin{tabular}{ccc}
  \small Image & \small Ground Truth & \small Prediction \\
    \includegraphics[width=0.129\linewidth, trim={4cm 1cm 3cm 1cm},clip]{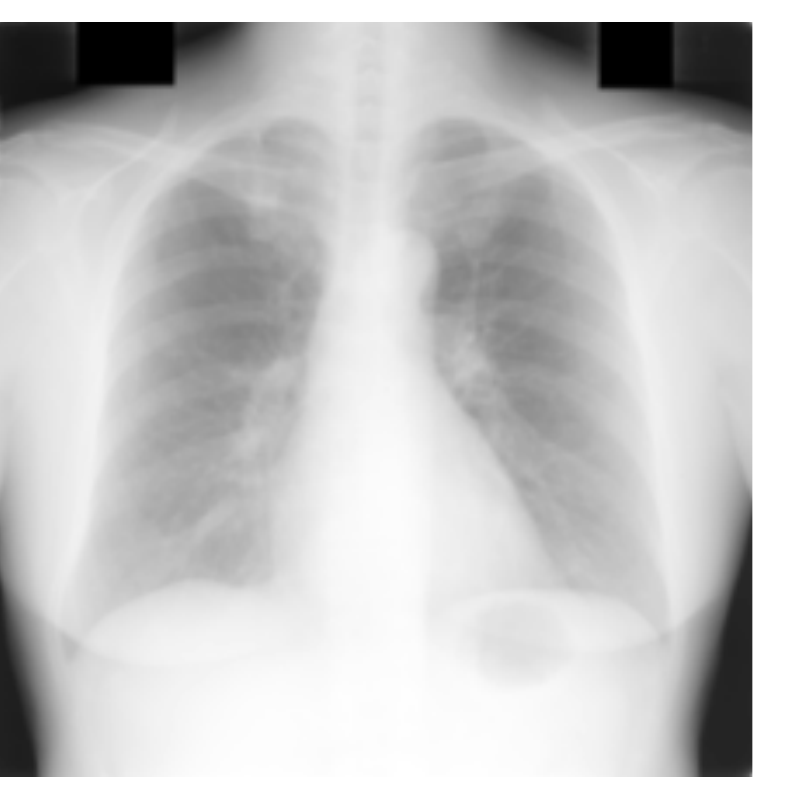} &
    \includegraphics[width=0.129\linewidth, trim={4cm 1cm 3cm 1cm},clip]{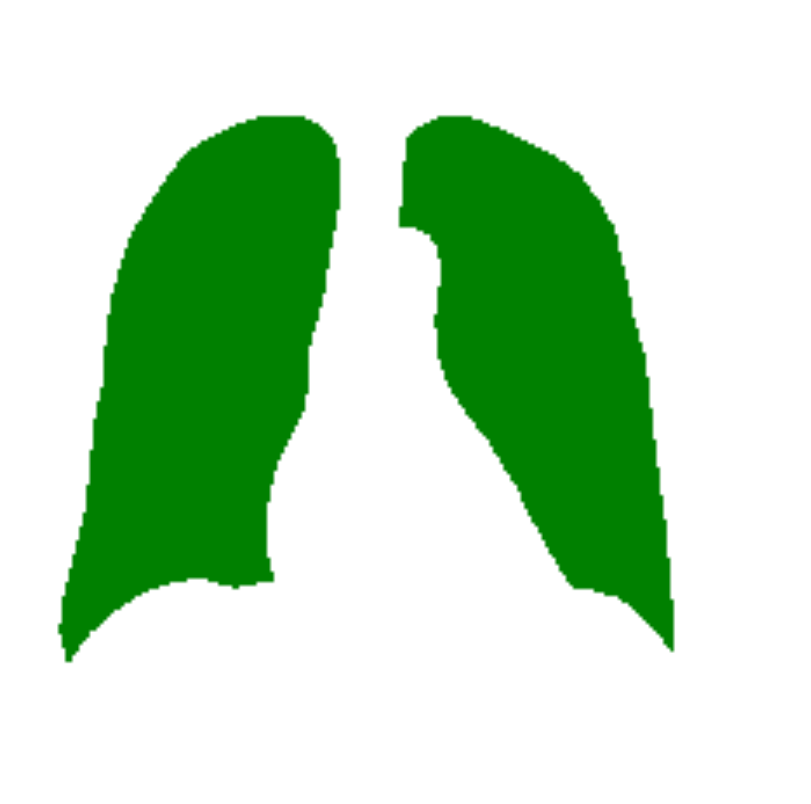} &
    \includegraphics[width=0.129\linewidth, trim={4cm 1cm 3cm 1cm},clip]{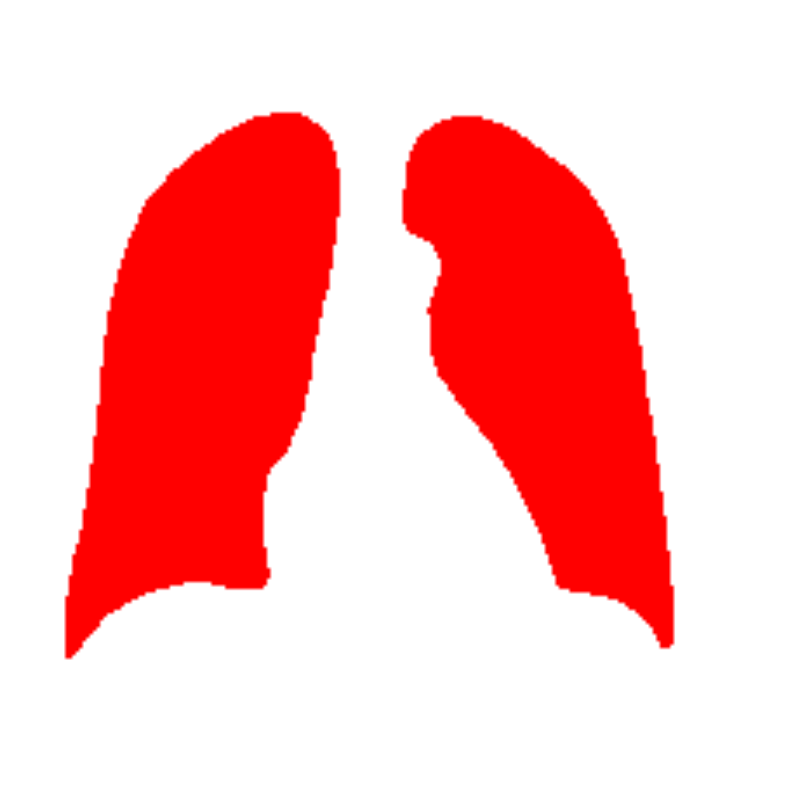}
    \\
    \includegraphics[width=0.129\linewidth, trim={4cm 1cm 3cm 1cm},clip]{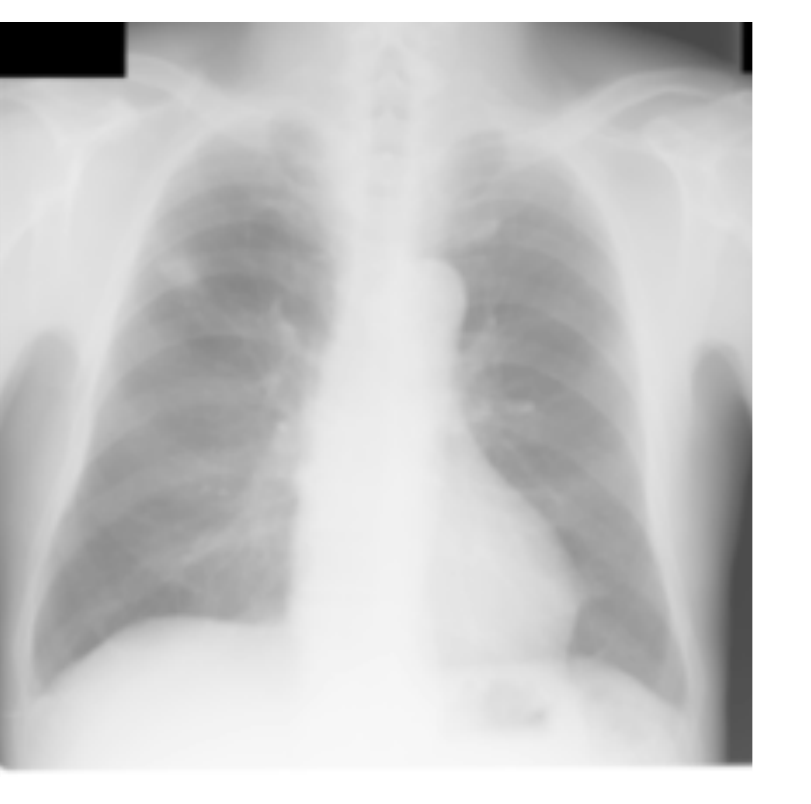} &
    \includegraphics[width=0.129\linewidth, trim={4cm 1cm 3cm 1cm},clip]{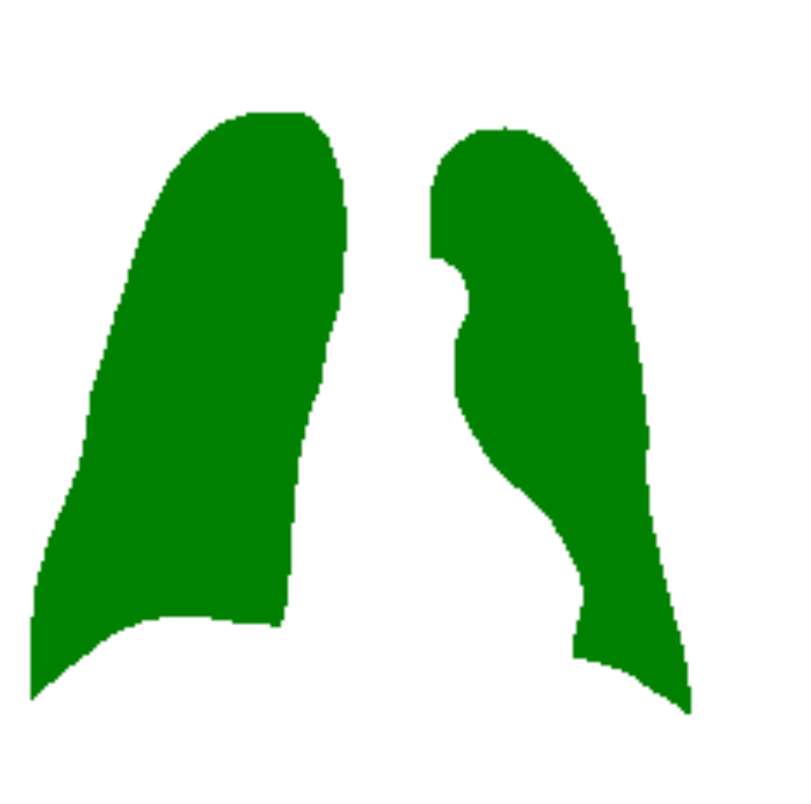} &
    \includegraphics[width=0.129\linewidth, trim={4cm 1cm 3cm 1cm},clip]{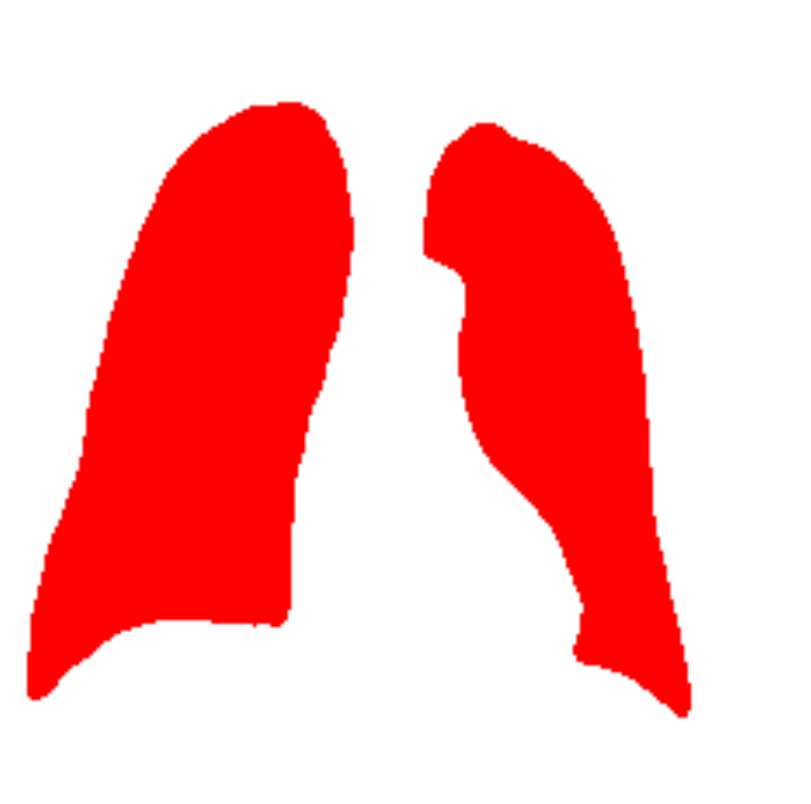}
    \\
    \includegraphics[width=0.129\linewidth, trim={4cm 1cm 3cm 1cm},clip]{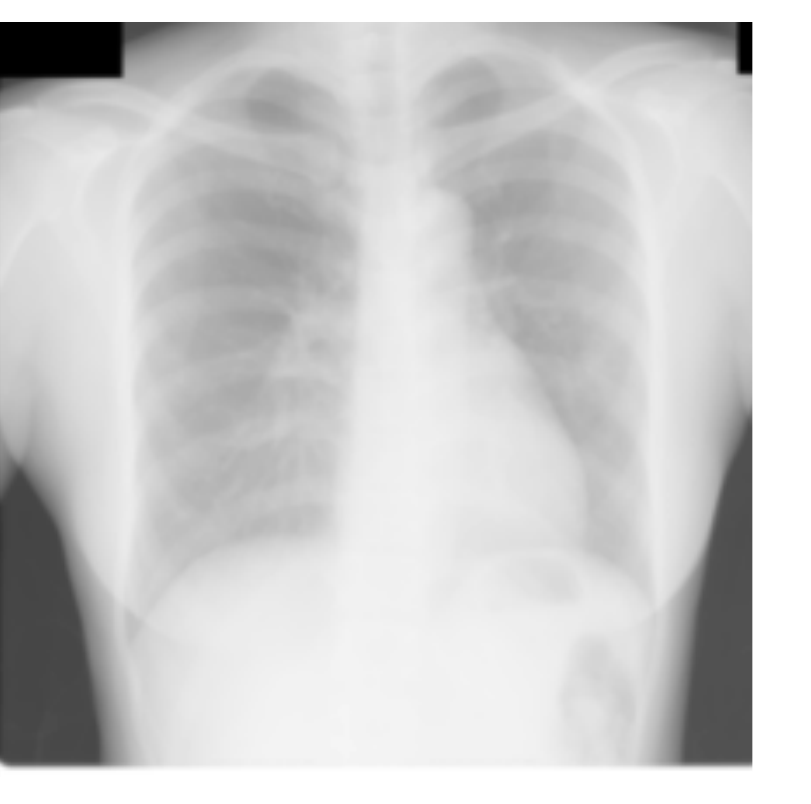} &
    \includegraphics[width=0.129\linewidth, trim={4cm 1cm 3cm 1cm},clip]{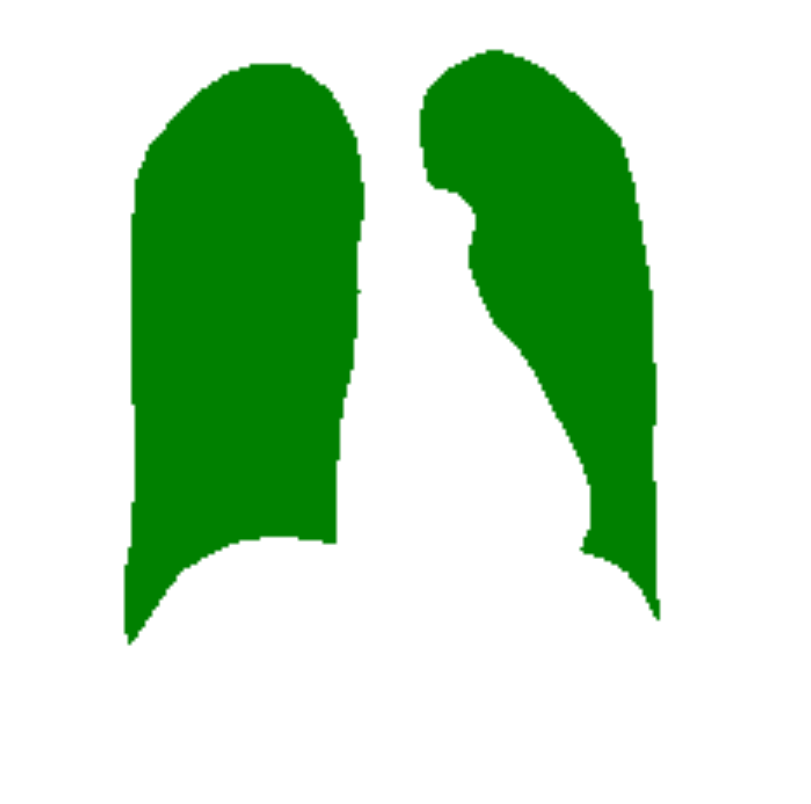} &
    \includegraphics[width=0.129\linewidth, trim={4cm 1cm 3cm 1cm},clip]{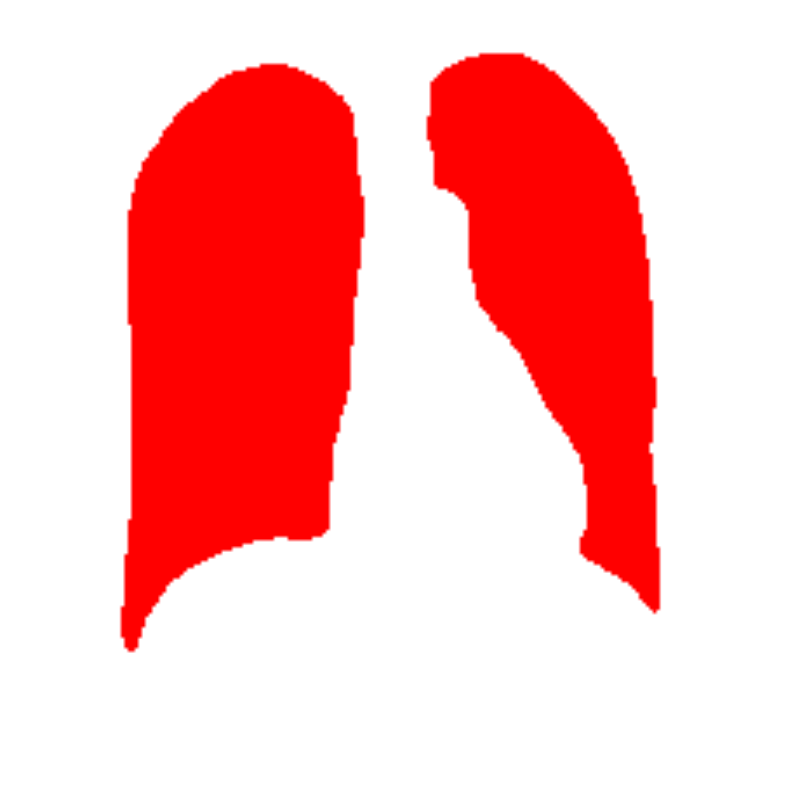}
    \\
    \includegraphics[width=0.129\linewidth, trim={4cm 1cm 3cm 1cm},clip]{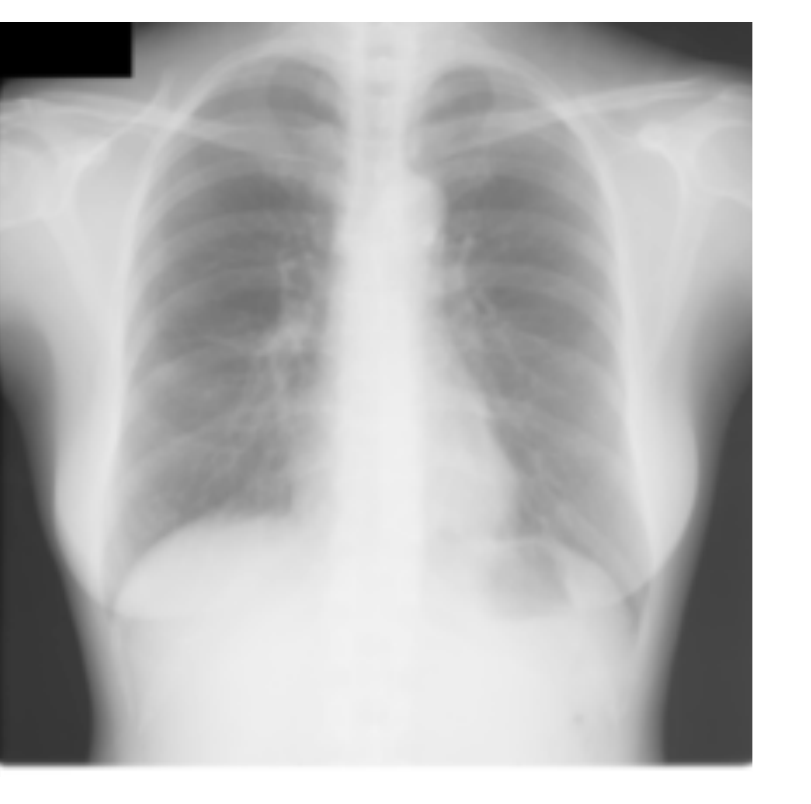} &
    \includegraphics[width=0.129\linewidth, trim={4cm 1cm 3cm 1cm},clip]{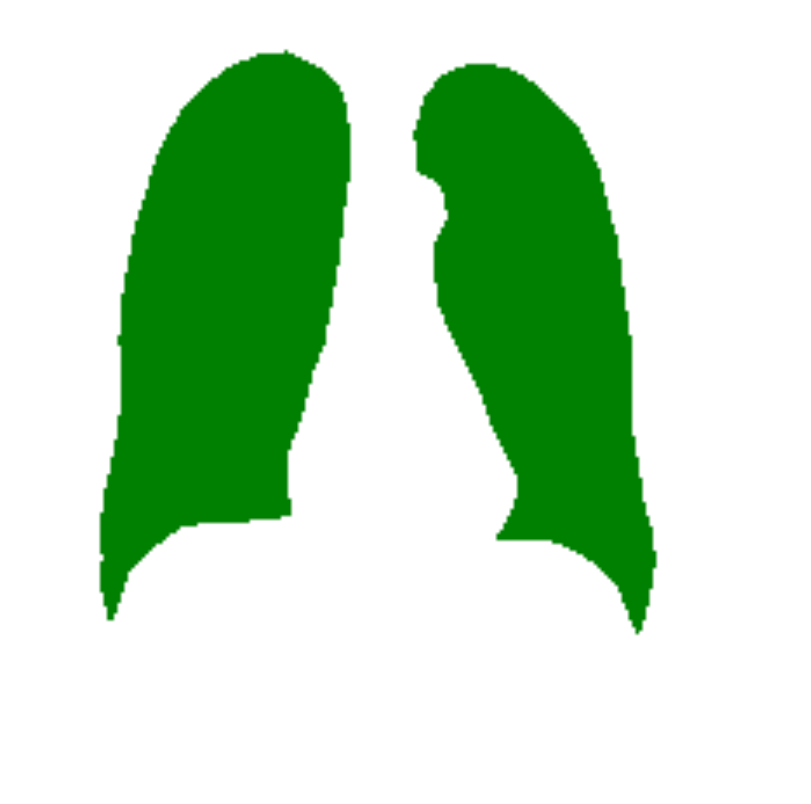} &
    \includegraphics[width=0.129\linewidth, trim={4cm 1cm 3cm 1cm},clip]{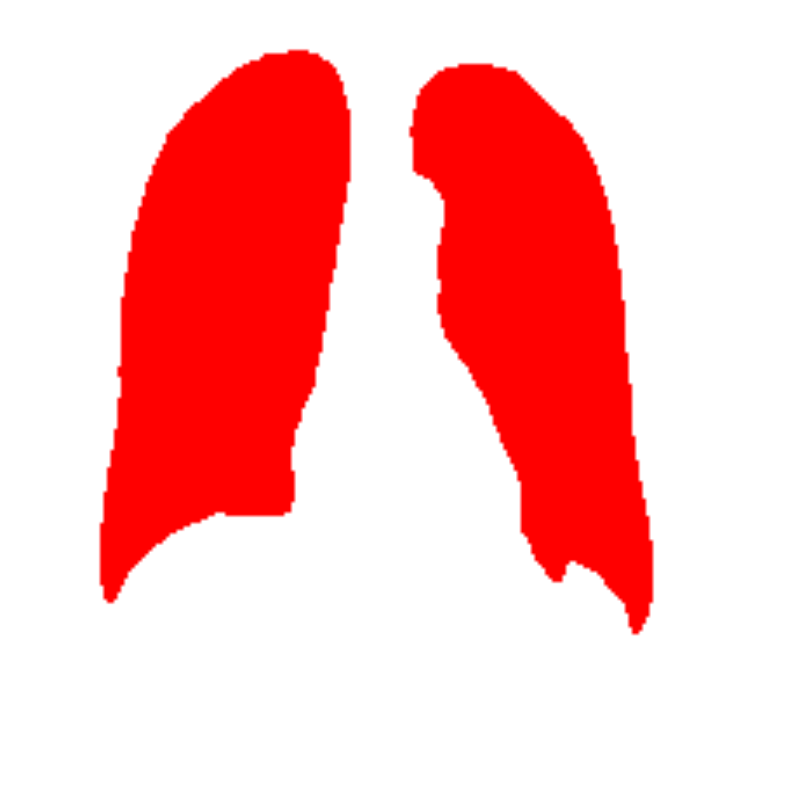}
    \\
    \includegraphics[width=0.129\linewidth, trim={4cm 1cm 3cm 1cm},clip]{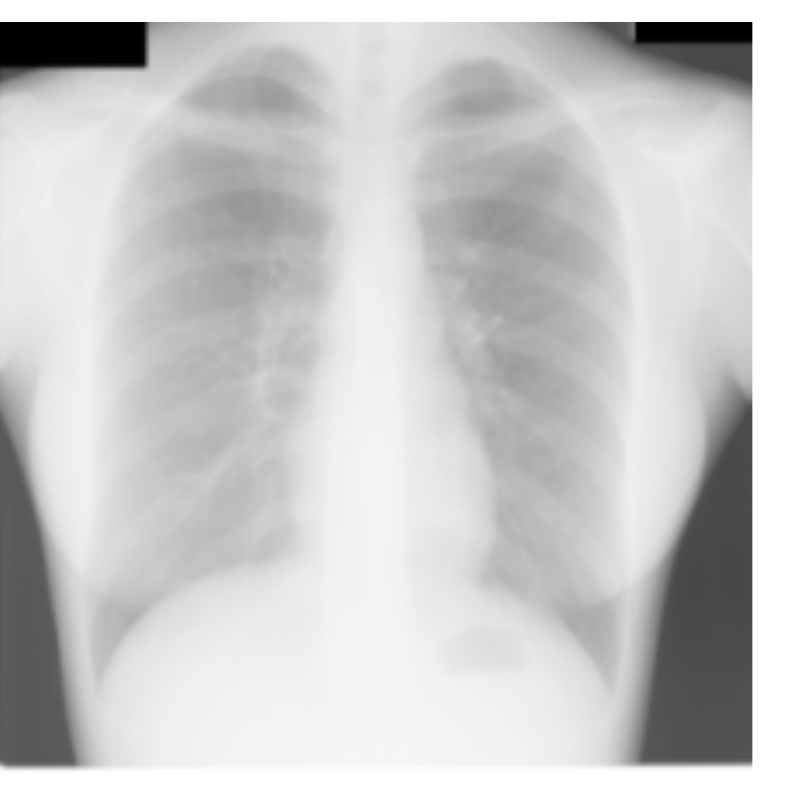} &
    \includegraphics[width=0.129\linewidth, trim={4cm 1cm 3cm 1cm},clip]{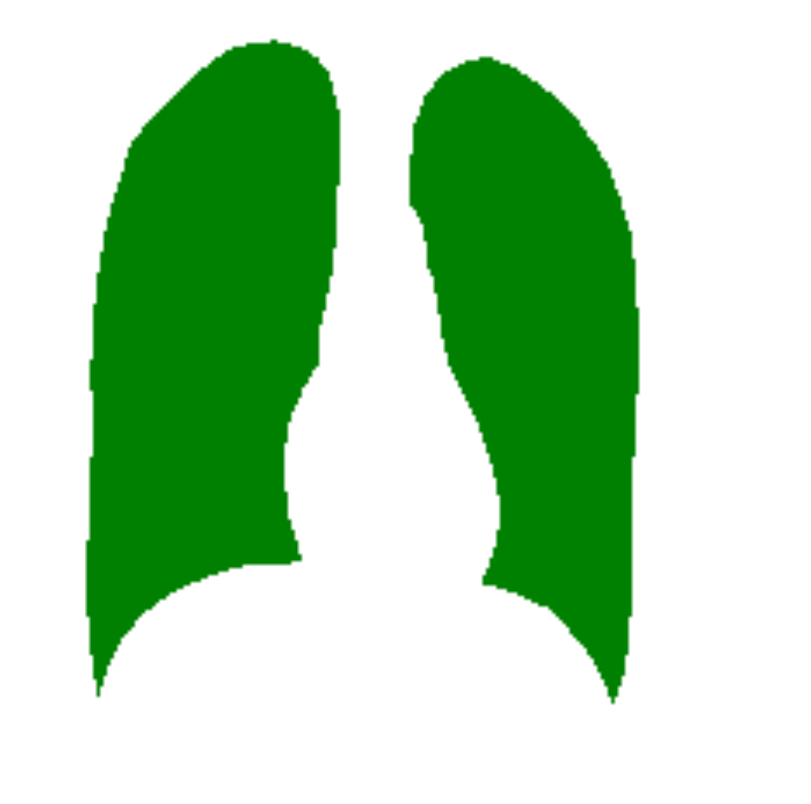} &
    \includegraphics[width=0.129\linewidth, trim={4cm 1cm 3cm 1cm},clip]{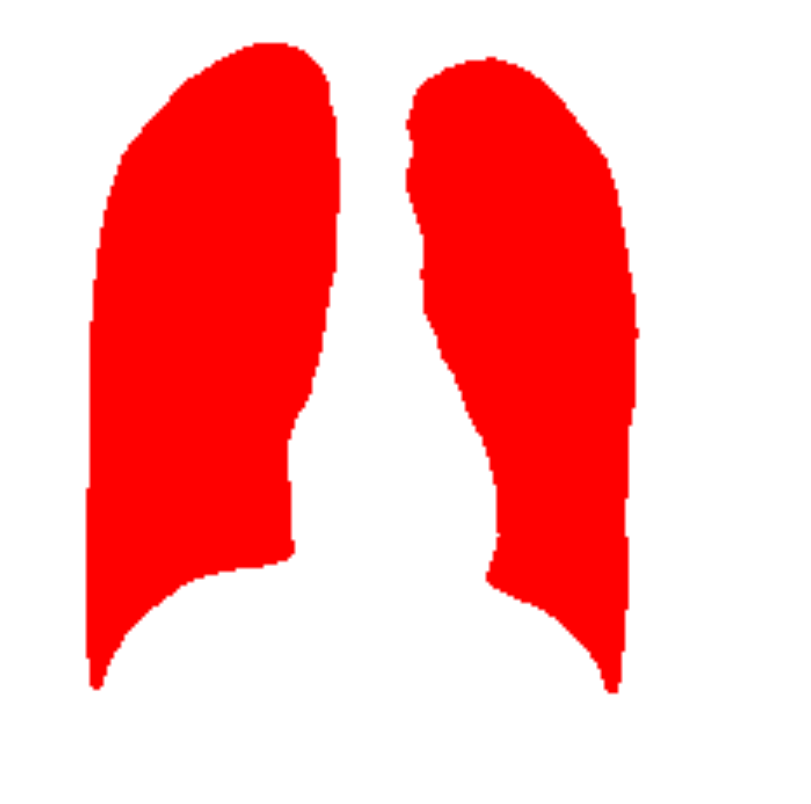}
    \\
    \includegraphics[width=0.129\linewidth, trim={4cm 1cm 3cm 1cm},clip]{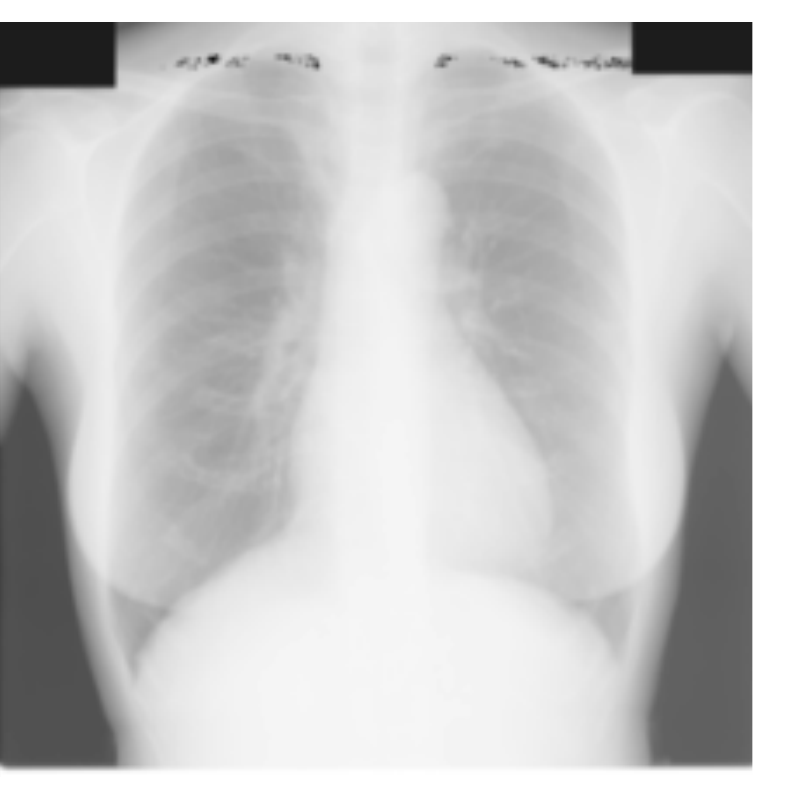} &
    \includegraphics[width=0.129\linewidth, trim={4cm 1cm 3cm 1cm},clip]{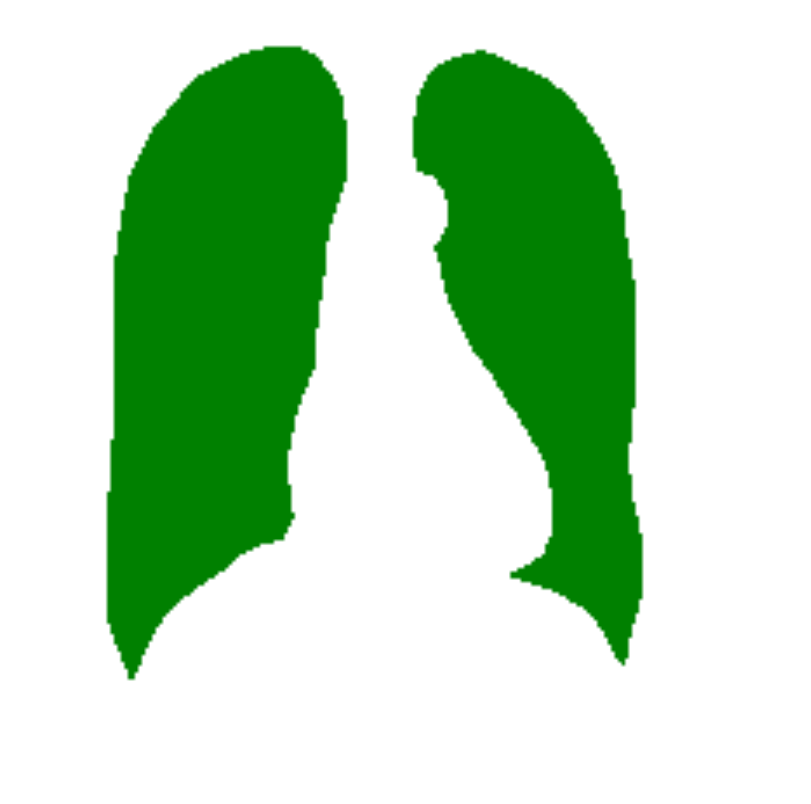} &
    \includegraphics[width=0.129\linewidth, trim={4cm 1cm 3cm 1cm},clip]{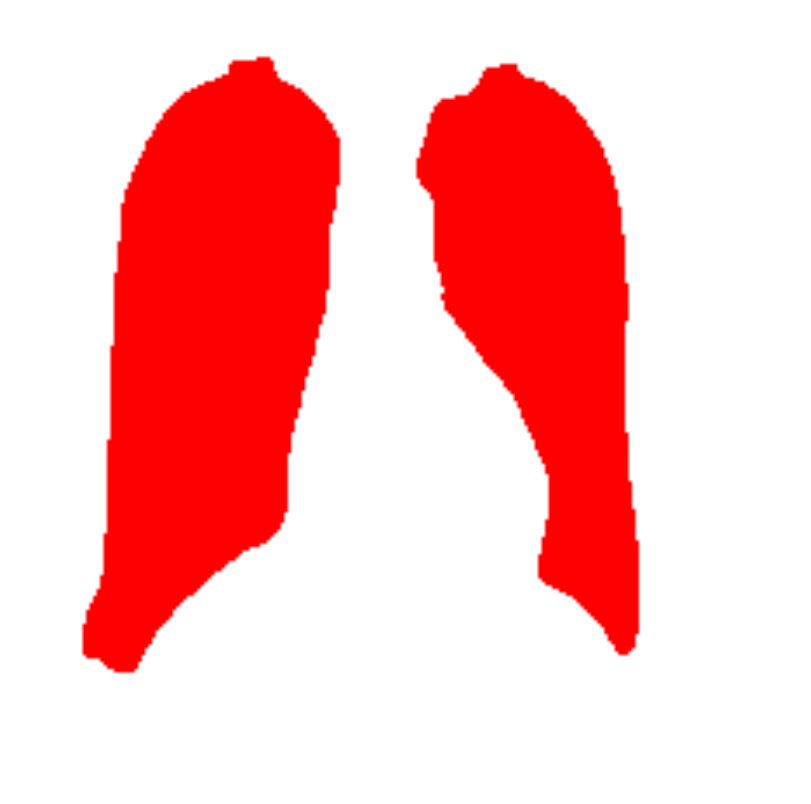}
    \\
    \includegraphics[width=0.129\linewidth, trim={4cm 1cm 3cm 1cm},clip]{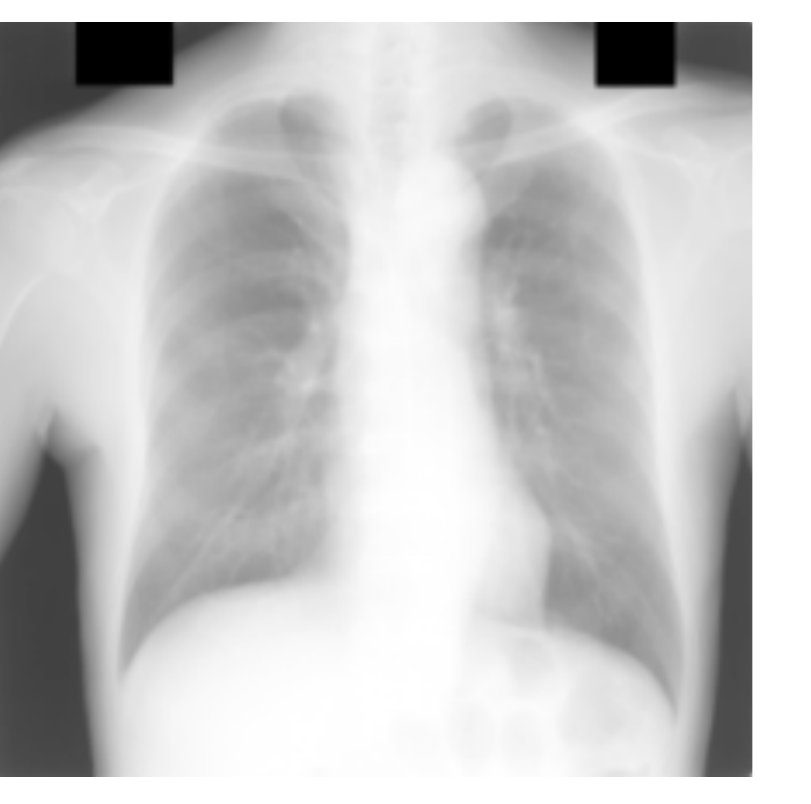} &
    \includegraphics[width=0.129\linewidth, trim={4cm 1cm 3cm 1cm},clip]{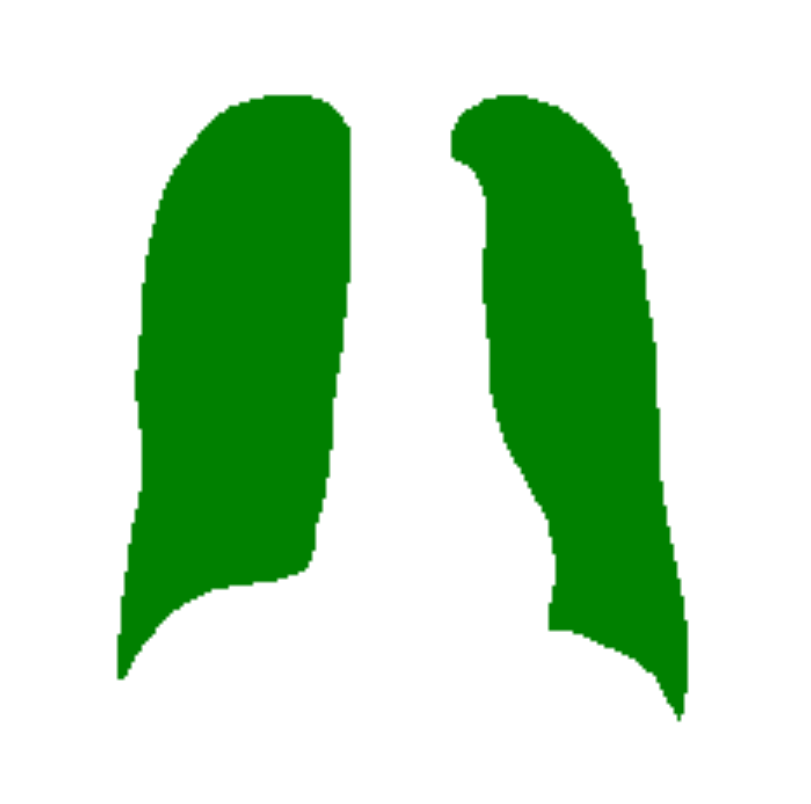} &
    \includegraphics[width=0.129\linewidth, trim={4cm 1cm 3cm 1cm},clip]{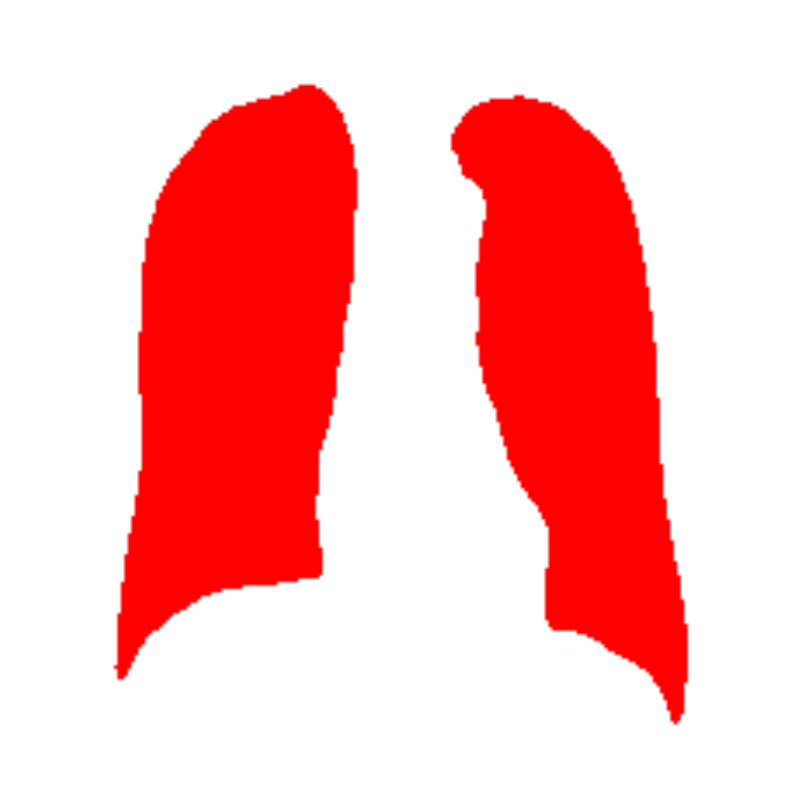}
\end{tabular}}
\hfill
\subcaptionbox{MCU (cross-domain)}{
  \begin{tabular}{ccc}
  \small Image & \small Ground Truth & \small Prediction \\
    \includegraphics[width=0.129\linewidth, trim={4cm 1cm 3cm 1cm},clip]{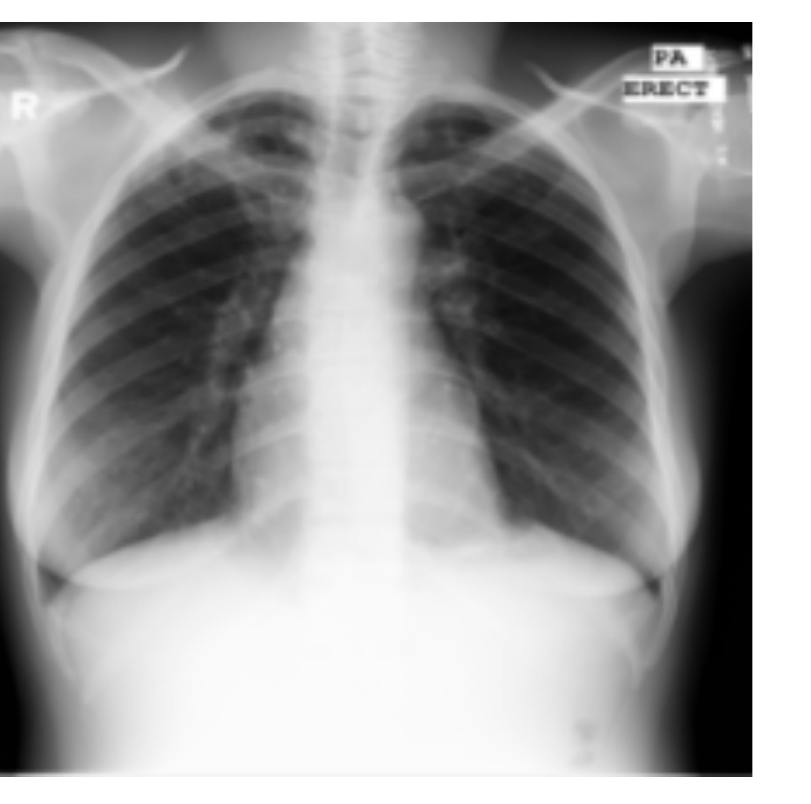} &
    \includegraphics[width=0.129\linewidth, trim={4cm 1cm 3cm 1cm},clip]{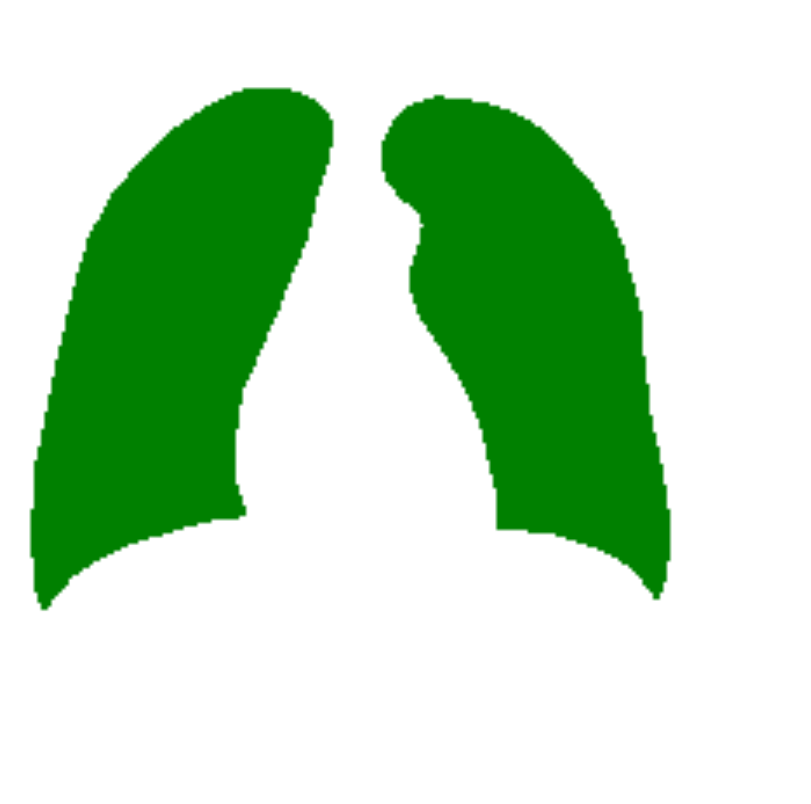} &
    \includegraphics[width=0.129\linewidth, trim={4cm 1cm 3cm 1cm},clip]{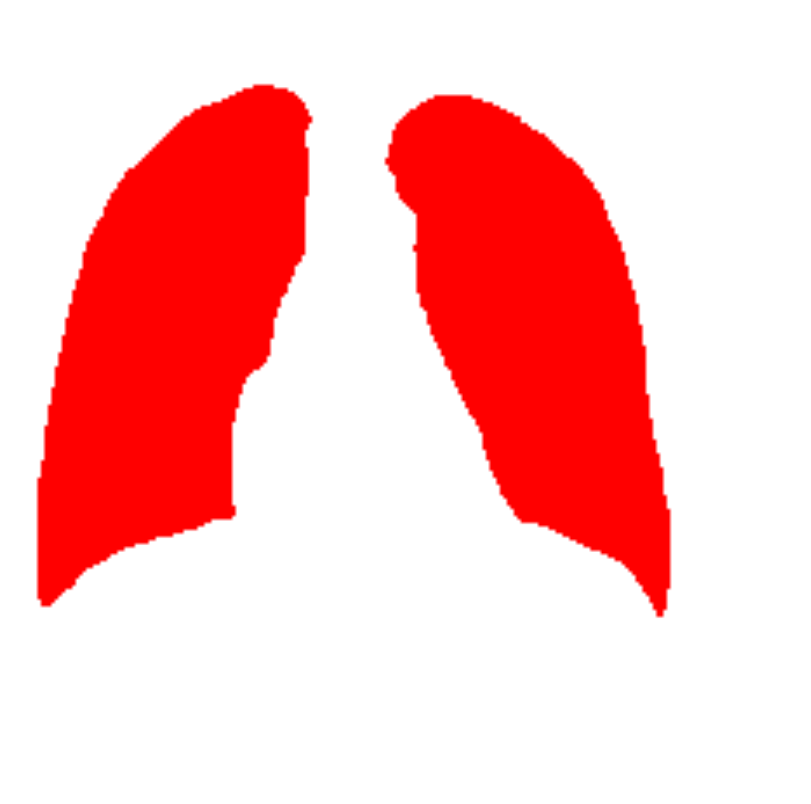}
    \\
    \includegraphics[width=0.129\linewidth, trim={4cm 1cm 3cm 1cm},clip]{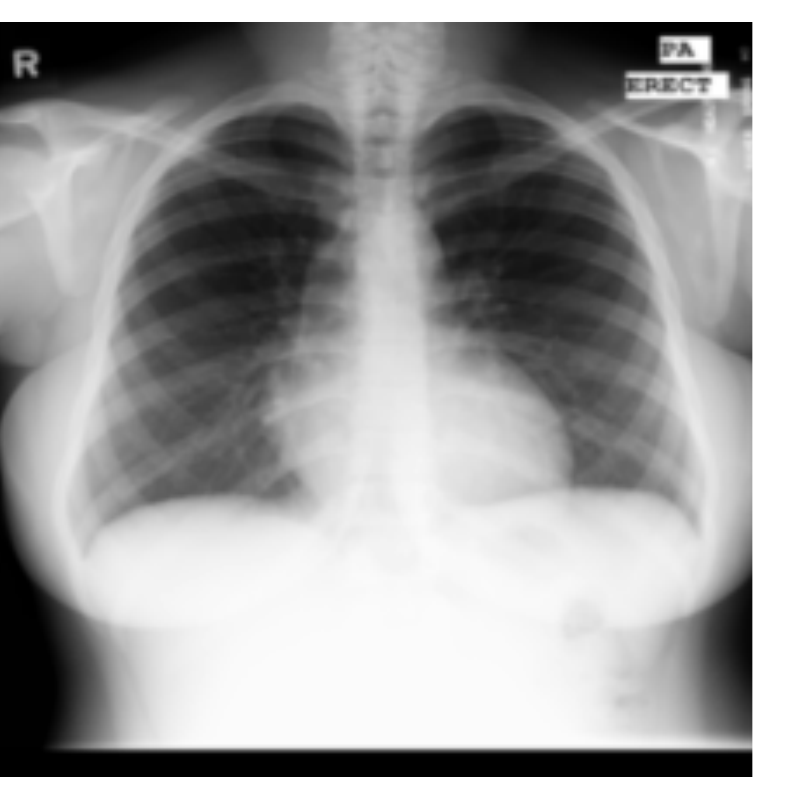} &
    \includegraphics[width=0.129\linewidth, trim={4cm 1cm 3cm 1cm},clip]{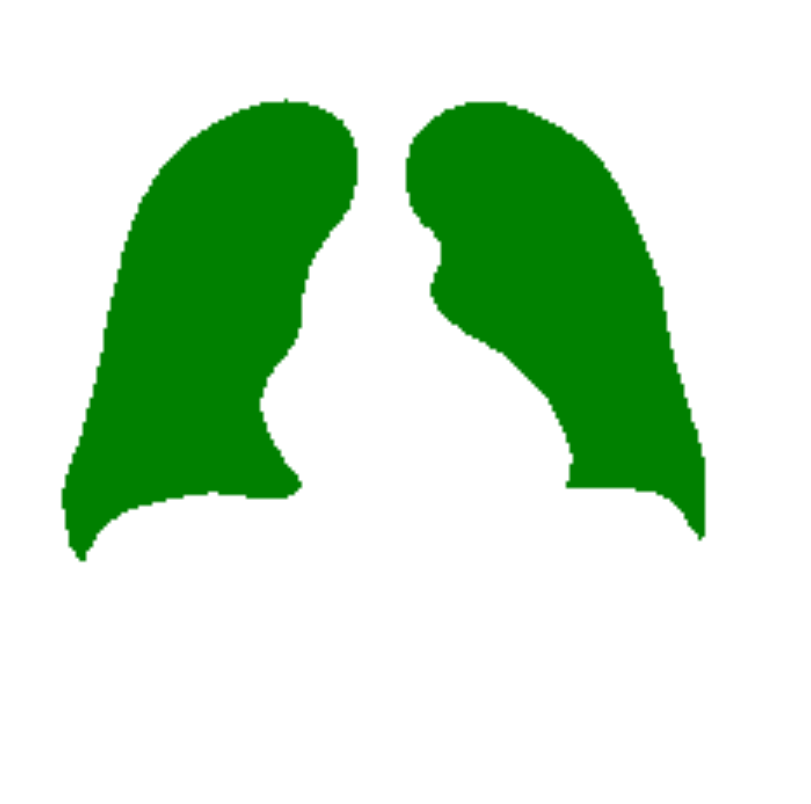} &
    \includegraphics[width=0.129\linewidth, trim={4cm 1cm 3cm 1cm},clip]{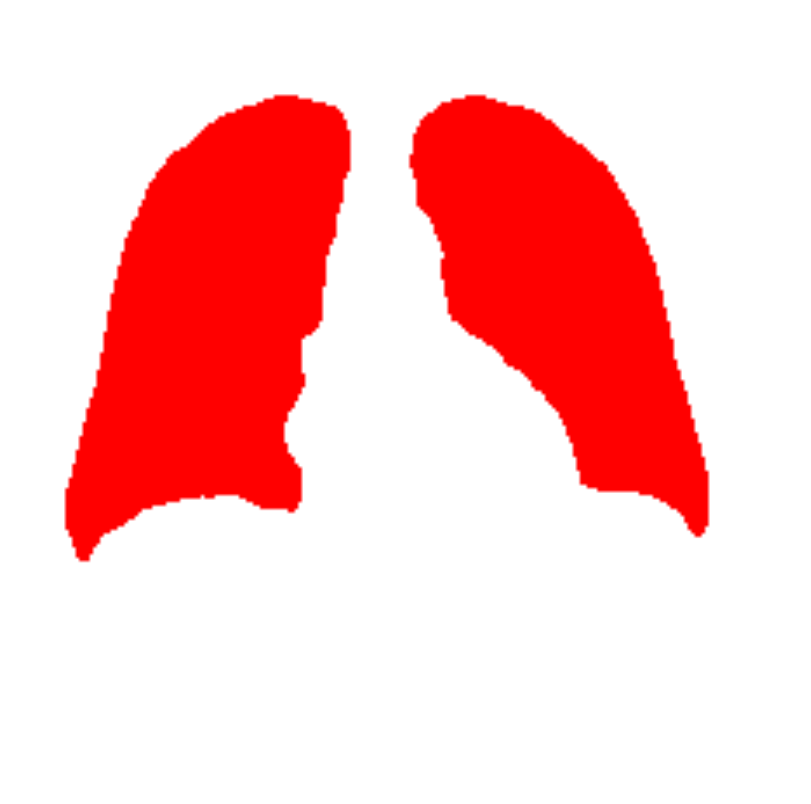}
    \\
    \includegraphics[width=0.129\linewidth, trim={4cm 1cm 3cm 1cm},clip]{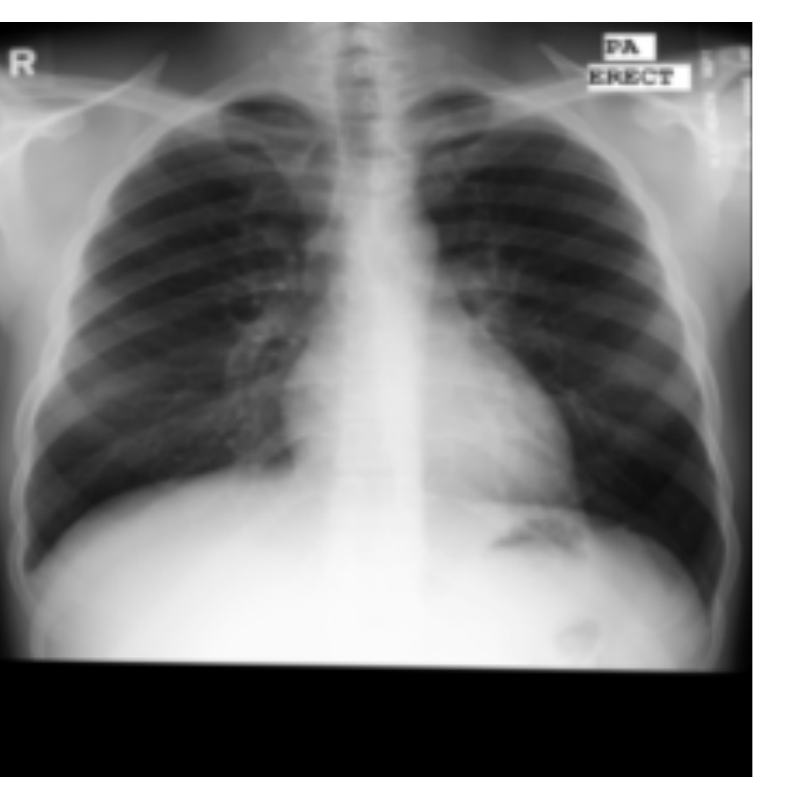} &
    \includegraphics[width=0.129\linewidth, trim={4cm 1cm 3cm 1cm},clip]{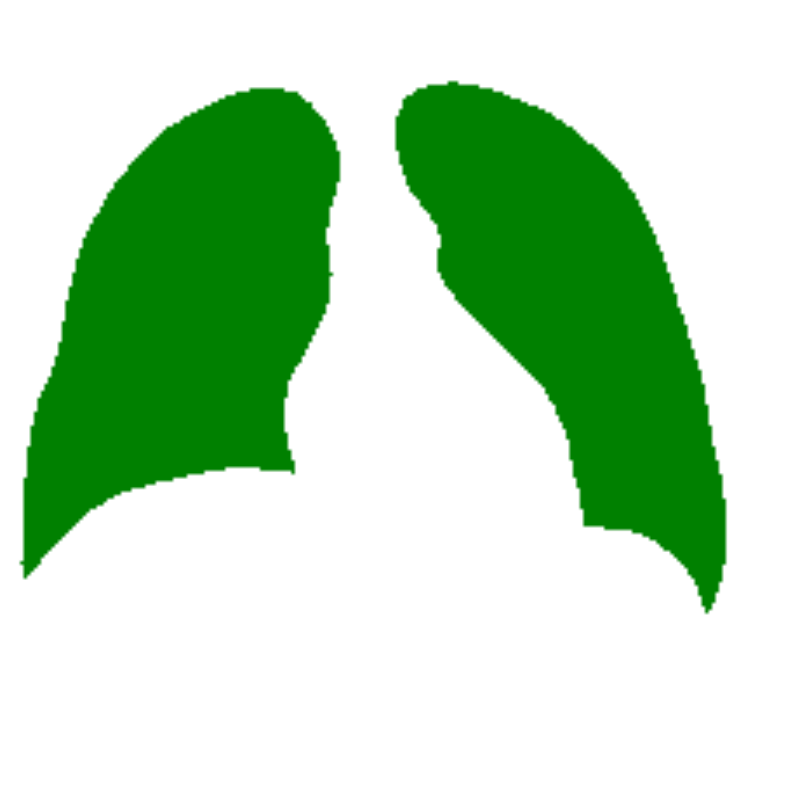} &
    \includegraphics[width=0.129\linewidth, trim={4cm 1cm 3cm 1cm},clip]{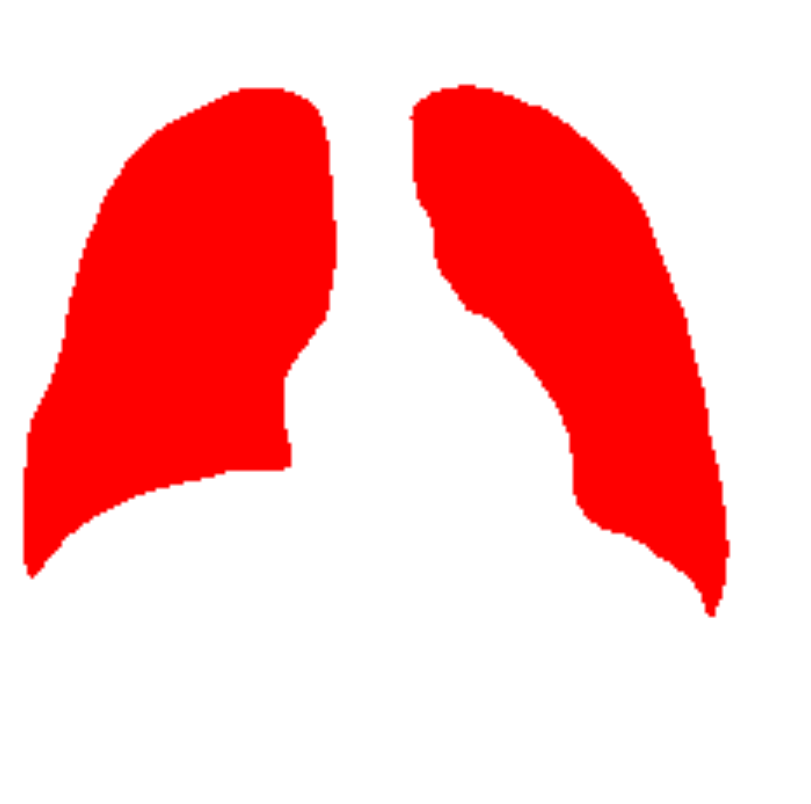}
    \\
    \includegraphics[width=0.129\linewidth, trim={4cm 1cm 3cm 1cm},clip]{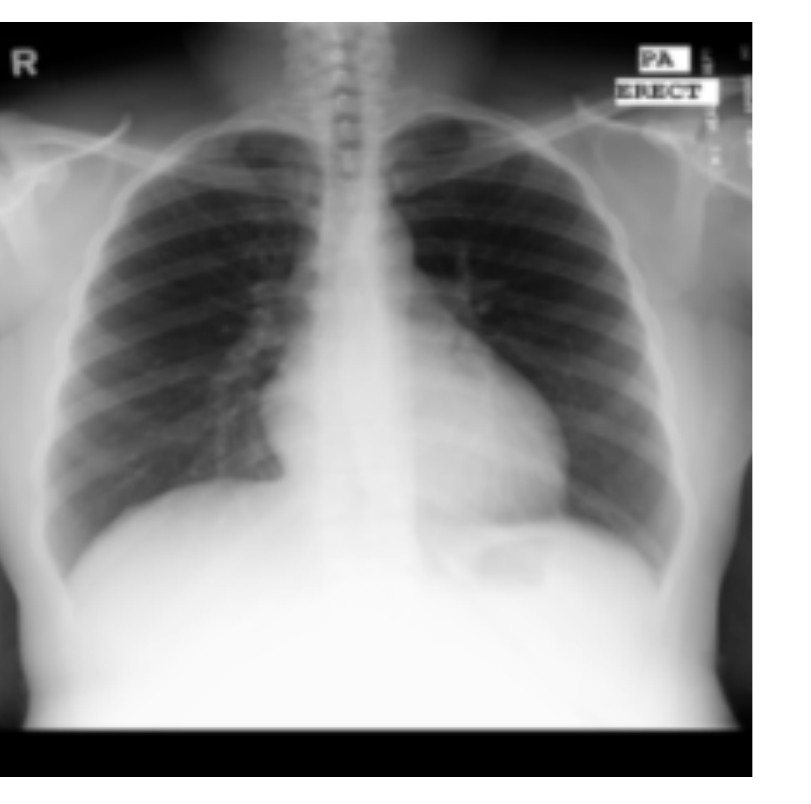} &
    \includegraphics[width=0.129\linewidth, trim={4cm 1cm 3cm 1cm},clip]{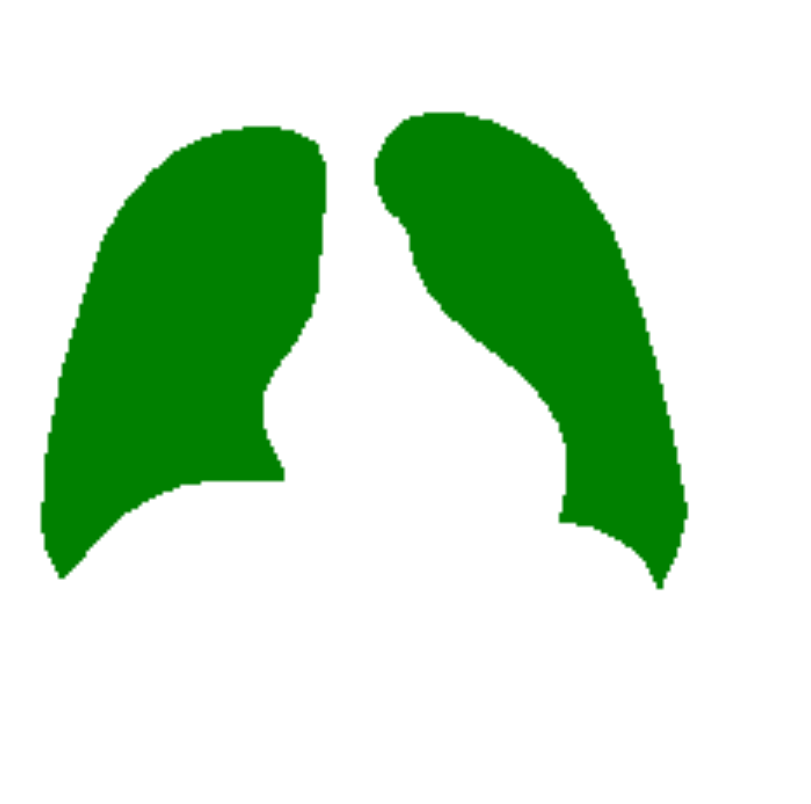} &
    \includegraphics[width=0.129\linewidth, trim={4cm 1cm 3cm 1cm},clip]{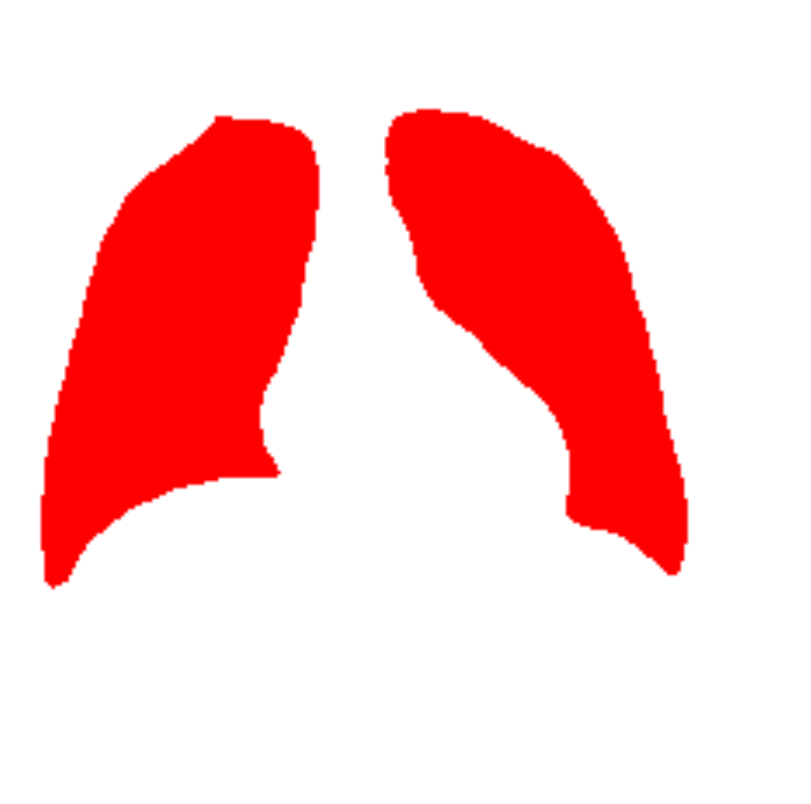}
    \\
    \includegraphics[width=0.129\linewidth, trim={4cm 1cm 3cm 1cm},clip]{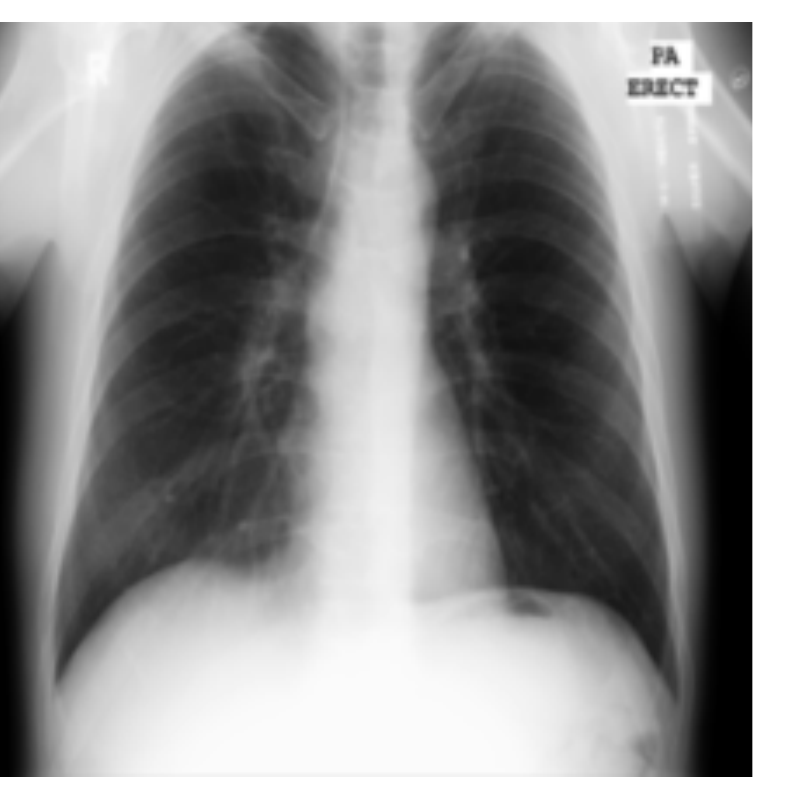} &
    \includegraphics[width=0.129\linewidth, trim={4cm 1cm 3cm 1cm},clip]{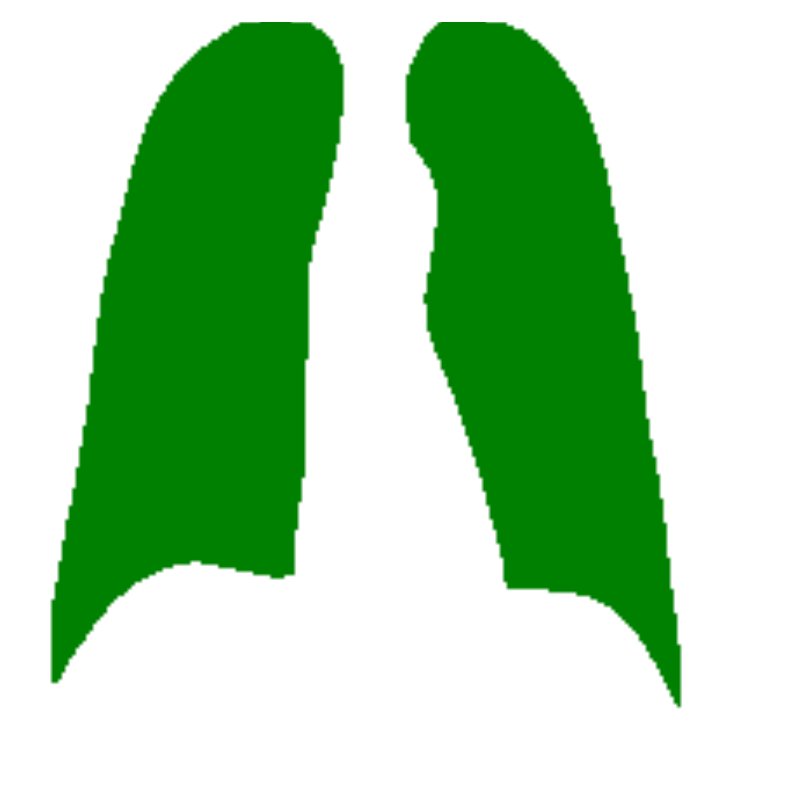} &
    \includegraphics[width=0.129\linewidth, trim={4cm 1cm 3cm 1cm},clip]{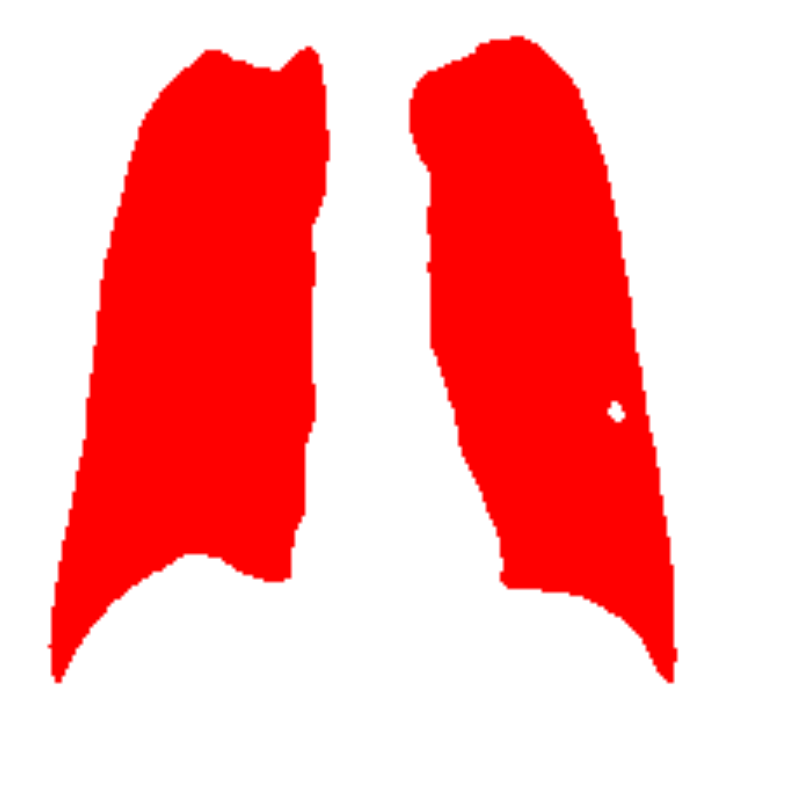}
    \\
    \includegraphics[width=0.129\linewidth, trim={4cm 1cm 3cm 1cm},clip]{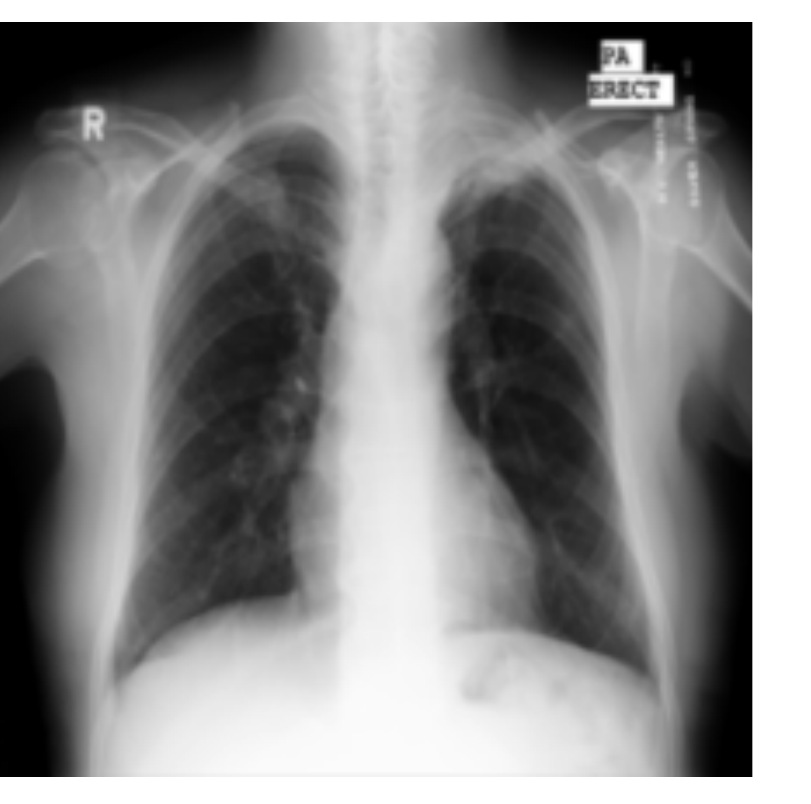} &
    \includegraphics[width=0.129\linewidth, trim={4cm 1cm 3cm 1cm},clip]{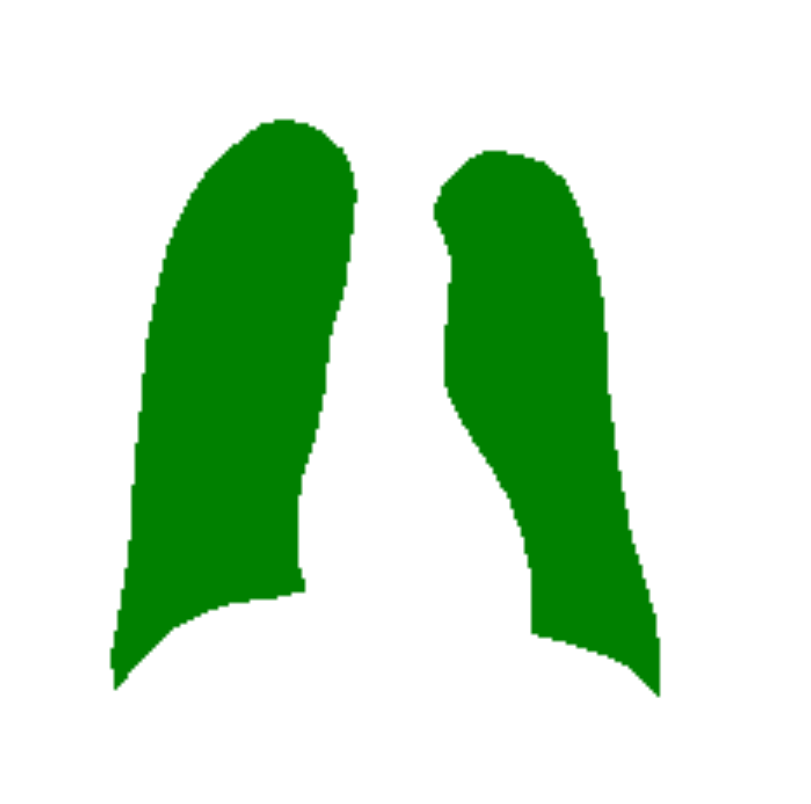} &
    \includegraphics[width=0.129\linewidth, trim={4cm 1cm 3cm 1cm},clip]{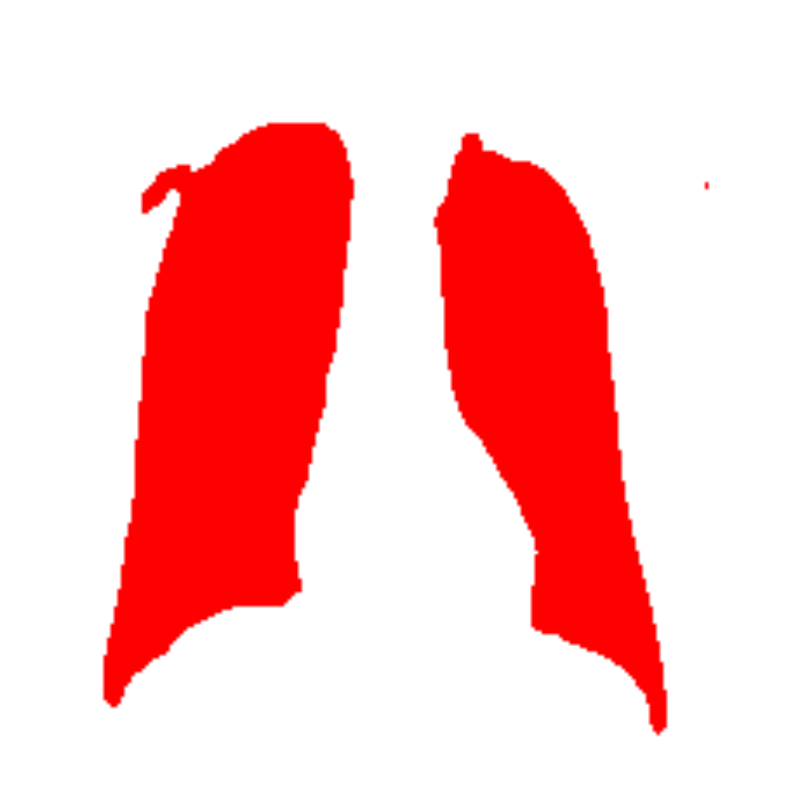}
    \\
    \includegraphics[width=0.129\linewidth, trim={4cm 1cm 3cm 1cm},clip]{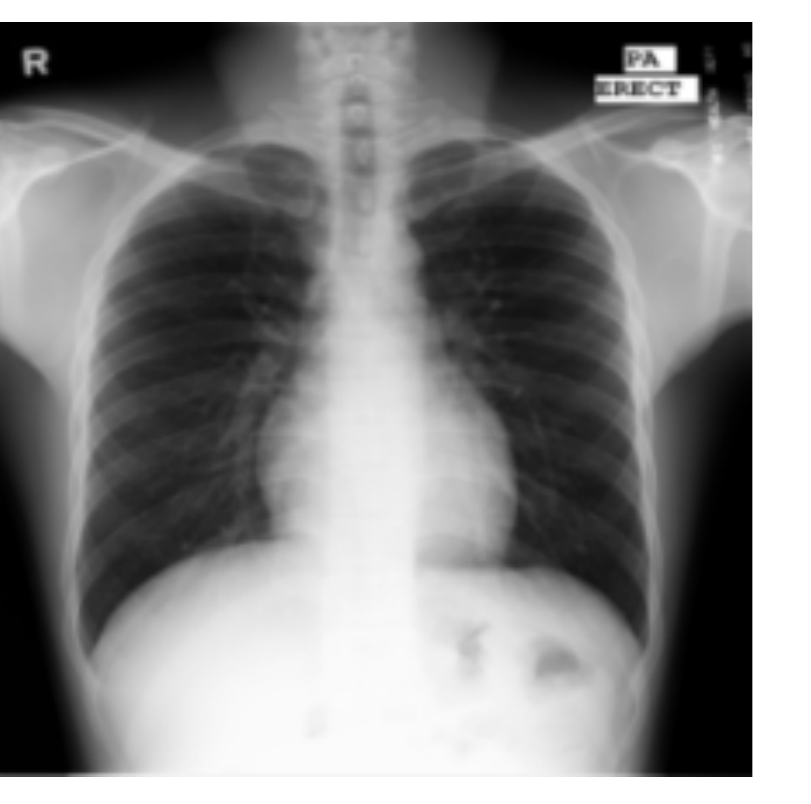} &
    \includegraphics[width=0.129\linewidth, trim={4cm 1cm 3cm 1cm},clip]{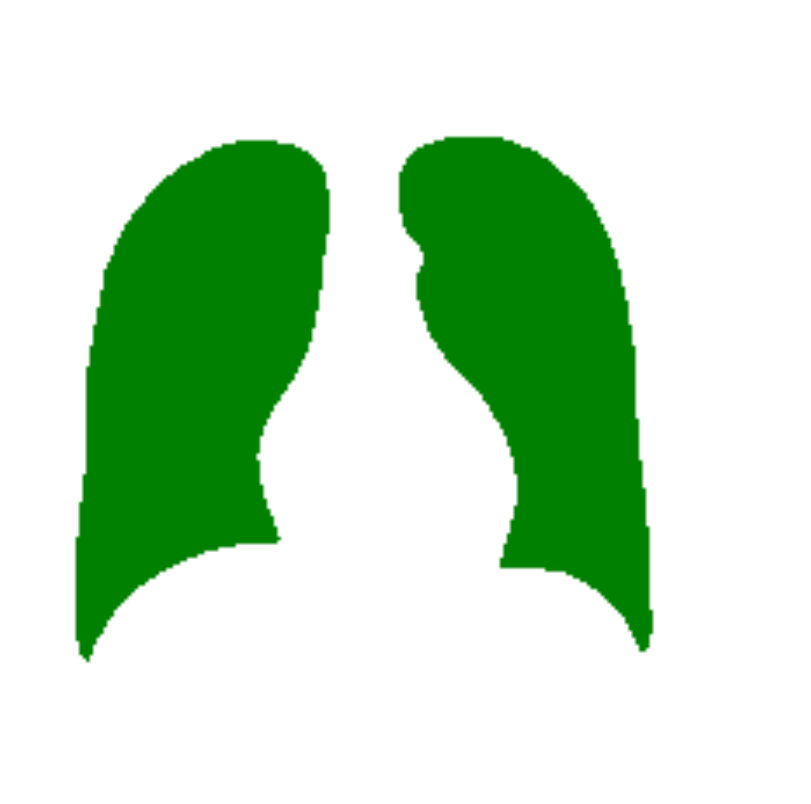} &
    \includegraphics[width=0.129\linewidth, trim={4cm 1cm 3cm 1cm},clip]{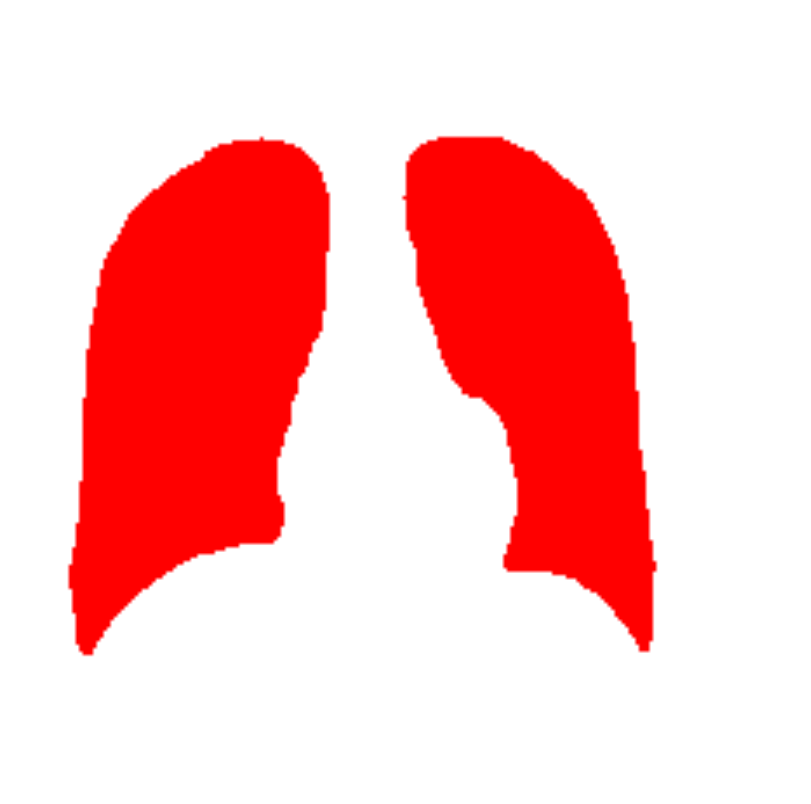}
\end{tabular}}
\caption{Visualizations of the segmented lung masks by MultiMix-50-1000 on the in-domain JSRT dataset and cross-domain MCU dataset. The results show good agreement between the groundtruth and predicted masks.}
\label{fig:pred_masks_in}
\end{figure}

Figure~\ref{fig:pred_masks_in} shows the ground truth lung masks and  masks predicted by the MultiMix model (MultiMix-50-1000) for a number of images from the JSRT dataset (in-domain) and MCU dataset (cross-domain). Both parts of the figure display the accuracy in the predicted segmentation masks, both in-domain and cross-domain, as there is almost no noise in these predictions, proving the effectiveness of our algorithm even when being trained with limited labeled data.

\section{Saliency Visualization}
\label{app:saliency-vis}

\begin{figure} \centering
\subcaptionbox{CheX (in-domain)}{
  \begin{tabular}{cc}
  \small Image & \small Saliency Map \\
  \multicolumn{2}{c}{\small Normal} \\
    \includegraphics[width=0.195\linewidth, trim={4cm 1cm 3cm 1cm},clip]{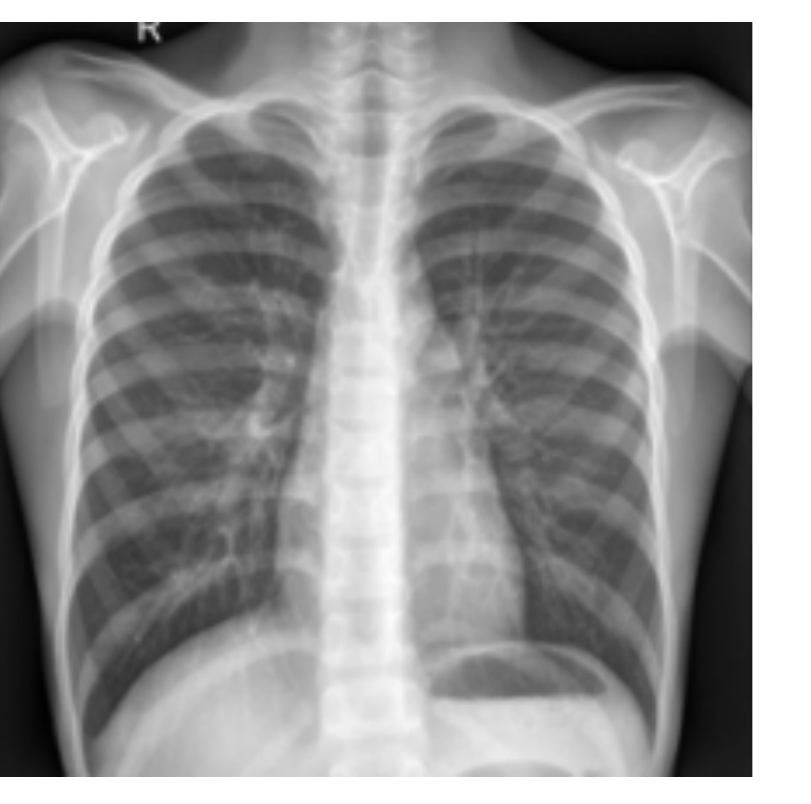} &
    \includegraphics[width=0.195\linewidth, trim={4cm 1cm 3cm 1cm},clip]{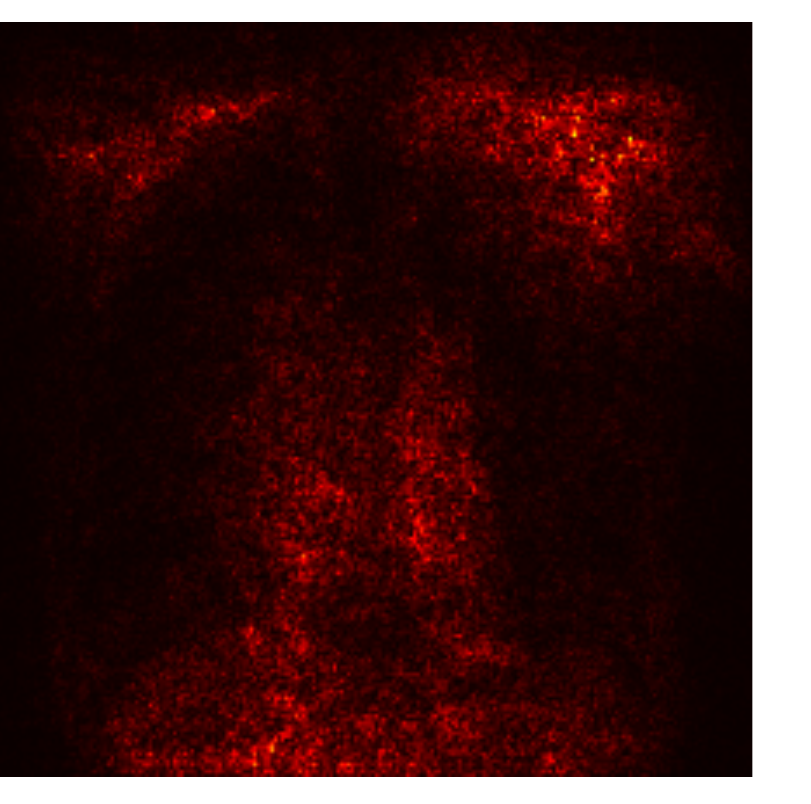}
    \\
    \includegraphics[width=0.195\linewidth, trim={4cm 1cm 3cm 1cm},clip]{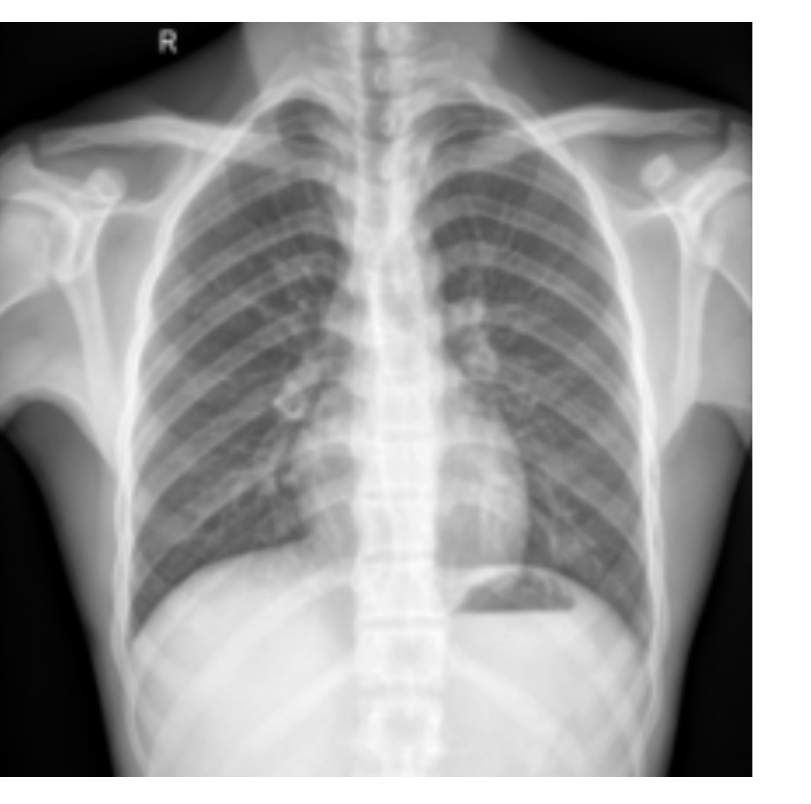} &
    \includegraphics[width=0.195\linewidth, trim={4cm 1cm 3cm 1cm},clip]{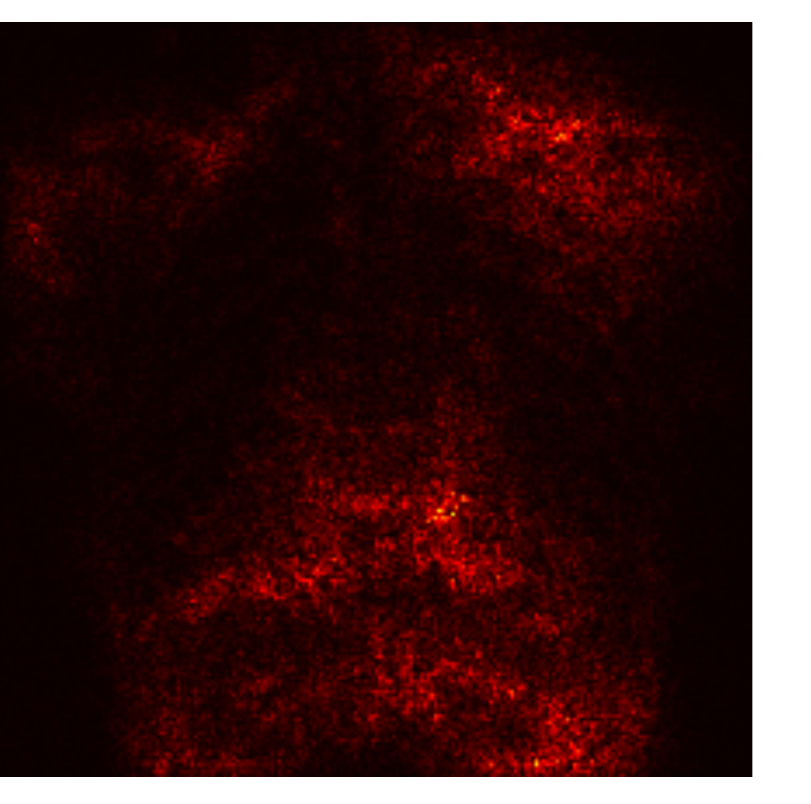}
    \\
    \includegraphics[width=0.195\linewidth, trim={4cm 1cm 3cm 1cm},clip]{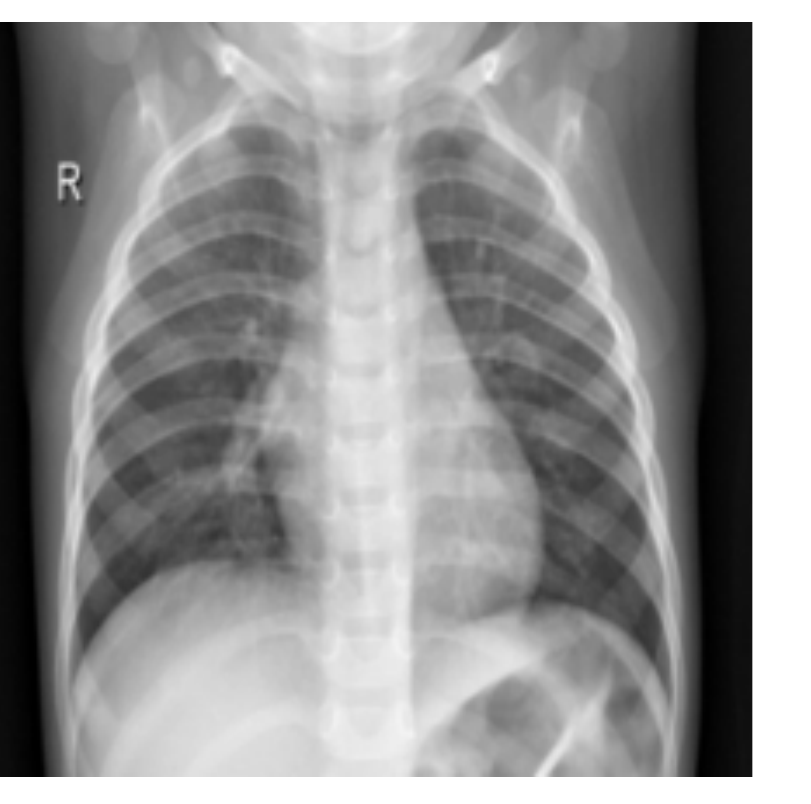} &
    \includegraphics[width=0.195\linewidth, trim={4cm 1cm 3cm 1cm},clip]{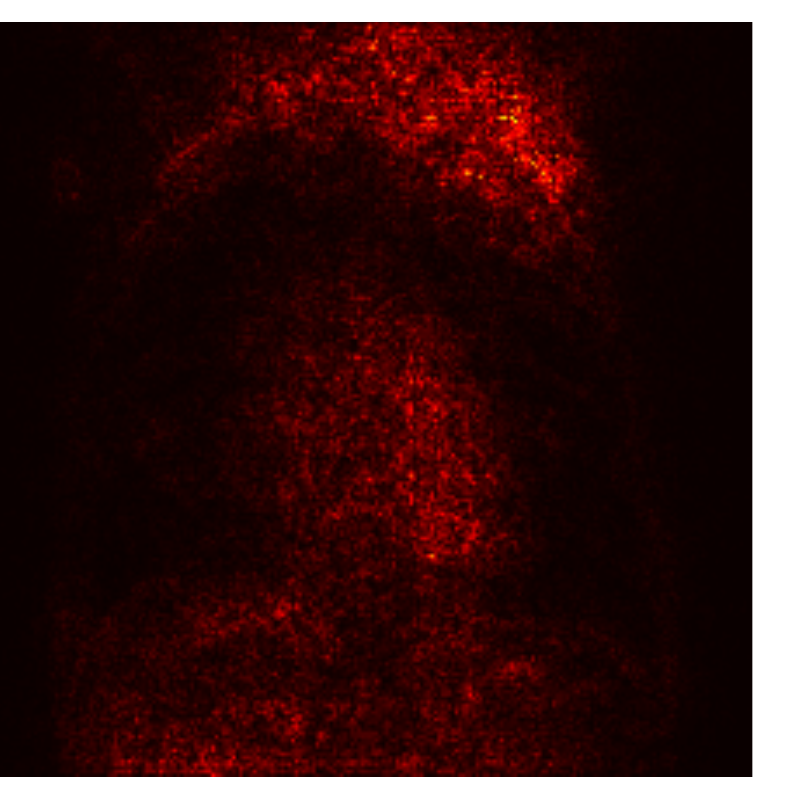}
    \\
    \multicolumn{2}{c}{\small Pneumonia} \\
    \includegraphics[width=0.195\linewidth, trim={4cm 1cm 3cm 1cm},clip]{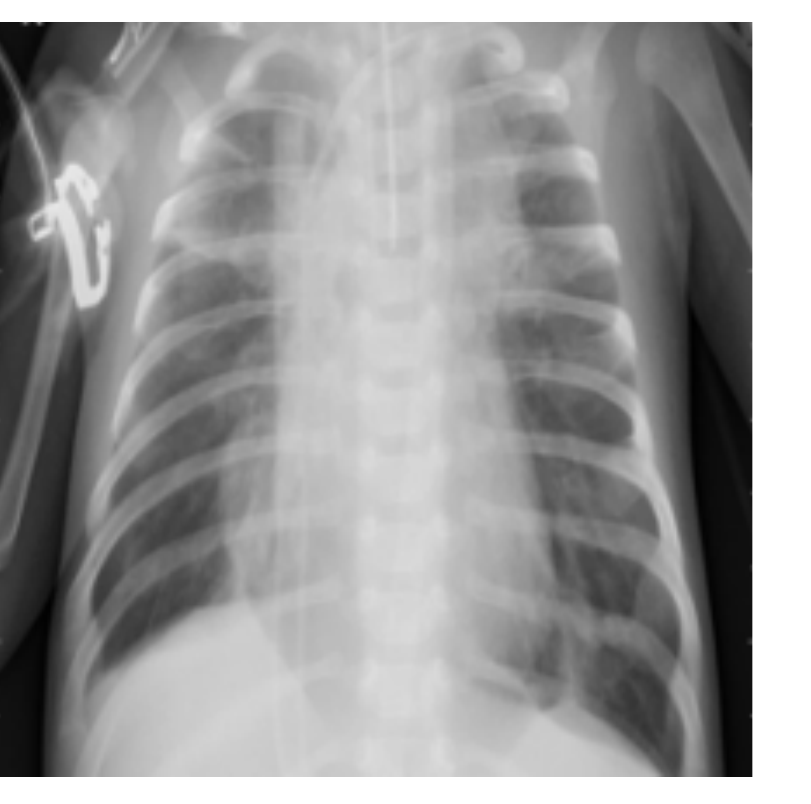} &
    \includegraphics[width=0.195\linewidth, trim={4cm 1cm 3cm 1cm},clip]{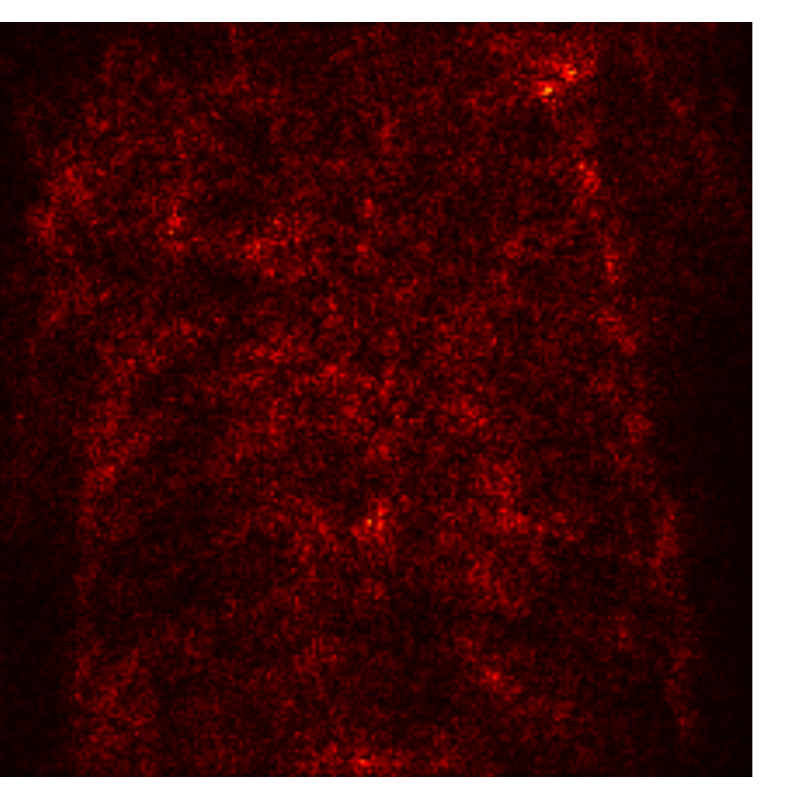}
    \\
    \includegraphics[width=0.195\linewidth, trim={4cm 1cm 3cm 1cm},clip]{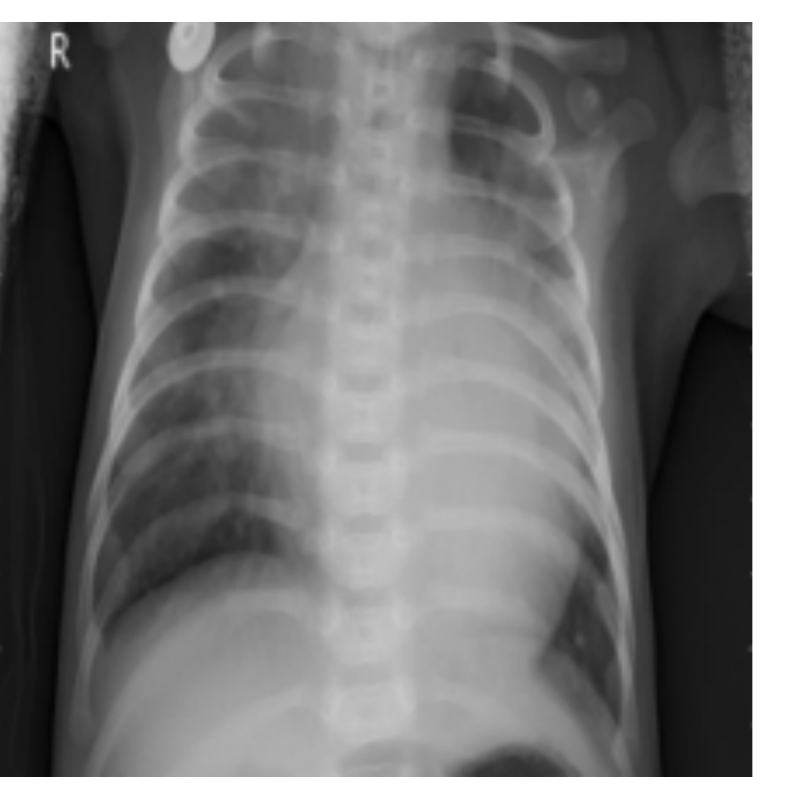} &
    \includegraphics[width=0.195\linewidth, trim={4cm 1cm 3cm 1cm},clip]{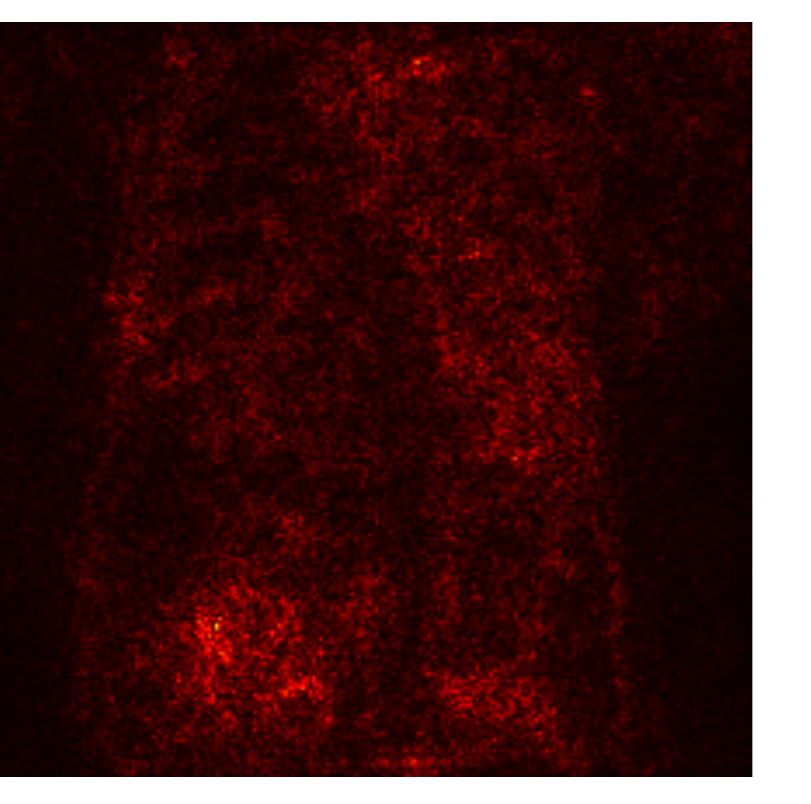}
    \\
    \includegraphics[width=0.195\linewidth, trim={4cm 1cm 3cm 1cm},clip]{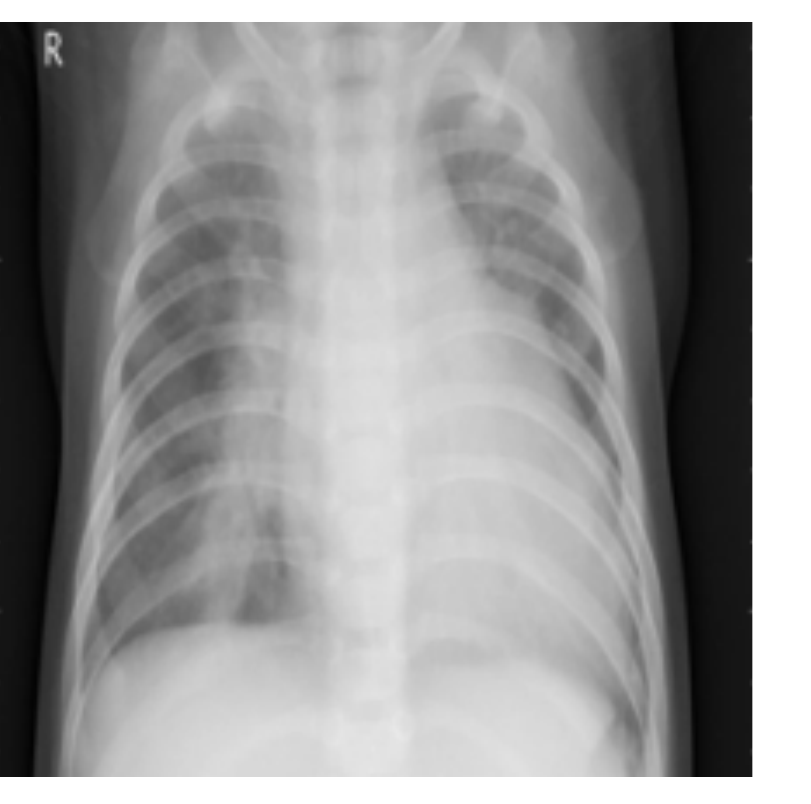} &
    \includegraphics[width=0.195\linewidth, trim={4cm 1cm 3cm 1cm},clip]{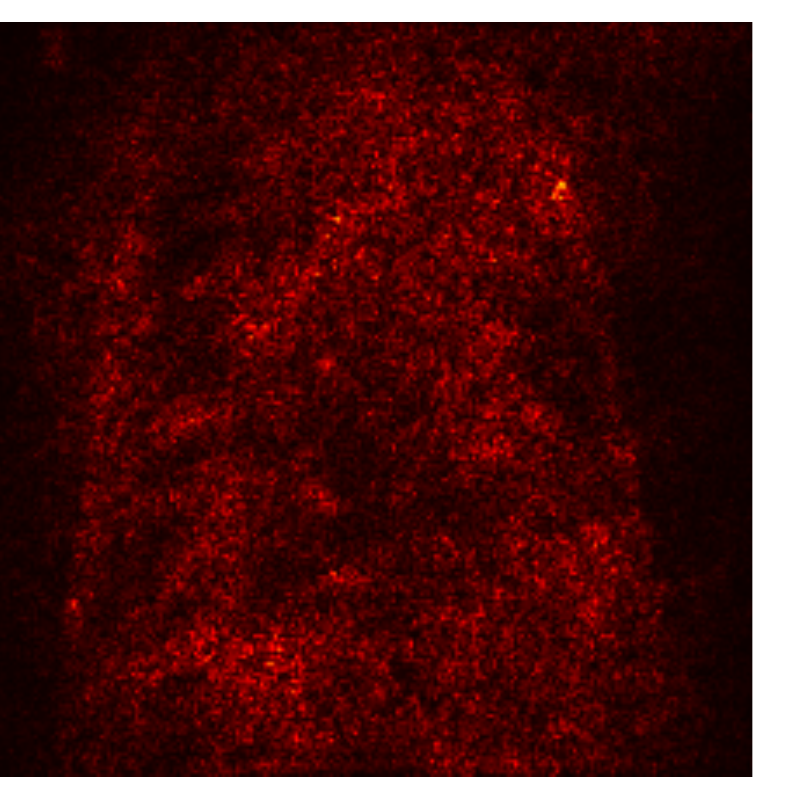}
\end{tabular}}
\hfill
\subcaptionbox{NIHX (cross-domain)}{
  \begin{tabular}{cc}
  \small Image & \small Saliency Map \\
  \multicolumn{2}{c}{\small Normal} \\
  \includegraphics[width=0.195\linewidth, trim={4cm 1cm 3cm 1cm},clip]{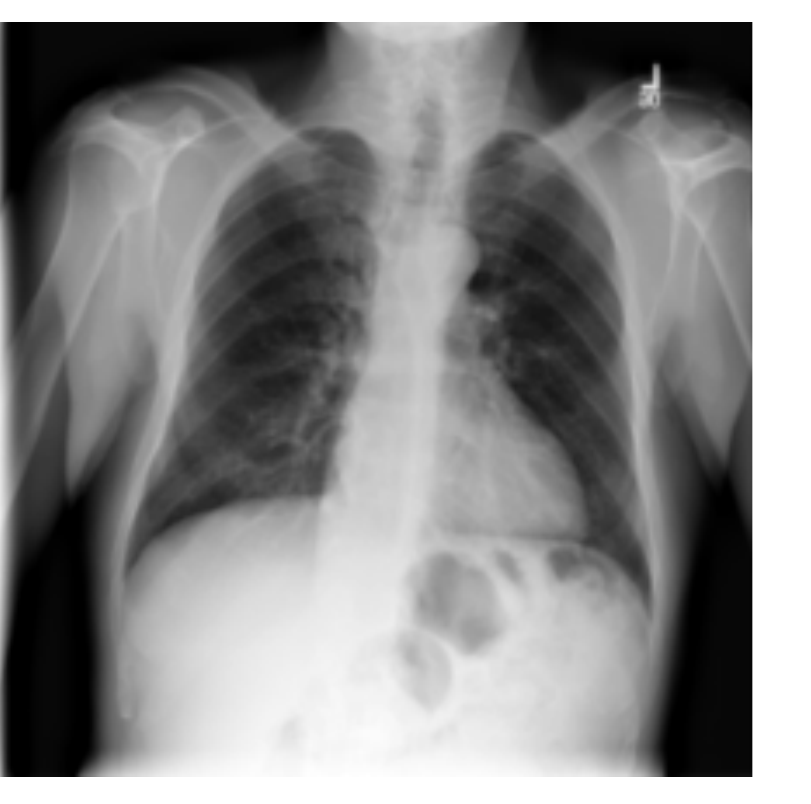} &
    \includegraphics[width=0.195\linewidth, trim={4cm 1cm 3cm 1cm},clip]{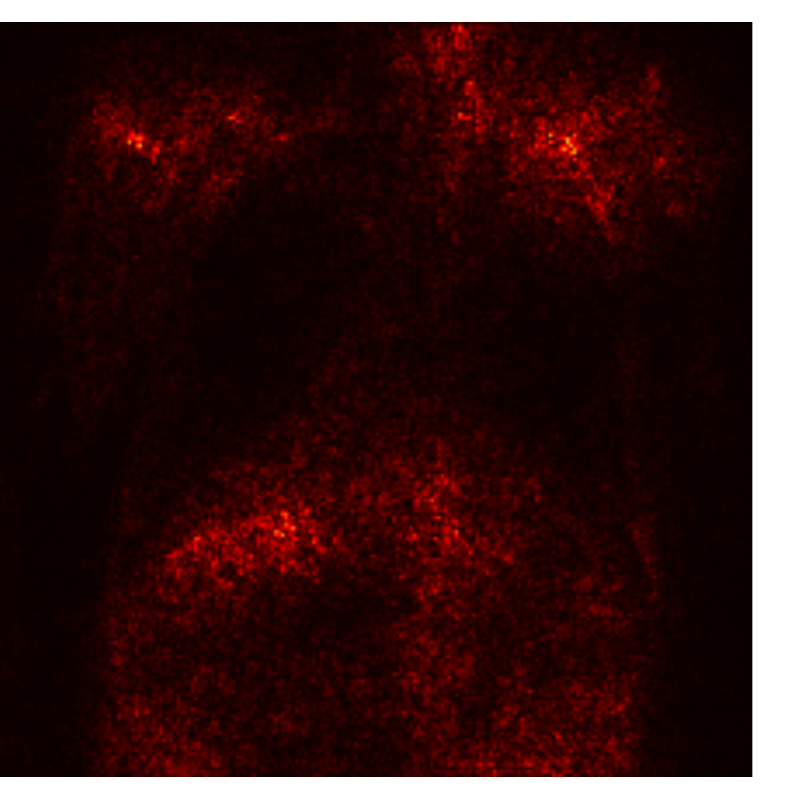}
    \\
    \includegraphics[width=0.195\linewidth, trim={4cm 1cm 3cm 1cm},clip]{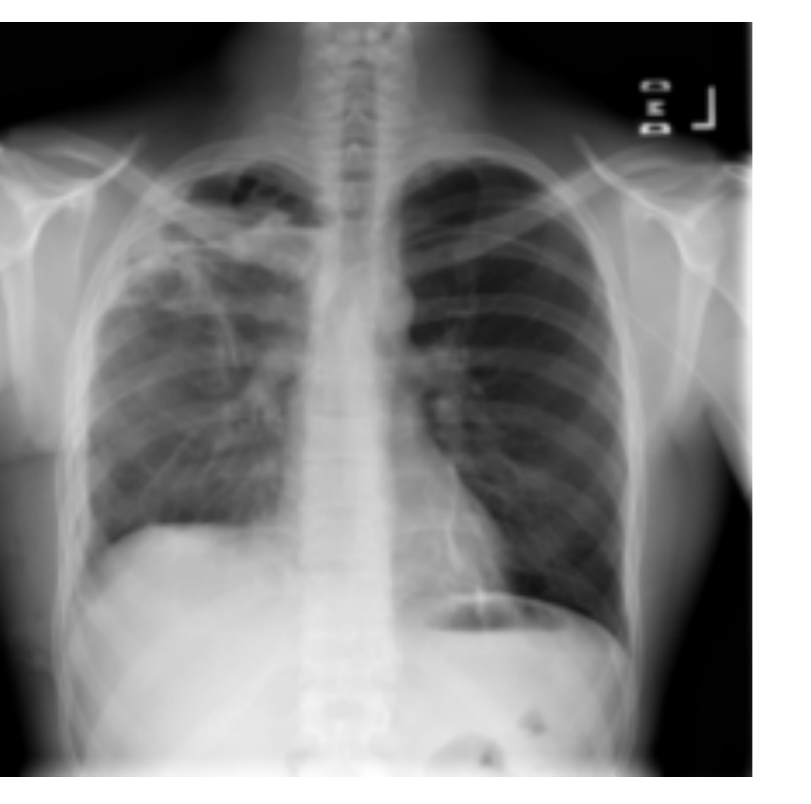} &
    \includegraphics[width=0.195\linewidth, trim={4cm 1cm 3cm 1cm},clip]{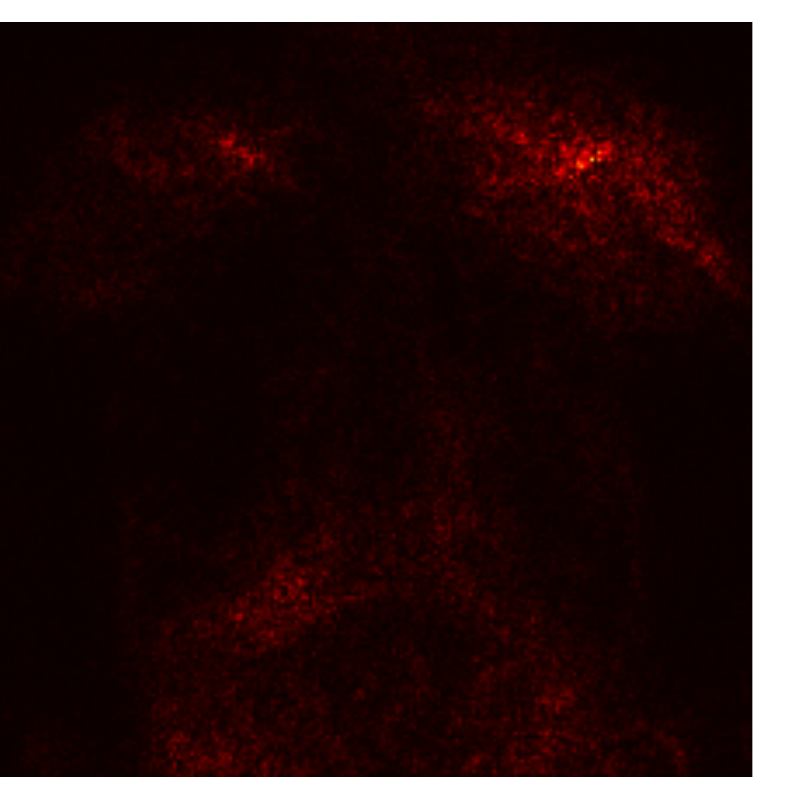}
    \\
    \includegraphics[width=0.195\linewidth, trim={4cm 1cm 3cm 1cm},clip]{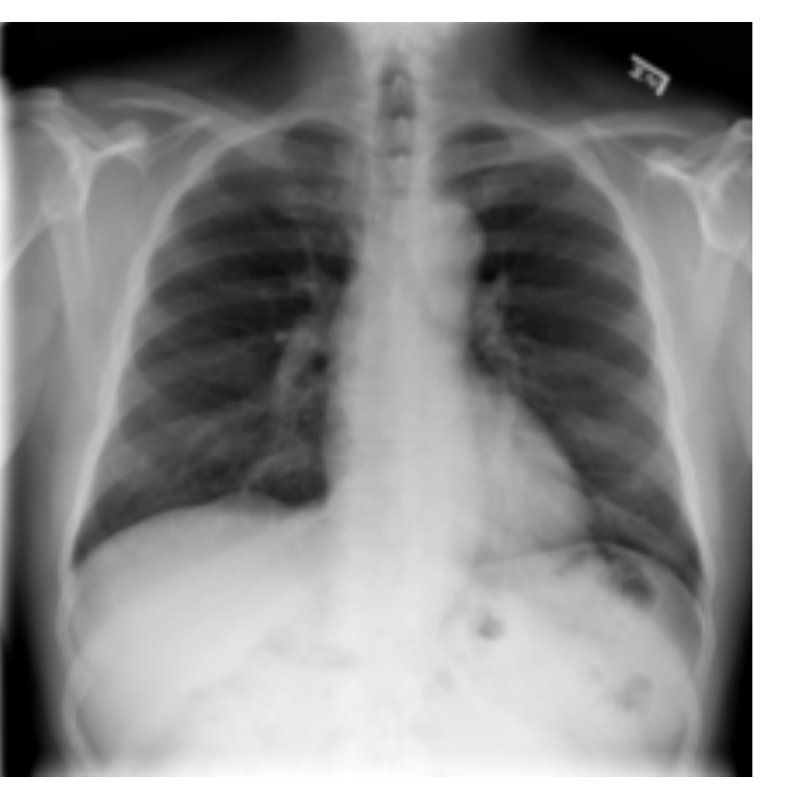} &
    \includegraphics[width=0.195\linewidth, trim={4cm 1cm 3cm 1cm},clip]{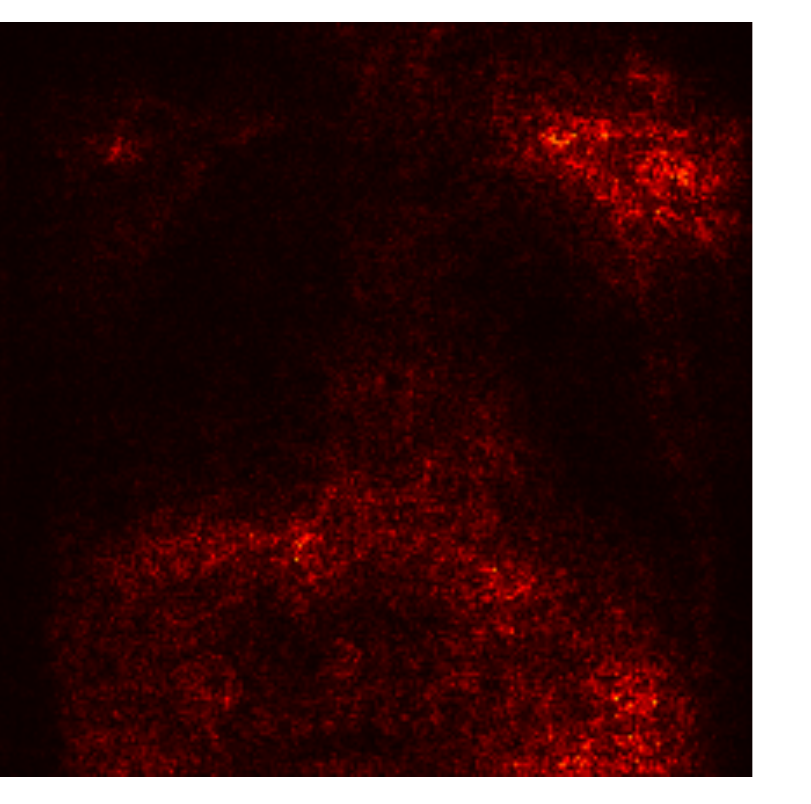}
    \\
    \multicolumn{2}{c}{\small Pneumonia} \\
    \includegraphics[width=0.195\linewidth, trim={4cm 1cm 3cm 1cm},clip]{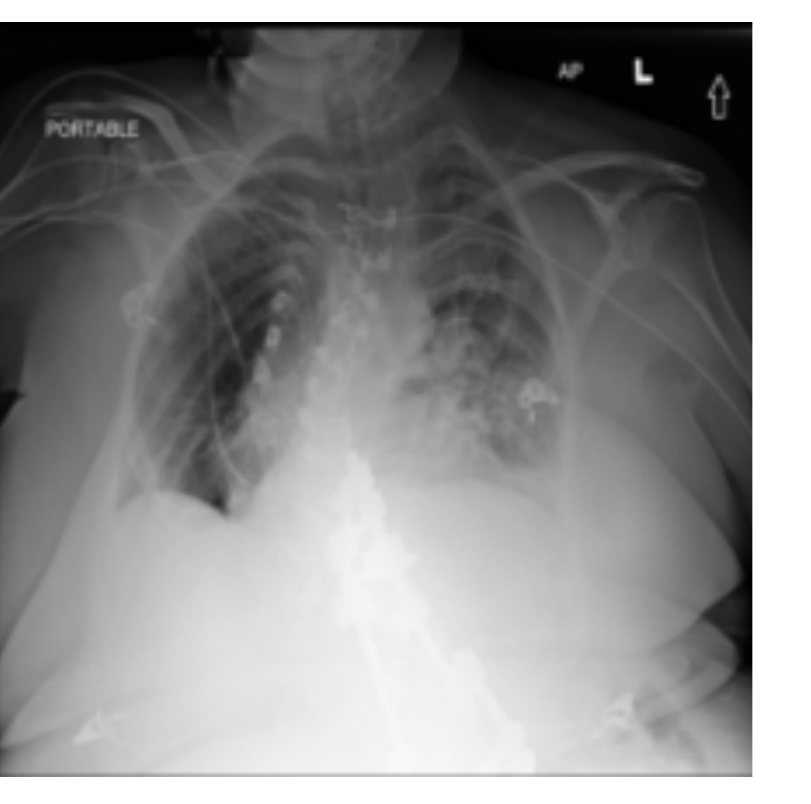} &
    \includegraphics[width=0.195\linewidth, trim={4cm 1cm 3cm 1cm},clip]{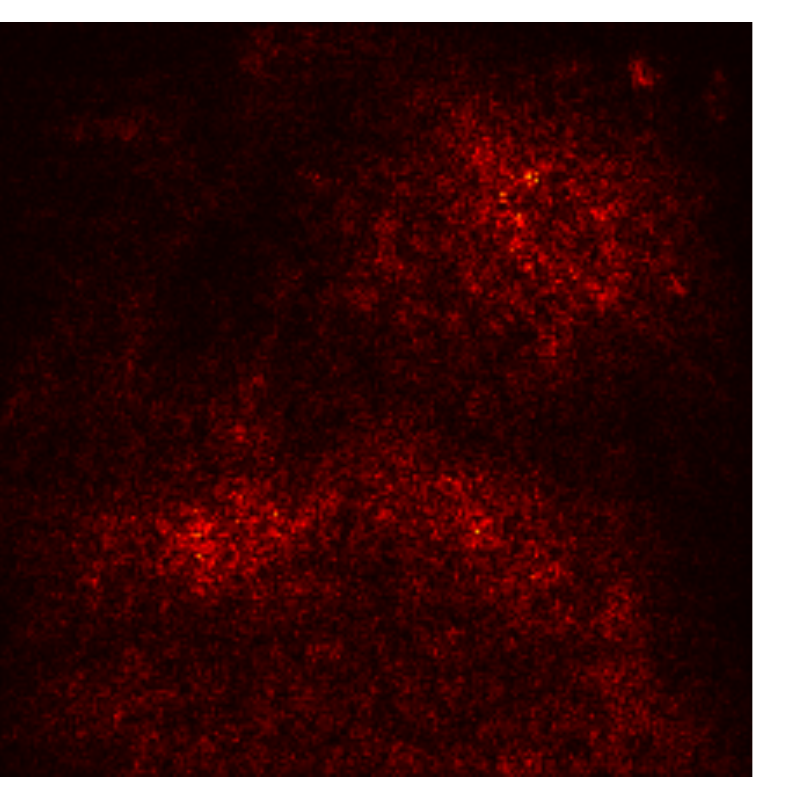}
    \\
    \includegraphics[width=0.195\linewidth, trim={4cm 1cm 3cm 1cm},clip]{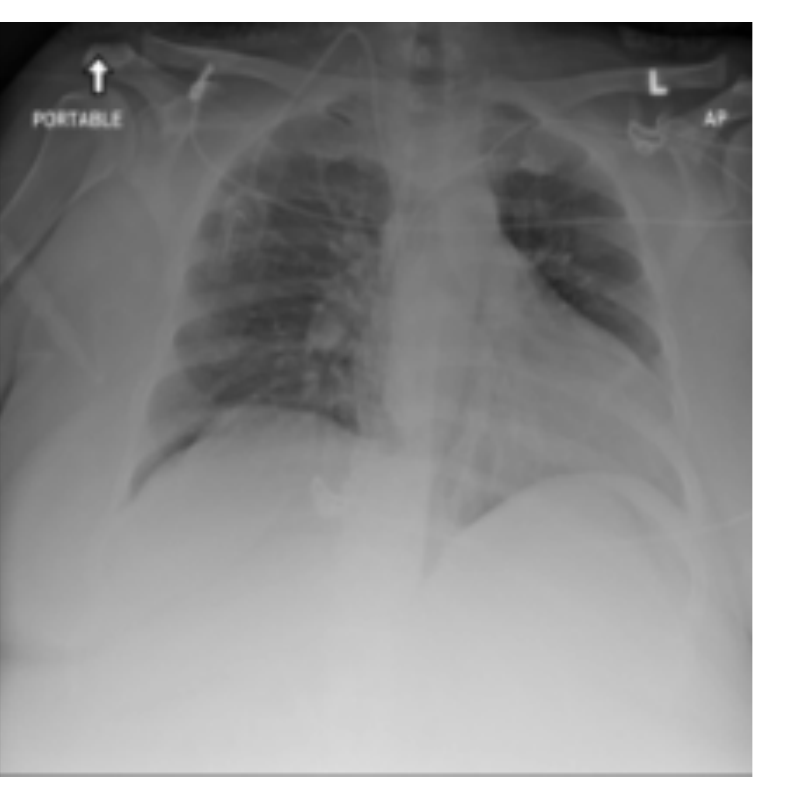} &
    \includegraphics[width=0.195\linewidth, trim={4cm 1cm 3cm 1cm},clip]{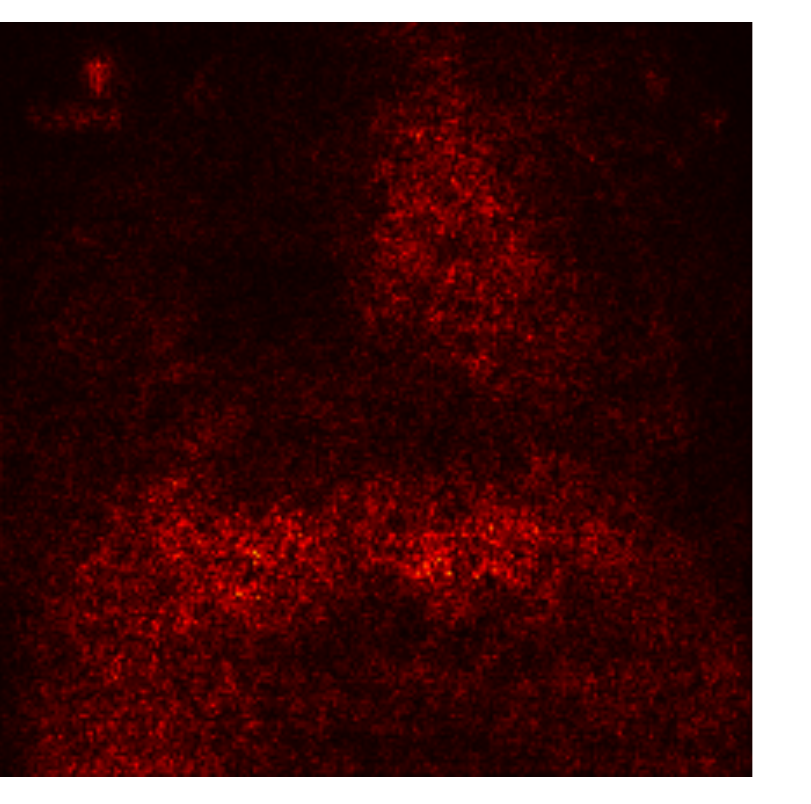}
    \\
    \includegraphics[width=0.195\linewidth, trim={4cm 1cm 3cm 1cm},clip]{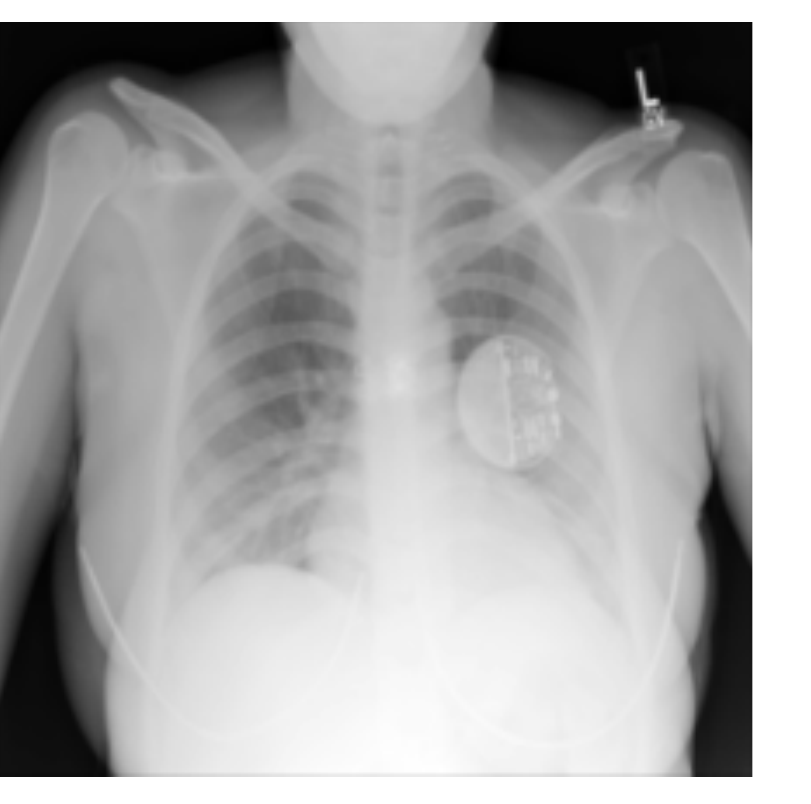} &
    \includegraphics[width=0.195\linewidth, trim={4cm 1cm 3cm 1cm},clip]{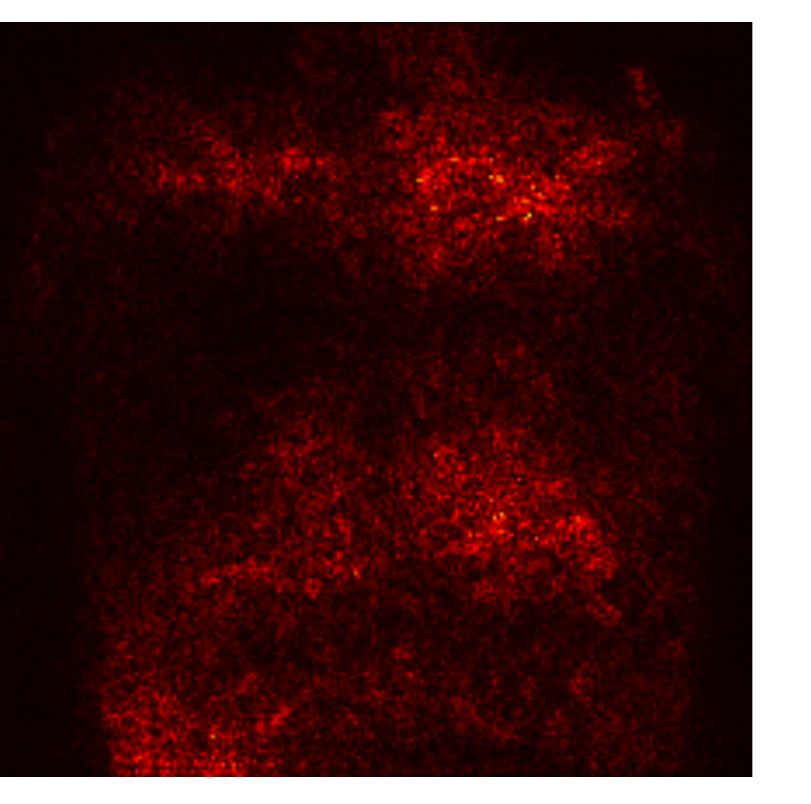}
\end{tabular}}
\caption{Examples from $X^c$. Class-specific MultiMix saliency maps highlight crucial regions in the input X-ray images in detecting pneumonia, demonstrating the effective predictions by the classifier and providing useful information for improved segmentation.}
\label{fig:class_saliency}
\end{figure}

\begin{figure} \centering
\subcaptionbox{JSRT (in-domain)}{
  \begin{tabular}{cc}
  \small Image & \small Saliency Map \\
  \includegraphics[width=0.195\linewidth, trim={4cm 1cm 3cm 1cm},clip]{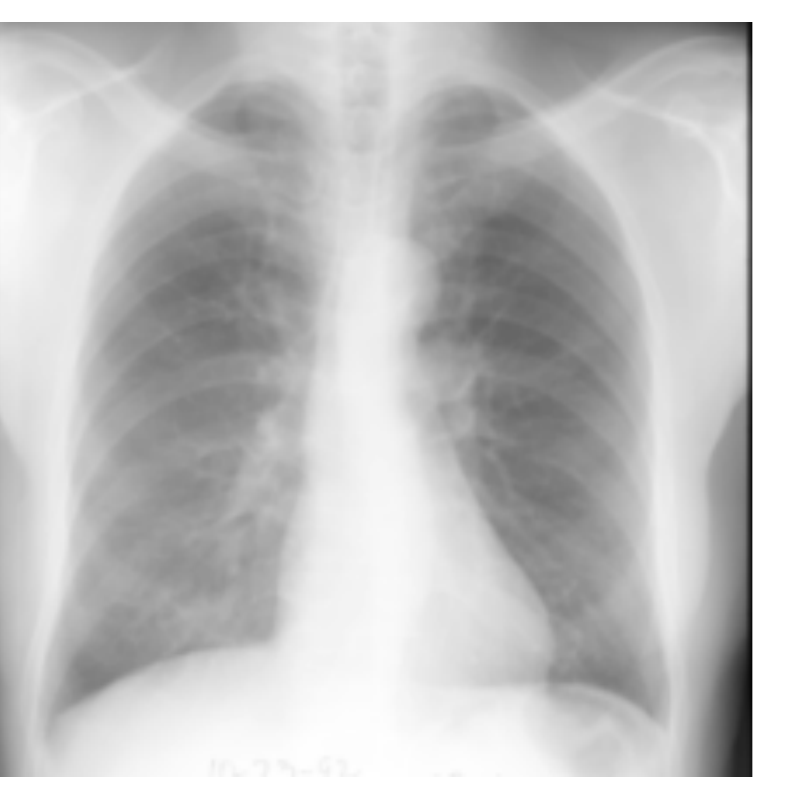} &
  \includegraphics[width=0.195\linewidth, trim={4cm 1cm 3cm 1cm},clip]{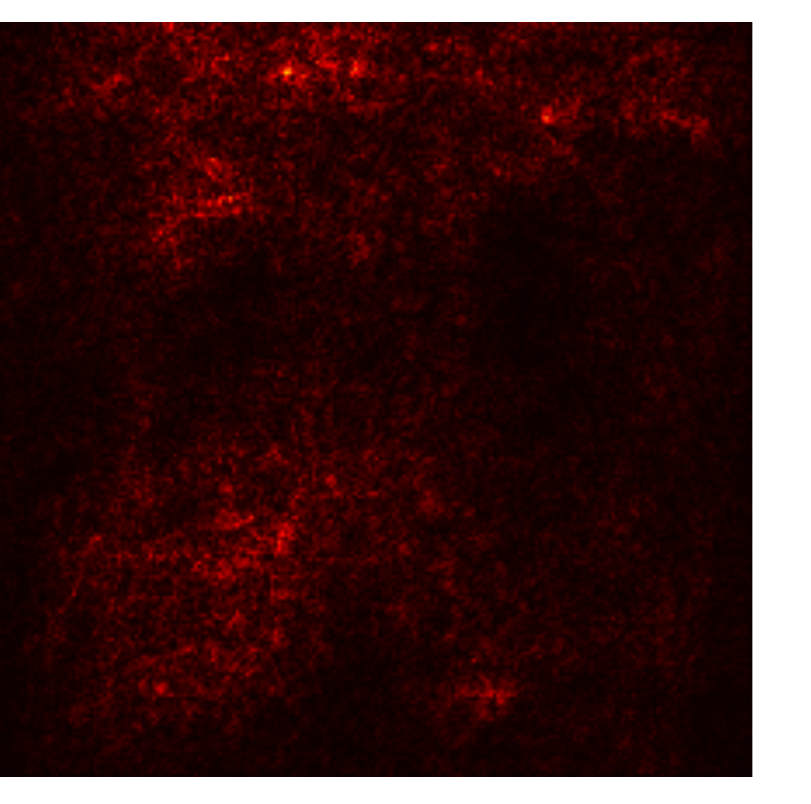}
  \\
  \includegraphics[width=0.195\linewidth, trim={4cm 1cm 3cm 1cm},clip]{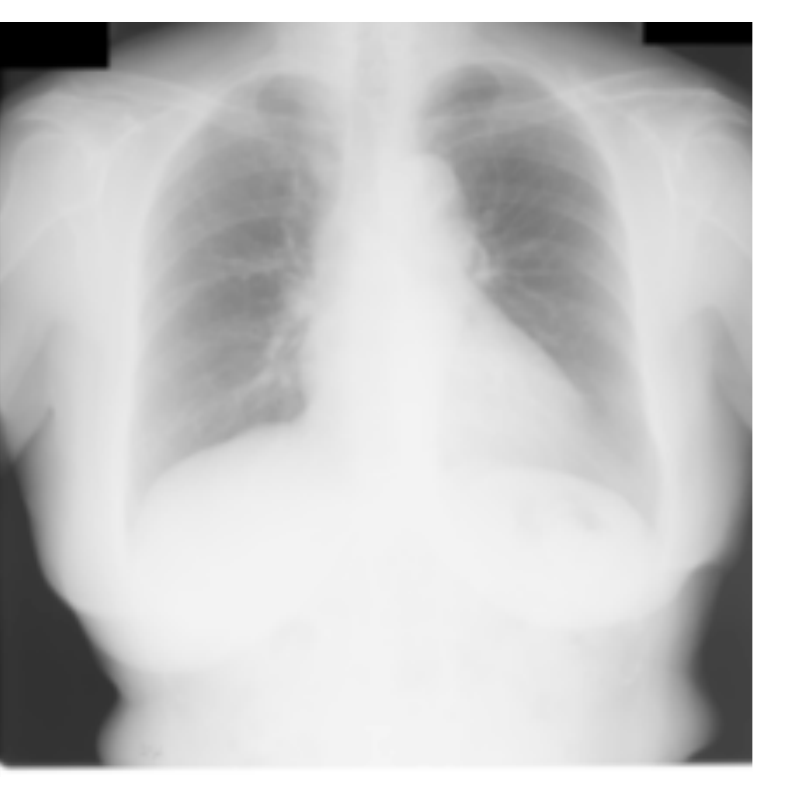} &
  \includegraphics[width=0.195\linewidth, trim={4cm 1cm 3cm 1cm},clip]{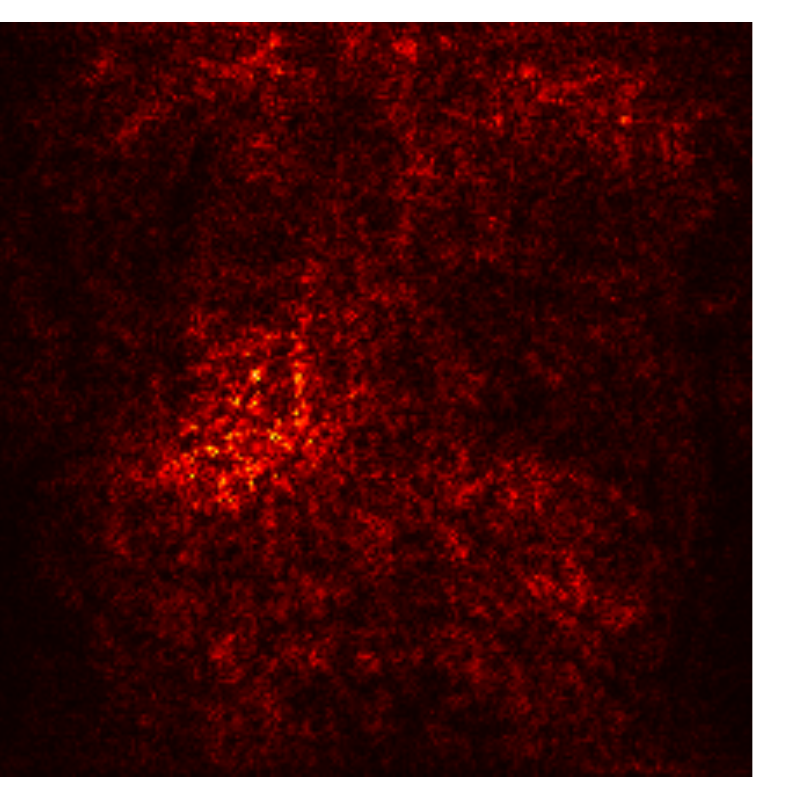}
  \\
  \includegraphics[width=0.195\linewidth, trim={4cm 1cm 3cm 1cm},clip]{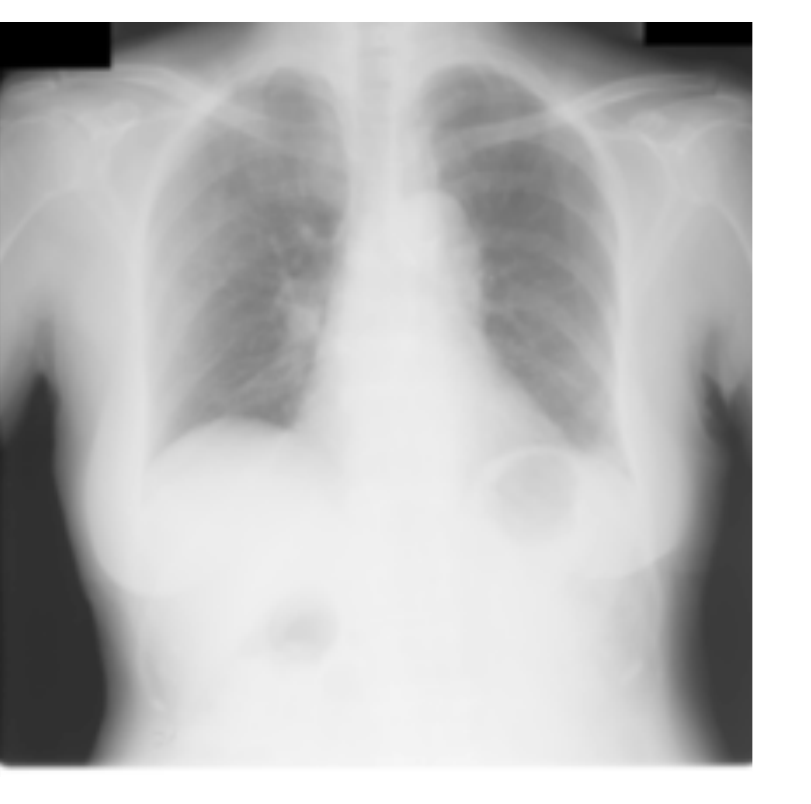} &
  \includegraphics[width=0.195\linewidth, trim={4cm 1cm 3cm 1cm},clip]{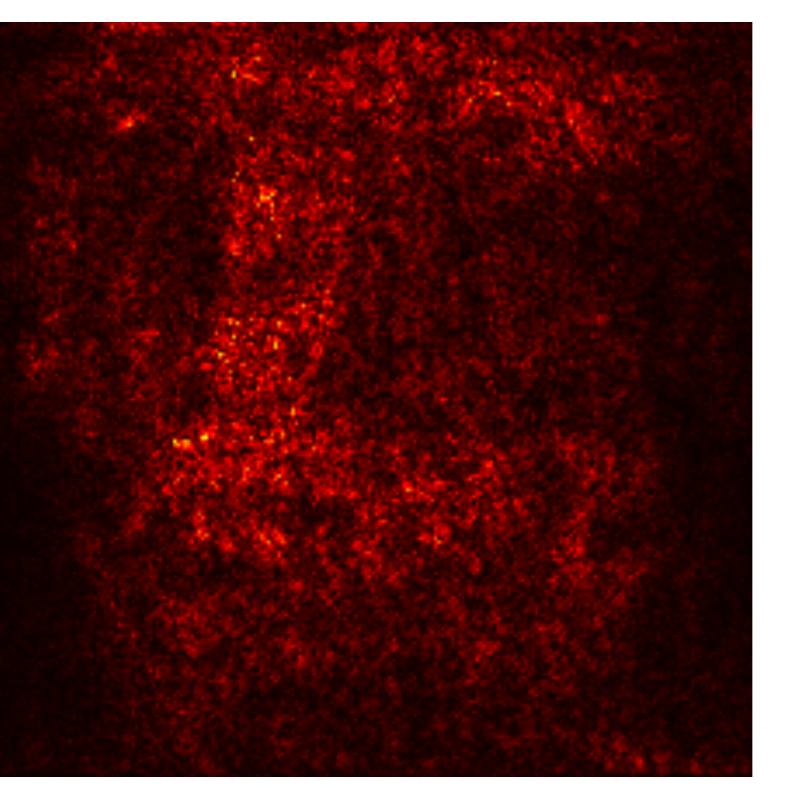}
  \\
  \includegraphics[width=0.195\linewidth, trim={4cm 1cm 3cm 1cm},clip]{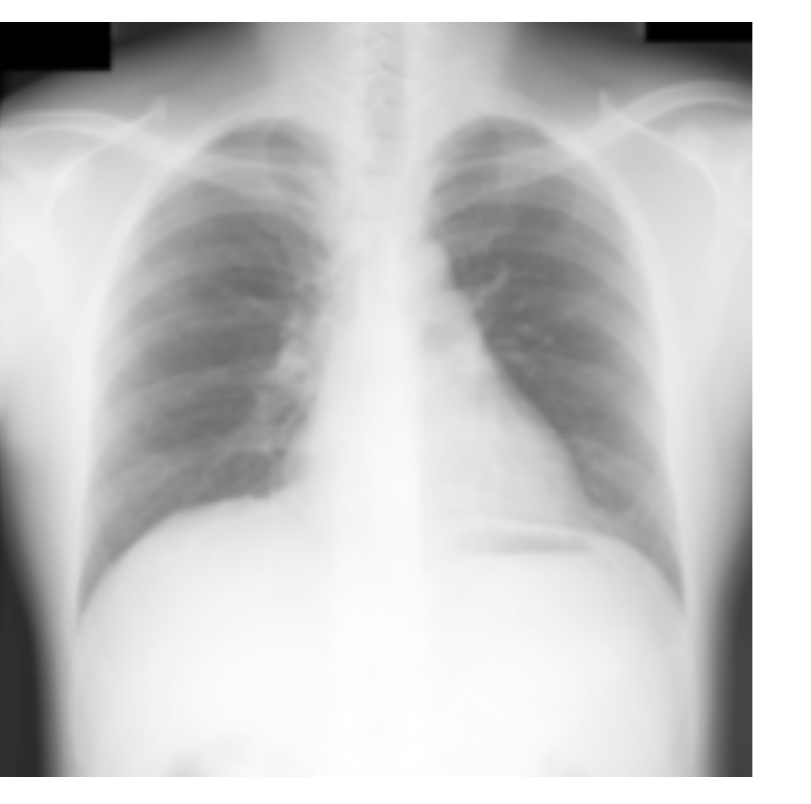} &
  \includegraphics[width=0.195\linewidth, trim={4cm 1cm 3cm 1cm},clip]{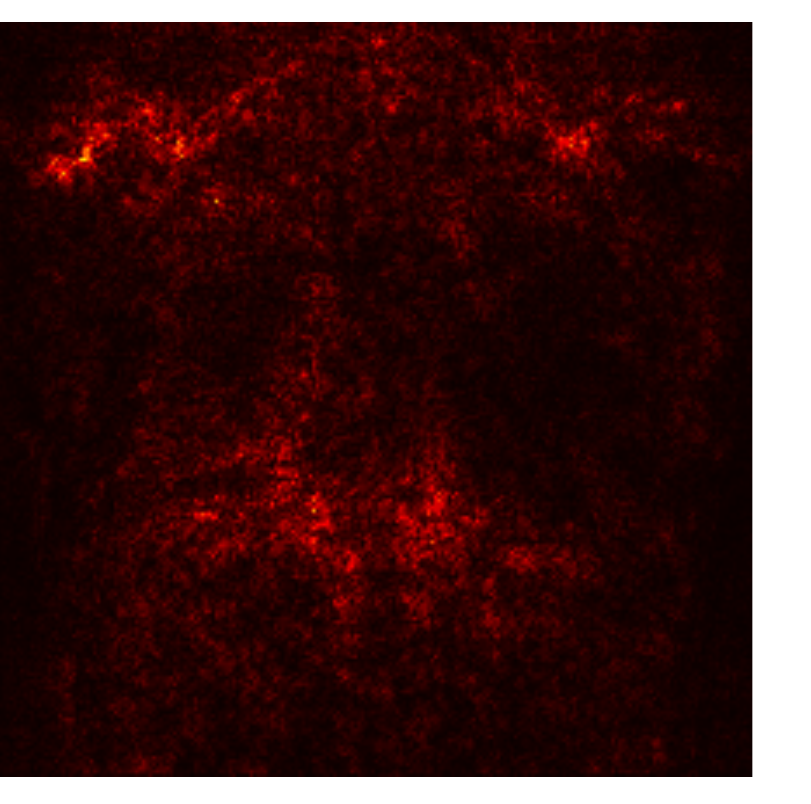}
  \\
  \includegraphics[width=0.195\linewidth, trim={4cm 1cm 3cm 1cm},clip]{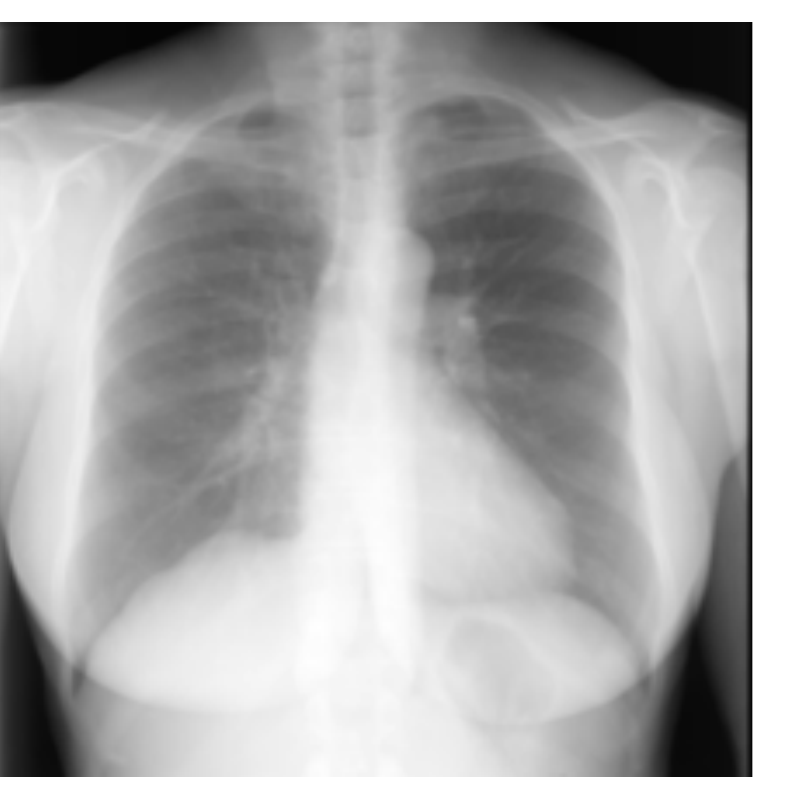} &
  \includegraphics[width=0.195\linewidth, trim={4cm 1cm 3cm 1cm},clip]{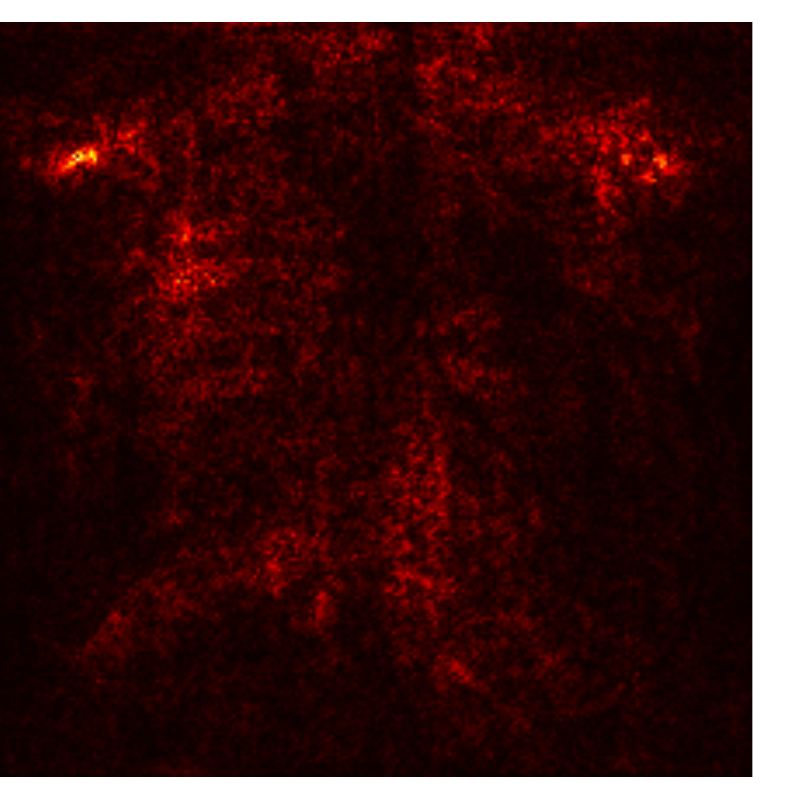}
  \\
  \includegraphics[width=0.195\linewidth, trim={4cm 1cm 3cm 1cm},clip]{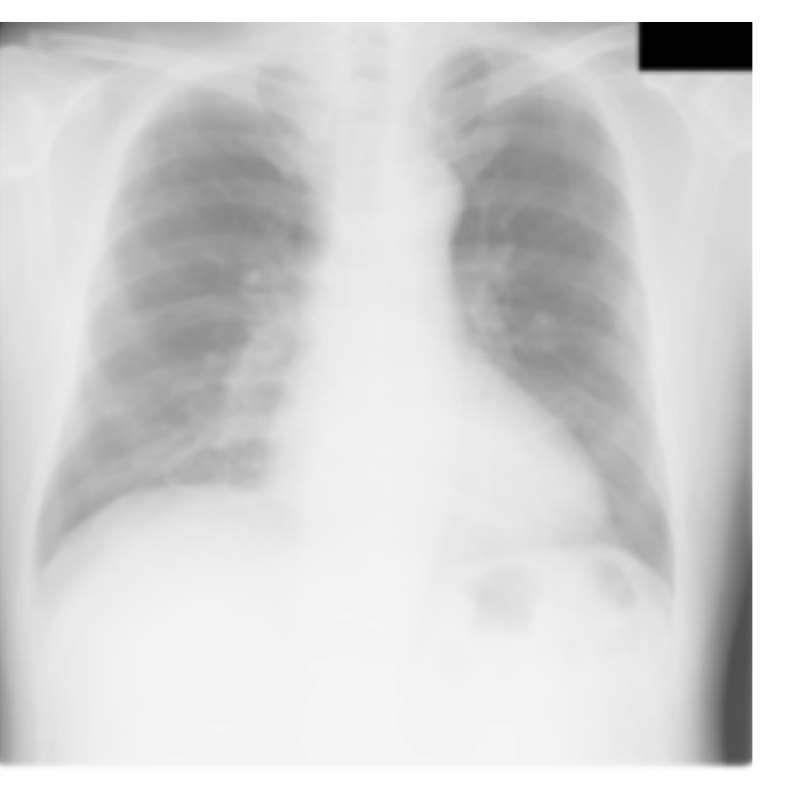} &
  \includegraphics[width=0.195\linewidth, trim={4cm 1cm 3cm 1cm},clip]{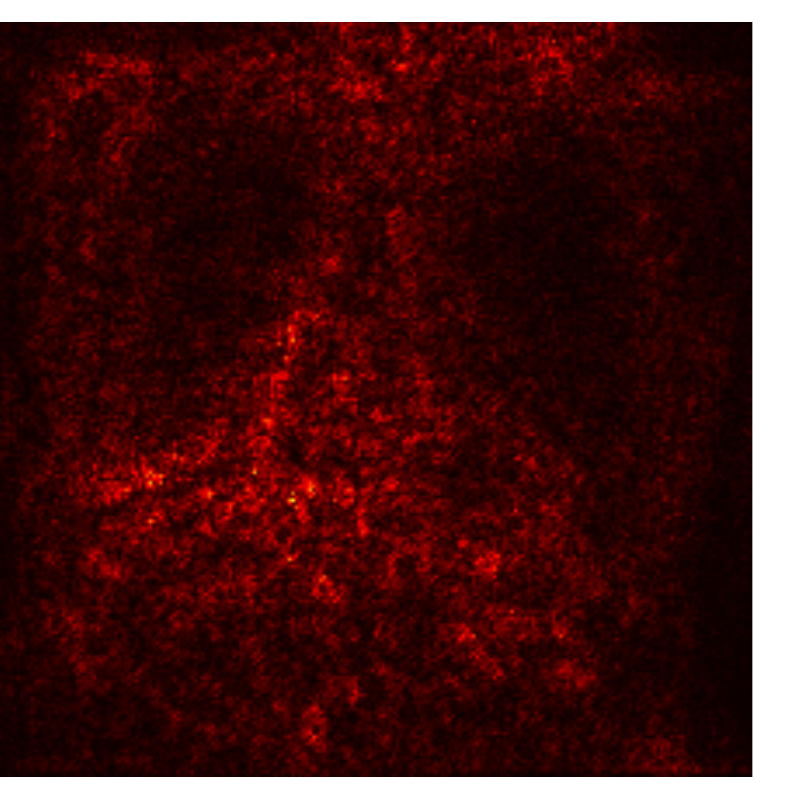}
\end{tabular}}
\hfill
\subcaptionbox{MCU (cross-domain)}{
  \begin{tabular}{cc}
  \small Image & \small Saliency Map \\
  \includegraphics[width=0.195\linewidth, trim={4cm 1cm 3cm 1cm},clip]{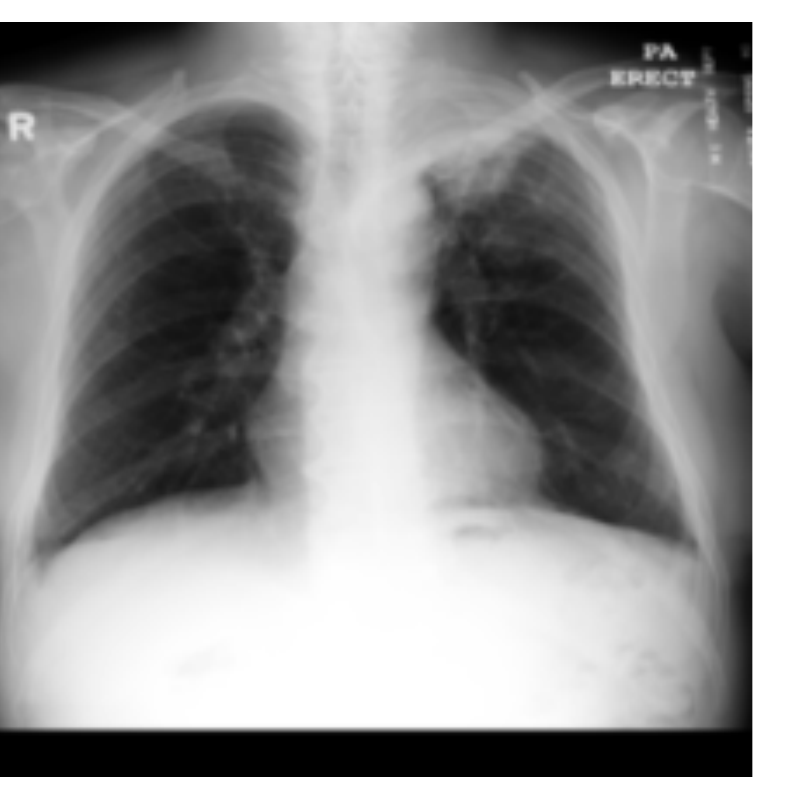} &
  \includegraphics[width=0.195\linewidth, trim={4cm 1cm 3cm 1cm},clip]{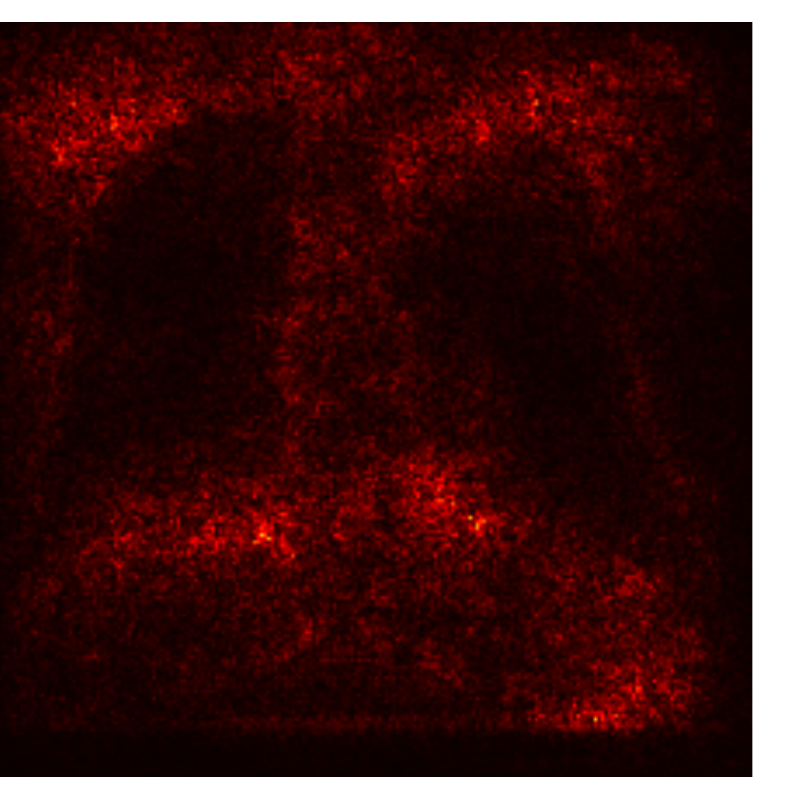}
  \\
  \includegraphics[width=0.195\linewidth, trim={4cm 1cm 3cm 1cm},clip]{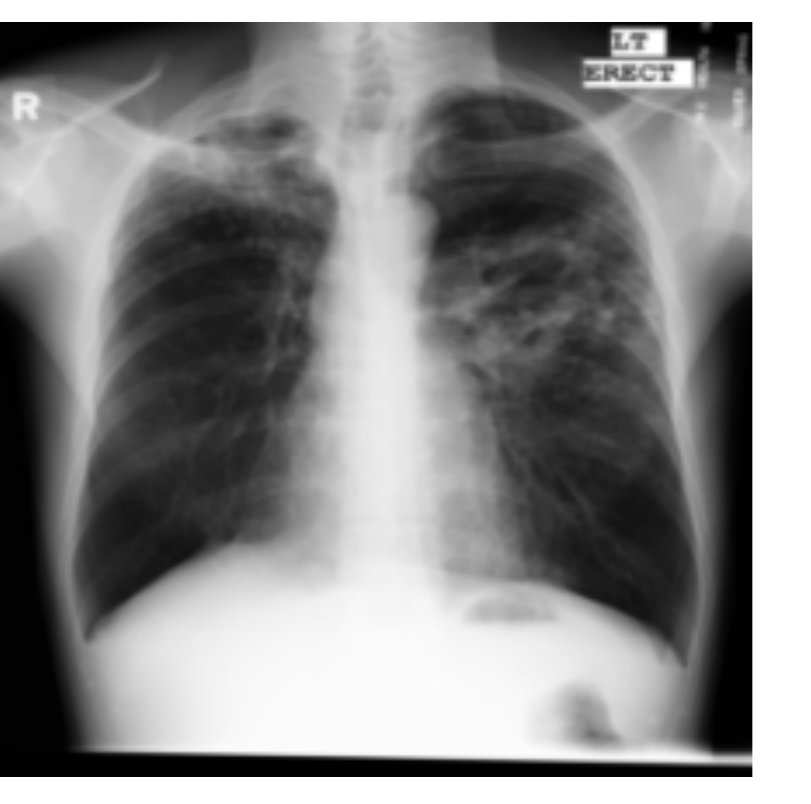} &
  \includegraphics[width=0.195\linewidth, trim={4cm 1cm 3cm 1cm},clip]{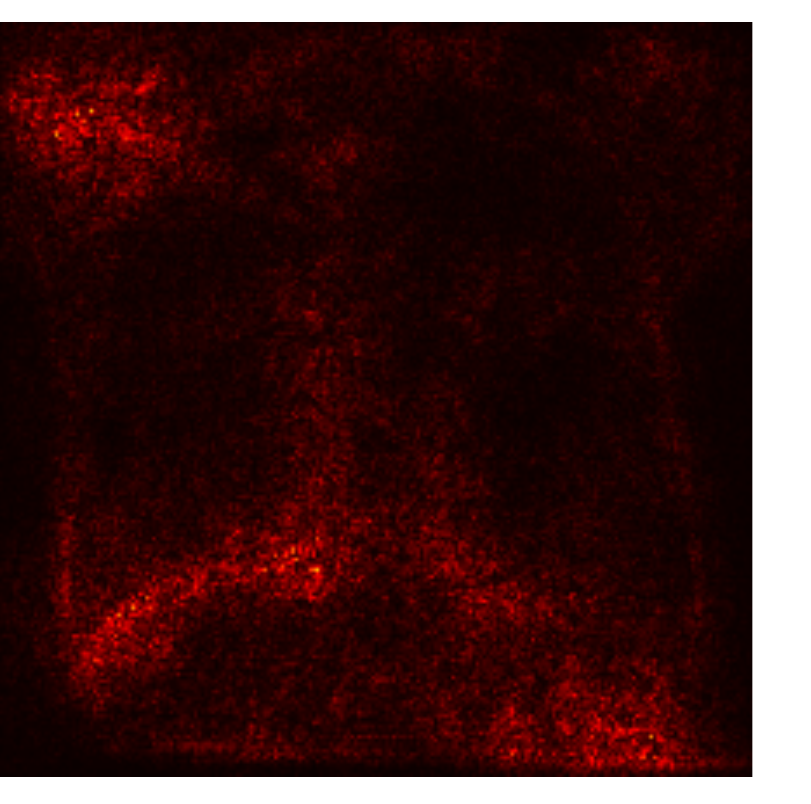}
  \\
  \includegraphics[width=0.195\linewidth, trim={4cm 1cm 3cm 1cm},clip]{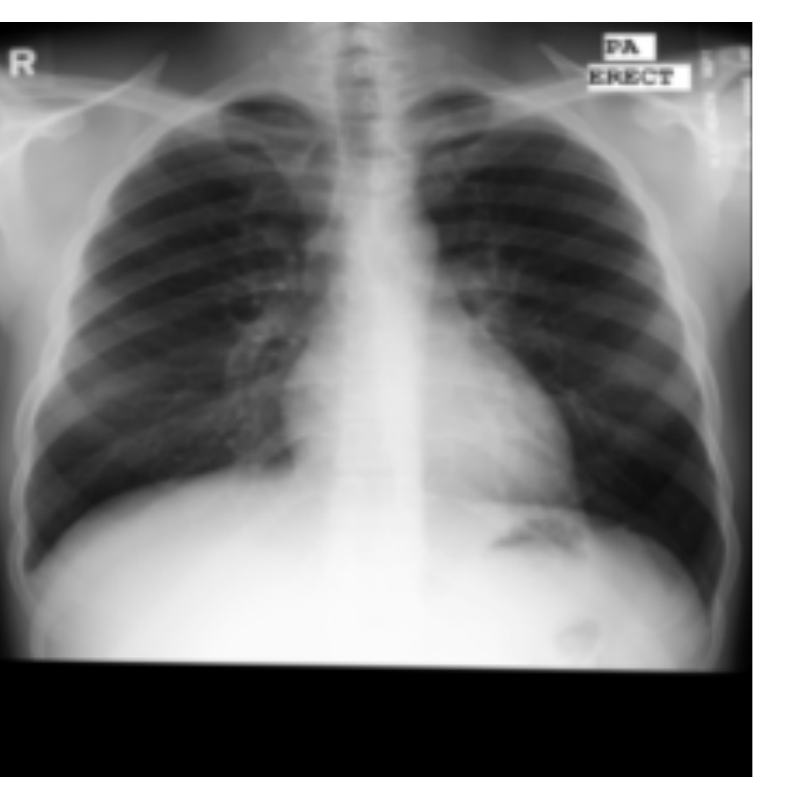} &
  \includegraphics[width=0.195\linewidth, trim={4cm 1cm 3cm 1cm},clip]{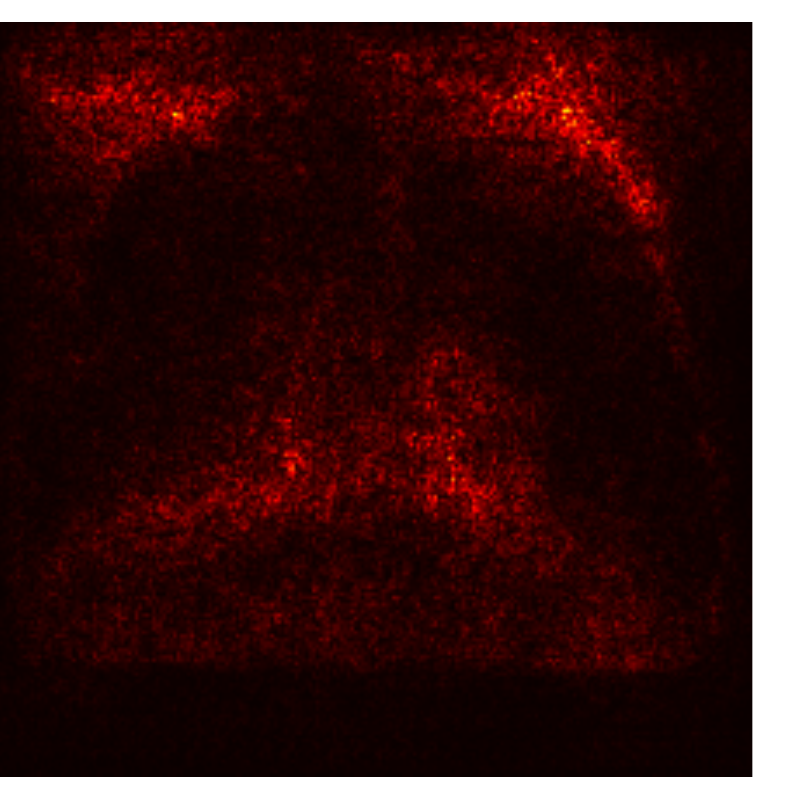}
  \\
  \includegraphics[width=0.195\linewidth, trim={4cm 1cm 3cm 1cm},clip]{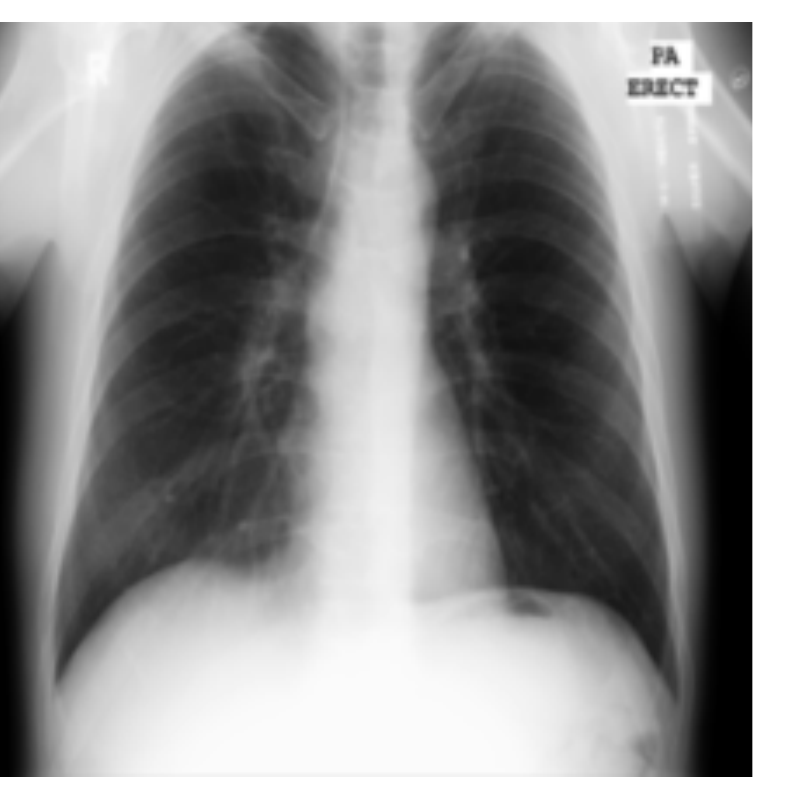} &
  \includegraphics[width=0.195\linewidth, trim={4cm 1cm 3cm 1cm},clip]{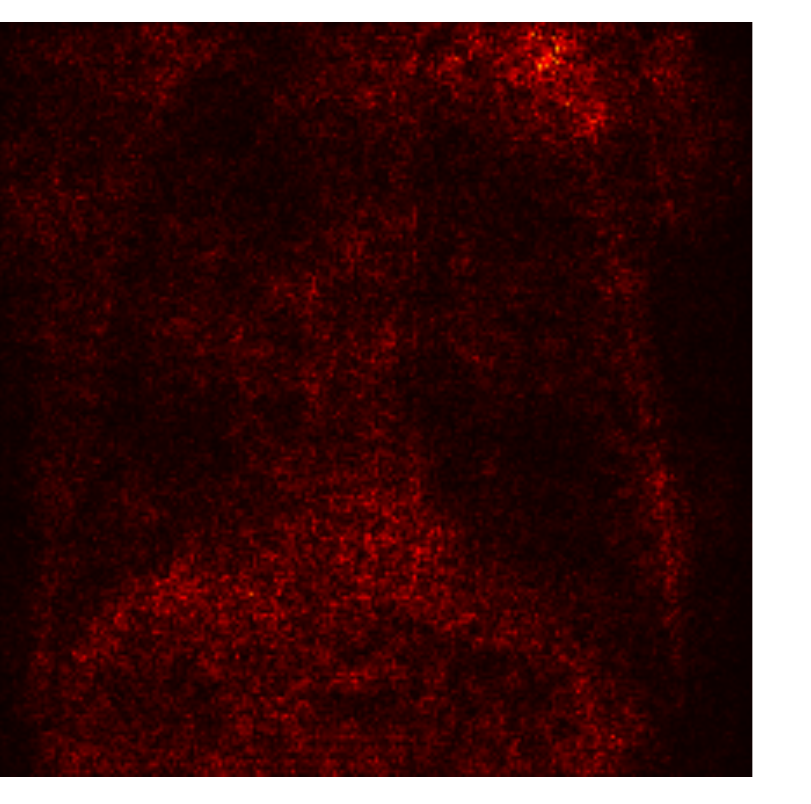}
  \\
  \includegraphics[width=0.195\linewidth, trim={4cm 1cm 3cm 1cm},clip]{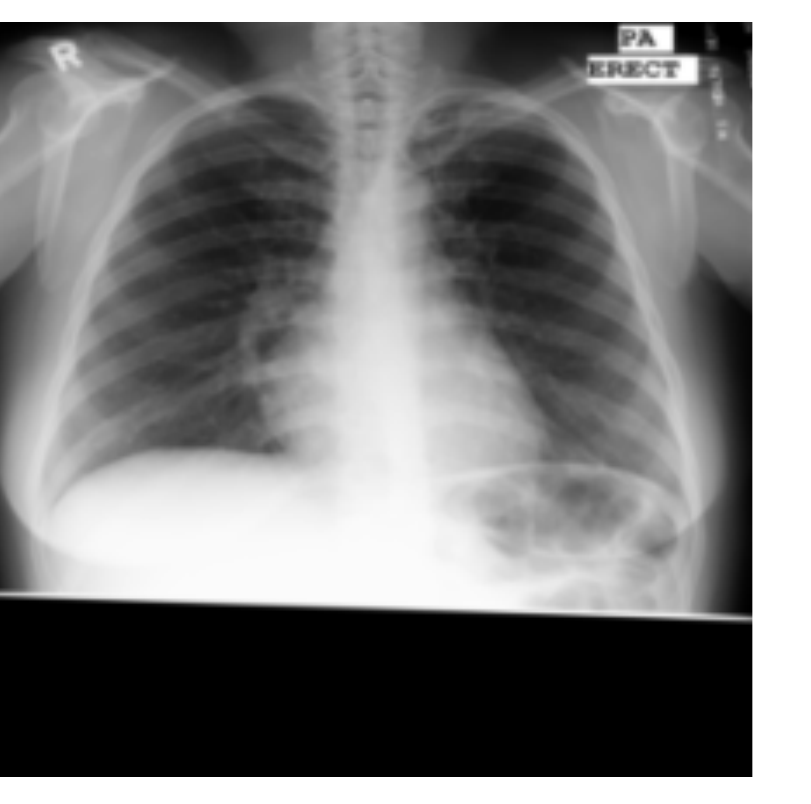} &
  \includegraphics[width=0.195\linewidth, trim={4cm 1cm 3cm 1cm},clip]{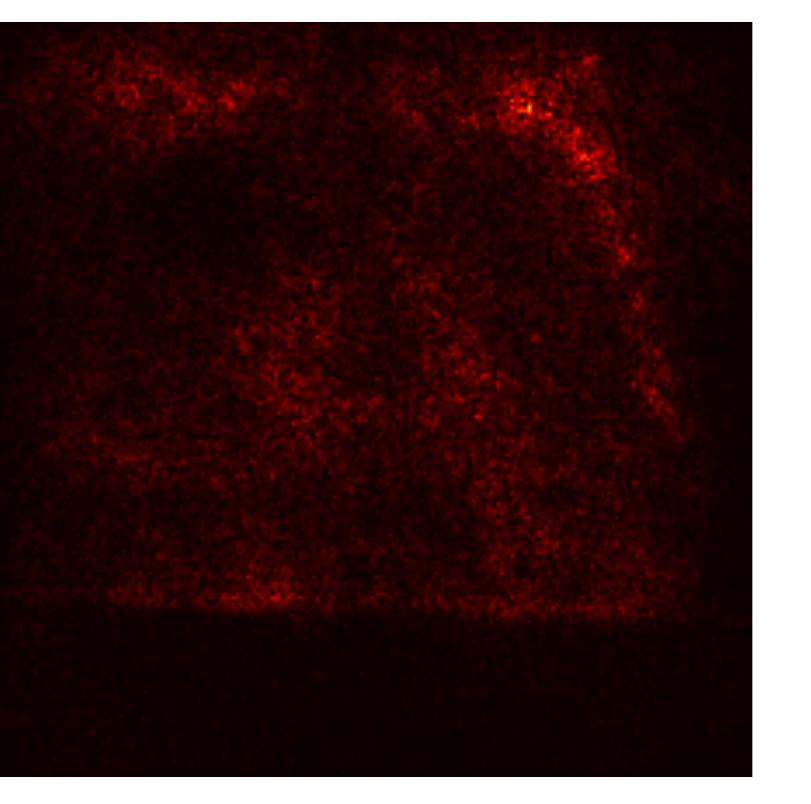}
  \\
  \includegraphics[width=0.195\linewidth, trim={4cm 1cm 3cm 1cm},clip]{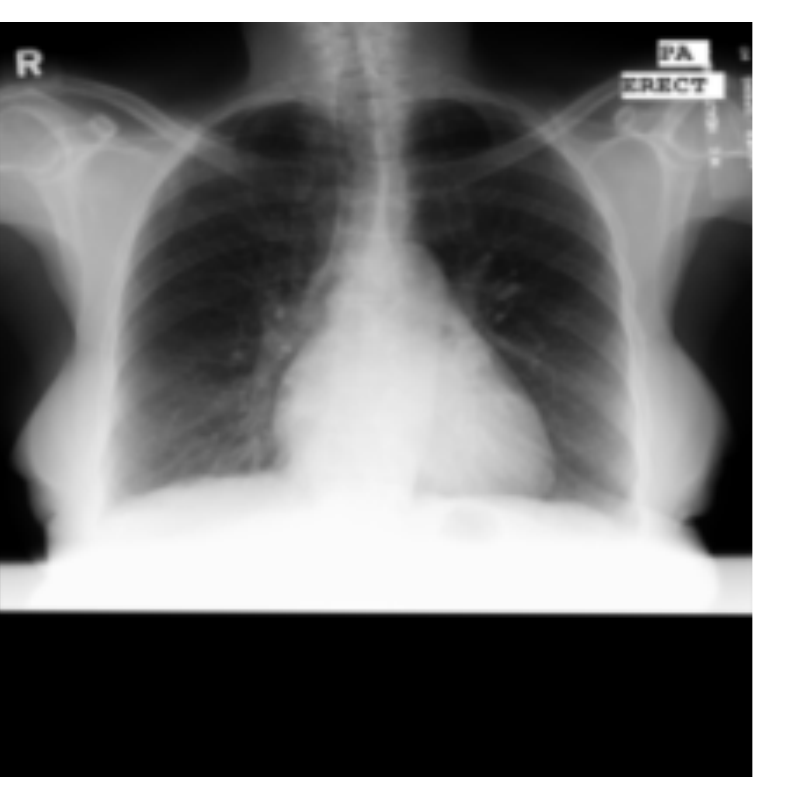} &
  \includegraphics[width=0.195\linewidth, trim={4cm 1cm 3cm 1cm},clip]{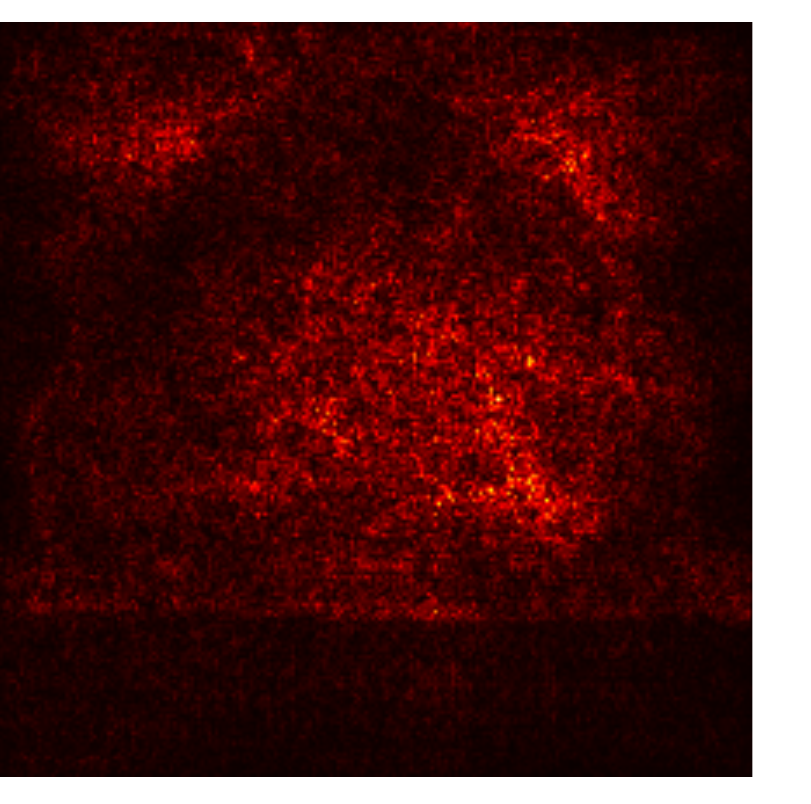}
\end{tabular}}
\caption{Examples from $X^s$. MultiMix saliency maps consistently highlight the crucial regions in the input X-ray images, thus providing useful information for improved segmentation.}
\label{fig:seg_saliency}
\end{figure}

Figure~\ref{fig:class_saliency} shows the class-specific saliency maps generated by our MultiMix-50-1000 model for both in-domain and cross-domain classification data ($X^c$). The maps consistently highlight particular regions in the input X-rays for the Normal and Pneumonia classes. Similarly, Figure~\ref{fig:seg_saliency} shows the saliency maps for the in-domain and cross-domain segmentation data ($X^s$). While the class labels are not available, two distinct types of saliency maps are generated like for the classification data.

Class-specific saliency maps generated for images in $X^c$ consistently highlight regions responsible for predicting the particular classes of the images (Figure~\ref{fig:class_saliency}), enabling the use of these maps to improve the segmentation of images in $X^s$ (Figure~\ref{fig:seg_saliency}).

\begin{table}
\setlength{\tabcolsep}{4pt}
    \centering
     \caption{Architectural details of the MultiMix Encoder for minibatch size  $m$.}
    \label{tab:arch-encoder}
    \begin{tabular}{lcc}
        \toprule
        Name & Input Feature Maps & Output Feature Maps
        \\
        \midrule
        Conv layer - 1 & $m\times256\times256\times1$  & $m\times256\times256\times16$ \\
        InstanceNorm - 1 & $m\times256\times256\times16$ & $m\times256\times256\times16$ \\
        LReLU - 1 & $m\times256\times256\times16$  & $m\times256\times256\times16$ \\
        Conv Layer - 2 & $m\times256\times256\times16$  & $m\times256\times256\times16$ \\
        InstanceNorm - 2 & $m\times256\times256\times16$ & $m\times256\times256\times16$ \\
        LReLU - 2 & $m\times256\times256\times16$  & $m\times256\times256\times16$ \\
        Dropout - 1 & $m\times256\times256\times16$  & $m\times256\times256\times16$ \\
        Maxpool - 1 & $m\times256\times256\times16$  & $m\times128\times128\times16$ \\
        
        Conv Layer - 3 & $m\times128\times128\times16$  & $m\times128\times128\times32$ \\
        InstanceNorm - 3 & $m\times128\times128\times32$ & $m\times128\times128\times32$ \\
        LReLU - 3 & $m\times128\times128\times32$ & $m\times128\times128\times32$ \\
        Conv Layer - 4 & $m\times128\times128\times32$ & $m\times128\times128\times32$ \\
        InstanceNorm - 4 & $m\times128\times128\times32$ & $m\times128\times128\times32$ \\
        LReLU - 4 & $m\times128\times128\times32$ & $m\times128\times128\times32$ \\
        Dropout - 2 & $m\times128\times128\times32$ & $m\times128\times128\times32$ \\
        Maxpool - 2 & $m\times128\times128\times32$ & $m\times64\times64\times32$ \\
        
        Conv Layer - 5 & $m\times64\times64\times32$  & $m\times64\times64\times64$ \\
        InstanceNorm - 5 & $m\times64\times64\times64$ & $m\times64\times64\times64$ \\
        LReLU - 5 & $m\times64\times64\times64$ & $m\times64\times64\times64$ \\
        Conv Layer - 6 & $m\times64\times64\times64$ & $m\times64\times64\times64$ \\
        InstanceNorm - 6 & $m\times64\times64\times64$ & $m\times64\times64\times64$ \\
        LReLU - 6 & $m\times64\times64\times64$ & $m\times64\times64\times64$ \\
        Dropout - 3 & $m\times64\times64\times64$ & $m\times64\times64\times64$ \\
        Maxpool - 3 & $m\times64\times64\times64$ & $m\times32\times32\times64$ \\
        
        Conv Layer - 7 & $m\times32\times32\times64$  & $m\times32\times32\times128$ \\
        InstanceNorm - 7 & $m\times32\times32\times128$ & $m\times32\times32\times128$ \\
        LReLU - 7 & $m\times32\times32\times128$ & $m\times32\times32\times128$ \\
        Conv Layer - 8 & $m\times32\times32\times128$ & $m\times32\times32\times128$ \\
        InstanceNorm - 8 & $m\times32\times32\times128$ & $m\times32\times32\times128$ \\
        LReLU - 8 & $m\times32\times32\times128$ & $m\times32\times32\times128$ \\
        Dropout - 4 & $m\times32\times32\times128$ & $m\times32\times32\times128$ \\
        Maxpool - 4 & $m\times32\times32\times128$ & $m\times16\times16\times128$ \\
        
        Conv Layer - 9 & $m\times16\times16\times128$  & $m\times16\times16\times256$ \\
        InstanceNorm - 9 & $m\times16\times16\times256$ & $m\times16\times16\times256$ \\
        LReLU - 9 & $m\times16\times16\times256$ & $m\times16\times16\times256$ \\
        Conv Layer - 10 & $m\times16\times16\times256$ & $m\times16\times16\times256$ \\
        InstanceNorm - 10 & $m\times16\times16\times256$ & $m\times16\times16\times256$ \\
        LReLU - 10 & $m\times16\times16\times256$ & $m\times16\times16\times256$ \\
        Dropout - 5 & $m\times16\times16\times256$ & $m\times16\times16\times256$ \\
        Maxpool - 5 & $m\times16\times16\times256$ & $m\times8\times8\times256$ \\
        
        Avgpool & $m\times8\times8\times256$ & $m\times1\times1\times256$ \\
        GAP & $m\times1\times1\times256$ & $m\times256$ \\
        Fully Connected Layer & $m\times256$ & $m\times2$ \\
        \bottomrule
    \end{tabular}
\end{table}

\begin{table}
\setlength{\tabcolsep}{4pt}
    \centering
     \caption{Architectural details of the MultiMix Decoder for minibatch size $m$.}
    \label{tab:arch-decoder}
    \begin{tabular}{lcc}
        \toprule
        Name & Input Feature Maps & Output Feature Maps
        \\
        \midrule
        Upsample - 1 & $m\times16\times16\times256$  & $m\times32\times32\times256$ \\
        Conv Layer - 1 & $m\times32\times32\times386$ & $m\times32\times32\times128$ \\
        InstanceNorm - 1 & $m\times32\times32\times128$ & $m\times32\times32\times128$ \\
        LReLU - 1 & $m\times32\times32\times128$  & $m\times32\times32\times128$ \\
        Conv Layer - 2 & $m\times32\times32\times128$  & $m\times32\times32\times128$ \\
        InstanceNorm - 2 & $m\times32\times32\times128$ & $m\times32\times32\times128$ \\
        LReLU - 2 & $m\times32\times32\times128$  & $m\times32\times32\times128$ \\
        Dropout - 1 & $m\times32\times32\times128$  & $m\times32\times32\times128$ \\
        
        Upsample - 2 & $m\times32\times32\times128$  & $m\times64\times64\times128$ \\
        Conv Layer - 3 & $m\times64\times64\times192$  & $m\times64\times64\times64$ \\
        InstanceNorm - 3 & $m\times64\times64\times64$ & $m\times64\times64\times64$ \\
        LReLU - 3 & $m\times64\times64\times64$ & $m\times64\times64\times64$ \\
        Conv Layer - 4 & $m\times64\times64\times64$ & $m\times64\times64\times64$ \\
        InstanceNorm - 4 & $m\times64\times64\times64$ & $m\times64\times64\times64$ \\
        LReLU - 4 & $m\times64\times64\times64$ & $m\times64\times64\times64$ \\
        Dropout - 2 & $m\times64\times64\times64$ & $m\times64\times64\times64$ \\
        
        Upsample - 3 & $m\times64\times64\times64$  & $m\times128\times128\times64$ \\
        Conv Layer - 5 & $m\times128\times128\times96$  & $m\times128\times128\times32$ \\
        InstanceNorm - 5 & $m\times128\times128\times32$ & $m\times128\times128\times32$ \\
        LReLU - 5 & $m\times128\times128\times32$ & $m\times128\times128\times32$ \\
        Conv Layer - 6 & $m\times128\times128\times32$ & $m\times128\times128\times32$ \\
        InstanceNorm - 6 & $m\times128\times128\times32$ & $m\times128\times128\times32$ \\
        LReLU - 6 & $m\times128\times128\times32$ & $m\times128\times128\times32$ \\
        Dropout - 3 & $m\times128\times128\times32$ & $m\times128\times128\times32$ \\
        
        Upsample - 4 & $m\times128\times128\times32$  & $m\times256\times256\times32$ \\
        Conv Layer - 7 & $m\times256\times256\times48$  & $m\times256\times256\times16$ \\
        InstanceNorm - 7 & $m\times256\times256\times16$ & $m\times256\times256\times16$ \\
        LReLU - 7 & $m\times256\times256\times16$ & $m\times256\times256\times16$ \\
        Conv Layer - 8 & $m\times256\times256\times16$ & $m\times256\times256\times16$ \\
        InstanceNorm - 8 & $m\times256\times256\times16$ & $m\times256\times256\times16$ \\
        LReLU - 8 & $m\times256\times256\times16$ & $m\times256\times256\times16$ \\
        Dropout - 4 & $m\times256\times256\times16$ & $m\times256\times256\times16$ \\
        
        Final Conv Layer & $m\times256\times256\times16$  & $m\times256\times256\times1$ \\
        \bottomrule
    \end{tabular}
\end{table}

\newpage

\bibliography{melba21}

\end{document}